\documentclass[lettersize,journal]{IEEEtran}
\usepackage{amsmath,amsfonts}
\usepackage{algorithmic}
\usepackage{algorithm}
\usepackage{array}
\usepackage[caption=false,font=normalsize,labelfont=sf,textfont=sf]{subfig}
\usepackage{textcomp}
\usepackage{stfloats}
\usepackage{url}
\usepackage{verbatim}
\usepackage{graphicx}
\usepackage{cite}

\usepackage{caption} % 提供 caption 命令
\usepackage{subcaption} % 处理子图的宏包

\usepackage{multirow}
\usepackage{makecell}
\usepackage{algorithm}
\usepackage{algorithmic}
\usepackage{amsmath}
\usepackage{amsthm}
\usepackage{bm}
\usepackage{color}
\usepackage{graphicx}
\usepackage{amssymb}
\usepackage{url}
\usepackage{booktabs}
\usepackage{multirow}
\usepackage[normalem]{ulem}
\usepackage[colorlinks,linkcolor=blue]{hyperref}
\usepackage{ragged2e}
\usepackage{soul}
\useunder{\uline}{\ul}{}

\ifCLASSOPTIONcompsoc
  \usepackage[nocompress]{cite}
\else
  \usepackage{cite}
\fi

\ifCLASSINFOpdf
\else
\fi

\begin{document}

% \chapter{chapter 1}

% Do not put math or special symbols in the title .From Transfer to Transformer: A Survey on Time-Series Pre-Trained Models
\title{A Survey on Time-Series Pre-Trained Models}

\author{Qianli Ma\textsuperscript{\dag}\thanks{{\dag} Corresponding author: Qianli Ma.},~\IEEEmembership{Member,~IEEE,}
        Zhen Liu, Zhenjing Zheng, Ziyang Huang, Siying Zhu, Zhongzhong Yu, \\ and James T. Kwok,~\IEEEmembership{Fellow,~IEEE}
        % , Garrison W. Cottrell

% \thanks{$^{\dagger}$Corresponding author} 
% $^{\dagger}$ $^*$ $^*$ ,$^*$Equal contribution

% \thanks{Qianli Ma is with the School of Computer Science and Engineering, South China University of Technology, Guangzhou 510006, China (e-mail: qianlima@scut.edu.cn, corresponding author).}
\thanks{Qianli Ma and Zhen Liu are co-first authors.} 
\thanks{Qianli Ma, Zhen Liu, Zhenjing Zheng, Ziyang Huang, Siying Zhu, and Zhongzhong Yu are with the School of Computer
Science and Engineering, South China University of Technology, Guangzhou 510006, China. E-mail: qianlima@scut.edu.cn; cszhenliu@mail.scut.edu.cn; 982360227@qq.com; stevenhuang12@outlook.com; 489531037@qq.com; yuzzhong2020@foxmail.com. 
% (Corresponding author: Qianli Ma.)
}

\thanks{James T. Kwok is with the Department of Computer Science and Engineering, The Hong Kong University of Science and Technology,
Hong Kong, China. E-mail: jamesk@cse.ust.hk.}

% \thanks{Garrison W. Cottrell is with the Department of Computer Science and Engineering, University of California, San Diego, CA 92093, USA (e-mail: gary@ucsd.edu)}

}

% \markboth{Journal of \LaTeX\ Class Files,~Vol.~14, No.~8, August~2015}%
% {Shell \MakeLowercase{\textit{et al.}}: Bare Demo of IEEEtran.cls for Computer Society Journals}

\markboth{ This paper is accepted in the \textbf{IEEE Transactions on Knowledge and Data Engineering (TKDE)}}%
{Shell \MakeLowercase{\textit{et al.}}: A Sample Article Using IEEEtran.cls for IEEE Journals}

\IEEEtitleabstractindextext{%
\begin{abstract}
\justifying
Time-Series Mining (TSM) is an important research area since it shows great potential in practical applications. Deep learning models that rely on massive labeled data have been utilized for TSM successfully. However, constructing a large-scale well-labeled dataset is difficult due to data annotation costs.  Recently, pre-trained models have gradually attracted attention in the time series domain due to their remarkable performance in computer vision and natural language processing. In this survey, we provide a comprehensive review of Time-Series Pre-Trained Models (TS-PTMs), aiming to guide the understanding, applying, and studying TS-PTMs. Specifically, we first briefly introduce the typical deep learning models employed in TSM. Then, we give an overview of TS-PTMs according to the pre-training techniques. The main categories we explore include supervised, unsupervised, and self-supervised TS-PTMs. Further, extensive experiments involving  27 methods, 434 datasets, and 679 transfer learning scenarios are conducted to analyze the advantages and disadvantages of transfer learning strategies, Transformer-based models, and representative TS-PTMs. Finally, we point out some potential directions of TS-PTMs for future work.
The source code is available at
\href{https://github.com/qianlima-lab/time-series-ptms}{https://github.com/qianlima-lab/time-series-ptms}.
\end{abstract}

}

\maketitle

\IEEEdisplaynontitleabstractindextext

% \fi
\IEEEpeerreviewmaketitle

\section{Introduction}\label{sec:introduction}
\IEEEPARstart{A}{s} an important research direction in the field of data mining, Time-Series Mining~(TSM) has been widely utilized in real-world applications, such as finance~\cite{wu2013dynamic}, speech analysis~\cite{moritz2020streaming}, action recognition~\cite{martinez2017human}, and traffic flow forecasting~\cite{wang2020deep, tedjopurnomo2020survey}. 
The fundamental problem of TSM is how to represent the time-series data~\cite{fu2011review}. Then, various mining tasks can be performed based on the given representations. Traditional time-series representations (e.g., shapelets~\cite{li2020efficient}) are time-consuming due to heavy reliance on domain or expert knowledge.
Therefore, it remains challenging to learn the appropriate time series representations automatically.

In recent years, deep learning models~\cite{che2018recurrent,tahan2022development,sen2019think,zhou2021informer} have achieved great success in a variety of TSM tasks.
% , such as Recurrent Neural Networks (RNNs)~\cite{che2018recurrent,kieu2019outlier}, Convolutional Neural Networks (CNNs)~\cite{liu2018time,tahan2022development}, Temporal Convolutional Networks (TCNs)~\cite{sen2019think}, and Transformers~\cite{zhou2021informer,xu2021autoformer}.
Unlike traditional machine learning methods, deep learning models do not require time-consuming feature engineering. Instead, they automatically learn time-series representations through a data-driven approach.
However, the success of deep learning models relies on the availability of massive labeled data.
In many real-world situations, it can be difficult to construct a large well-labeled dataset due to data acquisition and annotation costs.

To alleviate the reliance of deep
learning models on large datasets,
approaches 
based on data augmentation~\cite{wen2021time,iwana2021empirical} and semi-supervised learning~\cite{van2020survey,liu2023temporal}
have been commonly used.
Data augmentation can enhance the size and quality of the training
data, and has been used as an important component in many computer vision
tasks~\cite{shorten2019survey}.
However, different from image data augmentation, time-series data augmentation
also needs to consider properties 
such as temporal dependencies and multi-scale dependencies
in the time series.
Moreover, design of the time-series data augmentation techniques generally relies on expert knowledge. 
On the other hand, semi-supervised methods 
%only require a small amount of labeled data, and 
employ a large amount of unlabeled data to improve model performance. However, in
many cases,
even unlabeled time-series samples can be difficult to collect
(e.g.,  electrocardiogram time series data in
healthcare~\cite{yang2022spectral}).

%\footnote{ *** in fact, what u want to say is that the user may hv very limited labeled or unlabeled data. but the one producing the pretrained model should hv access to lots of labeled and/or unlabeled data, otherwise, the pretrained model cannot be trained. what are the possible scenarios?  {}{Answer: Yeah. Existing TS-PTMs are designed using domain-adaption [1] and meta-learning (e.g., few-shot [2] or zero-shot [3] learning) for scenarios where both labelled and unlabeled time series data are lacking. For domain adaptation, these studies [1] can use data from similar scenes as the source dataset to pre-train the encoder, thus alleviating the problem of insufficient data for the target scenes.\newline [1] Time series domain adaptation via sparse associative structure alignment. AAAI, 2021.\newline [2] Spectral Propagation Graph Network for Few-shot Time Series Classification. arXiv, 2022.\newline [3] Meta-Learning Framework with Applications to Zero-Shot Time-Series Forecasting. AAAI, 2021.  }}

Another effective solution 
to alleviate the problem of insufficient training data
is 
transfer learning~\cite{pan2009survey,zhuang2020comprehensive},
which relaxes the assumption that the training and test data must be independently
and identically distributed. Transfer learning usually has two stages: pre-training and fine-tuning. 
During pre-training,
the model is pre-trained on 
some source domains that 
contain a large amount of data,
and 
are
separate but relevant to the target domain.
On fine-tuning, the 
pre-trained model (PTM) is fine-tuned on the often limited data from the target domain.

Recently, PTMs,
particularly Transformer-based PTMs,
have 
achieved remarkable performance
in various Computer Vision
(CV)~\cite{krizhevsky2012imagenet,he2021masked} and Natural Language Processing
(NLP)~\cite{qiu2020pre} applications. 
Inspired by these, recent
studies consider the design of Time-Series Pre-Trained Models~(TS-PTMs) for
time-series data.
First,
a time-series model 
is pre-trained 
by \textit{supervised learning}~\cite{fawaz2018transfer,yang2021voice2series}, \textit{unsupervised learning}~\cite{malhotra2017timenet,zerveas2021transformer}, or
\textit{self-supervised learning}~\cite{deldari2022beyond,zhang2022self,zhang2022cross} so as
to obtain appropriate representations.
The TS-PTM is then fine-tuned on the target domain for improved
performance on the downstream TSM tasks~(e.g., time-series classification and
anomaly detection). 

Supervised TS-PTMs~\cite{fawaz2018transfer,ye2018novel} are typically
pre-trained by classification or forecasting tasks. However, the difficulty to
obtain massive labeled 
time-series 
datasets for pre-training often
limits the performance of supervised TS-PTMs. 
In addition, unsupervised TS-PTMs utilize unlabeled data for pre-training, which further tackle the limitation of insufficient labeled data. For example, the reconstruction-based TS-PTMs~\cite{malhotra2017timenet} employ auto-encoders and a reconstruction loss to pre-train time-series models.
Recently, self-supervised PTMs based on contrastive learning~\cite{he2020momentum,chen2020simple} have shown great potential in CV.
Therefore, some scholars~\cite{zerveas2021transformer,zhang2021sleeppriorcl} have started exploring the design of consistency-based tasks and pseudo-labeling techniques for mining the inherent properties of time series. 
% Supervised or unsupervised PTMs may present overfitting on downstream TSM tasks of some unique scenarios.
% To this end, domain ~\cite{zhang2021sleeppriorcl}adaption~\cite{csurka2017domain,cai2021time} and meta-learning~\cite{finn2017model,yang2022spectral} design pre-training paradigms in terms of domain-invariant and task adaptive, respectively, and gradually attract attention in the field of time series.
Nonetheless, the study of TS-PTMs remains a challenge.

% \begin{figure}[t]
%     \centering
%     \subfigure[supervised pre-training task]{
%         \begin{minipage}[t]{0.38\textwidth}
%         \centering
%         \includegraphics[width=\textwidth]{figures/taxonomy_SL.pdf}
%         \end{minipage}
%     }
    
%     \subfigure[unsupervised pre-training task]{
%         \begin{minipage}[t]{0.46\textwidth}
%         \centering
%         \includegraphics[width=\textwidth]{figures/taxonomy_UL.pdf}
%         \end{minipage}
%     }
%     \caption{Pre-training tasks for time series.}
%     \label{fig:taxonomy_intro}
% \end{figure}

\begin{figure}
	\centering 
	\includegraphics[width=0.46\textwidth]{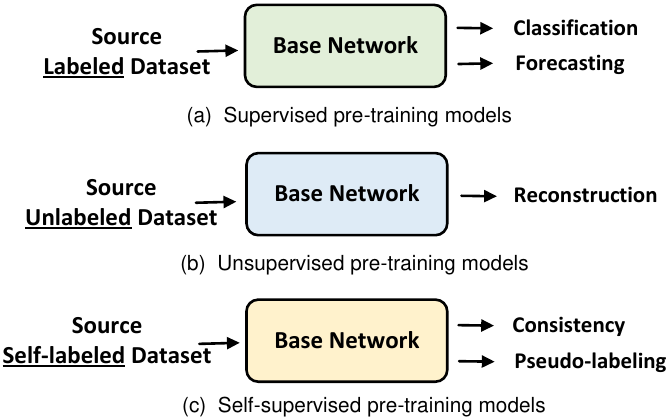}
	\caption{{}{Pre-training techniques for time series.}} 
	 \label{fig:taxonomy_intro}
\end{figure}

In this survey, we provide a comprehensive review of TS-PTMs. Specifically, we
first introduce various TSM tasks and deep learning models used in TSM. We then
propose a taxonomy of TS-PTMs 
%which systematically classifies existing TS-PTMs
based on the pre-training techniques
(Figure~\ref{fig:taxonomy_intro}). These include supervised pre-training
techniques (leading to classification-based and forecasting-based PTMs), 
unsupervised pre-training techniques (reconstruction-based PTMs) and
self-supervised pre-training techniques (consistency-based and
pseudo-labeling-based PTMs).
Note that some TS-PTMs may use multiple tasks (e.g., {}{forecasting and reconstruction in~\cite{laptev2018reconstruction}) for pre-training.
To simplify the review, we classify the 
TS-PTM
based on its core pre-training task.

{}{While existing surveys have analyzed time series pre-training~\cite{eldele2024label,zhang2024self,meng2023unsupervised,jin2023large,jin2024position,liang2024foundation}, our work was made available online before the majority of these surveys.
Our work specifically focuses on pre-training, whereas the majority of the aforementioned surveys primarily address representation learning.
In addition, the key difference is that our work conducts extensive experiments to analyze the performance of various TS-PTMs in a uniform experimental setting (e.g., consistent dataset preprocessing and Python version). Specifically, our experiments involve 27 methods, 434 datasets, and 679 sets of transfer learning.  The related experimental code and datasets have been open-sourced\footnote{\url{https://github.com/qianlima-lab/time-series-ptms}}. For a detailed comparison with existing surveys, please refer to Appendix~\ref{appendix_A}.}

% Extensive 
% experiments  on time-series classification, forecasting and anomaly detection 
% are performed 
% to study the pros and cons of various transfer learning strategies
% %, Transformer-based models, 
% and representative TS-PTMs.
% In addition, future directions of TS-PTMs are discussed. 
% This survey aims to give readers a comprehensive understanding of TS-PTMs.
% , ranging
% from the early transfer learning methods to the recent Transformer-based and consistency-based TS-PTMs. 
% on the pre-training techniques used. 
The main contributions {}{of our survey} can be summarized as follows:
\begin{itemize}
\item 
% We provide a taxonomy and comprehensive review of existing TS-PTMs from early transfer learning methods to the recent Transformer-based and consistency-based TS-PTMs. 
% Specially, we first systematically categorize existing TS-PTMs  according to supervised, unsupervised, and self-supervised pre-training techniques,  offering detailed summary for future research.
{}{We provide a taxonomy and a systematic review of existing TS-PTMs, ranging from early transfer learning methods to recent Transformer-based and consistency-based TS-PTMs. Specifically, we categorize TS-PTMs according to supervised, unsupervised, and self-supervised pre-training techniques, providing a detailed summary of each to guide future research.}
\item We perform extensive experiments 
to analyze the pros and cons of TS-PTMs. 
For time series classification, we find that transfer learning-based TS-PTMs
perform poorly on the UCR time series datasets~{}{(containing many small datasets)}, but achieve excellent performance
on other publicly-available {}{large} time series datasets.
%with similar domains.
For time series forecasting and anomaly detection, we find that {}{patch-based pre-training technique} should be the focus of future research on TS-PTMs.
\item 
{}{We present potential future directions with a detailed and thorough discussion. In particular, we analyze the limitations of current TS-PTMs and suggest future directions under (i) datasets, (ii) deep learning models, (iii) inherent properties, (iv) adversarial attacks, (v) noisy labels, and (vi) pre-trained large language models.}
% We present potential future directions with detailed and through discussion. Specially, we analyze the limitations of existing TS-PTMs and suggest potential 
% future directions under (i) datasets, (ii) deep learning models, (iii)
% inherent properties, (iv) adversarial attacks, (v) noisy labels, and (vi) large language models.
\end{itemize}

The remainder of this paper is organized as follows. Section~\ref{section2} 
provides background on the TS-PTM. A comprehensive review of the TS-PTMs is then
given 
in Section~\ref{section3}. 
Section~\ref{section4} presents
experiments on the various TS-PTMs. Section~\ref{section5} suggests some future directions. Finally, we summarize our findings in Section~\ref{section6}.

\section{Background~\label{section2}}
In this section, we first describe the TSM tasks in Section~\ref{section2.1}.
Section~\ref{section2.2} then introduces various deep learning models used in TSM. 
Finally, 
Section~\ref{section2.3}
discusses why we need to employ the PTMs.

\subsection{Time-Series Mining Tasks~\label{section2.1}}
%Before introducing various TSM tasks, we give a brief definition of time series.  Time-series usually refers to the data obtained by recording the statistical values of a certain phenomenon~(e.g., weather changes) ordered in time. Therefore,

A time series can be represented as a $T$-dimensional vector $\boldsymbol{X} =
[\boldsymbol{x}_1, \cdots, \boldsymbol{x}_t, \cdots, \boldsymbol{x}_T]$, where $T$
is the length of the time series, $\boldsymbol{x}_t \in \mathbb{R}^M$ is the value
at $t$-th time step, and $M$ is the number of variables.

\subsubsection{Time-Series Classification}
In time-series classification~\cite{ismail2019deep}, 
%one aims to train a classifier mapping from the space of possible time-series inputs to a probability distribution on categories~(labels). 
a labeled time series dataset is used to train a classifier, which can then be used to
classify unseen samples. A labeled time-series dataset with $N$ samples can be
denoted as $\mathcal{D}=\{(\boldsymbol{X}_1, \boldsymbol{y}_1), \dots,
(\boldsymbol{X}_t, \boldsymbol{y}_t), \dots, (\boldsymbol{X}_N, \boldsymbol{y}_N)
\}$, where $\boldsymbol{X}_t$ can be either a univariate or multivariate time
series, and
$\boldsymbol{y}_t$
is the corresponding one-hot label vector.

\subsubsection{Time-Series Forecasting}
Time-Series Forecasting~(TSF)~\cite{lim2021time} aims to analyze the
dynamics and correlations among historical temporal data to predict future
behavior. TSF models usually need to consider the trend and seasonal variations in
the time series, and also correlations between historical observed values. Let
$\boldsymbol{X} = [\boldsymbol{x}_1, \dots, \boldsymbol{x}_t, \dots,
\boldsymbol{x}_T]$ be the historical observations, and $H$ be the desired
forecasting horizon, the problem is to predict the future values $[\boldsymbol{x}_{T+1}, \boldsymbol{x}_{T+2}, \dots, \boldsymbol{x}_{T+H}]$.

\subsubsection{Time-Series Clustering}
{}{Time-series clustering}~\cite{lafabregue2021end} 
%is an unsupervised task. TSC 
aims to partition the time-series dataset $\mathcal{D}$ into 
a partition
$K$ clusters
%$C=
$\{c_1,\dots, c_K\}$,
such that both the similarities among
samples from the same cluster and 
the dissimilarities between samples of different clusters are
maximized. 
While clustering has been successfully used on static data,  the
clustering of time-series  data
is more difficult because of the presence of temporal and multi-scale
dependencies.
{}{Time-series clustering} helps to discover interesting patterns and enables the extraction of valuable information from massive time series datasets.

\subsubsection{Time-Series Anomaly Detection}
{}{Time-series anomaly detection}~\cite{liu2020anomaly} aims to
identify observations that significantly deviate from the other observations in
the time series. It has to learn informative representations from
the time series $\boldsymbol{X} = [\boldsymbol{x}_1, \dots, \boldsymbol{x}_t,
\dots, \boldsymbol{x}_T]$, and then derive an
anomaly score 
to determine whether a point
$\boldsymbol{x}_t$ or a subsequence $S=[x_p, \ldots, x_{p+n-1}]$~(where
$n \leq|T|$) is anomalous \cite{blazquez2021review}. 
%A higher score indicates that the point or subsequence is more likely to be an anomaly in the time series.

\subsubsection{Time Series Imputation} 
Time-Series Imputation (TSI)~\cite{cao2018brits} aims to replace
missing values in a time series with realistic values so as to facilitate TSM tasks. 
Given a time series $\boldsymbol{X} = [\boldsymbol{x}_1, \dots, \boldsymbol{x}_t,
\dots, \boldsymbol{x}_T]$  and a binary $\boldsymbol{M} = [m_1, \dots, m_t, \dots, m_T]$,
$\boldsymbol{x}_t$ is missing if $m_t = 0$, and
is observed
otherwise.
TSI imputes the missing values as:
 \[ \boldsymbol{X}_{imputed} = \boldsymbol{X} \odot \boldsymbol{M} +
 \boldsymbol{\hat{X}} \odot (1-\boldsymbol{M}), \]
% \label{eq:impute}
where $\boldsymbol{\hat{X}}$ is the predicted values generated by the TSI technique.
As a conditional generation model, TSI techniques have been
studied and applied to areas such as gene expression~\cite{de2008clustering} and healthcare~\cite{de2019deep}.

\subsubsection{{}{Time Series Extrinsic Regression}} 

{}{Time-Series Extrinsic Regression (TSER) is a task designed to learn the relationship between time series data and a continuous scalar variable~\cite{tan2021time}. The TSER model, represented as a function $\mathcal{T} \rightarrow \mathcal{R}$, is trained on a dataset $\mathcal{D}$ comprising pairs of time series $\boldsymbol{X}_t$ and corresponding scalar values $\boldsymbol{r}_t$. Unlike time series classification, which predicts categorical labels, TSER produces numerical outputs~\cite{xu2023reinforced}. TSER contrasts with traditional TSF tasks by focusing on the association between time series and an external variable sequence~\cite{li2023difformer}. For instance, in smart city applications, TSER can integrate various sensor readings (e.g., temperature, humidity, rain, voltage) to predict a continuous value such as power consumption~\cite{tan2020monash}.}

\begin{figure}
	\centering 
	\includegraphics[width=0.46\textwidth]{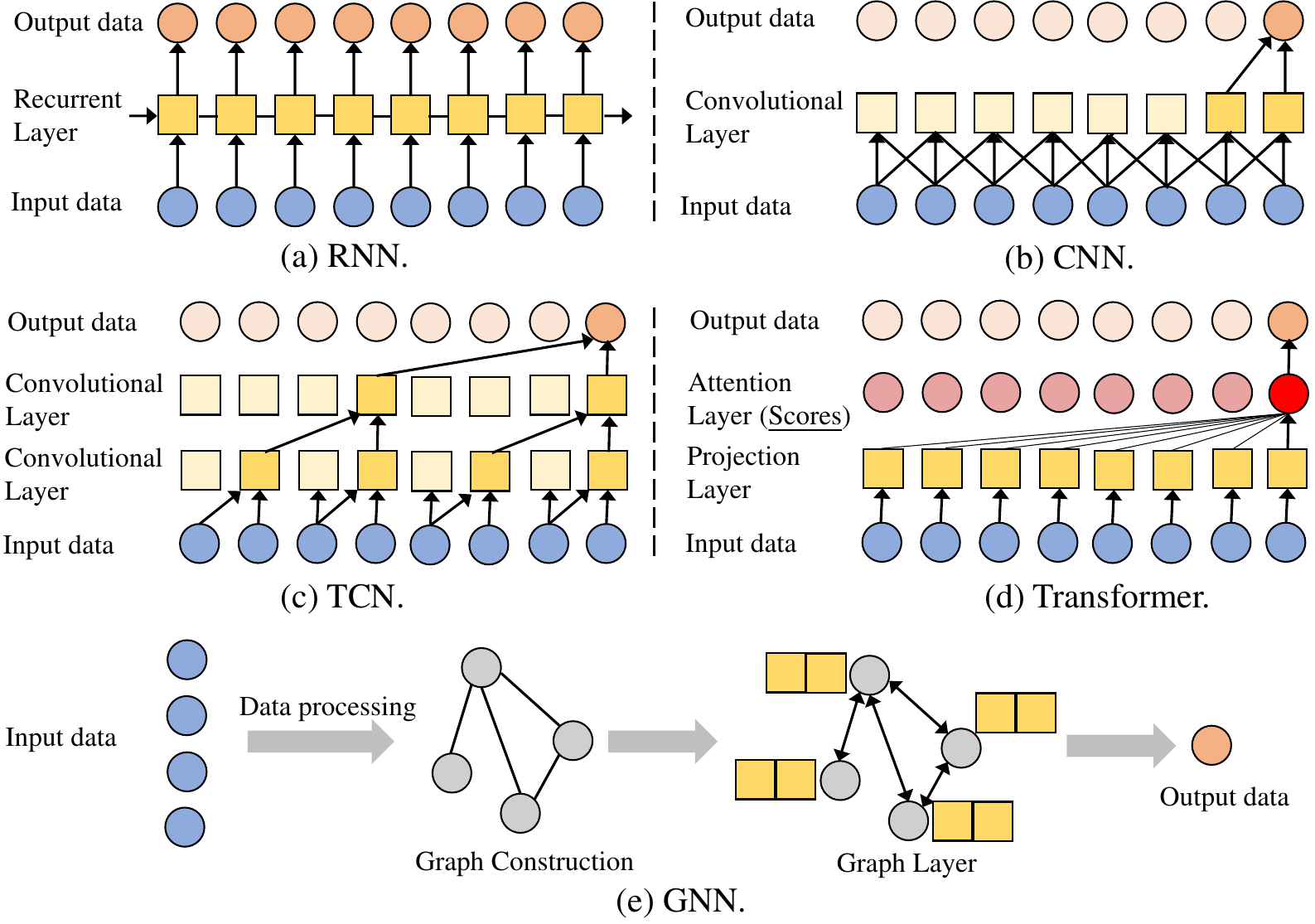}
	\caption{{}{Deep learning models used for time-series mining.}} 
	 \label{fig:DLModel}
\end{figure}

\subsection{Deep Learning Models for Time Series~\label{section2.2}}
% Here, we review deep learning models applied to time series, including RNNs, CNNs, TCNs, and Transformers.

\subsubsection{Recurrent Neural Networks}
{}{Recurrent Neural Networks~(RNNs)~\cite{chung2014empirical} usually consist} of an input layer, one or more
recurrent hidden layers, and an output layer
(Fig.~\ref{fig:DLModel}~(a)). In the past decade, RNNs and their variants (such as the long short-term
memory network~\cite{greff2016lstm} and gated recurrent
units~\cite{chung2014empirical}) have achieved remarkable success in
TSM.
%since they can model temporal dependencies. 
For example, Muralidhar et
al.~\cite{muralidhar2019dyat} combined dynamic attention and a 
RNN-based sequence-to-sequence model
for time-series forecasting. 
Ma et al.~\cite{ma2019learning} employed a multi-layer dilated RNN to extract multi-scale temporal dependencies for time-series clustering.

\subsubsection{Convolutional Neural Networks}
Convolutional Neural Networks~(CNNs)~\cite{gu2018recent} are originally designed for
computer vision tasks.
%since they provide an architectural prior for image processing tasks. 
A typical CNN is shown in Fig.~\ref{fig:DLModel}~(b).
To use CNNs for TSM, the data need to be first encoded in an image-like format.
%before feeding into the network. 
The CNN receives embedding of the value at each time step and then aggregates local
information from nearby time steps using convolution. CNNs have been shown to be very
effective for TSM~\cite{liu2022scinet}.
%due to their ability of learning complex feature representations. 
For example, the multi-scale convolutional neural network~\cite{cui2016multi} can automatically extract features at different scales by a multi-branch layer and convolutional layers. Kashiparekh et al.~\cite{kashiparekh2019convtimenet} incorporated filters of multiple lengths in all convolutional layers to capture multi-scale temporal features for time-series classification.

Unlike vanilla  CNNs, Temporal \label{TCN} Convolutional
Networks~(TCNs)~\cite{bai2018empirical} use a fully convolutional
network~\cite{long2015fully} so that all the layers are of the same length,
and employ causal convolutions with no information ``leakage" from
future to past. A typical TCNs is shown in
Fig.~\ref{fig:DLModel}~(c). 
Compared to recurrent networks,
TCNs have
recently
shown to be more accurate,
%such as LSTMs and GRUs but 
simpler, and more
efficient across a diverse range of sequence modeling
tasks~\cite{chen2020probabilistic}. For example, Sen et al.~\cite{sen2019think}
combined a local temporal network and a global matrix factorization model
regularized by a TCN for time-series forecasting.

\subsubsection{Transformers}

Transformers 
\cite{xu2021autoformer,zhou2022fedformer}
integrate information from data points in the time
series by
dynamically computing
the associations between representations 
with self-attention.
A typical Transformer is shown in Fig.~\ref{fig:DLModel}~(d). Transformers have
shown great power in TSM due to their powerful capacity to model
long-range dependencies. For example, Zhou et al.~\cite{zhou2021informer} combined
a self-attention mechanism (with 
$\mathcal{O}(L \log L)$
time and space complexities)
%an operation that highlights dominant attention,
and a generative decoder for long time-series forecasting.

\subsubsection{{}{Graph Neural Networks}}
{}{
Graph Neural Networks (GNNs) are highly effective at processing graph data consisting of nodes and edges~\cite{wu2020comprehensive}. Recently, some scholars have converted time series data into graphs by examining both intra-sample and inter-sample relationships and applying GNNs to TSM~(Fig.~\ref{fig:DLModel}~(e)). Intra-sample relationship approaches construct graphs by analyzing relationships between subsequences within a sequence~\cite{cheng2021time2graph} or between different variables~\cite{wang2024graph}. For inter-sample relationships, graphs are constructed based on sample similarities or category information~\cite{zha2022towards}.}

% model inter-temporal and inter-variable relationships
% 图神经网络擅长于处理具有节点和边的图数据。近年来，相关学者从样本内和样本间的关系将时间序列数据转换为图，然后将图网络用于TSM。在基于样本内关系的方法中，一方面可采用序列内的子序列间关系进行图构建，另一方面可采用不同变量间关系进行图构建。在样本间的关系，可依据样本间的相似度或者类别信息进行图构建。

\begin{figure*}
	\centering 
	\includegraphics[ width=1.0\textwidth]{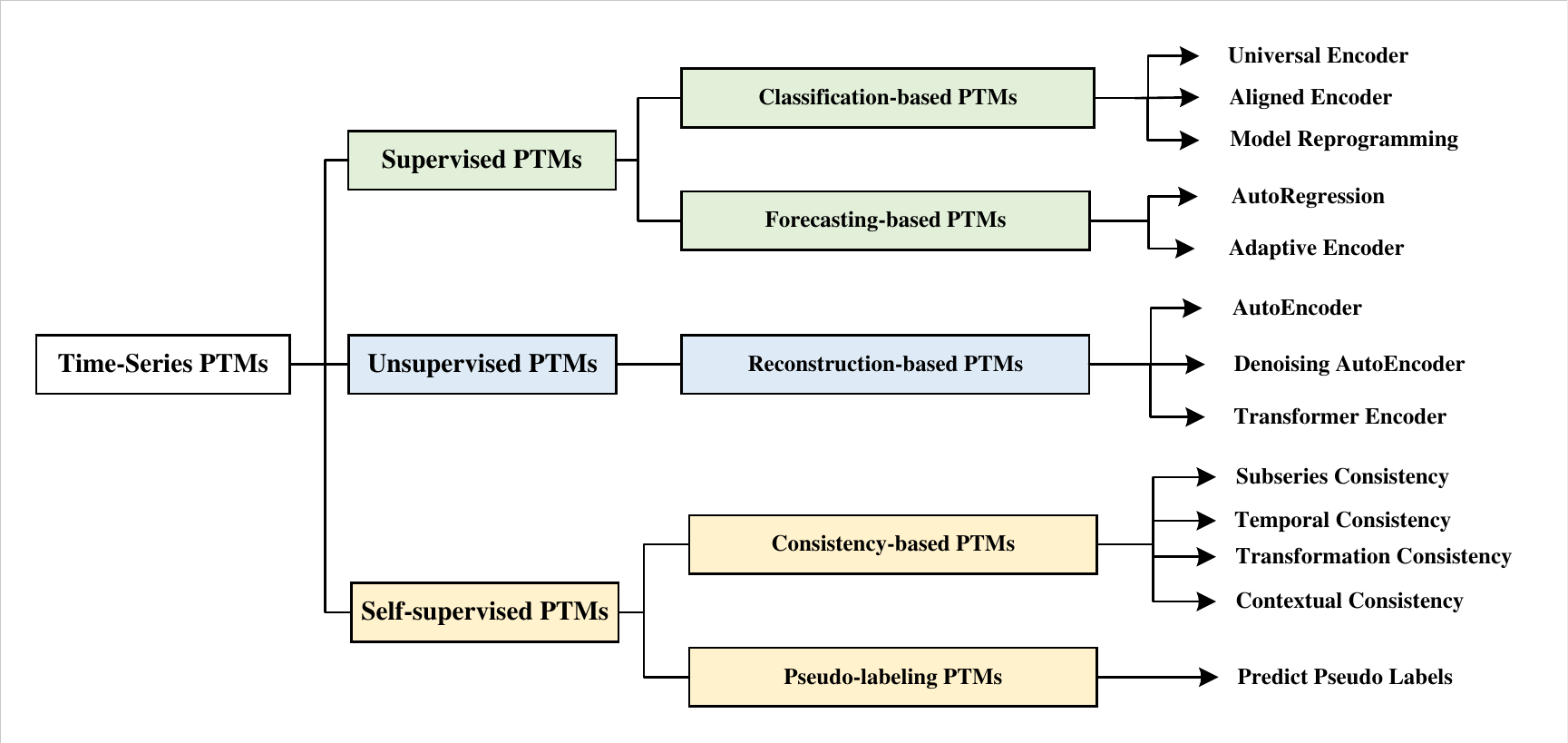}
	\caption{The taxonomy of Pre-Trained Models for time-series mining.
	} 
	\label{fig:taxonomy}
\end{figure*}

\subsection{Why Pre-Trained Models?~\label{section2.3}}
With the rapid development of deep learning, deep learning-based TSM has received
more and more attention. In recent years, deep learning models have been widely
used in TSM and have achieved great success. However, 
as data acquisition and annotation can be expensive,
the limited labeled time-series data often hinders sufficient training of deep
learning models.
{}{For example, in bioinformatics, time series classification involves assigning each sample to a specific category, such as clinical diagnosis results. Nonetheless, obtaining accurate labels for all samples is a challenge as it requires expert knowledge to classify the samples. Therefore, pre-training strategies have
been proposed to alleviate this data sparseness problem. The advantages of Pre-Trained Models (PTMs) for TSM can be summarized as follows:

\begin{itemize}
    \item PTMs provide better model initialization for the downstream TSM tasks, which generally results in better generalization performance.
    \item PTMs can automatically obtain appropriate time series representations by pre-training on source datasets, thus avoiding over-reliance on expert knowledge.
  %   \item PTMs can effectively alleviate over-fitting of deep learning models on
	 % small-scale time-series datasets.
 
\end{itemize}

\section{Overview of TS-PTMs~\label{section3}}
% Pre-training is an effective strategy to solve the problem of insufficient labeled data. 
% Recently, inspired by PTMs of CV and NLP, some scholars have begun to explore PTMs for Time-Series Mining (TSM). 
% The main difference between these studies is the pre-training task. 
In this section, we propose a new taxonomy of TS-PTMs, which systematically classifies existing TS-PTMs based on pre-training techniques. 
% Specifically, the TS-PTMs can be mainly divided into two categories. The first category is supervised PTMs, including classification-based and forecasting-based PTMs. 
% % In this category, the models are pre-trained using classification and prediction tasks, respectively. 
% The second category is unsupervised PTMs, including reconstruction-based, denoising-based, consistency-based, and othe-based PTMs. In addition, as an extension, we discuss domain-in and meta-learning.
The taxonomy of TS-PTMs is shown in Fig.~\ref{fig:taxonomy}, and please refer to Appendix~\ref{appendix_B1} for a literature summary of TS-PTMs. 

% The summary of TS-PTMs is shown in Table~\ref{tab:summary}.

\subsection{Supervised PTMs}
The early TS-PTMs are inspired by transfer learning applications in CV. Many
vision-based PTMs are trained on large labeled datasets such as the
ImageNet~\cite{deng2009imagenet}. The corresponding weights are then fine-tuned on the
target dataset, which is usually small. This strategy has been shown to improve the
generalization performance of deep learning models on many CV tasks. Naturally,
some 
also investigated whether this strategy is effective in the time-series domain
\cite{serra2018towards,fawaz2018transfer}.
Their experiments on the UCR time series datasets~\cite{UCRArchive} show that
transfer learning may improve or degrade downstream task performance, depending on
whether the source and target datasets are similar 
\cite{fawaz2018transfer}.

%Moreover, they find that a more similar source and target time series dataset may improve the transfer learning performance, which has been successfully applied to other domain data.

\subsubsection{Classification-based PTMs}

Time-series classification is the most common supervised learning task in TSM.
The loss function
is usually the cross-entropy,
which is defined as:
    \[ \mathcal{L}_{classification} = - \frac{1}{N} \sum_{i=1}^{N} \sum_{j=1}^{C}
	 y_{ij} \log(p_{ij}). \]
Here, $\boldsymbol{y}_i=[
	 y_{ij}]$
and $\boldsymbol{p}_i=
	 [p_{ij}]$
are the ground-truth label vector and
predicted label vector of the $i$-th input time series, respectively, $N$ is the number
of samples and $C$ is the number of categories. Recently, the CNN has also been used
for time-series classification \cite{wang2017time,ismail2020inceptiontime,tang2021omni},
and achieved better performance than traditional machine learning methods (such as
the nearest neighbor classifier with dynamic time warping distance). However, deep
learning models are prone to over-fitting on small-scale time-series datasets.

To prevent the over-fitting problem 
when training a new model from scratch
on the target dataset,
there have been attempts to
employ classification-based PTMs that pre-train a
classification base model on some labeled source datasets
\cite{fawaz2018transfer,li2020deep,yang2021voice2series}.
Existing
classification-based PTMs can be divided into three categories: (i) universal
encoder, (ii) model reprogramming, and (iii) aligned encoder.

\vspace{1ex}
\noindent \textbf{Universal Encoder} {\;}
For TS-PTMs,
a key issue 
is how to learn universal time-series representations~\cite{yue2022ts2vec} that can benefit a variety
of downstream TSM tasks. A common approach is to design a universal encoder 
%suitable for time-series data, 
which can be quickly adapted to new tasks with or without fine-tuning. 
Serr{\`a} et al.~\cite{serra2018towards} proposed a universal encoder that
combines CNN with attention mechanism, and pre-trained the encoder using the
supervised classification task. The encoder is jointly pre-trained on multiple
time series datasets. 
Using multi-head outputs, 
each dataset 
has an extra fully-connected layer for classification. In this way, the encoder
acts as a carrier to transfer knowledge from multiple related source domains to enhance learning in the target domain. Fig.~\ref{fig:Encoder} shows the schematic diagram of the universal encoder.

\begin{figure} 
	\centering 
	\includegraphics[width=0.48\textwidth]{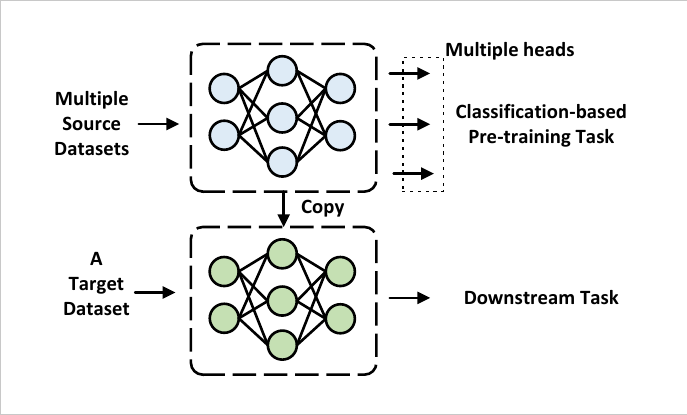}
\caption{Universal encoder aims to learn general time-series representations
through pre-training on various source datasets. The universal encoder is then fine-tuned on the target dataset for downstream TSM tasks.} 
	\label{fig:Encoder}
\end{figure}

Fawaz et al.~\cite{fawaz2018transfer} considered transfer learning on time-series
data from the shapelet perspective.
Shapelets~\cite{ye2009time} are discriminative subsequences
in the time series and can be used as an efficient time-series representation. 
Fawaz et al.~\cite{fawaz2018transfer}
hypothesized that the learned shapelets can generalize to unseen datasets via
transfer learning.
For each dataset in the UCR archive~\cite{UCRArchive}, 
they trained a fully-convolutional network~\cite{wang2017time} 
and then fine-tuned it on all the other datasets.
They found that pre-training can degrade (negative transfer) or improve (positive
transfer) the encoder's performance on the target dataset,
%Also, their experimental results indicated that 
and the likelihood of positive transfer is greater when the source dataset is similar to the target dataset. 
Due to privacy and annotation issues, it may be difficult to obtain source
datasets that are very similar to the target dataset. To alleviate this problem,
Meiseles et al.~\cite{meiseles2020source} utilized the clustering property between
latent encoding space categories as an indicator to select the best source dataset.
The above studies mainly focus on univariate time series. Li et
al.~\cite{li2020deep} proposed a general architecture that can be used for
transfer learning
on multivariate time series.

The aforementioned works employ CNN as backbone for time series transfer learning.
%which can effectively extract time-series representations beneficial to the classification task.
However, vanilla CNNs have difficulty in capturing multi-scale information and
long-term dependencies in the time series.
Studies~\cite{ cui2016multi,tang2021omni} have shown that using
different time scales or using LSTM with vanilla CNNs can further improve classification performance.
For example, Kashiparekh et al.~\cite{kashiparekh2019convtimenet} proposed a novel
pre-training deep CNN, in which 1-D convolutional filters of multiple lengths are
used to capture features at different time scales. Mutegeki et
al.~\cite{mutegeki2019feature} used CNN-LSTM as the base network to explore how
transfer learning can improve the performance of time-series classification with few labeled time series samples.
%Therefore, an important focus is on designing encoders that can fully exploit the inherent properties of time series data.

%\footnote{***  seems that some of these methods perform more than fine-tuning. they require all the source domain data and the source domain is trained together w\/ the target domain. in this sense, this doesnt really count as pretrained model?  {}{Answer: The time series domain adaption-based method transfers the knowledge from the source domain to the target domain [1]. That’s to say, domain adaption can be viewed as a sub-area of transfer learning which is the core of PTMs. In addition, the authors [2] write that domain adaptation is a rising field in transfer learning since it does not require labels in target domains.\newline [1] Time Series Domain Adaptation via Sparse Associative Structure Alignment. AAAI, 2021.\newline [2] Adversarial Spectral Kernel Matching for Unsupervised Time Series Domain Adaptation. IJCAI, 2021.  } } 

\begin{figure} 
	\centering 
	\includegraphics[width=3.4in,height=0.32\textwidth]{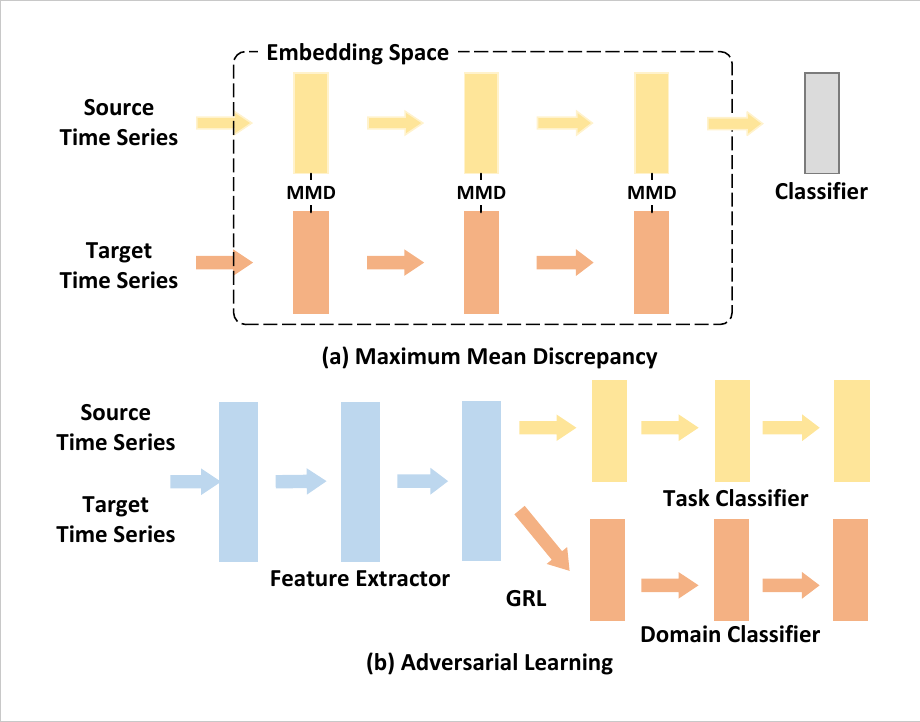}
	\caption{Aligned encoder aims to learn domain-invariant representations.} 
	\label{fig:domain_adaptation}
\end{figure}

\vspace{1ex}
\noindent \textbf{Aligned Encoder} {\;}
A universal encoder first pre-trains the model with the source dataset,
%via classification training, 
and then fine-tunes the model using the target dataset.
However, 
the difference between 
the source and target data
distributions 
is often not considered.
To address this issue, some recent works~\cite{cai2021time,wilson2020multi} first
map the source
and target datasets 
%(with different distributions) 
to a shared feature representation space, and then prompt the model to learn 
domain-invariant representations during pre-training. 
%thus obtaining an aligned pre-trained encoder.
%from the source and target datasets.
The pre-training strategy of the aligned encoder is at the core of domain
adaptation, and has been extensively validated studied on image data~\cite{csurka2017domain}.
For time series data,
extraction of domain-invariant representations 
is difficult
due to distribution shifts among timestamps and the associative structure
among variables. To this end,
existing 
pre-training techniques 
for time series aligned encoder 
are based on either
Maximum Mean Discrepancy~(MMD)~\cite{chen2020homm} or adversarial
learning~\cite{xie2018learning}.

MMD~\cite{yan2017mind} is a standard metric on distributions, and has
been employed to measure the dissimilarity of two distributions in domain
adaptation~\cite{you2019universal}. 
Given a representation $f(\cdot)$ on source data $\boldsymbol{X}_s \in \mathcal{D}_s$ and target data $\boldsymbol{X}_t \in \mathcal{D}_t$, the empirical approximation of MMD is:
%\resizebox{.85\hsize}{!} {
\[ \text{MMD}(\mathcal{D}_s, \mathcal{D}_t) =
\left\| \frac{1}{|\mathcal{D}_s|} \!\!\! \sum_{\boldsymbol{X}_s \in
	 \mathcal{D}_s} \!\!\!\! f(\boldsymbol{X}_s) - \frac{1}{|\mathcal{D}_t|}
\!\!\! 	 \sum_{\boldsymbol{X}_t \in \mathcal{D}_t} \!\!\!\!f(\boldsymbol{X}_t)
\right\|.  \]

MMD-based methods~\cite{wang2018stratified,li2021transferable,liu2021adversarial} learn
domain-invariant representations by minimizing the MMD between the source and
target domains in classification training (Fig.~\ref{fig:domain_adaptation}~(a)).
Khan et al.~\cite{khan2018scaling} used a CNN to extract features from the source and target domain data separately. The divergence of the source and target domains is reduced by minimizing the Kullback–Leibler divergence in each layer of the network. Wang et al.~\cite{wang2018stratified} proposed stratified transfer learning to improve the accuracy for cross-domain activity recognition. 
% Specifically, STL first employs the majority voting technique to obtain pseudo labels for the target domain. Then, an intra-class MMD distance is proposed to perform intra-class knowledge transfer. Finally, the labels of the target domain data can be obtained through the second annotation. 
Moreover, considering that time lags or offsets can influence the 
extraction of domain-invariant features,
Cai et al.~\cite{cai2021time} designed a sparse associative structure
alignment model which assumes that the causal structures are stable across domains. 
Li et al.~\cite{li2021transferable} considered the compact causal mechanisms among
variables and the variant strength of association,
and used the Granger
causality alignment model~\cite{tank2021neural} to discover the data's causal structure.
Ragab et al.~\cite{ragab2021self} proposed an autoregressive domain discriminator to explicitly address the temporal dependencies during both representation learning and domain alignment.
% Liu et al.~\cite{liu2021adversarial} minimized domain divergence on a reinforced MMD metric, which is embedded by a hybrid spectral kernel network in time-series distribution.
Despite all these advances, the use of MMD-based methods on time series
data is still challenging due to the underlying complex dynamics.

Another common approach is to learn a domain-invariant representation between the
source and target domains through adversarial learning~\cite{da2020remaining,wilson2021calda}. For example, 
Wilson et al.~\cite{wilson2020multi} proposed the Convolutional deep Domain
Adaptation model for Time Series data~(CoDATS), which consists of a feature
extractor, Gradient Reversal Layer~(GRL), task classifier, and domain classifier
(Fig.~\ref{fig:domain_adaptation}~(b)). The adversarial step is performed by the GRL placed 
between the feature extractor and domain classifier
in the network.
CoDATS first updates the feature extractor and task classifier to classify the
labeled source data. The domain classifier is then updated to distinguish which
domain each sample comes from. At the same time, the feature extractor is 
adversarially 
updated
to make it more difficult for the domain classifier to distinguish which domain each
sample comes from.
Li et al.~\cite{li2021causal} argued that temporal causal mechanisms should be
considered and proposed a time-series causal mechanism transfer network to obtain
a domain-invariant representation.
%through adversarial learning. 
%The above studies indicate the applicability of adversarial training to TS-PTMs.
However, the exploitation of inherent properties in the time series
(such as multi-scale and frequency properties)
still need to be explored
in adversarial training.

\vspace{1ex}
\noindent \textbf{Model Reprogramming} {\;}
The great success of PTMs in CV~\cite{he2020momentum} and NLP~\cite{qiu2020pre}
shows that using a large-scale labeled dataset can significantly benefit the downstream tasks.
However, most time-series datasets are not large. 
Recently, a novel TS-PTM called Voice2Series~\cite{yang2021voice2series}
was proposed for 
time-series classification. Voice2Series 
is based on model reprogramming~\cite{elsayed2018adversarial}
(Fig.~\ref{fig:Reprogram}).
It uses a large acoustic model pre-trained on massive human voice datasets~(e.g.,
spoken-term recognition). The voice data can be considered as univariate time
series, and therefore enormous voice data can be employed to pre-train an acoustic
model. 
To make the acoustic model suitable for general time-series data,
Voice2Series reprogrammed the model through input transformation learning and output label mapping.
The input transformation of a time-series sample
$\boldsymbol{X}$ is defined as:
\[ \boldsymbol{X}'=
%\mathcal{H}(\boldsymbol{X};\boldsymbol{\theta})=
{\rm
Pad}(\boldsymbol{X}) + \boldsymbol{M} \odot \boldsymbol{\theta}, \]
where ${\rm Pad}(\cdot)$ 
is a zero-padding function, 
$\boldsymbol{M}$ is
a binary mask, and $\boldsymbol{\theta}$ are reprogramming parameters for
aligning the data distributions of the source and target domains. 
A random (but non-overlapping) many-to-one
mapping between source and target labels is used as the output label mapping.
%since the authors~\cite{yang2021voice2series} find that it performs better than one-to-one label mapping for improving the classification accuracy.
Transformer-based attention mechanism~\cite{yang2021decentralizing}
is used as the acoustic model.
%in Voice2Series. 
Note that model reprogramming not only uses more labeled training samples during
pre-training, but also considers adapts the PTMs to the target tasks by constructing relationships between the source and target domain data.
% , which improves the performance on the target tasks.

\begin{figure}
	\centering 
	\includegraphics[width=0.43\textwidth]{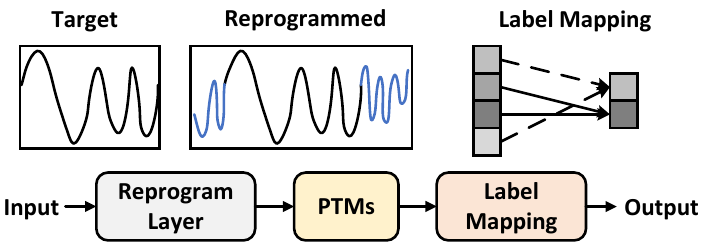}
	\caption{Model reprogramming can adapt PTMs to a new task by reprogramming the time-series from the target domain and label mapping.} 
	\label{fig:Reprogram}
\end{figure}
% jintime

\vspace{1ex}
\noindent \textbf{Summary} {\;}
The universal encoder first pre-trains a base network on the labeled source
datasets, and then the base network is transferred to the target domain. This 
usually requires a large amout of labeled source samples for pre-training, and can
be difficult to obtain in the time-series domain.
Positive (resp. negative)
transfer often occurs when the source and target datasets are similar (resp.
dissimilar).
Previous studies have explored how to select the source based on inter-dataset similarity or time series representations in the latent representation space.
In addition, the aligned encoder based on domain adaption considers the 
differences between source and target data
distributions.
%thus providing an optional direction for the study of scenario-specific TS-PTMs.
Voice2Serie~\cite{yang2021voice2series} provides a new approach for
classification-based PTMs. Some domain-specific time series data (e.g., voice
data) is used to pre-train a base network, which is then applied to general time series through model reprogramming. 
{}{Similarly, building on the success of pre-trained Large Language Models (LLMs) in time series modeling~\cite{zhou2023one,caotempo}, Time-LLM~\cite{jintime} converts time series data into text-like formats with patching strategy and use model reprogramming to transfer LLM knowledge to time series tasks.}
Nevertheless, how to construct a large-scale well-labeled time series dataset suitable for TS-PTMs {}{remains an open challenge.}

\subsubsection{Forecasting-based PTMs}
Time Series Forecasting~(TSF) aims to estimate the values at the future timesteps
using observed values from the current and past timesteps. The problem of one-step-ahead forecasting is defined as:
\[ \hat{y}_{i, t+1}=f(y_{i, t-k: t}, \boldsymbol{x}_{i, t-k: t}),\]
where $\hat{y}_{i, t+1}$ is the model's predicted value of the $i$th sample at time $t+1$, 
$y_{i, t-k: t}=\{y_{i, t-k}, \dots, y_{i, t}\}$ and $\boldsymbol{x}_{i, t-k: t}=\left\{\boldsymbol{x}_{i, t-k}, \ldots, \boldsymbol{x}_{i, t}\right\}
$
are the observed values and exogenous inputs, respectively, over a look-back
window of length $k$, and $f(\cdot)$ is the prediction function learnt by the model~\cite{lim2021time}. 
Unlike the classification task that employs manual labels as supervisory
information, TSF utilizes the observed value in the future as supervisory
information. In addition, the mean absolute error or mean squared
error is often adopted as the loss function for the TSF task.

A unique property of time-series data is the presence of temporal dependencies.
{}{Forecasting can utilize a time series of past and present values to estimate future values, and it is naturally employed as a time series pre-training task.}
An intuitive approach to produce forecast values is the recursive strategy, which
can be achieved by autoregression. In general, a PTM first pre-trains a forecasting-based model on the source dataset. The weights of the base model are then fine-tuned on the target dataset.

%\footnote{***  u’re suggesting that AutoRegression is the only approach for timeseries forecasting. however, this may not be the case?. the trans- former u mentioned at the end of this sec can be non-ar {}{Answer: That's a good question. For recent hot studies, transformer-based (i.e., Informer, AutoFormer, and Fedforemer) methods are non-AutoRegression models for time series forecasting. However, we aim to discuss TS-PTMs, not focus on time series forecasting without pre-training. Further, existing transformer-based models have not been used for time series pre-training. On the contrary, the key idea of AutoRegression-based models (such as RNN-based models predicting the next values via auto-regression, refer to Fig.7) for time series forecasting has been explored for TS-PTMs.} }
%\footnote{*** also, the desc in this part does not really rely on autoregression?  {}{Answer: Auto-Regression is a strategy that uses regression iterations to predict the future value of the next time step, thus capturing the time-series dependencies.  The autoregressive process can be implemented using the classical RNN-based model. However, some works use Transformer-based to predict the values of multiple future time steps (non-auto-regression) but will use the cross-autoregressive relationships between multiple time steps obtained from two different perspective Transformer models to capture the time series dependence properties.  } } 

\vspace{1ex}
\noindent \textbf{AutoRegression} {\;}
Deep learning models,
including RNNs~\cite{muralidhar2019dyat}, TCNs~\cite{jiangsequential} and the more recent Transformers~\cite{zhou2022fedformer},
have been used for TSF.
The RNN-based forecasting architecture is shown in Fig.~\ref{fig:prediction}. In
this 
example, given the historical observations
$[\boldsymbol{x}_1,\boldsymbol{x}_2,\boldsymbol{x}_3,\boldsymbol{x}_4,\boldsymbol{x}_5]$,
TSF tries to predict the future values
$[\boldsymbol{x}_6,\boldsymbol{x}_7,\boldsymbol{x}_8]$. Thus, TSF does not require
manual labels, but uses the values at future timesteps for supervision.

\begin{figure}
	\centering 
	\includegraphics[width=3.2in,height=0.13\textwidth]{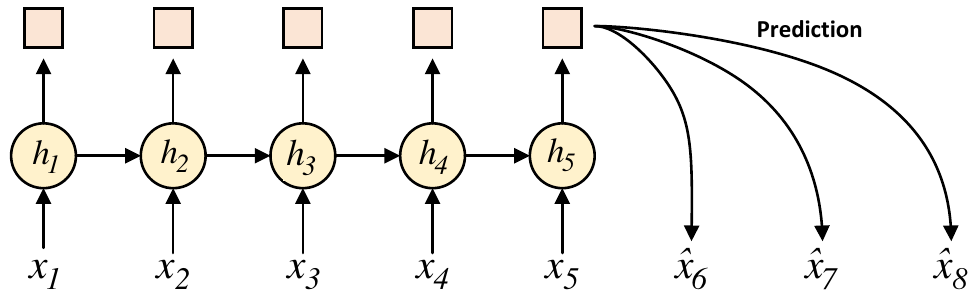}
	\caption{The RNN-based forecasting architecture.} 
	\label{fig:prediction}
\end{figure}

The dynamic nature 
and inherent temporal dependencies 
of the time series 
enable forecasting to assist {}{TS-PTMs in obtaining robust representations 
%downstream tasks~
\cite{liu2024timer,woo2024unified,dasdecoder}.}
{}{For instance, Mallick et al.~\cite{mallick2021transfer} developed a GNN-based model for pre-training short-term highway network forecasting models, employing a transfer learning strategy operation within a RNN.
In addition,} Du et al.~\cite{du2021adarnn} proposed Adaptive RNNs~(AdaRNN) to
solve the temporal covariate shift problem.
% {that temporal distribution can vary with the statistical properties of the time series.} 
The AdaRNN model consists of Temporal Distribution Characterization~(TDC) and Temporal Distribution Matching~(TDM) modules.
The TDC divides the training time series into least-similar
{}{$K$ different} subsequences, which fully characterize the distribution information of the time series in each period. 
The {}{$K$ different} subsequences are then used as source data to pre-train a generalized RNN model using the forecasting task. 
The TDM module can also be used with the Transformer architecture, further
improving the TSF performance.

% Some recent
% studies~\cite{oord2018representation,schneider2019wav2vec,eldele2021time} applied
% contrastive learning\footnote{*** ref} to TSF.\footnote{*** this is not related to
% autoregression?} 
Some recent
studies~\cite{oord2018representation,schneider2019wav2vec,eldele2021time} combines the autoregressive forecasting strategy of TSF with contrastive learning~\cite{chen2020simple} to obtain time series representations favorable for downstream tasks.
For example, Oord et al.~\cite{oord2018representation} proposed 
%a representation learning method called 
contrastive predictive coding by employing model-predicted timesteps as positive samples and randomly-sampled timesteps as negative samples.
Eldele et al.~\cite{eldele2021time} used the autoregressive model's predicted
values as positive sample pairs for contrastive learning, thus enabling the model
to capture temporal dependencies of the time series. In particular, they fed both
strong and weak augmentation samples to the encoder, therefore using a cross-view
TSF task as the PTM objective.

\vspace{1ex}
\noindent \textbf{Adaptive Encoder} {\;}
{}{Unlike transfer learning which focuses on the current learning ability of the model, meta-learning~\cite{vinyals2016matching,jiangsequential} focuses on the future learning potential of the model to obtain an \textit{adaptive encoder} via task-adaptive pre-training paradigm~(as shown in Fig.~\ref{fig:meta-learning}).
Especially, transfer learning-based PTMs are prone to overfitting on downstream tasks when the number of samples in the target dataset is small.}
Inspired by the fact that humans are good at learning a priori knowledge from a
few new samples, 
task-adaptive pre-training, or
meta-learning ,
has also been used.
Fig.~\ref{fig:meta-learning} shows an adaptive encoder via a classic task-adaptive
pre-training algorithm called model agnostic meta-learning~\cite{finn2017model,lu2022spatio,wang2022meta}.
Recently, task-adaptive pre-training strategies have been receiving increasing attention in the field of time series.
Existing studies focus on how to perform cross-task learning using properties such
as temporal and multivariate dependencies in the time series.

\begin{figure}
	\centering 
	\includegraphics[width=2.6in,height=0.14\textwidth]{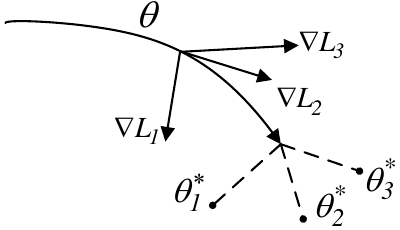}
	\caption{Adaptive encoder aims to obtain a better initialization model parameters $\boldsymbol{\theta}$ so that the model can quickly generalize to new tasks using only a small number of samples~(e.g., $\boldsymbol{\theta}^{*}_{1}$), where  $\boldsymbol{\theta}^{*}_{i} (i \in {1,2,3})$ denote task adaptive parameters obtained using gradient descent on $\boldsymbol{\theta}$ (e.g., $\nabla L_1$) based on the task adaptive data.} 
	\label{fig:meta-learning}
\end{figure}

A task-adaptive pre-training paradigm based on time-series forecasting has been applied to TS-PTMs~\cite{iwata2020few,oreshkin2021meta}. For example, Oreshkin et al.~\cite{oreshkin2021meta} proposed a meta-learning framework based on deep stacking of fully-connected networks for zero-shot time-series forecasting. 
The meta-learning procedure consists of a meta-initialization function, an update
function, and a meta-learner. The meta-initialization function defines the
initial predictor parameters for {}{a given task}.
The update function 
then
iteratively updates the predictor parameters based on the previous value of the predictor parameters and the given task. The final predictor parameters are obtained by performing a Bayesian ensemble~(or weighted sum) on the whole sequence of parameters. 
The meta-learner learns shared knowledge across tasks by training on different tasks, thus providing suitable initial predictor parameters for new tasks. 
Brinkmeyer and Rego~\cite{brinkmeyer2022few} developed a few-shot multivariate
time series forecasting model working across tasks with heterogeneous channels.
Experimental results showed that it provides a good generalization for the target datasets.
Also,
Autoformer~\cite{xu2021autoformer} and FEDformer~\cite{zhou2022fedformer} show that frequency domain information and Transformer architecture can improve time-series forecasting performance.
Hence, it would be an interesting direction to explore task adaptive Transformer-based models for TS-PTMs.

%\footnote{***  ts forecasting does not necessarily involve data augmentation; and data augmentation can also be used in ts classification {}{Answer: That's a good question. Recently, some works have begun to use time series representations for ts forecasting. To learn better representations for ts forecasting, CoST [1] and TS2Vec [2] have adopted  data augmentation combined with contrastive learning to pretrain the encoder. In addition, data augmentation can be used for ts classification, and we have added a discussion about data augmentation in ts classification.  Overall, data augmentation is an efficient technology for PTMs (i.e., contrastive learning-based PTMs in CV and NLP), which can also be used for various downstream ts tasks (i.e., TS2Vec [2]).  \newline [1] CoST: contrastive learning of disentangled seasonal-trend representations for time series forecasting. ICLR, 2022.\newline [2] Ts2vec: Towards universal representation of time series. AAAI, 2022.  } } 

\vspace{1ex}
\noindent \textbf{Summary} {\;}
TSF-based PTMs can exploit the complex dynamics in the time series and use that to
guide the model in capturing temporal dependencies.
Autoregression-based models use the dependencies between subseries and the
consistency of future predicted values of the same time series, thus pre-training the time series data using TSF.
Unlike classification-based PTMs that use manual labels for pre-training,
avoiding sampling bias
among subseries (e.g., outliers)~\cite{yue2022ts2vec} 
for pre-training based on the TSF task remain challenging.
Meanwhile, the adaptive encoder based on meta-learning allows
%provides an optional pre-training strategy 
for scenarios with small time series samples in the target dataset.
In addition, regression-based one-step forecasting models~(e.g., RNNs) potentially
lead to bad performance due to accumulated error~\cite{che2018recurrent}. Instead, some studies~\cite{zhou2021informer, xu2021autoformer} employ Transformer-based models to generate all predictions in one forward operation. 
Therefore, designing efficient TSF encoders would be a basis for studying TSF-based PTMs.

\subsection{Unsupervised PTMs}
This section introduces unsupervised TS-PTMs,
which are often pre-trained by reconstruction techniques. 
Compared with supervised TS-PTMs, unsupervised TS-PTMs are more widely applicable since they do not require labeled samples. 
%Next, we present unsupervised TS-PTMs.

\subsubsection{Reconstruction-based PTMs}
Reconstruction is a common unsupervised task, and is usually implemented by
an encoder-decoder architecture
\cite{ma2019learning}. 
The encoder maps the
original time series to a latent space of representations, which is 
then used 
by the decoder
to reconstruct the input time series. The mean square
error is often used as the reconstruction loss. 
For example, Castellani et al.~\cite{castellani2021estimating} 
used reconstruction to learn robust representations for detecting noisy labels in the time series. 
Naturally, 
%as an efficient unsupervised learning method, 
reconstruction is also
utilized for TS-PTMs. In the following, we introduce
TS-PTMs based on  the
%reconstruction techniques, including 
(i) autoencoder, (ii)
denoising autoencoder, and (iii) transformer encoder.

\begin{figure}
	\centering 
	\includegraphics[width=0.48\textwidth]{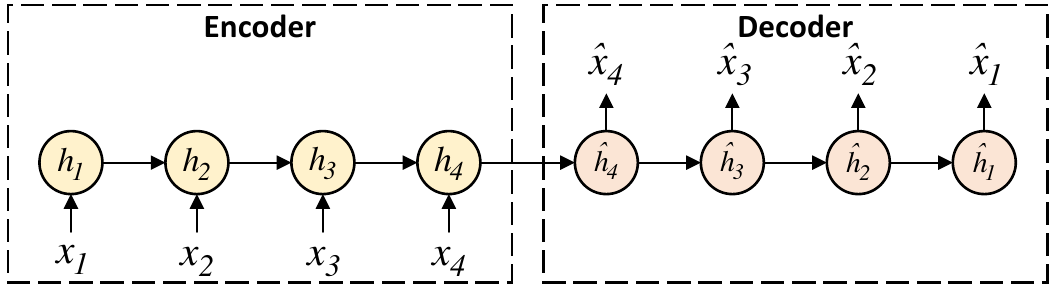}
	\caption{The RNN-based encoder-decoder architecture.} 
	\label{fig:reconstruction}
\end{figure}

\vspace{1ex}
\noindent \textbf{AutoEncoder} {\;}
For variable-length NLP sequences such as sentences, paragraphs, and documents,
it has been very useful
to first convert them to fixed-dimensional vectors \cite{qiu2020pre}. Inspired by this, Malhotra et al.
proposed the time-series model Timenet~\cite{malhotra2017timenet}, which encodes a
univariate time series to a fixed-dimensional vector via the
Sequence-to-Sequence~(Seq2Seq) framework~\cite{sutskever2014sequence}. This consists of an encoder and a decoder.
An example RNN-based encoder-decoder 
is shown
in Fig.~\ref{fig:reconstruction}. The encoder converts the input sequence to a fixed-dimensional vector representation, and the decoder reconstructs it to another sequence as output. 
This process can be written as:
$ \boldsymbol{H} = f_{enc}(\boldsymbol{X})$ and $ \boldsymbol{\hat{X}} = f_{dec}(\boldsymbol{H})$,
where 
$\boldsymbol{H}$ is the representation,
$\boldsymbol{X}$ is the input, $\boldsymbol{\hat{X}}$ is the reconstructed input,
$f_{enc}(\cdot)$ and $f_{dec}(\cdot)$ are the encoder and decoder functions, respectively. 
By 
combining
the pre-trained encoder 
with a classifier, 
Timenet has shown good performance on time-series classification.
However, using autoencoder-based PTMs for other TSM tasks,
such as forecasting and anomaly detection, has been less explored.

% \subsubsection{Denoising-based PTMs}

\vspace{1ex}
\noindent \textbf{Denoising AutoEncoder} {\;}
%Due to the success of autoencoder-based models for unsupervised representation learning, 
To learn more robust representations
for the downstream tasks,
the Denoising AutoEncoder~(DAE)~\cite{vincent2010stacked} has been used
\cite{baevski2019vq}. As shown in
Fig.~\ref{fig:DAE}, the input time series $\boldsymbol{X}$ is corrupted  to
$\boldsymbol{\tilde{X}}$
by adding noise
or random masking.
The DAE is then trained to reconstruct the original $\boldsymbol{X}$. Compared to
the vanilla AutoEncoder, the DAE makes learning more difficult, thus enhancing
robustness of the time-series representations. Since the DAE has achieved good
performance on representation 
learning~\cite{vincent2010stacked,ma2021learning}, it 
is also used in the {}{TS-PTMs~\cite{nietime,dong2024simmtm,zhang2023trid}.}

A pioneering work that uses DAE in TS-PTM is the Audio Word2Vec~\cite{chung2016audio}.
Given variable-length audio segments,
Audio Word2Vec
obtains fixed-dimensional vectors,
which are then used in applications such as query-by-example Spoken Term Detection~(STD). 
To learn robust representations, Audio Word2Vec consists of two stages: offline and online~\cite{vincent2010stacked}. 
In the offline phase,
a pre-trained model is 
obtained using all the audio segments. This is
then
used to encode online language sequences to fixed-dimensional vectors.
Inspired by Audio Word2Vec, Hu et al.~\cite{hu2016transfer} employed a DAE to pre-train the model for short-term wind speed time-series forecasting. 
% The DAE-based model
% is pre-trained on massive data from older farms,
% %In this way, the model can
% which is then fine-tuned on data from newly-built farms.
%for improved performance on wind speed forecasting.
{}{Also, Shao et al.~\cite{shao2022pre} integrated Graph Neural Networks (GNNs) with an unsupervised masked auto-encoder strategy for pre-training multivariate time series forecasting models. Specifically, they employed a series of Transformer blocks as the encoder-decoder architecture and used GNNs to capture the spatio-temporal relationships between different variables.}

\begin{figure}
	\centering 
	\includegraphics[width=0.48\textwidth]{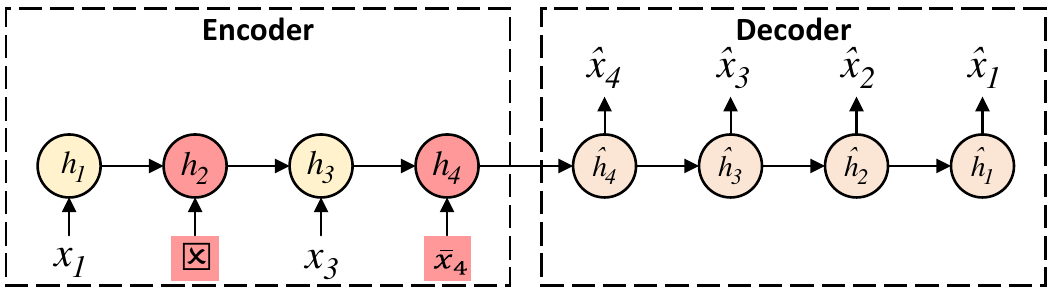}
	\caption{The RNN-based denoising encoder-decoder architecture.} 
	\label{fig:DAE}
\end{figure}

DAE's ``mask and predict" mechanism has been
successfully used in pre-training NLP \cite{devlin2019bert} and CV
models~\cite{he2021masked}. 
%For example, He et al.~\cite{he2021masked} proposed a simple and efficient extensible masked autoencoder CV model.
%based on the ``mask and predict" mechanism.
In the time-series domain, Ma et al.~\cite{ma2021learning} proposed a joint
imputation and clustering DAE-based model for incomplete time-series
representation learning. They showed that the joint learning strategy enables the
imputation task to be optimized in the direction favoring the downstream
clustering task, resulting in better downstream clustering performance. {}{Furthermore, Nie et al.~\cite{nietime} segment time series into patches and successfully apply the ``mask and predict" mechanism for time series pre-training. Additionally, studies~\cite{dong2024timesiam,eldeletslanet,zhangup2me,goswamimoment} indicate that this mechanism can be effectively used for TS-PTMs.}
%using the missing information.
%Hence, the "mask and predict" mechanism based on the DAE architecture is a potential solution for TS-PTMs.

\vspace{1ex}
\noindent \textbf{Transformer Encoder} {\;}
Transformers {}{based on ``mask and predict" training paradigms}
can be regarded as reconstruction-based models, which have been wildly applied in NLP for studying PTMs~\cite{devlin2019bert}.
Transformers have received much attention because of their ability to capture input data's contextual (future-past) dependencies, making them a good choice for modeling sequence data~\cite{chowdhury2022tarnet,liu2021tera}.
Inspired by Transformer-based PTMs in NLP~\cite{kalyan2021ammus}, Zerveas et al.~\cite{zerveas2021transformer} proposed a multivariate time-series unsupervised representation learning model called Time-Series Transformer~(TST).
Specifically, TST employs the original transformer encoder~\cite{vaswani2017attention} to learn multivariate time series representations through a designed masking strategy.
% To design a suitable masking strategy, each masked segment in TST has a length that follows a geometric distribution with mean length $l_m$, and an unmasked segment of mean length is $l_u=\frac{1-r}{r}l_m$. In this way, the masked segment that is too short (e.g., the length is $1$.) will not be generated, thus ensuring the effect of pre-training.
Further, Shi et al.~\cite{shi2021self} proposed an unsupervised pre-training method based on the self-attention mechanism of Transformers. In particular, the authors~\cite{shi2021self} utilize a denoising pretext task that captures the local dependencies and changing trends in the time series by reconstructing corrupted time series. 
Unlike the above studies, Zhang et al.~\cite{zhang2022cross} proposed a cross reconstruction transformer pre-trained by a cross-domain dropping reconstruction task, thereby modeling temporal-spectral relations of time series. 

In addition, Transformer-based PTMs have been applied to traffic flow time series, tabular time series, and speech series data. For example, Hou et al.~\cite{hou2022masked} proposed a token-based PTM for traffic flow time series. Concretely, the authors~\cite{hou2022masked} designed a random mask tokens strategy, and then the transformer encoder was pre-trained to predict these tokens.
% , thereby capturing contextual information of time series. 
Further, Zhao et al.~\cite{zhao2022st} designed a bidirectional Transformer encoder to learn relationships between two-time intervals across different scales.
% , thus modeling temporal dependencies in traffic flow time series.
% proposed a spatial-temporal global semantic representation learning method for modeling spatial and temporal information 
% of traffic flow time series. 
% The authors employed a semantic flow encoder consisting of ResNet and multi-layer perceptron to capture the spatial dependencies. 
% Specially,
% In addition, a bidirectional Transformer encoder was designed to learn relationships between any two-time intervals across different scales to model temporal dependencies. 
For tabular time-series data, Padhi et al.~\cite{padhi2021tabular} proposed hierarchical tabular BERT~\cite{devlin2019bert} as the backbone to pre-train by predicting masked tokens. Also, Shankaranarayana et al.~\cite{shankaranarayana2021attention} utilized a 1-D convolution model combined with the transformer as the backbone, which was pre-trained using the ``mask and predict" mechanism.
In terms of speech series data, Liu et al.~\cite{liu2020mockingjay} proposed an unsupervised speech representation learning method based on a multilayer transformer encoder, pre-trained by recovering the random 15\% zero-masked input frames. 
Unlike the PTMs mentioned above, Liu et al.~\cite{liu2021tera} proposed a novel PTM called Transformer Encoder Representations from Alteration~(TERA) for speech series data. Specifically, TERA utilized three self-supervised training schemes~(time alteration, frequency alteration, and magnitude alteration) 
based on a transformer encoder through reconstructing altered counterpart speech data for pre-training.

\vspace{1ex}
\noindent \textbf{Summary} {\;}
DAE-based TS-PTMs add noise or masks to the original time series for pre-training, and have recently been gaining attention compared to autoencoder-based PTMs.
Using DAE to study PTMs~\cite{he2021masked} demonstrates the potential of denoising strategies in pre-training.
However, designing DAE-based PTMs applicable to time series is still in the exploratory stage.
Meanwhile, Transformer-based PTMs inherit the advantages of unsupervised training from the denoising strategy and have achieved good pre-training results in NLP in recent years~\cite{kalyan2021ammus}.
Nevertheless, existing Transformer-based TS-PTMs mainly focus on the time series classification and forecasting tasks, and their performance on other downstream tasks warrants further exploration.
At the same time, the time series of different domains may vary widely, resulting in poor model transferability across different domain datasets.
Therefore, how to design reconstruction-based unsupervised learning mechanisms~(i.e, mask and predict) applicable to different domains is a challenge for studying  TS-PTMs.

\subsection{Self-supervised PTMs}
This section presents self-supervised TS-PTMs based on consistency and pseudo-labeling training strategies which are commonly utilized in self-supervised learning.
Compared with unsupervised learning (e.g., reconstruction), self-supervised learning employs self-provided supervisory information (e.g., pseudo-labels) during the training process.

\subsubsection{Consistency-based PTMs}
Self-supervised learning strategies based on the consistency of different augmented (transformed) views from the same sample have been studied for PTMs in CV~\cite{wang2021dense} with great success. 
Naturally, {}{consistency-based strategies~\cite{oord2018representation,yang2023dcdetector,hu2023self,liu2024timesurl} have recently} attracted attention in the time series field.
Specifically, consistency-based PTMs keep the distances of different view representations from the same sample (positive pairs) close to each other, while keeping the distances of view representations (negative pairs) from different samples far from each other. 
The above learning strategy motivates the model to learn representations beneficial for downstream tasks.
Based on the above idea, contrastive learning~\cite{he2020momentum,chen2020simple} is commonly employed as a training loss for PTMs, which is defined as:
\begin{equation}
% \begin{split}
\resizebox{.9\hsize}{!}
{$\mathcal{L}_{CL} = 
- \frac{1}{N} \sum_{i=1}^{N} \text{log} \frac{\text{exp}(f(\boldsymbol{X}_i)^{\text{T}}f(\boldsymbol{X}_i^p))}{\text{exp}(f(\boldsymbol{X}_i)^{\text{T}}f(\boldsymbol{X}_i^p)) + \sum_{j=1}^{B-1} \text{exp}(f(\boldsymbol{X}_i)^{\text{T}}f(\boldsymbol{X}_i^j))}$},
\label{InfoNCE}
\end{equation}
where $N$ denotes the number of training sample pairs and each anchor sample~($\boldsymbol{X}_i$) has $1$ positive sample~($\boldsymbol{X}_i^p$) and $B-1$ negative samples. $\boldsymbol{X}_i^j$ denotes the $j$-th negative sample of the $i$-th anchor sample.

Consistency-based PTMs need to consider how to construct positive and negative samples so that the model can effectively exploit the complex dynamic properties of time series data.
It is worth noting that there is no uniform large-scale well-labeled 
time series dataset for pre-training.
Most existing consistency-based PTMs first pre-train the encoder using the target dataset via self-supervised representation learning, and then fine-tune the pre-trained encoder using the supervised information from the target dataset or directly utilize the representations for downstream tasks.
Using contrastive learning for time-series consistency-based PTMs can be roughly divided into four categories~\cite{yue2022ts2vec}: subseries consistency, temporal consistency, transformation consistency, and contextual consistency. 

\begin{figure}
	\centering 
	\includegraphics[width=0.48\textwidth]{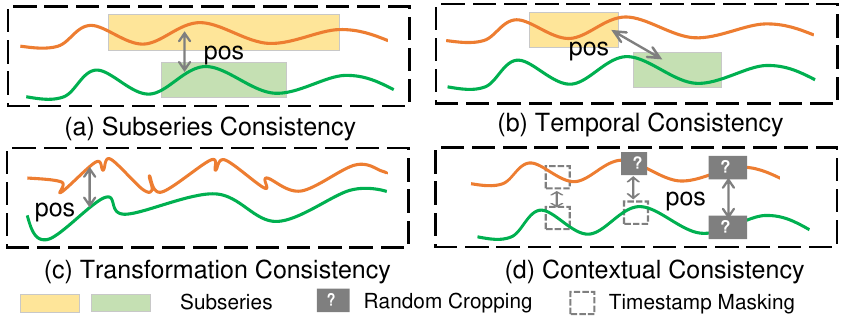}
	\caption{Positive pair selection strategies in contrastive learning.} 
	\label{fig:contrastive}
\end{figure}

\vspace{1ex}
\noindent \textbf{Subseries Consistency} {\;}
% \vspace{0.5ex}
Two subseries belonging to an inclusion relationship sampled from the same time series sample are utilized as positive pair, which is called subseries consistency, as shown in Fig.~\ref{fig:contrastive}~(a).
Utilizing Word2Vec~\cite{mikolov2013distributed} as an analogy, Franceschi et al. first ~\cite{franceschi2019unsupervised} employed a sufficiently long and non-stationary subseries in a time series sample as the context. Then, a subseries selected from the context corresponds to a word (positive subseries), while a subseries collected from a different context in another time series sample represents a random word (negative subseries).
Further, the authors~\cite{franceschi2019unsupervised} employed the Triplet Loss~(T-Loss) to keep the context and positive subsequences close, while making the context and negative subsequences far away for representation learning of time series.
Experimental results on UCR and UEA archives indicated that the
representations obtained by T-Loss can be beneficial for the downstream classification task.

Moreover, Fan et al.~\cite{fan2020self} proposed a self-supervised
representation learning framework named SelfTime, by exploring the inter-sample
and intra-temporal relations of time series.
In terms of the inter-sample relationships, unlike SimCLR~\cite{chen2020simple}, 
the authors employed the cross-entropy loss to guide the reasoning of different levels of entity relationships.
As for the intra-temporal relationships, the authors attempted to capture the temporal pattern by reasoning the multi-level relation among subseries sampled from the same sample, thus obtaining representations for the downstream classification task.

\vspace{1ex}
\noindent \textbf{Temporal Consistency} {\;}
% \vspace{0.5ex}
Two adjacent subseries from the same time series sample are selected as a positive pair, which is called temporal consistency, as shown in Fig.~\ref{fig:contrastive}~(b).
Based on the assumption of temporal consistency, 
Tonekaboni et al.~\cite{tonekaboni2020unsupervised} proposed a representation learning framework named Temporal Neighborhood Coding~(TNC).
TNC learns time series representations by ensuring that the distribution of adjacent signals in the coding space is distinguishable from the distribution of non-neighboring signals.
Experimental results on multiple time series datasets demonstrate that TNC
performs better on downstream classification and clustering tasks compared with
the T-Loss~\cite{franceschi2019unsupervised}.

Further, Woo et al.~\cite{woo2022cost} proposed a contrastive learning pre-training method for long sequence time series forecasting, called CoST. Specifically, CoST utilizes temporal consistencies in the time domain to learn the discriminative trend of a sequence, while transforming the data to the frequency domain to discover the seasonal representation. 
Deldari et al.~\cite{deldari2021time} proposed a self-supervised time series change point detection method based on contrast predictive coding. Particularly, the authors exploit local correlations in the time series to drive the maximization of shared information across consecutive time intervals while minimizing shared information across time intervals.
Luo et al.~\cite{luo2021information} applied the objective of contrastive learning to both the local subseries and global instance levels, thus obtaining feature representations favorable to downstream time series forecasting and classification tasks.

\vspace{1ex}
\noindent \textbf{Transformation Consistency} {\;}
% \vspace{0.5ex} 
Two augmented sequences of the same time series by different transformations are selected as positive pair, which is called transformation consistency, as shown in Fig.~\ref{fig:contrastive}~(c).
Inspired by the successful application of transformation consistency in CV, Eldele et al.~\cite{eldele2021time} proposed a Time-Series representation learning framework via Temporal and Contextual Contrasting (TS-TCC).
TS-TCC first transforms the original time series data in two different ways to obtain weak augmentation (jitter-and-scaling) and strong augmentation (permutation-and-jitter) variants, respectively.
Then, the temporal and contextual contrastive learning modules are designed in TS-TCC for learning discriminative representations. 

In addition, Hyvärinen et al.~\cite{hyvarinen2016unsupervised} 
% proposed an unsupervised feature extraction by time-contrastive learning and nonlinear independent component analysis. Specifically, the authors 
analyzed the non-stationarity of time series by performing a nonlinear mixture transformation of subseries from different time windows to ensure that the same subseries are consistent.
% , and employ a multinomial logistic regression to try to distinguish the differences between subseries. 
Unlike~\cite{hyvarinen2016unsupervised}, Hyvarinen and Morioka~\cite{hyvarinen2017nonlinear} developed a method based on a logistic regression estimation model to learn the dependencies of time dimensions by distinguishing the differences between subseries of the original time series and those randomly transformed time points. 
Lately, Zhang et al.~\cite{zhang2022self} utilized a time-based augmentation bank~(jittering, scaling, time-shifts, and neighborhood segments) and frequency-based augmentation methods~(adding or removing frequency components) for time series self-supervised contrastive pre-training via time-frequency consistency.

\vspace{1ex}
\noindent \textbf{Contextual Consistency} {\;}
% \vspace{0.5ex} 
Two augmented contexts with the same timestamp inside the same time series are
selected as a positive pair, which is called contextual consistency, as shown in Fig.~\ref{fig:contrastive}~(d).
By designing a timestamp masking and random cropping strategy to obtain augmented contexts, Yue et al.~\cite{yue2022ts2vec} proposed a universal framework for learning time series representations named TS2Vec.
Specifically, 
% unlike existing methods~\cite{franceschi2019unsupervised, tonekaboni2020unsupervised,eldele2021time}, 
TS2Vec distinguishes positive and negative samples from the temporal dimensions
and {}{instance level}  hierarchically.
In terms of temporal consistency, the augmented contexts with the same timestamp within the same time series are treated as positive pairs, while the augmented contexts in different timestamps are treated as negative pairs. Unlike temporal consistency, {}{instance level consistency}
treats the augmented contexts of other instances within a batch as negative pairs.
Experiments on various time series datasets indicate that TS2Vec achieves excellent performance on downstream classification, forecasting, and anomaly detection tasks.

Recently, Yang et al.~\cite{yang2022unsupervised} proposed a novel representation learning framework for time series, namely Bilinear Temporal-Spectral Fusion~(BTSF).
Unlike TS2Vec~\cite{yue2022ts2vec}, BTSF utilizes instance-level augmentation by simply applying 
a dropout strategy to obtain positive/negative samples for unsupervised contrastive learning, thus preserving the global context and capturing long-term dependencies of time series.
Further, an iterative bilinear temporal-spectral fusion module is designed in BTSF to iteratively refine the representation of time series by encoding the affinities of abundant time-frequency pairs. 
Extensive experiments on classification, forecasting, and anomaly detection tasks demonstrate the superiority of BTSF.

\begin{table*}[]
\caption{Test classification results of transfer learning on 675 sets~(15 source datasets multiplied by 45 target datasets) UCR time series datasets. Positive represents the classification performance is better than direct classification without using a transfer strategy.}
\label{tab:transfer_result_sum}
\centering
\resizebox{\linewidth}{!}{
\begin{tabular}{@{}c|cccc|cccc|cccc@{}}
\toprule
\multicolumn{1}{c|}{} & \multicolumn{4}{c|}{15 Minimum Target Datasets} & \multicolumn{4}{c|}{15 Medium Target Datasets} & \multicolumn{4}{c}{15 Maximum Target Datasets} \\ \midrule
Transfer Strategy & \multicolumn{1}{c}{Avg. Acc} & Avg. Rank & P-value & Positive & \multicolumn{1}{c}{Avg. Acc} & Avg. Rank & P-value & Positive & \multicolumn{1}{c}{Avg. Acc} & Avg. Rank & P-value & Positive \\ \hline
Sup CLS & {\ul \textbf{0.7720 (0.0738)}} & {{1.75}} & - & {\ul \textbf{48 (225)}} & {\ul \textbf{0.7196 (0.0398)}} & {\ul \textbf{1.45}} & - & {\ul \textbf{9 (225)}} & {\ul \textbf{0.7885 (0.0231)}} & {\ul \textbf{1.73}} & - & {\ul \textbf{17 (225)}} \\
Unsup FCN Decoder & 0.7616 (0.0756) & {\ul \textbf{1.59}} & 5.53E-02 & 41 (225) & {{0.7001 (0.0380)}} & {{1.92}} & 7.64E-08 & {{4 (225)}} & {{0.7715 (0.0240)}} & {{2.02}} & 1.27E-03 & {{15 (225)}} \\
Unsup RNN Decoder & 0.7292 (0.0823) & 2.16 & 2.28E-9 & 29 (225) & 0.6655 (0.0355) & 2.20 & 1.87E-15 & 2 (225) & 0.7532 (0.0264) & 2.20 & 9.38E-07 & 14 (225) \\ \bottomrule
\end{tabular}}
\end{table*}

\vspace{1ex}
\noindent \textbf{Summary} {\;}
% \vspace{0.5ex}
The aforementioned studies demonstrate that consistency-based PTMs can obtain robust representations beneficial for TSM tasks.
Although subseries and temporal consistency strategies have achieved good performance on the classification task, sampling biases between subseries (e.g., outliers and pattern shifts)~\cite{yue2022ts2vec} tend to introduce false positive pairs.
Meanwhile, the transformation consistency strategy relies on effective instance-level data augmentation techniques.
However, designing uniform data augmentation techniques for time
series from different domains remains a
challenge~\cite{wen2020time,iwana2021empirical}. One alternative is to utilize expert features of time series~\cite{nonnenmacher2022utilizing} instead of the commonly used data transformation for contrastive learning.
The contextual consistency strategy utilizes a mask~(or dropout) 
mechanism~\cite{yue2022ts2vec,yang2022unsupervised} 
to obtain contrastive pairs by capturing temporal dependencies, which can alleviate the problem of sampling bias among subsequences and achieve excellent performance on forecasting, classification, and anomaly detection tasks.
Nevertheless, designing consistency-based strategies using the multi-scale property~\cite{tang2021omni} of time series has not been fully explored.

\subsubsection{Pseudo-labeling PTMs}
In addition to consistency-based PTMs mentioned above, some other self-supervised
TS-PTMs have been explored to improve the performance of TSM. We briefly review
these methods in this section.

\vspace{1ex}
\noindent \textbf{Predict Pseudo Labels} {\;}
% \vspace{0.5ex}
A large amount of correctly labeled information is the basis for the success of deep neural network models.
However, labeling time series data generally requires manual expert knowledge assistance, resulting in high annotation costs.
Meanwhile, representation learning using pseudo-label-guided models has yielded rich results in CV~\cite{asano2020self}.
In addition, some studies~\cite{gidaris2018unsupervised,chung2018speech2vec} employed self-supervised
%\footnote{*** contrastive learning is also self-supervised. maybe merge this para w/ sec 3.2.3?  {}{Answer: That's a good question. We add a new Section Self-supervised-based PTMs, and we have let the contrastive learning-based PTMs into Self-supervised-baed PTMs. In addition, we modify the name of contrastive learning-based PTMs as Consistency-based PTMs.} } 
learning as an auxiliary task to help the primary task learn better by training the auxiliary task alongside the primary task. In the auxiliary task, when given the transformed sample, the model predicts which transformation is applied to the input~(i.e., predicts pseudo labels). This idea has been applied in a few TS-PTMs~\cite{fan2020self}. For example, Fan et al.~\cite{fan2020self} 
% proposed a self-supervised time-series representation learning model that performed inter-intra relational reasoning on time series. Specifically, the authors
randomly selected two 
length-$L$
subsequences from the same time series and assigned a pseudo-label based on their temporal distance, and then the proposed model is pre-trained by predicting the pseudo-label of subsequence pairs. Also, Zhang et al.~\cite{zhang2021sleeppriorcl} incorporated expert features to create pseudo-labels for time series self-supervised contrastive representation learning.
Despite these progresses, predicting pseudo labels inevitably contain incorrect labels. How to mitigate the negative impact of incorrect labels will be the focus of studying TS-PTMs.

% \vspace{1ex}
% \noindent \textbf{Predict Contextual Information} {\;}
% % \vspace{0.5ex} 
% The contextual information of a time series can play a guiding role in unsupervised representation learning. 
% For example, speech data is rich in contextual information, which can provide some reference for studying time series unsupervised pre-training.
% Inspired by Word2Vec~\cite{mikolov2013distributed} in NLP, Chung and Glass~\cite{chung2018speech2vec} {}{treated each audio segment as a word in a speech corpus. Then, they utilized an encoder-decoder framework to predict the contextual semantic information audio segments (corresponding to nearby words) within a nearby range $k$ for pre-training. } Specifically, the authors employed skipgrams and cbow to predict the adjacency contextual information of audio segments. The skipgrams capture the semantic information of the current audio segment by minimizing the gap between the output sequence and its corresponding nearby audio segment. Further, cbow\footnote{***  cbow is more like mask-and-predict in sec3.2.2?
% {}{Answer: }
% } performs an inference task using the contextual information of nearby audio segments to obtain robust representations for downstream tasks. Therefore, it is worth exploring contextual information of the subsequence in the study of TS-PTMs.

\section{Experimental Results and Analysis~\label{section4}}
In this section, following~\cite{yue2022ts2vec,yang2022unsupervised}, we evaluate TS-PTMs on three TSM tasks, including classification, forecasting, and anomaly detection. 
Like~\cite{yue2022ts2vec}, we select  a range of time-series benchmark datasets
employed in the corresponding TSM tasks for evaluation.
We first analyze the performance of TS-PTMs on the classification task using UCR~\cite{dau2019ucr} and UEA~\cite{bagnall2018uea} archives time series datasets. Also, following~\cite{zhang2022self}, we select four time series scenarios datasets for transfer learning analysis. Second, the performance of TS-PTMs and related baselines on the forecasting task are compared using {}{nine benchmark datasets~\cite{zhou2021informer,wutimesnet}.}
% the ETT~\cite{zhou2021informer} and Electricity~\cite{Dua:2019} datasets. 
Finally, we analyze the performance of TS-PTMs on the
anomaly detection by using the univariate Yahoo~\cite{YahooTime}, KPI~\cite{ren2019time}, {}{UCR anomaly detection archive~\cite{wu2021current}, and seven multivariate datasets~\cite{yang2023dcdetector}.} 
For information about datasets, baselines, and implementation details, please refer to Appendix~\ref{appendix_B}. 

% Since there is no available benchmark large-scale time series dataset, the above TS-PTMs are pre-trained and analyzed on the same dataset. {}{In other words, it is difficult to select a suitable source dataset to pre-train the encoder to obtain a positive transfer performance on the target dataset.}

% Also, we provide our experimental open-source code\footnote{https://github.com/qianlima-lab/transfer-to-transformer-tsm}.

\subsection{Performance of PTMs on Time-Series Classification}
The datasets in the UCR and UEA archives do not specify a validation set for hyperparameter selection, while some datasets have a much higher number of test set samples than the training set. As suggested by~\cite{dau2019ucr}, we merge the training and test sets for each dataset in UCR and UEA archives. Then, we adopt the five-fold cross-validation strategy to divide the training, validation, and test sets in the ratio of 60\%-20\%-20\%. 
For the four independent time series scenarios datasets, we utilize the datasets processed by~\cite{zhang2022self} for experimental analysis.
Finally, we use the average accuracy on the test set as the evaluation metric.

\begin{table*}[]
\caption{Test classification results of transfer learning on four independent time series scenarios. Supervised (FCN) means that the FCN encoder is directly used for supervised classification training on the target dataset.}
\centering
\resizebox{\linewidth}{!}{
\begin{tabular}{@{}ccccccc@{}}
\toprule
Scenario & Source Dataset & Target Dataset & Supervised (FCN) & Sup CLS & Unsup FCN Decoder & Unsup RNN Decoder \\ \midrule
Neurological Stage Detection & SleepEEG & Epilepsy & 0.7766 (0.0552) & {\ul \textbf{0.8021 (0.0010)}} & 0.6813 (0.2483) & 0.6813 (0.2486) \\
Mechanical Device Diagnosis & FD-A & FD-B & 0.5781 (0.0292) & {\ul \textbf{0.6718 (0.0783)}} & 0.6541 (0.0547) & 0.6541 (0.0543) \\
Activity Recognition & HAR & Gesture & 0.3904 (0.0281) & 0.3796 (0.0257) & 0.3517 (0.0363) & {\ul \textbf{0.4933 (0.0196)}} \\
Physical Status Monitoring & ECG & EMG & {\ul \textbf{0.9756 (0.0322)}} & 0.4634 (0.0010) & 0.8439 (0.0892) & 0.8731 (0.0103) \\ \bottomrule
\end{tabular}}
\label{tab:transfer_result_new}
\end{table*}

% \usepackage{booktabs}
% \usepackage{multirow}
% \usepackage[normalem]{ulem}
\begin{table*}[]
\centering
\caption{{}{Comparisons of classification test accuracy on different TS-PTMs~(standard deviations are in parentheses).}}
\label{tab:cls_ptms}
\resizebox{\linewidth}{!}{
\begin{tabular}{@{}c|cccc|cccc@{}}
\toprule
\multirow{2}{*}{Models} & \multicolumn{4}{c|}{128 UCR datasets} & \multicolumn{4}{c}{30 UEA datasets} \\ \cmidrule(l){2-9} 
 & Avg. Acc & Avg. Rank & P-value & Training Time (hours) & Avg. Acc & Avg. Rank & P-value & Training Time (hours) \\ \cmidrule(r){1-9}
Supervised (FCN) & 0.8296 (0.0520) & 4.15 & 1.24E-02 & 4.87 & 0.7718 (0.0722) & 3.07 & 4.80E-02 & 0.73 \\
T-Loss & 0.8325 (0.0302) & 5.07 & 1.36E-02 & 37.56 & 0.5863 (0.0459) & 7.07 & 2.78E-09 & 22.03 \\
Selftime & 0.8017 (0.0339) & 6.59 & 9.69E-07 & - & - & - & - & - \\
TS-TCC & 0.7807 (0.0430) & 6.59 & 8.79E-12 & 9.45 & 0.7059 (0.0432) & 5.27 & 5.64E-05 & 34.35 \\
TST & 0.7755 (0.0365) & 6.39 & 4.53E-13 & 196.72 & 0.6921 (0.0382) & 5.37 & 3.29E-05 & 52.1 \\
TS2Vec & {\ul \textbf{0.8691 (0.0265)}} & \textbf{3.28} & 1.28E-01 & {\ul \textbf{0.87}} & 0.7101 (0.0411) & 4.27 & 1.04E-03 & {\ul \textbf{2.07}} \\
{}{TimesNet} & {}{0.8367 (0.0687)} & {}{4.4} & {}{4.78E-05} & {}{336.62} & {}{0.7570 (0.0833)} & {}{4.13} & {}{2.58E-03} & {}{137.72} \\
{}{PatchTST} & {}{0.8265 (0.0706)} & {}{4.16} & {}{2.60E-03} & {}{11.94} & {}{0.7504 (0.0898)} & {}{3.9} & {}{3.13E-03} & {}{120.47} \\
{}{GPT4TS} & {}{0.8593 (0.0761)} & {}{3.73} & - & {}{86.02} & {}{{\ul \textbf{0.8355 (0.0978)}}} & {}{{\ul \textbf{2.7}}} & - & {}{16.84} \\ \bottomrule
\end{tabular}}
\end{table*}

\subsubsection{Comparison of Transfer Learning PTMs based on Supervised Classification and Unsupervised Reconstruction}

% Considering the training time and transferability of the models, we choose FCN~\cite{wang2017time} for the experimental analysis of transfer learning.

% Based on the results in Table~\ref{tab:cls_supervised}, we utilize FCN combined with a nonlinear classifier as the encoder for transfer learning.

% Time series classification methods such as  have achieved remarkable performance. 

Due to space constraints, we report the transfer learning classification results in Table~\ref{tab:transfer_result_sum}.
The P-value is calculated for the transfer classification results between the supervised classification and unsupervised reconstruction transfer strategy.
As shown in Table~\ref{tab:transfer_result_sum}, the supervised classification transfer (SUP CLS) strategy has the best average Acc and the number of positive transfer results on the minimum, medium and maximum target datasets. The above results indicate that the SUP CLS strategy is more suitable for time series pre-training than the unsupervised transfer strategy.
% From Table~\ref{tab:transfer_result_sum}, on the smallest target
% datasets,
% the average accuracy and rank of the 225 sets of test accuracies obtained using
% the supervised classification transfer strategy are the best, 
On the smallest target
datasets, the overall performance of the SUP CLS strategy and the unsupervised reconstruction strategy utilizing the symmetric FCN decoder is insignificant (P-value greater than 0.05).
Also,
the unsupervised reconstruction strategy utilizing the symmetric FCN decoder is better than the supervised classification transfer strategy.
% The above results may be because the small difference in the number of samples between the source and target datasets affects the classification accuracy of the transfer strategy.
The above results indicate that a symmetric FCN decoder is more suitable for transfer learning in the time-series classification task than the asymmetric RNN decoder.
% The target datasets with the minimum sample sizes contain the largest number of positive transfer results, indicating that the sample size of the target dataset may affect the classification accuracy of transfer learning. 
Overall, the number of positive transfer classification results obtained on the target datasets is unsatisfactory, which may be related to the small number of samples in the UCR source datasets~(the number of samples in most source datasets is less than 8000, and please refer to Table~\ref{tab:my-table-source} in the Appendix).
% However, obtaining a large-scale time-series dataset for pre-training is still challenging.

% Further, we select four independent time series scenarios datasets for transfer
% learning PTMs analysis. The samples of source datasets for each scenario are
% relatively large, and the test classification results are shown in
% Table~\ref{tab:transfer_result_new}. Compared with the supervised
% approach\footnote{*** is this what u mean?

% {}{Answer: We use FCN as the backbone for classification without using transfer learning. In other words, we directly use target datasets for classification without using a source dataset to pretrain the FCN.}
% } (FCN) {}{without pre-training}, the transfer learning strategy
% based on supervised classification (Sup CLS) can achieve significant positive
% transfer performance in the neural stage detection and mechanical device diagnosis
% scenarios. In addition, the transfer learning strategy based on unsupervised RNN
% Decoder can achieve obvious positive transfer performance in the activity
% recognition scenario. However, in the physical status monitoring scenario, the
% three transfer learning strategies all suffer from the negative transfer, which may be
% related to the small number of EMG samples. Compared with using the UCR time series datasets, transfer learning has a better pre-training effect on independent time series scenarios datasets with large source datasets.

Further, we select four independent time series datasets for transfer learning PTMs analysis. The number of samples in each source dataset is large, and the test classification results are shown in Table~\ref{tab:transfer_result_new}. Compared with the supervised approach (FCN) {}{without pre-training}, the transfer learning strategy based on supervised classification (Sup CLS) achieves significant positive transfer performance in the neural stage detection and mechanical device diagnosis scenarios. In addition, the transfer learning strategy based on unsupervised RNN Decoders achieves obvious positive transfer performance in the activity recognition scenario. However, in the physical status monitoring scenario, the three transfer learning strategies all result in negative transfer, which may be related to the small number of EMG samples. Compared with using UCR time series datasets, transfer learning has a better pre-training effect on independent time series datasets with large source datasets.

\subsubsection{Comparison of {}{TS-PTMs for Classification}}

We select {}{one general time series model and seven TS-PTMs} for performance comparison of the downstream classification task. {}{TimesNet~\cite{wutimesnet} is a general model based on CNNs for time series analysis.}
TST~\cite{zerveas2021transformer} is a transformer-based PTM. T-loss~\cite{franceschi2019unsupervised}, Selftime~\cite{fan2020self}, TS-TCC~\cite{eldele2021time}, and TS2Vec~\cite{yue2022ts2vec} are consistency-based PTMs.
{}{PatchTST~\cite{nietime} utilizes patches based on the transformer for time series modeling, employing a reconstruction strategy for pre-training. GPT4TS~\cite{zhou2023one} adopts the same patch strategy in PatchTST to fine-tune pre-trained LLMs for time series analysis.}
% Also, the encoder of the above transformer-based and consistency-based methods is first pre-trained on the target dataset using an unsupervised/self-supervised pre-learning strategy. Then the label information of the target dataset is utilized to fine-tune the pre-trained encoder or to perform downstream classification tasks using the low-dimensional representations obtained from the pre-trained encoder.
Related work in CV~\cite{chen2020simple} and NLP~\cite{devlin2019bert} generally utilizes a selected backbone combined with a linear classifier as a supervised method to analyze the classification performance compared with PTMs.
Therefore, the FCN combined with a linear classifier is selected as a baseline without pre-training. 
The averaged results on UCR and UEA archives are reported in Table~\ref{tab:cls_ptms}.
Detailed results are in
{}{Appendix~\ref{appendix_C2}. Also, we provide the visualization analysis in Appendix~\ref{appendix_C2}.}
{}{Since the training of Selftime on the multivariate time series UEA datasets is too time-consuming, we only report its performance on the univariate UCR datasets.}
Also, we spend about a month on four 1080Ti GPUs to obtain the classification results of SelfTime on the UCR archive.
% For the detailed test classification accuracy of each dataset in Table~\ref{tab:cls_ptms}, please refer to Appendix 2.2.

As shown in Table~\ref{tab:cls_ptms}, the classification performance of TS2Vec {}{and GPT4TS achieve the best and second-best on the 128 UCR datasets. Also, the P-value is calculated to compare the classification results between GPT4TS and other methods. 
In terms of the P-value for significance tests, TS2Vec is insignificant compared to GPT4TS (P-value greater than 0.05).
% Since the P-value between GPT4TS and TS2Vec is greater than 0.05, it indicates no significant difference in classification performance between the two . 
However, GPT4TS requires more training time for fine-tuning LLMs compared to TS2Vec.}
The above results show that {}{both consistency-based TS2Vec and LLM-based GPT4TS} can effectively learn robust representations beneficial for the time-series classification task on the UCR time series classification archive.
On the 30 UEA datasets{}{, GPT4TS achieves the best performance. 
The model using FCN} for direct classification is
the second-best, while TS2Vec is
inferior in terms of average accuracy and rank. 
% Further, in terms of the P-value for significance tests, TS2Vec is insignificant compared to TS-TCC and TST (p-value greater than 0.05).
{}{Also, PatchTST outperforms TS2Vec, highlighting the effectiveness of the patch strategy combined with a vanilla Transformer for multivariate time series classification. Additionally, GPT4TS shows faster training times on the UEA datasets compared to most other methods, such as PatchTST, due to its ability to converge quickly with fewer training epochs. In contrast, TimesNet requires significantly more training time on both the UCR and UEA datasets. In summary, consistency-based strategies, patch-based strategies, and LLM-based fine-tuning demonstrate significant potential for pre-training in time series classification tasks.}

% The UEA archive is the benchmark for multivariate time series classification, while TS2Vec does not have a suitable pre-training strategy designed specifically for time series variables.
% In addition, the TST model based on the vanilla Transformer has poorer
% classification results than Supervised (FCN) on both UCR and UEA archives,
% indicating that the vanilla Transformer architecture is still a
% challenge for pre-training in the time series classification task.

% In addition, comparing the direct classification results in Table~\ref{tab:cls_supervised} using Transformer combined with a linear classifier on 128 UCR datasets, we find that TST can effectively improve the classification performance of Transformer.

% \usepackage{booktabs}
% \usepackage[normalem]{ulem}
\begin{table*}[]
\caption{{}{Comparison of average test results for time-series forecasting with prediction lengths of \( S \in \{24, 48, 168, 336, 720\} \) for ETTh1, ETTh2, and Electricity datasets; \( S \in \{24, 48, 96, 288, 672\} \) for ETTm1 dataset; \( S \in \{24, 36, 48, 60\} \) for ILI dataset; and \( S \in \{96, 192, 336, 720\} \) for other datasets. ``-" indicates that the results could not be obtained due to memory errors or excessive training time.}}
\label{tab:my-table7}
\resizebox{\linewidth}{!}{ 
\begin{tabular}{@{}c|cccccccccccccccccccccccc@{}}
\toprule
Models & \multicolumn{2}{c}{LogTrans} & \multicolumn{2}{c}{TCN} & \multicolumn{2}{c}{Informer} & \multicolumn{2}{c}{Autoformer} & \multicolumn{2}{c}{TS2Vec} & \multicolumn{2}{c}{CoST} & \multicolumn{2}{c}{{}{TimesNet}} & \multicolumn{2}{c}{{}{PatchTST}} & \multicolumn{2}{c}{{}{DLinear}} & \multicolumn{2}{c}{{}{GPT4TS}} & \multicolumn{2}{c}{{}{TEMPO}} & \multicolumn{2}{c}{{}{iTransformer}} \\ \midrule
Metric & \multicolumn{1}{c}{MSE} & \multicolumn{1}{c}{MAE} & \multicolumn{1}{c}{MSE} & \multicolumn{1}{c}{MAE} & \multicolumn{1}{c}{MSE} & \multicolumn{1}{c}{MAE} & \multicolumn{1}{c}{MSE} & \multicolumn{1}{c}{MAE} & \multicolumn{1}{c}{MSE} & \multicolumn{1}{c}{MAE} & \multicolumn{1}{c}{MSE} & \multicolumn{1}{c}{MAE} & \multicolumn{1}{c}{MSE} & \multicolumn{1}{c}{MAE} & \multicolumn{1}{c}{MSE} & \multicolumn{1}{c}{MAE} & \multicolumn{1}{c}{MSE} & \multicolumn{1}{c}{MAE} & \multicolumn{1}{c}{MSE} & \multicolumn{1}{c}{MAE} & \multicolumn{1}{c}{MSE} & \multicolumn{1}{c}{MAE} & \multicolumn{1}{c}{MSE} & \multicolumn{1}{c}{MAE} \\ \hline
ETTh1 & 0.9166 & 0.7521 & 0.8815 & 0.7098 & 0.9174 & 0.7317 & 0.4574 & 0.4639 & 0.8061 & 0.6543 & 0.6482 & 0.5843 & {}{0.4758} & {}{0.4635} & {}{0.5121} & {}{0.4804} & {}{0.4697} & {}{0.4633} & {}{0.4655} & {}{0.4602} & {}{0.5341} & {}{0.4972} & {\ul \textbf{{}{0.4179}}} & {\ul \textbf{{}{0.4259}}} \\
ETTh2 & 3.0287 & 1.3299 & 4.1595 & 1.5241 & 3.3405 & 1.4479 & 0.4126 & 0.4412 & 1.5138 & 0.9413 & 1.2838 & 0.8533 & {}{0.3896} & {}{0.4060} & {}{0.3676} & {}{0.3989} & {}{0.4929} & {}{0.4816} & {}{0.3917} & {}{0.4172} & {}{0.4542} & {}{0.4544} & {\ul \textbf{{}{0.3363}}} & {\ul \textbf{{}{0.3759}}} \\
ETTm1 & 1.0399 & 0.6569 & 0.7323 & 0.6077 & 1.0390 & 0.6837 & {\ul \textbf{0.2911}} & {\ul \textbf{0.3415}} & 0.6834 & 0.5717 & 0.5957 & 0.5253 & {}{0.4029} & {}{0.4049} & {}{0.3672} & {}{0.3852} & {}{0.3693} & {}{0.3876} & {}{0.3260} & {}{0.3621} & {}{0.3698} & {}{0.3907} & {}{0.3538} & {}{0.3784} \\
{}{ETTm2} & {}{1.2367} & {}{0.8489} & {}{0.9831} & {}{0.7286} & {}{1.4381} & {}{0.8539} & {}{0.3137} & {}{0.3565} & {}{1.1120} & {}{0.7532} & {}{0.9154} & {}{0.6846} & {}{0.3256} & {}{0.3493} & {}{0.2943} & {}{0.3367} & {}{0.3550} & {}{0.4041} & {}{0.2974} & {}{0.3508} & {}{0.3032} & {}{0.3492} & {\ul \textbf{{}{0.2919}}} & {\ul \textbf{{}{0.3342}}} \\
Electricity & 0.3921 & 0.4210 & 0.4998 & 0.4889 & 0.6151 & 0.5550 & 0.2401 & 0.3370 & 0.3622 & 0.4206 & 0.1916 & 0.2845 & {}{0.1915} & {}{0.2891} & {}{0.2136} & {}{0.2913} & {}{0.2750} & {}{0.3661} & {\ul \textbf{{}{0.1497}}} & {\ul \textbf{{}{0.2409}}} & \multicolumn{1}{c}{-} & \multicolumn{1}{c}{-} & {}{0.2005} & {}{0.2818} \\
{}{Traffic} & {}{0.9500} & {}{0.5396} & {}{0.7040} & {}{0.4360} & {}{0.7291} & {}{0.4089} & {}{0.6398} & {}{0.3942} & \multicolumn{1}{c}{-} & \multicolumn{1}{c}{-} & \multicolumn{1}{c}{-} & \multicolumn{1}{c}{-} & {}{0.6417} & {}{0.3362} & {}{0.6308} & {}{0.3937} & {}{0.7918} & {}{0.4835} & {\ul \textbf{{}{0.3953}}} & {\ul \textbf{{}{0.2677}}} & {}{0.4194} & {}{0.3035} & {}{0.4518} & {}{0.3002} \\
{}{Weather} & {}{0.3074} & {}{0.3730} & {}{0.2543} & {}{0.3169} & {}{0.7977} & {}{0.6144} & {}{0.3508} & {}{0.3853} & {}{0.2620} & {}{0.3288} & {\ul \textbf{{}{0.2347}}} & {}{0.3026} & {}{0.2777} & {}{0.2992} & {}{0.2791} & {}{0.2973} & {}{0.2662} & {}{0.3184} & {}{0.2353} & {\ul \textbf{{}{0.2755}}} & {}{0.2453} & {}{0.2802} & {}{0.2625} & {}{0.2825} \\
{}{Exchange} & {}{1.5922} & {}{1.0458} & {}{1.0117} & {}{0.7443} & {}{1.2631} & {}{0.9041} & {}{0.5071} & {}{0.5011} & {}{0.6890} & {}{0.6061} & {}{0.7597} & {}{0.6614} & {}{0.5506} & {}{0.4878} & {}{0.3856} & {}{0.4242} & {\ul \textbf{{}{0.2943}}} & {\ul \textbf{{}{0.4053}}} & {}{0.4430} & {}{0.4525} & \multicolumn{1}{c}{-} & \multicolumn{1}{c}{-} & {}{0.3724} & {}{0.4136} \\
{}{ILI} & {}{7.2016} & {}{1.9188} & {}{7.2306} & {}{1.9251} & {}{5.3215} & {}{1.5495} & {}{3.3649} & {}{1.3129} & {}{3.2403} & {}{1.1139} & {\ul \textbf{{}{2.0877}}} & {\ul \textbf{{}{0.9073}}} & {}{2.2351} & {}{0.9849} & {}{2.7266} & {}{1.1015} & {}{3.9794} & {}{1.4557} & {}{2.9957} & {}{1.2795} & {}{4.3664} & {}{1.5527} & {}{2.1335} & {}{1.0102} \\ \bottomrule
\end{tabular}}
\end{table*}

% \usepackage{booktabs}
% \usepackage{multirow}
% \usepackage[normalem]{ulem}
\begin{table*}[]
\centering
\caption{Comparisions test results of time-series anomaly detection. {}{Aff-P and Aff-R refer to the precision and recall of the affiliation metric~\cite{huet2022local}. R\_A\_R and R\_A\_P represent Range-AUC-ROC and Range-AUC-PR~\cite{paparrizos2022volume}, indicating scores based on label transformation under the ROC and PR curves, respectively. V\_ROC and V\_RR denote the volumes under the surfaces created by the ROC and PR curves~\cite{paparrizos2022volume}, respectively.}}
\label{tab:ts_anamaly_result}
\resizebox{\linewidth}{!}{ 
\begin{tabular}{@{}c|ccccccccc|ccccccccc@{}}
\toprule
\multirow{2}{*}{Models} & \multicolumn{9}{c|}{Yahoo} & \multicolumn{9}{c}{KPI} \\ \cmidrule(l){2-19} 
 & \multicolumn{1}{c}{F1} & \multicolumn{1}{c}{P} & \multicolumn{1}{c}{R} & \multicolumn{1}{c}{{}{Aff-P}} & \multicolumn{1}{c}{{}{Aff-R}} & \multicolumn{1}{c}{{}{R\_A\_R}} & \multicolumn{1}{c}{{}{R\_A\_P}} & \multicolumn{1}{c}{{}{V\_ROC}} & \multicolumn{1}{c|}{{}{V\_PR}} & \multicolumn{1}{c}{F1} & \multicolumn{1}{c}{P} & \multicolumn{1}{c}{R} & \multicolumn{1}{c}{{}{Aff-P}} & \multicolumn{1}{c}{{}{Aff-R}} & \multicolumn{1}{c}{{}{R\_A\_R}} & \multicolumn{1}{c}{{}{R\_A\_P}} & \multicolumn{1}{c}{{}{V\_ROC}} & \multicolumn{1}{c}{{}{V\_PR}} \\ \cmidrule(r){1-19}
SPOT & 0.5037 & 0.4972 & 0.5105 & {}{0.9413} & {}{0.5425} & {}{0.5758} & {}{0.4389} & {}{0.5828} & {}{0.4420} & {}{0.1911} & {}{0.7628} & {}{0.1092} & {}{{\ul \textbf{0.9310}}} & {}{0.4090} & {}{0.5431} & {}{0.1211} & {}{0.5535} & {}{0.1314} \\
DSPOT & 0.3952 & 0.3222 & 0.5109 & {}{{\ul \textbf{0.9468}}} & {}{0.2889} & {}{0.5500} & {}{0.2814} & {}{0.5585} & {}{0.2890} & {}{0.1250} & {}{0.0889} & {}{0.2106} & {}{0.9092} & {}{0.3696} & {}{0.5224} & {}{0.1136} & {}{0.5243} & {}{0.1154} \\
LSTM-VAE & 0.5386 & 0.4764 & 0.6196 & {}{0.7533} & {}{0.7818} & {}{0.6791} & {}{0.6662} & {}{0.6709} & {}{0.6392} & {}{0.1833} & {}{0.3715} & {}{0.1216} & {}{0.5488} & {}{0.6642} & {}{{\ul \textbf{0.6619}}} & {}{{\ul \textbf{0.4229}}} & {}{0.6506} & {}{0.3939} \\
DONUT & 0.4352 & 0.3795 & 0.5101 & {}{0.6429} & {}{0.6930} & {}{{\ul \textbf{0.7100}}} & {}{{\ul \textbf{0.6834}}} & {}{{\ul \textbf{0.6978}}} & {}{{\ul \textbf{0.6413}}} & {}{0.4044} & {}{0.5423} & {}{0.3225} & {}{0.5411} & {}{{\ul \textbf{0.7006}}} & {}{0.6615} & {}{0.4200} & {}{{\ul \textbf{0.6516}}} & {}{{\ul \textbf{0.3943}}} \\
SR$^*$ & 0.5630 & 0.4510 & 0.7470 & - & - & - & - & - & - & 0.6220 & 0.6570 & {\ul \textbf{0.5980}} & - & - & - & - & - & - \\
AT & 0.7529 & {\ul \textbf{0.8173}} & 0.6980 & - & - & - & - & - & - & 0.4444 & 0.7272 & 0.3200 & - & - & - & - & - & - \\
TS2Vec & {\ul \textbf{0.7574}} & 0.7543 & {\ul \textbf{0.7605}} & {}{0.9331} & {}{{\ul \textbf{0.8788}}} & {}{0.6150} & {}{0.6056} & {}{0.6207} & {}{0.6026} & {}{{\ul \textbf{0.6904}}} & {}{{\ul \textbf{0.9221}}} & {}{0.5517} & {}{0.8730} & {}{0.5922} & {}{0.5659} & {}{0.3753} & {}{0.5709} & {}{0.3782} \\ \bottomrule
\end{tabular}}
\end{table*}

\subsection{Performance of PTMs on Time-Series Forecasting}

% Also, we compare them with supervised benchmark methods in time-series forecasting, including LogTrans~\cite{li2019enhancing}, TCN~\cite{bai2018empirical}, Informer~\cite{zhou2021informer}, and Autoformer~\cite{xu2021autoformer}, where LogTrans, Informer, and Autoformer are end-to-end models based on the Transformer. TS2Vec~\cite{yue2022ts2vec} and CoST~\cite{woo2022cost} 

We further evaluate the performance of PTMs including TS2Vec~\cite{yue2022ts2vec}, CoST~\cite{woo2022cost}, {}{GPT4TS~\cite{zhou2023one}, and TEMPO~\cite{caotempo}} on time-series forecasting. 
Experiments on state-of-the-art direct forecasting methods are also conducted for comparison. These approaches include {}{five} Transformer based models, LogTrans~\cite{li2019enhancing}, Informer~\cite{zhou2021informer}, Autoformer~\cite{xu2021autoformer}, {}{PatchTST~\cite{nietime}, iTransformer~\cite{liuitransformer}, one simple liner layer-based model named DLinear~\cite{zeng2023transformers}}, Temporal Convolutional Network~(TCN)~\cite{bai2018empirical}, {}{and TimesNet~\cite{wutimesnet}.}
% TS2Vec and CoST performs pre-training the downstream forecasting task on the same dataset, while other baselines are trained directly on the dataset without pre-training.
We employ the Mean Square Error (MSE) and Mean Absolute Error (MAE) as evaluation metrics following \cite{woo2022cost}. Also, we follow \cite{yue2022ts2vec,wutimesnet} to preprocess the datasets. Table~\ref{tab:my-table7} presents {}{the average test results with different prediction lengths on nine public datasets} for multivariate time-series forecasting. {}{For detailed results of Table~\ref{tab:my-table7}, please refer to Appendix~\ref{appendix_C3}.}

{}{As can be seen in Table~\ref{tab:my-table7}, iTransformer and GPT4TS generally outperform other methods. Also, CoST, Autoformer, and DLinear demonstrate notable improvements, with better long-term robustness on two, one, and one datasets, respectively.
The performance of GPT4TS suggests that fine-tuning pre-trained LLMs with patching process data for time series forecasting is a promising strategy. Furthermore, iTransformer and Autoformer demonstrate the potential of Transformer frameworks for time series pre-training.
Note that CoST achieves the best performance on
the Weather and ILI datasets. We credit it to the modeling of trend and seasonal features with self-supervised pre-training, which is also verified to be effective in the decomposition architecture of Autoformer.
% Most existing studies are end-to-end supervised methods, while few utilize pre-training and downstream fine-tuning paradigms for time-series forecasting.  
The above empirical results demonstrate that designing an efficient TS-PTM is a promising paradigm for time series forecasting.}

\subsection{Performance of PTMs on Time-Series Anomaly Detection}

%  For the time-series anomaly detection task, we select TS2Vec~\cite{yue2022ts2vec} as the benchmark TS-PTM method. We compare it with the benchmark methods SPOT~\cite{siffer2017anomaly}, DSPOT~\cite{siffer2017anomaly}, LSTM-VAE~\cite{park2018multimodal}, DONUT~\cite{xu2018unsupervised}, Spectral Residual (SR)~\cite{ren2019time}, and Anomaly Transformer (AT)~\cite{xu2021anomaly} in time-series anomaly detection

For time-series anomaly detection, we follow the settings of~\cite{ren2019time, yue2022ts2vec} to determine whether the last time point $x_{t}$ of a sequence fragment $[x_{1}, x_{2}, \ldots, x_{t}]$ is an anomaly. {}{Also, we select various evaluation metrics for a comprehensive comparison, including the commonly-used F1-Score (F1), Precision (P), and Recall (R) metrics~\cite{yue2022ts2vec}, as well as recently proposed evaluation measures: f1-score Point Adjustment (PA)\%K~\cite{kim2022towards}, affiliation precision/recall pair~\cite{huet2022local}, and volume under the surface~\cite{paparrizos2022volume}.}
% and preprocess the datasets in the manner of~\cite{yue2022ts2vec}. 
The TS2Vec~\cite{yue2022ts2vec}, {}{TimesNet~\cite{wutimesnet}, GPT4TS~\cite{zhou2023one}, DCdetector~\cite{yang2023dcdetector},} and the benchmark methods SPOT~\cite{siffer2017anomaly}, DSPOT~\cite{siffer2017anomaly}, LSTM-VAE~\cite{park2018multimodal}, DONUT~\cite{xu2018unsupervised}, Spectral Residual~(SR)~\cite{ren2019time}, and Anomaly Transformer (AT)~\cite{xu2021anomaly} are employed as comparison methods. 
% Like the classification and forecasting tasks, 
% TS2Vec performs pre-training and the downstream  anomaly detection task on the same dataset, while other baselines are trained without pre-training.
The test results on Yahoo and KPI datasets are shown in Table~\ref{tab:ts_anamaly_result}. 
The authors of SR do not provide open-source code, so we use the anomaly detection results (F1, P, and R) reported in the original SR article~\cite{ren2019time} for comparison. {}{Also, the ``-" in the AT method indicates that the metric could not be obtained due to excessive computation time.}
% Among them, the authors of SR do not provide open-source code. We follow the anomaly detection results (F1, P, and R) provided in the original SR article~\cite{ren2019time} for comparative analysis.
% Also, ”-” in the AT method indicates that the metric could not be obtained due to the excessive time required to compute it
% from the model’s output.

As shown in Table~\ref{tab:ts_anamaly_result}, 
{}{on the traditional F1, P, and R metrics, TS2Vec and AT achieve the best performance compared to other baselines. However, on the other six metrics, the overall performance of DOUNT is better than TS2Vec.
The above results suggest that there are still challenges in designing suitable PTMs over traditional methods for the latest evaluation metrics on Yahoo and KPI datasets.
The test results on the UCR anomaly detection archive in Table~\ref{tab:ts_anamaly_result_ucr250} indicate that TS2Vec outperforms DOUNT on the F1, P, F1-PA-10, F1-PA-50, and F1-PA-90 metrics, while DOUNT excels in the other six metrics. Overall, the consistency-based DCdetector achieves the best average performance across all twelve metrics. Due to space constraints, the detailed test results of Table~\ref{tab:ts_anamaly_result_ucr250} and results for seven multivariate datasets are provided in Appendix~\ref{appendix_C4}. For the seven multivariate datasets, DCdetector performs best on five datasets, while TS2Vec and GPT4TS each lead in one dataset. Notably, DCdetector uses a patch strategy similar to PatchTST to preprocess raw time series data before inputting it into the Transformer for pre-training. In short, the patch-based strategy and Transformer framework present promising avenues for further exploration in pre-training for time series anomaly detection tasks.}

% Meanwhile, Table 6 indicates the test results on the UCR anomaly detection archive. 
% TS2Vec outperforms other comparison
% methods on F1-Score on both the Yahoo and KPI datasets, indicating that the pre-training strategy based on contrastive learning can effectively improve the performance of the time-series anomaly detection task.
% However, on the Yahoo dataset, AT significantly outperforms TS2Vec in the Precision metric, and the F1-Score is also very close to TS2Vec, indicating that the AT model based on Transformer architecture has certain advantages in the time-series anomaly detection.
% In addition, the performance of AT demonstrates that Transformer has excellent potential for studying time-series anomaly detection.

% \usepackage{booktabs}
% \usepackage[normalem]{ulem}
\begin{table*}[]
\centering
\caption{{}{Comparison of optimal dataset count for time-series anomaly detection in the UCR anomaly detection archive with 250 datasets. F1-PA-10, F1-PA-50, and F1-PA-90 represent the F1-scores for PA\%K~\cite{kim2022towards} with \( k \) set to 10, 50, and 90, respectively. ``-" indicates that the metric could not be obtained due to excessive computation time.}}
\label{tab:ts_anamaly_result_ucr250}
\begin{tabular}{@{}c|cccccccccccc|c@{}}
\toprule
{}{Models} & {}{F1} & {}{P} & {}{R} & {}{F1-PA-10} & {}{F1-PA-50} & {}{F1-PA-90} & {}{Aff-P} & {}{Aff-R} & {}{R\_A\_R} & {}{R\_A\_P} & {}{V\_ROC} & {}{V\_PR} & {}{Avg. Count}\\ \midrule
{}{SPOT} & {}{7} & {}{10} & {}{10} & {}{68} & {}{73} & {}{70} & {}{16} & {}{6} & {}{4} & {}{4} & {}{4} & {}{4} & {}{23.00} 
\\
{}{DSPOT} & {}{6} & {}{9} & {}{12} & {}{{\ul \textbf{71}}} & {}{{\ul \textbf{92}}} & {}{{\ul \textbf{77}}} & {}{12} & {}{9} & {}{5} & {}{7} & {}{5} & {}{7} & {}{26.00} 
\\
{}{LSTM-VAE} & {}{21} & {}{19} & {}{{\ul \textbf{169}}} & {}{49} & {}{31} & {}{34} & {}{10} & {}{{\ul \textbf{110}}} & {}{53} & {}{{\ul \textbf{84}}} & {}{52} & {}{78} & {}{59.17} 
\\
{}{DONUT} & {}{21} & {}{18} & {}{{\ul \textbf{169}}} & {}{46} & {}{28} & {}{31} & {}{7} & {}{87} & {}{33} & {}{79} & {}{31} & {}{66} & {}{51.33} 
\\
{}{AT} & {}{23} & {}{34} & {}{17} & - & - & - & {}{81} & {}{4} & {}{20} & {}{9} & {}{19} & {}{10} & {}{24.11} 
\\
{}{TS2Vec} & {}{73} & {}{67} & {}{108} & {}{62} & {}{82} & {}{74} & {}{13} & {}{28} & {}{14} & {}{22} & {}{13} & {}{19} & {}{47.92} 
\\
{}{TimesNet} & {}{23} & {}{36} & {}{19} & - & - & - & {}{54} & {}{8} & {}{13} & {}{6} & {}{15} & {}{6} & {}{20.00}
\\
{}{GPT4TS} & {}{16} & {}{22} & {}{17} & - & - & - & {}{{\ul \textbf{83}}} & {}{4} & {}{13} & {}{5} & {}{9} & {}{5} &  {}{19.33} 
\\
{}{DCdetector} & {}{{\ul \textbf{103}}} & {}{{\ul \textbf{99}}} & {}{146} & {}{59} & {}{49} & {}{69} & {}{6} & {}{45} & {}{{\ul \textbf{127}}} & {}{66} & {}{{\ul \textbf{134}}} & {}{{\ul \textbf{87}}} & {}{\ul \textbf{82.50}} 
 \\ \bottomrule
\end{tabular}
\end{table*}

\section{Future Directions~\label{section5}}
% Although TS-PTMs have proven their effectiveness in various TSM tasks, there are still major challenges due to the complexity of time series data. In this section, we suggest some potential directions of TS-PTMs for future work.

\subsection{Large-scale Time Series Datasets}

Large-scale benchmark datasets are critical to the success of PTMs in CV and NLP.
Most existing TS-PTMs are pre-trained on datasets from archives such as UCR~\cite{dau2019ucr} and UEA~\cite{bagnall2018uea}, most of which have small sample sizes (only a few thousand or even hundreds of samples).
Although these time-series benchmark datasets have contributed significantly to the development of the time-series community, it is challenging to utilize them to pre-train deep learning models due to limitations of sample sizes and generality.

Recently, Yang et al.~\cite{yang2021voice2series} proposed a TS-PTM called Voice2Series that pre-trains deep learning models using a large-scale sequence dataset from the speech domain. Then, the pre-training models are transferred to the general time-series datasets (UCR time-series archive) through model reprogramming and label mapping. 
Experimental results indicate that pre-training with large-scale sequence datasets in the speech domain can achieve state-of-the-art on the time-series classification task. 
{}{Further, several studies~\cite{goswamimoment,liu2024timer,woo2024unified} have constructed large-scale time series datasets by integrating sequences from various existing time series datasets across multiple domains.  These synthesized datasets have been effectively employed for the purpose of time series pre-training.
% Therefore, how to employ large-scale sequence data in related fields is a worthy research direction for TS-PTMs.
However,} the construction of large-scale generic time-series datasets like ImageNet~\cite{deng2009imagenet} is a crucial focus, which will significantly facilitate further research on TS-PTMs.

\subsection{Inherent Properties of Time Series}

Time-series representation learning has attracted much attention in recent years.
% as deep learning has been extensively applied to the field of time series. 
% The key to time-series representation learning is to mine the inherent properties (such as temporal dependencies and multi-scale dependencies) in time-series data.
Existing studies have explored the inherent properties of time series for representation learning, such as using CNNs to capture multi-scale dependencies, RNNs to model temporal dependencies, and Transformers to model long-term temporal dependencies. Also, the context-dependencies and frequency domain~(or seasonal-trend~\cite{woo2022cost}) information of time series have been explored in recent contrastive learning studies.
Although mining the inherent properties of time series can learn representations beneficial for downstream TSM tasks, the transferability of time series from different domains remains weakly interpretable.

Due to the 
inherent properties~(i.e., frequency domain information) of time series, the pre-training techniques applied to image data are difficult to transfer directly to time series. Compared to NLP, TS-PTMs are challenging to learn universal time-series representations due to the absence of a large-scale unified semantic sequence dataset.
For example, each word in a text sequence dataset has similar semantics in different sentences with high probability. Therefore, the word embeddings learned by the model can transfer knowledge across different text sequence data scenarios.
However, time series datasets are challenging to obtain subsequences (corresponding to words in text sequences) that have consistent semantics across scenarios, making it difficult to transfer the knowledge learned by the model.
Hence, exploiting the inherent properties of time series to mine transferable segments in time series data is a challenge for future research on TS-PTMs.

\subsection{{}{Deep Learning Models} in Time Series}

% In time series domain, RNNs and CNNs are two commonly used deep learning models, where RNNs are often used to model the temporal dependency of time series, and CNNs are used to extract local features of time series. Therefore, TS-PTMs often employ RNNs or CNNs as the backbone. % However, it is generally difficult to obtain excellent performance by directly utilizing the vaillina transformer for time series pertaining. 

{}{Recently, deep learning models such as CNNs and Transformers~\cite{wutimesnet,chen2023contiformer} have garnered significant attention in time series mining. In particular, Transformer-based models have achieved excellent pre-training performance in fields like NLP and CV, prompting their exploration in time series studies~\cite{liu2024timer}. Leveraging multi-head attention mechanisms to capture long-term dependencies, Transformers have been successfully employed to design large foundational pre-training models for time series forecasting tasks~\cite{dasdecoder,woo2024unified}. However, there are relatively few Transformer models with competitive advantages for time series classification, possibly because this task focuses more on capturing discriminative subseries (e.g., shapelets) or multiscale features~\cite{tang2021omni}. For classification tasks, the superior performance of consistency-based PTMs with TCNs and patching-based strategies on UCR and UEA archives validates their advantages. Furthermore, our experimental results across various downstream tasks suggest that the patch strategy~\cite{nietime} and the iTransformer architecture~\cite{liuitransformer} offer promising research avenues.}

{}{Although significant work has applied transformers to time series pre-training~\cite{nietime,goswamimoment,woo2024unified}, some CNN-based models also deserve attention~\cite{luo2024moderntcn}. Notably, Eldele et al.~\cite{eldeletslanet} demonstrated that their CNN-based model outperforms existing Transformer-based approaches on various time series downstream tasks. Additionally, the Mamba model, based on state space models, shows strong sequence learning capabilities for pre-training~\cite{gu2023mamba}, outperforming Transformers in many CV domain tasks~\cite{zhuvision}. Consequently, researchers in time series analysis are increasingly exploring Mamba for time series mining~\cite{patro2024simba,wang2024mamba}. Compared to these deep learning models, GNNs offer the ability to learn spatio-temporal dependencies in time series pre-training, especially when combined with models like Transformers~\cite{shao2022pre}. In summary, selecting the appropriate deep learning architectures will be crucial for future designs of time series pre-training models.}

% for time-series forecasting, and Anomaly Transformer~\cite{xu2021anomaly} for time-series anomaly detection are well suited for analyzing time series.
% However, there are relatively few existing Transformer models with competitive advantages for the time-series classification task, which may be the fact that the time-series classification task is more focused on capturing discriminative subseries (e.g., shapelets) or multiscale features of time-series, where existing CNN-based models~\cite{tang2021omni} may be more advantageous.

% In addition, the large-scale pre-training transformer models have been widely studied and achieved remarkable results in various tasks of NLP and CV. 
% Although there has been some work applying transformers to TSM tasks~\cite{wen2022transformers}, there is little research on PTMs for time series. We believe that the current challenges with pre-training Transformer models for time series are two-fold. First, pre-training transformers usually require 
% massive data for pre-training to learn the universal representations. However, there is currently a lack of large-scale datasets in the time series field. Second, how to design an effective pre-trained transformer model combined with the inherent properties of time series still needs further exploration. Therefore, exploring pre-training transformer models for time series is an exciting research direction.

\subsection{Adversarial Attacks on Time Series}

Adversarial example attacks have recently received extensive attention in various fields because they have significant security risks. 
Naturally, scholars in the time-series domain have also begun to consider the impact of adversarial sample attacks on time-series models~\cite{wu2022small,liurobust2023}. 
For example, Karim et al.~\cite{karim2020adversarial} utilized a distilled model as a surrogate that simulated the behavior of the attacked time-series classification model. Then, an adversarial transformation network is applied to the distilled model to generate time-series adversarial examples. Experiments on $42$ UCR datasets show they are vulnerable to adversarial examples.

Adversarial examples are generated by adding perturbations to the original examples to cross the classification boundaries of the classifier.
In general, adding random perturbations is difficult to generate adversarial examples. 
Also, adversarial examples are not easy to generate when each cluster is far from the classification boundary. 
Recently, Hendrycks et al.~\cite{hendrycks2019using} found that self-supervised learning could effectively improve the robustness of deep learning models to adversarial examples. Therefore, improving the robustness of time series models to adversarial examples by utilizing TS-PTMs is a direction worth exploring.

\subsection{Pre-Training Models for Time-Series Noisy Labels}

The acquisition cost of large-scale labeled datasets is very expensive. Therefore, various low-cost surrogate strategies are provided to collect labels automatically. For example, many weakly labeled images can be collected with the help of search engines and crawling algorithms. In the time-series domain,  Castellani et al.~\cite{castellani2021estimating} employed sensor readings to generate labels for time series. Although these strategies enable obtaining large-scale labeled data, they also inevitably lead to label noise. Training deep learning models efficiently with noisy labels is challenging since deep learning models have a high capacity for fitting noisy labels~\cite{song2022learning}.
To this end, many studies on learning with label noise~\cite{han2018co} have emerged. As an effective representation learning method, PTMs can be effectively employed to solve the problem of noisy labels~\cite{li2020dividemix}, which has been studied in CV. However, only a few studies are investigating the time-series noisy label~\cite{liu2023scale}, and PTMs for time-series noisy labels have not been studied.

\subsection{{}{Pre-Trained LLMs for Time-Series Mining}}

{}{Pre-trained LLMs have achieved significant success in NLP~\cite{kalyan2021ammus}, particularly with GPT-based models like ChatGPT, which are widely used in real life. Scholars~\cite{jin2023large,jintime,liang2024foundation} have demonstrated that LLMs can be effectively applied to time series analysis, with GPT4TS~\cite{zhou2023one} showing for the first time that combining the patching mechanism and fine-tuning some parameters of the GPT2 model can yield strong performance across various time series tasks. However, the reasons behind the effectiveness of fine-tuning LLMs directly on time series data require further investigation. Notably, Tan et al.~\cite{tan2024language} found that pre-trained LLMs do not outperform models trained from scratch in time series forecasting tasks, and a simple model with patching and attention as an encoder can achieve comparable results.}

{}{Unlike approaches that directly fine-tune LLMs on time series tasks, we tend to think that the pre-training and fine-tuning paradigm using time series datasets for designing TS-PTMs will remain the dominant research approach~\cite{liu2024timer,woo2024unified}. This paradigm aligns with human learning processes and has been validated in CV and NLP fields. However, the potential of LLMs in time series mining should not be overlooked. To better leverage LLMs for time series analysis, converting time series into textual descriptions for fine-tuning might more effectively transfer the knowledge acquired during LLM pre-training to time series tasks~\cite{xue2023promptcast}. Also, integrating textual descriptions with the time series data for multi-modal pre-training~\cite{liu2024diffusion,radford2021learning} is a promising direction worth exploring.}

\section{Conclusion~\label{section6}}

In this survey, we provide a systematic review and analysis of the development of
TS-PTMs. In the early research on TS-PTMs, related studies were
mainly based on CNN and RNN models for transfer learning on PTMs.
In recent years, Transformer-based, consistency-based models {}{and patching-based strategies} have achieved remarkable performance in time-series downstream tasks, and have been utilized for time series pre-training.
Hence, we conducted a large-scale experimental analysis of existing TS-PTMs, transfer learning strategies, Transformer-based time series methods, and related representative methods on the three main tasks of time-series classification, forecasting, and anomaly detection.
{}{The experimental results suggest that LLM-based fine-tuning PTMs, when combined with patching strategies and Transformer-based models, hold significant potential for time-series classification and forecasting tasks. Additionally, consistency-based PTMs using patching strategies demonstrate promising results for time-series anomaly detection.
% designing suitable Transformer-based models for the time-series classification task remains challenging.
Meanwhile, the pre-training strategy involving the selection of an appropriate deep learning model, such as CNNs or Mamba, represents a promising direction for the future development of TS-PTMs.}

% or


% \appendices

% \section{}

\ifCLASSOPTIONcompsoc

\section*{Acknowledgments}
The work described in this paper was partially funded by the National Natural Science Foundation of China (Grant Nos. 62272173, 61872148), the Natural Science Foundation of Guangdong Province (Grant Nos. 2022A1515010179, 2019A1515010768).
The authors would like to thank 
Professor Garrison W. Cottrell from UCSD, and Yu Chen, Peitian Ma from SCUT for the review and helpful suggestions.
\else

  \section*{Acknowledgments}
We thank the anonymous reviewers for their helpful feed-
back. 
  We thank Professor Eamonn Keogh from UCR and all the people who have contributed to the UCR\&UEA time series archives and other time series datasets.
The authors would like to thank 
Professor Garrison W. Cottrell from UCSD, and Chuxin Chen, Xidi Cai, Yu Chen, and Peitian Ma from SCUT for the helpful suggestions.
\fi


\ifCLASSOPTIONcaptionsoff
  \newpage
\fi



%\bibliographystyle{IEEEtran}


% \end{thebibliography}

\bibliographystyle{IEEEtran}
\bibliography{reference}

\begin{thebibliography}{100}
\providecommand{\url}[1]{#1}
\csname url@samestyle\endcsname
\providecommand{\newblock}{\relax}
\providecommand{\bibinfo}[2]{#2}
\providecommand{\BIBentrySTDinterwordspacing}{\spaceskip=0pt\relax}
\providecommand{\BIBentryALTinterwordstretchfactor}{4}
\providecommand{\BIBentryALTinterwordspacing}{\spaceskip=\fontdimen2\font plus
\BIBentryALTinterwordstretchfactor\fontdimen3\font minus \fontdimen4\font\relax}
\providecommand{\BIBforeignlanguage}[2]{{%
\expandafter\ifx\csname l@#1\endcsname\relax
\typeout{** WARNING: IEEEtran.bst: No hyphenation pattern has been}%
\typeout{** loaded for the language `#1'. Using the pattern for}%
\typeout{** the default language instead.}%
\else
\language=\csname l@#1\endcsname
\fi
#2}}
\providecommand{\BIBdecl}{\relax}
\BIBdecl

\bibitem{wu2013dynamic}
Y.~Wu, J.~M. Hern{\'a}ndez-Lobato, and G.~Zoubin, ``Dynamic covariance models for multivariate financial time series,'' in \emph{International Conference on Machine Learning}.\hskip 1em plus 0.5em minus 0.4em\relax PMLR, 2013, pp. 558--566.

\bibitem{moritz2020streaming}
N.~Moritz, T.~Hori, and J.~Le, ``Streaming automatic speech recognition with the transformer model,'' in \emph{IEEE International Conference on Acoustics, Speech and Signal Processing}, 2020, pp. 6074--6078.

\bibitem{martinez2017human}
J.~Martinez, M.~J. Black, and J.~Romero, ``On human motion prediction using recurrent neural networks,'' in \emph{Proceedings of the IEEE Conference on Computer Vision and Pattern Recognition}, 2017, pp. 2891--2900.

\bibitem{wang2020deep}
S.~Wang, J.~Cao, and P.~S. Yu, ``Deep learning for spatio-temporal data mining: A survey,'' \emph{IEEE Transactions on Knowledge and Data Engineering}, vol.~34, no.~8, pp. 3681--3700, 2022.

\bibitem{tedjopurnomo2020survey}
D.~A. Tedjopurnomo, Z.~Bao, B.~Zheng, F.~Choudhury, and A.~K. Qin, ``A survey on modern deep neural network for traffic prediction: Trends, methods and challenges,'' \emph{IEEE Transactions on Knowledge and Data Engineering}, vol.~34, no.~04, pp. 1544--1561, 2022.

\bibitem{fu2011review}
T.-c. Fu, ``A review on time series data mining,'' \emph{Engineering Applications of Artificial Intelligence}, vol.~24, no.~1, pp. 164--181, 2011.

\bibitem{li2020efficient}
G.~Li, B.~K.~K. Choi, J.~Xu, S.~S. Bhowmick, K.-P. Chun, and G.~L. Wong, ``Efficient shapelet discovery for time series classification,'' \emph{IEEE Transactions on Knowledge and Data Engineering}, vol.~34, no.~03, pp. 1149--1163, 2022.

\bibitem{che2018recurrent}
Z.~Che, S.~Purushotham, K.~Cho, D.~Sontag, and Y.~Liu, ``Recurrent neural networks for multivariate time series with missing values,'' \emph{Scientific Reports}, vol.~8, no.~1, pp. 1--12, 2018.

\bibitem{tahan2022development}
M.~H. Tahan, M.~Ghasemzadeh, and S.~Asadi, ``Development of fully convolutional neural networks based on discretization in time series classification,'' \emph{IEEE Transactions on Knowledge and Data Engineering}, vol.~35, no.~7, pp. 6827--6838, 2023.

\bibitem{sen2019think}
R.~Sen, H.-F. Yu, and I.~S. Dhillon, ``Think globally, act locally: {A} deep neural network approach to high-dimensional time series forecasting,'' in \emph{Advances in Neural Information Processing Systems}, 2019, pp. 1--10.

\bibitem{zhou2021informer}
H.~Zhou, S.~Zhang, J.~Peng, S.~Zhang, J.~Li, H.~Xiong, and W.~Zhang, ``Informer: {B}eyond efficient transformer for long sequence time-series forecasting,'' in \emph{Proceedings of AAAI Conference on Artificial Intelligence}, vol.~35, no.~12, 2021, pp. 11\,106--11\,115.

\bibitem{wen2021time}
Q.~Wen, L.~Sun, F.~Yang, X.~Song, J.~Gao, X.~Wang, and H.~Xu, ``Time series data augmentation for deep learning: {A} survey,'' in \emph{Proceedings of the Thirtieth International Joint Conference on Artificial Intelligence}, 2021, pp. 4653--4660.

\bibitem{iwana2021empirical}
B.~K. Iwana and S.~Uchida, ``An empirical survey of data augmentation for time series classification with neural networks,'' \emph{Plos one}, vol.~16, no.~7, p. e0254841, 2021.

\bibitem{van2020survey}
J.~E. Van~Engelen and H.~H. Hoos, ``A survey on semi-supervised learning,'' \emph{Machine Learning}, vol. 109, no.~2, pp. 373--440, 2020.

\bibitem{liu2023temporal}
Z.~Liu, Q.~Ma, P.~Ma, and L.~Wang, ``Temporal-frequency co-training for time series semi-supervised learning,'' in \emph{Proceedings of the AAAI Conference on Artificial Intelligence}, vol.~37, no.~7, 2023, pp. 8923--8931.

\bibitem{shorten2019survey}
C.~Shorten and T.~M. Khoshgoftaar, ``A survey on image data augmentation for deep learning,'' \emph{Journal of big data}, vol.~6, no.~1, pp. 1--48, 2019.

\bibitem{yang2022spectral}
L.~Yang, S.~Hong, and L.~Zhang, ``Spectral propagation graph network for few-shot time series classification,'' \emph{arXiv preprint arXiv:2202.04769}, 2022.

\bibitem{pan2009survey}
S.~J. Pan and Q.~Yang, ``A survey on transfer learning,'' \emph{IEEE Transactions on Knowledge and Data Engineering}, vol.~22, no.~10, pp. 1345--1359, 2009.

\bibitem{zhuang2020comprehensive}
F.~Zhuang, Z.~Qi, K.~Duan, D.~Xi, Y.~Zhu, H.~Zhu, H.~Xiong, and Q.~He, ``A comprehensive survey on transfer learning,'' \emph{Proceedings of the IEEE}, vol. 109, no.~1, pp. 43--76, 2020.

\bibitem{krizhevsky2012imagenet}
A.~Krizhevsky, I.~Sutskever, and G.~E. Hinton, ``Imagenet classification with deep convolutional neural networks,'' \emph{Advances in Neural Information Processing Systems}, vol.~25, pp. 1097--1105, 2012.

\bibitem{he2021masked}
K.~He, X.~Chen, S.~Xie, Y.~Li, P.~Doll{\'a}r, and R.~Girshick, ``Masked autoencoders are scalable vision learners,'' \emph{arXiv preprint arXiv:2111.06377}, 2021.

\bibitem{qiu2020pre}
X.~Qiu, T.~Sun, Y.~Xu, Y.~Shao, N.~Dai, and X.~Huang, ``Pre-trained models for natural language processing: {A} survey,'' \emph{Science China Technological Sciences}, pp. 1--26, 2020.

\bibitem{fawaz2018transfer}
H.~I. Fawaz, G.~Forestier, J.~Weber, L.~Idoumghar, and P.-A. Muller, ``Transfer learning for time series classification,'' in \emph{2018 IEEE International Conference on Big Data}, 2018, pp. 1367--1376.

\bibitem{yang2021voice2series}
C.-H.~H. Yang, Y.-Y. Tsai, and P.-Y. Chen, ``Voice2series: Reprogramming acoustic models for time series classification,'' in \emph{Proceedings of the 38th International Conference on Machine Learning}, vol. 139.\hskip 1em plus 0.5em minus 0.4em\relax PMLR, 2021, pp. 11\,808--11\,819.

\bibitem{malhotra2017timenet}
P.~Malhotra, V.~TV, L.~Vig, P.~Agarwal, and G.~Shroff, ``Timenet: Pre-trained deep recurrent neural network for time series classification,'' \emph{arXiv preprint arXiv:1706.08838}, 2017.

\bibitem{zerveas2021transformer}
G.~Zerveas, S.~Jayaraman, D.~Patel, A.~Bhamidipaty, and C.~Eickhoff, ``A transformer-based framework for multivariate time series representation learning,'' in \emph{Proceedings of the 27th ACM SIGKDD Conference on Knowledge Discovery \& Data Mining}, 2021, pp. 2114--2124.

\bibitem{deldari2022beyond}
S.~Deldari, H.~Xue, A.~Saeed, J.~He, D.~V. Smith, and F.~D. Salim, ``Beyond just vision: A review on self-supervised representation learning on multimodal and temporal data,'' \emph{arXiv preprint arXiv:2206.02353}, 2022.

\bibitem{zhang2022self}
X.~Zhang, Z.~Zhao, T.~Tsiligkaridis, and M.~Zitnik, ``Self-supervised contrastive pre-training for time series via time-frequency consistency,'' in \emph{Advances in neural information processing systems}, 2022, pp. 1--16.

\bibitem{zhang2022cross}
W.~Zhang, L.~Yang, S.~Geng, and S.~Hong, ``Cross reconstruction transformer for self-supervised time series representation learning,'' \emph{arXiv preprint arXiv:2205.09928}, 2022.

\bibitem{ye2018novel}
R.~Ye and Q.~Dai, ``A novel transfer learning framework for time series forecasting,'' \emph{Knowledge-Based Systems}, vol. 156, pp. 74--99, 2018.

\bibitem{he2020momentum}
K.~He, H.~Fan, Y.~Wu, S.~Xie, and R.~Girshick, ``Momentum contrast for unsupervised visual representation learning,'' in \emph{Proceedings of the IEEE/CVF Conference on Computer Vision and Pattern Recognition}, 2020, pp. 9729--9738.

\bibitem{chen2020simple}
T.~Chen, S.~Kornblith, M.~Norouzi, and G.~Hinton, ``A simple framework for contrastive learning of visual representations,'' in \emph{International Conference on Machine Learning}.\hskip 1em plus 0.5em minus 0.4em\relax PMLR, 2020, pp. 1597--1607.

\bibitem{zhang2021sleeppriorcl}
H.~Zhang, J.~Wang, Q.~Xiao, J.~Deng, and Y.~Lin, ``Sleeppriorcl/: Contrastive representation learning with prior knowledge-based positive mining and adaptive temperature for sleep staging,'' \emph{arXiv preprint arXiv:2110.09966}, 2021.

\bibitem{laptev2018reconstruction}
N.~Laptev, J.~Yu, and R.~Rajagopal, ``Reconstruction and regression loss for time-series transfer learning,'' in \emph{Proceedings of the Special Interest Group on Knowledge Discovery and Data Mining (SIGKDD) and the 4th Workshop on the Mining and LEarning from Time Series (MiLeTS)}, 2018, pp. 1--8.

\bibitem{eldele2024label}
E.~Eldele, M.~Ragab, Z.~Chen, M.~Wu, C.-K. Kwoh, and X.~Li, ``Label-efficient time series representation learning: A review,'' \emph{IEEE Transactions on Artificial Intelligence}, pp. 1--16, 2024.

\bibitem{zhang2024self}
K.~Zhang, Q.~Wen, C.~Zhang, R.~Cai, M.~Jin, Y.~Liu, J.~Y. Zhang, Y.~Liang, G.~Pang, D.~Song, and S.~Pan, ``Self-supervised learning for time series analysis: Taxonomy, progress, and prospects,'' \emph{IEEE Transactions on Pattern Analysis and Machine Intelligence}, vol.~46, no.~10, pp. 6775--6794, 2024.

\bibitem{meng2023unsupervised}
Q.~Meng, H.~Qian, Y.~Liu, Y.~Xu, Z.~Shen, and L.~Cui, ``Unsupervised representation learning for time series: A review,'' \emph{arXiv preprint arXiv:2308.01578}, 2023.

\bibitem{jin2023large}
M.~Jin, Q.~Wen, Y.~Liang, C.~Zhang, S.~Xue, X.~Wang, J.~Zhang, Y.~Wang, H.~Chen, X.~Li \emph{et~al.}, ``Large models for time series and spatio-temporal data: A survey and outlook,'' \emph{arXiv preprint arXiv:2310.10196}, 2023.

\bibitem{jin2024position}
M.~Jin, Y.~Zhang, W.~Chen, K.~Zhang, Y.~Liang, B.~Yang, J.~Wang, S.~Pan, and Q.~Wen, ``Position paper: What can large language models tell us about time series analysis,'' in \emph{In Forty-first International Conference on Machine Learning}, 2024, pp. 1--17.

\bibitem{liang2024foundation}
Y.~Liang, H.~Wen, Y.~Nie, Y.~Jiang, M.~Jin, D.~Song, S.~Pan, and Q.~Wen, ``Foundation models for time series analysis: A tutorial and survey,'' in \emph{In Proceedings of the 30th ACM SIGKDD Conference on Knowledge Discovery and Data Mining}, 2024, p. 6555–6565.

\bibitem{ismail2019deep}
H.~Ismail~Fawaz, G.~Forestier, J.~Weber, L.~Idoumghar, and P.-A. Muller, ``Deep learning for time series classification: a review,'' \emph{Data mining and knowledge discovery}, vol.~33, no.~4, pp. 917--963, 2019.

\bibitem{lim2021time}
B.~Lim and S.~Zohren, ``Time-series forecasting with deep learning: {A} survey,'' \emph{Philosophical Transactions of the Royal Society A}, vol. 379, no. 2194, p. 20200209, 2021.

\bibitem{lafabregue2021end}
B.~Lafabregue, J.~Weber, P.~Gan{\c{c}}arski, and G.~Forestier, ``End-to-end deep representation learning for time series clustering: {A} comparative study,'' \emph{Data Mining and Knowledge Discovery}, pp. 1--53, 2021.

\bibitem{liu2020anomaly}
F.~Liu, X.~Zhou, J.~Cao, Z.~Wang, T.~Wang, H.~Wang, and Y.~Zhang, ``Anomaly detection in quasi-periodic time series based on automatic data segmentation and attentional lstm-cnn,'' \emph{IEEE Transactions on Knowledge and Data Engineering}, vol.~34, no.~06, pp. 2626--2640, 2022.

\bibitem{blazquez2021review}
A.~Bl{\'a}zquez-Garc{\'\i}a, A.~Conde, U.~Mori, and J.~A. Lozano, ``A review on outlier/anomaly detection in time series data,'' \emph{ACM Computing Surveys (CSUR)}, vol.~54, no.~3, pp. 1--33, 2021.

\bibitem{cao2018brits}
W.~Cao, D.~Wang, J.~Li, H.~Zhou, L.~Li, and Y.~Li, ``{BRITS}: {B}idirectional recurrent imputation for time series,'' \emph{arXiv preprint arXiv:1805.10572}, 2018.

\bibitem{de2008clustering}
M.~C. De~Souto, I.~G. Costa, D.~S. De~Araujo, T.~B. Ludermir, and A.~Schliep, ``Clustering cancer gene expression data: {A} comparative study,'' \emph{BMC Bioinformatics}, vol.~9, no.~1, pp. 1--14, 2008.

\bibitem{de2019deep}
J.~de~Jong, M.~A. Emon, P.~Wu, R.~Karki, M.~Sood, P.~Godard, A.~Ahmad, H.~Vrooman, M.~Hofmann-Apitius, and H.~Fr{\"o}hlich, ``Deep learning for clustering of multivariate clinical patient trajectories with missing values,'' \emph{GigaScience}, vol.~8, no.~11, p. giz134, 2019.

\bibitem{tan2021time}
C.~W. Tan, C.~Bergmeir, F.~Petitjean, and G.~I. Webb, ``Time series extrinsic regression: Predicting numeric values from time series data,'' \emph{Data Mining and Knowledge Discovery}, vol.~35, no.~3, pp. 1032--1060, 2021.

\bibitem{xu2023reinforced}
Q.~Xu, K.~Wu, M.~Wu, K.~Mao, X.~Li, and Z.~Chen, ``Reinforced knowledge distillation for time series regression,'' \emph{IEEE Transactions on Artificial Intelligence}, vol.~5, no.~6, pp. 3184--3194, 2024.

\bibitem{li2023difformer}
B.~Li, W.~Cui, L.~Zhang, C.~Zhu, W.~Wang, I.~W. Tsang, and J.~T. Zhou, ``Difformer: Multi-resolutional differencing transformer with dynamic ranging for time series analysis,'' \emph{IEEE Transactions on Pattern Analysis and Machine Intelligence}, vol.~45, no.~11, pp. 13\,586--13\,598, 2023.

\bibitem{tan2020monash}
C.~W. Tan, C.~Bergmeir, F.~Petitjean, and G.~I. Webb, ``Monash university, uea, ucr time series extrinsic regression archive,'' \emph{arXiv preprint arXiv:2006.10996}, 2020.

\bibitem{chung2014empirical}
J.~Chung, C.~Gulcehre, K.~Cho, and Y.~Bengio, ``Empirical evaluation of gated recurrent neural networks on sequence modeling,'' \emph{arXiv preprint arXiv:1412.3555}, 2014.

\bibitem{greff2016lstm}
K.~Greff, R.~K. Srivastava, J.~Koutn{\'\i}k, B.~R. Steunebrink, and J.~Schmidhuber, ``Lstm: A search space odyssey,'' \emph{IEEE transactions on neural networks and learning systems}, vol.~28, no.~10, pp. 2222--2232, 2016.

\bibitem{muralidhar2019dyat}
N.~Muralidhar, S.~Muthiah, and N.~Ramakrishnan, ``Dyat nets: Dynamic attention networks for state forecasting in cyber-physical systems.'' in \emph{International Joint Conference on Artificial Intelligence}, 2019, pp. 3180--3186.

\bibitem{ma2019learning}
Q.~Ma, J.~Zheng, S.~Li, and G.~W. Cottrell, ``Learning representations for time series clustering,'' \emph{Advances in Neural Information Processing Systems}, vol.~32, pp. 3781--3791, 2019.

\bibitem{gu2018recent}
J.~Gu, Z.~Wang, J.~Kuen, L.~Ma, A.~Shahroudy, B.~Shuai, T.~Liu, X.~Wang, G.~Wang, J.~Cai \emph{et~al.}, ``Recent advances in convolutional neural networks,'' \emph{Pattern recognition}, vol.~77, pp. 354--377, 2018.

\bibitem{liu2022scinet}
M.~Liu, A.~Zeng, M.~Chen, Z.~Xu, Q.~Lai, L.~Ma, and Q.~Xu, ``Scinet: Time series modeling and forecasting with sample convolution and interaction,'' \emph{Advances in Neural Information Processing Systems}, vol.~35, pp. 5816--5828, 2022.

\bibitem{cui2016multi}
Z.~Cui, W.~Chen, and Y.~Chen, ``Multi-scale convolutional neural networks for time series classification,'' \emph{arXiv preprint arXiv:1603.06995}, 2016.

\bibitem{kashiparekh2019convtimenet}
K.~Kashiparekh, J.~Narwariya, P.~Malhotra, L.~Vig, and G.~Shroff, ``Convtimenet: A pre-trained deep convolutional neural network for time series classification,'' in \emph{2019 International Joint Conference on Neural Networks (IJCNN)}.\hskip 1em plus 0.5em minus 0.4em\relax IEEE, 2019, pp. 1--8.

\bibitem{bai2018empirical}
S.~Bai, J.~Z. Kolter, and V.~Koltun, ``An empirical evaluation of generic convolutional and recurrent networks for sequence modeling,'' \emph{arXiv preprint arXiv:1803.01271}, 2018.

\bibitem{long2015fully}
J.~Long, E.~Shelhamer, and T.~Darrell, ``Fully convolutional networks for semantic segmentation,'' in \emph{Proceedings of the IEEE conference on computer vision and pattern recognition}, 2015, pp. 3431--3440.

\bibitem{chen2020probabilistic}
Y.~Chen, Y.~Kang, Y.~Chen, and Z.~Wang, ``Probabilistic forecasting with temporal convolutional neural network,'' \emph{Neurocomputing}, vol. 399, pp. 491--501, 2020.

\bibitem{xu2021autoformer}
J.~Xu, J.~Wang, M.~Long \emph{et~al.}, ``Autoformer: {D}ecomposition transformers with auto-correlation for long-term series forecasting,'' \emph{Advances in Neural Information Processing Systems}, vol.~34, pp. 1--12, 2021.

\bibitem{zhou2022fedformer}
T.~Zhou, Z.~Ma, Q.~Wen, X.~Wang, L.~Sun, and R.~Jin, ``Fedformer: Frequency enhanced decomposed transformer for long-term series forecasting,'' \emph{arXiv preprint arXiv:2201.12740}, 2022.

\bibitem{wu2020comprehensive}
Z.~Wu, S.~Pan, F.~Chen, G.~Long, C.~Zhang, and S.~Y. Philip, ``A comprehensive survey on graph neural networks,'' \emph{IEEE transactions on neural networks and learning systems}, vol.~32, no.~1, pp. 4--24, 2020.

\bibitem{cheng2021time2graph}
Z.~Cheng, Y.~Yang, S.~Jiang, W.~Hu, Z.~Ying, Z.~Chai, and C.~Wang, ``Time2graph+: Bridging time series and graph representation learning via multiple attentions,'' \emph{IEEE Transactions on Knowledge and Data Engineering}, vol.~35, no.~2, pp. 2078--2090, 2021.

\bibitem{wang2024graph}
Y.~Wang, Y.~Xu, J.~Yang, M.~Wu, X.~Li, L.~Xie, and Z.~Chen, ``Graph-aware contrasting for multivariate time-series classification,'' in \emph{Proceedings of the AAAI Conference on Artificial Intelligence}, vol.~38, no.~14, 2024, pp. 15\,725--15\,734.

\bibitem{zha2022towards}
D.~Zha, K.-H. Lai, K.~Zhou, and X.~Hu, ``Towards similarity-aware time-series classification,'' in \emph{Proceedings of the 2022 SIAM International Conference on Data Mining (SDM)}.\hskip 1em plus 0.5em minus 0.4em\relax SIAM, 2022, pp. 199--207.

\bibitem{deng2009imagenet}
J.~Deng, W.~Dong, R.~Socher, L.-J. Li, K.~Li, and L.~Fei-Fei, ``Imagenet: {A} large-scale hierarchical image database,'' in \emph{IEEE Conference on Computer Vision and Pattern Recognition}, 2009, pp. 248--255.

\bibitem{serra2018towards}
J.~Serr{\`a}, S.~Pascual, and A.~Karatzoglou, ``Towards a universal neural network encoder for time series.'' in \emph{Artificial Intelligence Research and Development}, 2018, pp. 120--129.

\bibitem{UCRArchive}
Y.~Chen, E.~Keogh, B.~Hu, N.~Begum, A.~Bagnall, A.~Mueen, and G.~Batista, ``The {UCR} time series classification archive,'' July 2015, \url{www.cs.ucr.edu/~eamonn/time_series_data/}.

\bibitem{wang2017time}
Z.~Wang, W.~Yan, and T.~Oates, ``Time series classification from scratch with deep neural networks: A strong baseline,'' in \emph{2017 International joint conference on neural networks (IJCNN)}, 2017, pp. 1578--1585.

\bibitem{ismail2020inceptiontime}
H.~Ismail~Fawaz, B.~Lucas, G.~Forestier, C.~Pelletier, D.~F. Schmidt, J.~Weber, G.~I. Webb, L.~Idoumghar, P.-A. Muller, and F.~Petitjean, ``Inceptiontime: Finding alexnet for time series classification,'' \emph{Data Mining and Knowledge Discovery}, vol.~34, no.~6, pp. 1936--1962, 2020.

\bibitem{tang2021omni}
W.~Tang, G.~Long, L.~Liu, T.~Zhou, M.~Blumenstein, and J.~Jiang, ``Omni-scale cnns: a simple and effective kernel size configuration for time series classification,'' in \emph{International Conference on Learning Representations}, 2022, pp. 1--17.

\bibitem{li2020deep}
F.~Li, K.~Shirahama, M.~A. Nisar, X.~Huang, and M.~Grzegorzek, ``Deep transfer learning for time series data based on sensor modality classification,'' \emph{Sensors}, vol.~20, no.~15, p. 4271, 2020.

\bibitem{yue2022ts2vec}
Z.~Yue, Y.~Wang, J.~Duan, T.~Yang, C.~Huang, Y.~Tong, and B.~Xu, ``Ts2vec: Towards universal representation of time series,'' \emph{Proceedings of Advancement of Artificial Intelligence Conference on Artificial Intelligence}, vol.~36, no.~8, pp. 8980--8987, 2022.

\bibitem{ye2009time}
L.~Ye and E.~Keogh, ``Time series shapelets: a new primitive for data mining,'' in \emph{Proceedings of the ACM SIGKDD international conference on Knowledge discovery and data mining}, 2009, pp. 947--956.

\bibitem{meiseles2020source}
A.~Meiseles and L.~Rokach, ``Source model selection for deep learning in the time series domain,'' \emph{IEEE Access}, vol.~8, pp. 6190--6200, 2020.

\bibitem{mutegeki2019feature}
R.~Mutegeki and D.~S. Han, ``Feature-representation transfer learning for human activity recognition,'' in \emph{2019 International Conference on Information and Communication Technology Convergence (ICTC)}.\hskip 1em plus 0.5em minus 0.4em\relax IEEE, 2019, pp. 18--20.

\bibitem{cai2021time}
R.~Cai, J.~Chen, Z.~Li, W.~Chen, K.~Zhang, J.~Ye, Z.~Li, X.~Yang, and Z.~Zhang, ``Time series domain adaptation via sparse associative structure alignment,'' in \emph{Proceedings of the AAAI Conference on Artificial Intelligence}, vol.~35, no.~8, 2021, pp. 6859--6867.

\bibitem{wilson2020multi}
G.~Wilson, J.~R. Doppa, and D.~J. Cook, ``Multi-source deep domain adaptation with weak supervision for time-series sensor data,'' in \emph{Proceedings of the 26th ACM SIGKDD International Conference on Knowledge Discovery \& Data Mining}, 2020, pp. 1768--1778.

\bibitem{csurka2017domain}
G.~Csurka, ``Domain adaptation for visual applications: A comprehensive survey,'' \emph{arXiv preprint arXiv:1702.05374}, 2017.

\bibitem{chen2020homm}
C.~Chen, Z.~Fu, Z.~Chen, S.~Jin, Z.~Cheng, X.~Jin, and X.-S. Hua, ``Homm: Higher-order moment matching for unsupervised domain adaptation,'' in \emph{Proceedings of the AAAI conference on artificial intelligence}, vol.~34, no.~04, 2020, pp. 3422--3429.

\bibitem{xie2018learning}
S.~Xie, Z.~Zheng, L.~Chen, and C.~Chen, ``Learning semantic representations for unsupervised domain adaptation,'' in \emph{International conference on machine learning}.\hskip 1em plus 0.5em minus 0.4em\relax PMLR, 2018, pp. 5423--5432.

\bibitem{yan2017mind}
H.~Yan, Y.~Ding, P.~Li, Q.~Wang, Y.~Xu, and W.~Zuo, ``Mind the class weight bias: Weighted maximum mean discrepancy for unsupervised domain adaptation,'' in \emph{Proceedings of the IEEE conference on computer vision and pattern recognition}, 2017, pp. 2272--2281.

\bibitem{you2019universal}
K.~You, M.~Long, Z.~Cao, J.~Wang, and M.~I. Jordan, ``Universal domain adaptation,'' in \emph{Proceedings of the IEEE/CVF conference on computer vision and pattern recognition}, 2019, pp. 2720--2729.

\bibitem{wang2018stratified}
J.~Wang, Y.~Chen, L.~Hu, X.~Peng, and S.~Y. Philip, ``Stratified transfer learning for cross-domain activity recognition,'' in \emph{2018 IEEE International Conference on Pervasive Computing and Communications (PerCom)}.\hskip 1em plus 0.5em minus 0.4em\relax IEEE, 2018, pp. 1--10.

\bibitem{li2021transferable}
Z.~Li, R.~Cai, T.~Z. Fu, and K.~Zhang, ``Transferable time-series forecasting under causal conditional shift,'' \emph{arXiv preprint arXiv:2111.03422}, 2021.

\bibitem{liu2021adversarial}
Q.~Liu and H.~Xue, ``Adversarial spectral kernel matching for unsupervised time series domain adaptation,'' in \emph{International Joint Conference on Artificial Intelligence}, 2021, pp. 2744--2750.

\bibitem{khan2018scaling}
M.~A. A.~H. Khan, N.~Roy, and A.~Misra, ``Scaling human activity recognition via deep learning-based domain adaptation,'' in \emph{2018 IEEE international conference on pervasive computing and communications (PerCom)}, 2018, pp. 1--9.

\bibitem{tank2021neural}
A.~Tank, I.~Covert, N.~Foti, A.~Shojaie, and E.~B. Fox, ``Neural granger causality,'' \emph{IEEE Transactions on Pattern Analysis and Machine Intelligence}, vol.~44, no.~8, pp. 4267--4279, 2021.

\bibitem{ragab2021self}
M.~Ragab, E.~Eldele, Z.~Chen, M.~Wu, C.-K. Kwoh, and X.~Li, ``Self-supervised autoregressive domain adaptation for time series data,'' \emph{arXiv preprint arXiv:2111.14834}, 2021.

\bibitem{da2020remaining}
P.~R. d.~O. da~Costa, A.~Ak{\c{c}}ay, Y.~Zhang, and U.~Kaymak, ``Remaining useful lifetime prediction via deep domain adaptation,'' \emph{Reliability Engineering \& System Safety}, vol. 195, p. 106682, 2020.

\bibitem{wilson2021calda}
W.~Garrett, R.~D. Janardhan, and J.~C. Diane, ``Calda: {I}mproving multi-source time series domain adaptation with contrastive adversarial learning,'' \emph{arXiv preprint arXiv:2109.14778}, 2021.

\bibitem{li2021causal}
Z.~Li, R.~Cai, H.~W. Ng, M.~Winslett, T.~Z. Fu, B.~Xu, X.~Yang, and Z.~Zhang, ``Causal mechanism transfer network for time series domain adaptation in mechanical systems,'' \emph{ACM Transactions on Intelligent Systems and Technology (TIST)}, vol.~12, no.~2, pp. 1--21, 2021.

\bibitem{elsayed2018adversarial}
G.~F. Elsayed, I.~Goodfellow, and J.~Sohl-Dickstein, ``Adversarial reprogramming of neural networks,'' in \emph{International Conference on Learning Representations}, 2019, pp. 1--14.

\bibitem{yang2021decentralizing}
C.-H.~H. Yang, J.~Qi, S.~Y.-C. Chen, P.-Y. Chen, S.~M. Siniscalchi, X.~Ma, and C.-H. Lee, ``Decentralizing feature extraction with quantum convolutional neural network for automatic speech recognition,'' in \emph{IEEE International Conference on Acoustics, Speech and Signal Processing}.\hskip 1em plus 0.5em minus 0.4em\relax IEEE, 2021, pp. 6523--6527.

\bibitem{zhou2023one}
T.~Zhou, P.~Niu, L.~Sun, R.~Jin \emph{et~al.}, ``One fits all: Power general time series analysis by pretrained lm,'' \emph{Advances in neural information processing systems}, vol.~36, pp. 43\,322--43\,355, 2023.

\bibitem{caotempo}
D.~Cao, F.~Jia, S.~O. Arik, T.~Pfister, Y.~Zheng, W.~Ye, and Y.~Liu, ``Tempo: Prompt-based generative pre-trained transformer for time series forecasting,'' in \emph{The Twelfth International Conference on Learning Representations}, 2024, pp. 1--33.

\bibitem{jintime}
M.~Jin, S.~Wang, L.~Ma, Z.~Chu, J.~Y. Zhang, X.~Shi, P.-Y. Chen, Y.~Liang, Y.-F. Li, S.~Pan \emph{et~al.}, ``Time-llm: Time series forecasting by reprogramming large language models,'' in \emph{The Twelfth International Conference on Learning Representations}, 2024, pp. 1--24.

\bibitem{jiangsequential}
X.~Jiang, R.~Missel, Z.~Li, and L.~Wang, ``Sequential latent variable models for few-shot high-dimensional time-series forecasting,'' in \emph{The Eleventh International Conference on Learning Representations}, 2023, pp. 1--21.

\bibitem{liu2024timer}
Y.~Liu, H.~Zhang, C.~Li, X.~Huang, J.~Wang, and M.~Long, ``Timer: Generative pre-trained transformers are large time series models,'' in \emph{Forty-first International Conference on Machine Learning}, 2024, pp. 1--31.

\bibitem{woo2024unified}
G.~Woo, C.~Liu, A.~Kumar, C.~Xiong, S.~Savarese, and D.~Sahoo, ``Unified training of universal time series forecasting transformers,'' in \emph{Forty-first International Conference on Machine Learning}, 2024, pp. 1--25.

\bibitem{dasdecoder}
A.~Das, W.~Kong, R.~Sen, and Y.~Zhou, ``A decoder-only foundation model for time-series forecasting,'' in \emph{Forty-first International Conference on Machine Learning}, 2024, pp. 1--21.

\bibitem{mallick2021transfer}
T.~Mallick, P.~Balaprakash, E.~Rask, and J.~Macfarlane, ``Transfer learning with graph neural networks for short-term highway traffic forecasting,'' in \emph{2020 25th International Conference on Pattern Recognition (ICPR)}.\hskip 1em plus 0.5em minus 0.4em\relax IEEE, 2021, pp. 10\,367--10\,374.

\bibitem{du2021adarnn}
Y.~Du, J.~Wang, W.~Feng, S.~Pan, T.~Qin, R.~Xu, and C.~Wang, ``Adarnn: Adaptive learning and forecasting of time series,'' in \emph{Proceedings of the 30th ACM International Conference on Information \& Knowledge Management}, 2021, pp. 402--411.

\bibitem{oord2018representation}
A.~v.~d. Oord, Y.~Li, and O.~Vinyals, ``Representation learning with contrastive predictive coding,'' \emph{arXiv preprint arXiv:1807.03748}, 2018.

\bibitem{schneider2019wav2vec}
S.~Schneider, A.~Baevski, R.~Collobert, and M.~Auli, ``wav2vec: {U}nsupervised pre-training for speech recognition,'' \emph{Proc. Interspeech 2019}, pp. 3465--3469, 2019.

\bibitem{eldele2021time}
E.~Eldele, M.~Ragab, Z.~Chen, M.~Wu, C.~K. Kwoh, X.~Li, and C.~Guan, ``Time-series representation learning via temporal and contextual contrasting,'' in \emph{Proceedings of the Thirtieth International Joint Conference on Artificial Intelligence}, 2021, pp. 2352--2359.

\bibitem{vinyals2016matching}
O.~Vinyals, C.~Blundell, T.~Lillicrap, D.~Wierstra \emph{et~al.}, ``Matching networks for one shot learning,'' \emph{Advances in Neural Information Processing Systems}, vol.~29, pp. 1--9, 2016.

\bibitem{finn2017model}
C.~Finn, P.~Abbeel, and S.~Levine, ``Model-agnostic meta-learning for fast adaptation of deep networks,'' in \emph{International Conference on Machine Learning}.\hskip 1em plus 0.5em minus 0.4em\relax PMLR, 2017, pp. 1126--1135.

\bibitem{lu2022spatio}
B.~Lu, X.~Gan, W.~Zhang, H.~Yao, L.~Fu, and X.~Wang, ``Spatio-temporal graph few-shot learning with cross-city knowledge transfer,'' in \emph{Proceedings of the 28th ACM SIGKDD Conference on Knowledge Discovery and Data Mining}, 2022, pp. 1162--1172.

\bibitem{wang2022meta}
R.~Wang, R.~Walters, and R.~Yu, ``Meta-learning dynamics forecasting using task inference,'' \emph{Advances in Neural Information Processing Systems}, vol.~35, pp. 21\,640--21\,653, 2022.

\bibitem{iwata2020few}
T.~Iwata and A.~Kumagai, ``Few-shot learning for time-series forecasting,'' \emph{arXiv preprint arXiv:2009.14379}, 2020.

\bibitem{oreshkin2021meta}
B.~N. Oreshkin, D.~Carpov, N.~Chapados, and Y.~Bengio, ``Meta-learning framework with applications to zero-shot time-series forecasting,'' in \emph{Proceedings of the AAAI Conference on Artificial Intelligence}, vol.~35, no.~10, 2021, pp. 9242--9250.

\bibitem{brinkmeyer2022few}
L.~Brinkmeyer, R.~R. Drumond, J.~Burchert, and L.~Schmidt-Thieme, ``Few-shot forecasting of time-series with heterogeneous channels,'' \emph{arXiv preprint arXiv:2204.03456}, 2022.

\bibitem{castellani2021estimating}
A.~Castellani, S.~Schmitt, and B.~Hammer, ``Estimating the electrical power output of industrial devices with end-to-end time-series classification in the presence of label noise,'' in \emph{European Conference on Machine Learning}, 2021, pp. 1--32.

\bibitem{sutskever2014sequence}
I.~Sutskever, O.~Vinyals, and Q.~V. Le, ``Sequence to sequence learning with neural networks,'' \emph{Advances in Neural Information Processing Systems}, vol.~27, pp. 1--9, 2014.

\bibitem{vincent2010stacked}
P.~Vincent, H.~Larochelle, I.~Lajoie, Y.~Bengio, P.-A. Manzagol, and L.~Bottou, ``Stacked denoising autoencoders: Learning useful representations in a deep network with a local denoising criterion.'' \emph{Journal of Machine Learning Research}, vol.~11, no.~12, 2010.

\bibitem{baevski2019vq}
A.~Baevski, S.~Schneider, and M.~Auli, ``vq-wav2vec: {S}elf-supervised learning of discrete speech representations,'' \emph{arXiv preprint arXiv:1910.05453}, 2019.

\bibitem{ma2021learning}
Q.~Ma, C.~Chen, S.~Li, and G.~W. Cottrell, ``Learning representations for incomplete time series clustering,'' in \emph{Proceedings of the Advancement of Artificial Intelligence Conference on Artificial Intelligence}, vol.~35, no.~10, 2021, pp. 8837--8846.

\bibitem{nietime}
Y.~Nie, N.~H. Nguyen, P.~Sinthong, and J.~Kalagnanam, ``A time series is worth 64 words: Long-term forecasting with transformers,'' in \emph{The Eleventh International Conference on Learning Representations}, 2023, pp. 1--24.

\bibitem{dong2024simmtm}
J.~Dong, H.~Wu, H.~Zhang, L.~Zhang, J.~Wang, and M.~Long, ``Simmtm: A simple pre-training framework for masked time-series modeling,'' \emph{Advances in Neural Information Processing Systems}, vol.~36, pp. 29\,996--30\,025, 2023.

\bibitem{zhang2023trid}
K.~Zhang, C.~Li, and Q.~Yang, ``Trid-mae: A generic pre-trained model for multivariate time series with missing values,'' in \emph{Proceedings of the 32nd ACM International Conference on Information and Knowledge Management}, 2023, pp. 3164--3173.

\bibitem{chung2016audio}
Y.-A. Chung, C.-C. Wu, C.-H. Shen, H.-Y. Lee, and L.-S. Lee, ``Audio word2vec: Unsupervised learning of audio segment representations using sequence-to-sequence autoencoder,'' \emph{arXiv preprint arXiv:1603.00982}, 2016.

\bibitem{hu2016transfer}
Q.~Hu, R.~Zhang, and Y.~Zhou, ``Transfer learning for short-term wind speed prediction with deep neural networks,'' \emph{Renewable Energy}, vol.~85, pp. 83--95, 2016.

\bibitem{shao2022pre}
Z.~Shao, Z.~Zhang, F.~Wang, and Y.~Xu, ``Pre-training enhanced spatial-temporal graph neural network for multivariate time series forecasting,'' in \emph{Proceedings of the 28th ACM SIGKDD Conference on Knowledge Discovery and Data Mining}, 2022, pp. 1567--1577.

\bibitem{devlin2019bert}
J.~Devlin, M.-W. Chang, K.~Lee, and K.~Toutanova, ``Bert: Pre-training of deep bidirectional transformers for language understanding,'' in \emph{Proceedings of the 2019 Conference of the North American Chapter of the Association for Computational Linguistics: Human Language Technologies}, 2019, pp. 4171--4186.

\bibitem{dong2024timesiam}
J.~Dong, H.~Wu, Y.~Wang, Y.~Qiu, L.~Zhang, J.~Wang, and M.~Long, ``Timesiam: A pre-training framework for siamese time-series modeling,'' in \emph{Forty-first International Conference on Machine Learning}, 2024, pp. 1--25.

\bibitem{eldeletslanet}
E.~Eldele, M.~Ragab, Z.~Chen, M.~Wu, and X.~Li, ``Tslanet: Rethinking transformers for time series representation learning,'' in \emph{Forty-first International Conference on Machine Learning}, 2024, pp. 1--20.

\bibitem{zhangup2me}
Y.~Zhang, M.~Liu, S.~Zhou, and J.~Yan, ``Up2me: Univariate pre-training to multivariate fine-tuning as a general-purpose framework for multivariate time series analysis,'' in \emph{Forty-first International Conference on Machine Learning}, 2024, pp. 1--24.

\bibitem{goswamimoment}
M.~Goswami, K.~Szafer, A.~Choudhry, Y.~Cai, S.~Li, and A.~Dubrawski, ``Moment: A family of open time-series foundation models,'' in \emph{Forty-first International Conference on Machine Learning}, 2024, pp. 1--38.

\bibitem{chowdhury2022tarnet}
R.~R. Chowdhury, X.~Zhang, J.~Shang, R.~K. Gupta, and D.~Hong, ``Tarnet: Task-aware reconstruction for time-series transformer,'' in \emph{Proceedings of the 28th ACM SIGKDD Conference on Knowledge Discovery and Data Mining, Washington, DC, USA}, 2022, pp. 14--18.

\bibitem{liu2021tera}
A.~T. Liu, S.-W. Li, and H.-y. Lee, ``{TERA}: Self-supervised learning of transformer encoder representation for speech,'' \emph{IEEE/ACM Transactions on Audio, Speech, and Language Processing}, vol.~29, pp. 2351--2366, 2021.

\bibitem{kalyan2021ammus}
K.~S. Kalyan, A.~Rajasekharan, and S.~Sangeetha, ``Ammus: A survey of transformer-based pretrained models in natural language processing,'' \emph{arXiv preprint arXiv:2108.05542}, 2021.

\bibitem{vaswani2017attention}
A.~Vaswani, N.~Shazeer, N.~Parmar, J.~Uszkoreit, L.~Jones, A.~N. Gomez, {\L}.~Kaiser, and I.~Polosukhin, ``Attention is all you need,'' \emph{Advances in neural information processing systems}, vol.~30, pp. 1--11, 2017.

\bibitem{shi2021self}
P.~Shi, W.~Ye, and Z.~Qin, ``Self-supervised pre-training for time series classification,'' in \emph{2021 International Joint Conference on Neural Networks (IJCNN)}.\hskip 1em plus 0.5em minus 0.4em\relax IEEE, 2021, pp. 1--8.

\bibitem{hou2022masked}
L.~Hou, Y.~Geng, L.~Han, H.~Yang, K.~Zheng, and X.~Wang, ``Masked token enabled pre-training: A task-agnostic approach for understanding complex traffic flow,'' \emph{IEEE Transactions on Mobile Computing}, pp. 1--12, 2024.

\bibitem{zhao2022st}
L.~Zhao, M.~Gao, and Z.~Wang, ``St-gsp: Spatial-temporal global semantic representation learning for urban flow prediction,'' in \emph{Proceedings of the Fifteenth ACM International Conference on Web Search and Data Mining}, 2022, pp. 1443--1451.

\bibitem{padhi2021tabular}
I.~Padhi, Y.~Schiff, I.~Melnyk, M.~Rigotti, Y.~Mroueh, P.~Dognin, J.~Ross, R.~Nair, and E.~Altman, ``Tabular transformers for modeling multivariate time series,'' in \emph{IEEE International Conference on Acoustics, Speech and Signal Processing}.\hskip 1em plus 0.5em minus 0.4em\relax IEEE, 2021, pp. 3565--3569.

\bibitem{shankaranarayana2021attention}
S.~M. Shankaranarayana and D.~Runje, ``Attention augmented convolutional transformer for tabular time-series,'' in \emph{2021 International Conference on Data Mining Workshops (ICDMW)}.\hskip 1em plus 0.5em minus 0.4em\relax IEEE, 2021, pp. 537--541.

\bibitem{liu2020mockingjay}
A.~T. Liu, S.-w. Yang, P.-H. Chi, P.-c. Hsu, and H.-y. Lee, ``Mockingjay: Unsupervised speech representation learning with deep bidirectional transformer encoders,'' in \emph{IEEE International Conference on Acoustics, Speech and Signal Processing}, 2020, pp. 6419--6423.

\bibitem{wang2021dense}
X.~Wang, R.~Zhang, C.~Shen, T.~Kong, and L.~Li, ``Dense contrastive learning for self-supervised visual pre-training,'' in \emph{Proceedings of the IEEE/CVF Conference on Computer Vision and Pattern Recognition}, 2021, pp. 3024--3033.

\bibitem{yang2023dcdetector}
Y.~Yang, C.~Zhang, T.~Zhou, Q.~Wen, and L.~Sun, ``Dcdetector: Dual attention contrastive representation learning for time series anomaly detection,'' in \emph{Proceedings of the 29th ACM SIGKDD Conference on Knowledge Discovery and Data Mining}, 2023, pp. 3033--3045.

\bibitem{hu2023self}
M.~Hu, Z.~Zhong, X.~Zhang, Y.~Li, Y.~Xie, X.~Jia, X.~Zhou, and J.~Luo, ``Self-supervised pre-training for robust and generic spatial-temporal representations,'' in \emph{2023 IEEE International Conference on Data Mining (ICDM)}.\hskip 1em plus 0.5em minus 0.4em\relax IEEE, 2023, pp. 150--159.

\bibitem{liu2024timesurl}
J.~Liu and S.~Chen, ``Timesurl: Self-supervised contrastive learning for universal time series representation learning,'' in \emph{Proceedings of the AAAI Conference on Artificial Intelligence}, vol.~38, no.~12, 2024, pp. 13\,918--13\,926.

\bibitem{mikolov2013distributed}
T.~Mikolov, I.~Sutskever, K.~Chen, G.~S. Corrado, and J.~Dean, ``Distributed representations of words and phrases and their compositionality,'' \emph{Advances in Neural Information Processing Systems}, vol.~26, pp. 1--9, 2013.

\bibitem{franceschi2019unsupervised}
J.-Y. Franceschi, A.~Dieuleveut, and M.~Jaggi, ``Unsupervised scalable representation learning for multivariate time series,'' \emph{Advances in Neural Information Processing Systems}, vol.~32, pp. 1--12, 2019.

\bibitem{fan2020self}
H.~Fan, F.~Zhang, and Y.~Gao, ``Self-supervised time series representation learning by inter-intra relational reasoning,'' \emph{arXiv preprint arXiv:2011.13548}, 2020.

\bibitem{tonekaboni2020unsupervised}
S.~Tonekaboni, D.~Eytan, and A.~Goldenberg, ``Unsupervised representation learning for time series with temporal neighborhood coding,'' in \emph{International Conference on Learning Representations}, 2021, pp. 1--17.

\bibitem{woo2022cost}
G.~Woo, C.~Liu, D.~Sahoo, A.~Kumar, and S.~Hoi, ``Co{ST}: Contrastive learning of disentangled seasonal-trend representations for time series forecasting,'' in \emph{International Conference on Learning Representations}, 2022, pp. 1--18.

\bibitem{deldari2021time}
S.~Deldari, D.~V. Smith, H.~Xue, and F.~D. Salim, ``Time series change point detection with self-supervised contrastive predictive coding,'' in \emph{Proceedings of the Web Conference 2021}, 2021, pp. 3124--3135.

\bibitem{luo2021information}
D.~Luo, W.~Cheng, Y.~Wang, D.~Xu, J.~Ni, W.~Yu, X.~Zhang, Y.~Liu, H.~Chen, and X.~Zhang, ``Information-aware time series meta-contrastive learning,'' \emph{Submitted to International Conference on Learning Representations}, pp. 1--23, 2022.

\bibitem{hyvarinen2016unsupervised}
A.~Hyvarinen and H.~Morioka, ``Unsupervised feature extraction by time-contrastive learning and nonlinear ica,'' \emph{Advances in Neural Information Processing Systems}, vol.~29, pp. 1--9, 2016.

\bibitem{hyvarinen2017nonlinear}
H.~M. Aapo~Hyvarinen, ``Nonlinear ica of temporally dependent stationary sources,'' in \emph{Artificial Intelligence and Statistics}.\hskip 1em plus 0.5em minus 0.4em\relax PMLR, 2017, pp. 460--469.

\bibitem{yang2022unsupervised}
L.~Yang, S.~Hong, and L.~Zhang, ``Unsupervised time-series representation learning with iterative bilinear temporal-spectral fusion,'' \emph{arXiv preprint arXiv:2202.04770}, 2022.

\bibitem{wen2020time}
Q.~Wen, L.~Sun, F.~Yang, X.~Song, J.~Gao, X.~Wang, and H.~Xu, ``Time series data augmentation for deep learning: A survey,'' \emph{arXiv preprint arXiv:2002.12478}, 2020.

\bibitem{nonnenmacher2022utilizing}
M.~T. Nonnenmacher, L.~Oldenburg, I.~Steinwart, and D.~Reeb, ``Utilizing expert features for contrastive learning of time-series representations,'' in \emph{International Conference on Machine Learning}.\hskip 1em plus 0.5em minus 0.4em\relax PMLR, 2022, pp. 16\,969--16\,989.

\bibitem{asano2020self}
Y.~Asano, C.~Rupprecht, and A.~Vedaldi, ``Self-labelling via simultaneous clustering and representation learning,'' in \emph{International Conference on Learning Representations}, 2020, pp. 1--22.

\bibitem{gidaris2018unsupervised}
S.~Gidaris, P.~Singh, and N.~Komodakis, ``Unsupervised representation learning by predicting image rotations,'' in \emph{Proceedings of International Conference on Learning Representations}, 2018, pp. 1--17.

\bibitem{chung2018speech2vec}
Y.-A. Chung and J.~Glass, ``Speech2vec: A sequence-to-sequence framework for learning word embeddings from speech,'' \emph{arXiv preprint arXiv:1803.08976}, 2018.

\bibitem{dau2019ucr}
H.~A. Dau, A.~Bagnall, K.~Kamgar, C.-C.~M. Yeh, Y.~Zhu, S.~Gharghabi, C.~A. Ratanamahatana, and E.~Keogh, ``The ucr time series archive,'' \emph{IEEE/CAA Journal of Automatica Sinica}, vol.~6, no.~6, pp. 1293--1305, 2019.

\bibitem{bagnall2018uea}
A.~Bagnall, H.~A. Dau, J.~Lines, M.~Flynn, J.~Large, A.~Bostrom, P.~Southam, and E.~Keogh, ``The uea multivariate time series classification archive\, 2018,'' \emph{arXiv preprint arXiv:1811.00075}, 2018.

\bibitem{wutimesnet}
H.~Wu, T.~Hu, Y.~Liu, H.~Zhou, J.~Wang, and M.~Long, ``Timesnet: Temporal 2d-variation modeling for general time series analysis,'' in \emph{The Eleventh International Conference on Learning Representations}, 2023, pp. 1--23.

\bibitem{YahooTime}
Y.~B. Nikolay~Laptev, Saeed~Amizadeh, ``A benchmark dataset for time series anomaly detection,'' 2015, \url{https://yahooresearch.tumblr.com/post/114590420346/a-benchmark-dataset-for-time-series-anomaly}.

\bibitem{ren2019time}
H.~Ren, B.~Xu, Y.~Wang, C.~Yi, C.~Huang, X.~Kou, T.~Xing, M.~Yang, J.~Tong, and Q.~Zhang, ``Time-series anomaly detection service at microsoft,'' in \emph{Proceedings of the 25th ACM SIGKDD international conference on knowledge discovery \& data mining}, 2019, pp. 3009--3017.

\bibitem{wu2021current}
R.~Wu and E.~J. Keogh, ``Current time series anomaly detection benchmarks are flawed and are creating the illusion of progress,'' \emph{IEEE transactions on knowledge and data engineering}, vol.~35, no.~3, pp. 2421--2429, 2021.

\bibitem{huet2022local}
A.~Huet, J.~M. Navarro, and D.~Rossi, ``Local evaluation of time series anomaly detection algorithms,'' in \emph{Proceedings of the 28th ACM SIGKDD Conference on Knowledge Discovery and Data Mining}, 2022, pp. 635--645.

\bibitem{paparrizos2022volume}
J.~Paparrizos, P.~Boniol, T.~Palpanas, R.~S. Tsay, A.~Elmore, and M.~J. Franklin, ``Volume under the surface: a new accuracy evaluation measure for time-series anomaly detection,'' \emph{Proceedings of the VLDB Endowment}, vol.~15, no.~11, pp. 2774--2787, 2022.

\bibitem{li2019enhancing}
S.~Li, X.~Jin, Y.~Xuan, X.~Zhou, W.~Chen, Y.-X. Wang, and X.~Yan, ``Enhancing the locality and breaking the memory bottleneck of transformer on time series forecasting,'' \emph{Advances in Neural Information Processing Systems}, vol.~32, pp. 1--11, 2019.

\bibitem{liuitransformer}
Y.~Liu, T.~Hu, H.~Zhang, H.~Wu, S.~Wang, L.~Ma, and M.~Long, ``itransformer: Inverted transformers are effective for time series forecasting,'' in \emph{The Twelfth International Conference on Learning Representations}, 2024, pp. 1--25.

\bibitem{zeng2023transformers}
A.~Zeng, M.~Chen, L.~Zhang, and Q.~Xu, ``Are transformers effective for time series forecasting?'' in \emph{Proceedings of the AAAI conference on artificial intelligence}, vol.~37, no.~9, 2023, pp. 11\,121--11\,128.

\bibitem{kim2022towards}
S.~Kim, K.~Choi, H.-S. Choi, B.~Lee, and S.~Yoon, ``Towards a rigorous evaluation of time-series anomaly detection,'' in \emph{Proceedings of the AAAI Conference on Artificial Intelligence}, vol.~36, no.~7, 2022, pp. 7194--7201.

\bibitem{siffer2017anomaly}
A.~Siffer, P.-A. Fouque, A.~Termier, and C.~Largouet, ``Anomaly detection in streams with extreme value theory,'' in \emph{Proceedings of the 23rd ACM SIGKDD International Conference on Knowledge Discovery and Data Mining}, 2017, pp. 1067--1075.

\bibitem{park2018multimodal}
D.~Park, Y.~Hoshi, and C.~C. Kemp, ``A multimodal anomaly detector for robot-assisted feeding using an {LSTM}-based variational autoencoder,'' \emph{IEEE Robotics and Automation Letters}, vol.~3, no.~3, pp. 1544--1551, 2018.

\bibitem{xu2018unsupervised}
H.~Xu, W.~Chen, N.~Zhao, Z.~Li, J.~Bu, Z.~Li, Y.~Liu, Y.~Zhao, D.~Pei, Y.~Feng \emph{et~al.}, ``Unsupervised anomaly detection via variational auto-encoder for seasonal kpis in web applications,'' in \emph{Proceedings of the 2018 world wide web conference}, 2018, pp. 187--196.

\bibitem{xu2021anomaly}
J.~Xu, H.~Wu, J.~Wang, and M.~Long, ``Anomaly transformer: Time series anomaly detection with association discrepancy,'' in \emph{International Conference on Learning Representations}, 2022, pp. 1--20.

\bibitem{chen2023contiformer}
Y.~Chen, K.~Ren, Y.~Wang, Y.~Fang, W.~Sun, and D.~Li, ``Contiformer: continuous-time transformer for irregular time series modeling,'' in \emph{Proceedings of the 37th International Conference on Neural Information Processing Systems}, 2023, pp. 47\,143--47\,175.

\bibitem{luo2024moderntcn}
D.~Luo and X.~Wang, ``Moderntcn: A modern pure convolution structure for general time series analysis,'' in \emph{The Twelfth International Conference on Learning Representations}, 2024, pp. 1--43.

\bibitem{gu2023mamba}
A.~Gu and T.~Dao, ``Mamba: Linear-time sequence modeling with selective state spaces,'' \emph{arXiv preprint arXiv:2312.00752}, 2023.

\bibitem{zhuvision}
L.~Zhu, B.~Liao, Q.~Zhang, X.~Wang, W.~Liu, and X.~Wang, ``Vision mamba: Efficient visual representation learning with bidirectional state space model,'' in \emph{Forty-first International Conference on Machine Learning}, 2024, pp. 1--11.

\bibitem{patro2024simba}
B.~N. Patro and V.~S. Agneeswaran, ``Simba: Simplified mamba-based architecture for vision and multivariate time series,'' \emph{arXiv preprint arXiv:2403.15360}, 2024.

\bibitem{wang2024mamba}
Z.~Wang, F.~Kong, S.~Feng, M.~Wang, H.~Zhao, D.~Wang, and Y.~Zhang, ``Is mamba effective for time series forecasting?'' \emph{arXiv preprint arXiv:2403.11144}, 2024.

\bibitem{wu2022small}
T.~Wu, X.~Wang, S.~Qiao, X.~Xian, Y.~Liu, and L.~Zhang, ``Small perturbations are enough: Adversarial attacks on time series prediction,'' \emph{Information Sciences}, vol. 587, pp. 794--812, 2022.

\bibitem{liurobust2023}
L.~Liu, Y.~Park, T.~N. Hoang, H.~Hasson, and L.~Huan, ``Robust multivariate time-series forecasting: Adversarial attacks and defense mechanisms,'' in \emph{The Eleventh International Conference on Learning Representations}, 2023, pp. 1--18.

\bibitem{karim2020adversarial}
F.~Karim, S.~Majumdar, and H.~Darabi, ``Adversarial attacks on time series,'' \emph{IEEE transactions on pattern analysis and machine intelligence}, vol.~43, no.~10, pp. 3309--3320, 2020.

\bibitem{hendrycks2019using}
D.~Hendrycks, M.~Mazeika, S.~Kadavath, and D.~Song, ``Using self-supervised learning can improve model robustness and uncertainty,'' \emph{Advances in Neural Information Processing Systems}, vol.~32, pp. 1--12, 2019.

\bibitem{song2022learning}
H.~Song, M.~Kim, D.~Park, Y.~Shin, and J.-G. Lee, ``Learning from noisy labels with deep neural networks: A survey,'' \emph{IEEE Transactions on Neural Networks and Learning Systems}, vol.~34, no.~11, pp. 8135--8153, 2023.

\bibitem{han2018co}
B.~Han, Q.~Yao, X.~Yu, G.~Niu, M.~Xu, W.~Hu, I.~Tsang, and M.~Sugiyama, ``Co-teaching: Robust training of deep neural networks with extremely noisy labels,'' in \emph{Advances in neural information processing systems}, vol.~31, 2018, pp. 1--11.

\bibitem{li2020dividemix}
J.~Li, R.~Socher, and S.~C. Hoi, ``Dividemix: Learning with noisy labels as semi-supervised learning,'' in \emph{International Conference on Learning Representations}, 2020, pp. 1--14.

\bibitem{liu2023scale}
Z.~Liu, P.~Ma, D.~Chen, W.~Pei, and Q.~Ma, ``Scale-teaching: robust multi-scale training for time series classification with noisy labels,'' in \emph{Proceedings of the 37th International Conference on Neural Information Processing Systems}, 2023, pp. 33\,726--33\,757.

\bibitem{tan2024language}
M.~Tan, M.~A. Merrill, V.~Gupta, T.~Althoff, and T.~Hartvigsen, ``Are language models actually useful for time series forecasting?'' \emph{arXiv preprint arXiv:2406.16964}, 2024.

\bibitem{xue2023promptcast}
H.~Xue and F.~D. Salim, ``Promptcast: A new prompt-based learning paradigm for time series forecasting,'' \emph{IEEE Transactions on Knowledge and Data Engineering}, pp. 1--14, 2023.

\bibitem{liu2024diffusion}
Z.~Liu, W.~Pei, D.~Lan, and Q.~Ma, ``Diffusion language-shapelets for semi-supervised time-series classification,'' in \emph{Proceedings of the AAAI Conference on Artificial Intelligence}, vol.~38, no.~13, 2024, pp. 14\,079--14\,087.

\bibitem{radford2021learning}
A.~Radford, J.~W. Kim, C.~Hallacy, A.~Ramesh, G.~Goh, S.~Agarwal, G.~Sastry, A.~Askell, P.~Mishkin, J.~Clark \emph{et~al.}, ``Learning transferable visual models from natural language supervision,'' in \emph{International conference on machine learning}.\hskip 1em plus 0.5em minus 0.4em\relax PMLR, 2021, pp. 8748--8763.

\bibitem{ozyurt2022contrastive}
Y.~Ozyurt, S.~Feuerriegel, and C.~Zhang, ``Contrastive learning for unsupervised domain adaptation of time series,'' in \emph{The Eleventh International Conference on Learning Representations}, 2023, pp. 1--40.

\bibitem{morales2016deep}
F.~J.~O. Morales and D.~Roggen, ``Deep convolutional feature transfer across mobile activity recognition domains, sensor modalities and locations,'' in \emph{Proceedings of the 2016 ACM International Symposium on Wearable Computers}, 2016, pp. 92--99.

\bibitem{purushotham2017variational}
S.~Purushotham, W.~Carvalho, T.~Nilanon, and Y.~Liu, ``Variational recurrent adversarial deep domain adaptation,'' in \emph{International Conference on Learning Representations}, 2017, pp. 1--15.

\bibitem{xiong2018application}
P.~Xiong, Y.~Zhu, Z.~Sun, Z.~Cao, M.~Wang, Y.~Zheng, J.~Hou, T.~Huang, and Z.~Que, ``Application of transfer learning in continuous time series for anomaly detection in commercial aircraft flight data,'' in \emph{2018 IEEE International Conference on Smart Cloud (SmartCloud)}.\hskip 1em plus 0.5em minus 0.4em\relax IEEE, 2018, pp. 13--18.

\bibitem{narwariya2020meta}
J.~Narwariya, P.~Malhotra, L.~Vig, G.~Shroff, and T.~Vishnu, ``Meta-learning for few-shot time series classification,'' in \emph{Proceedings of the 7th ACM IKDD CoDS and 25th COMAD}, 2020, pp. 28--36.

\bibitem{baireddy2021spacecraft}
S.~Baireddy, S.~R. Desai, J.~L. Mathieson, R.~H. Foster, M.~W. Chan, M.~L. Comer, and E.~J. Delp, ``Spacecraft time-series anomaly detection using transfer learning,'' in \emph{Proceedings of the IEEE/CVF Conference on Computer Vision and Pattern Recognition}, 2021, pp. 1951--1960.

\bibitem{shifaz2020ts}
A.~Shifaz, C.~Pelletier, F.~Petitjean, and G.~I. Webb, ``Ts-chief: a scalable and accurate forest algorithm for time series classification,'' \emph{Data Mining and Knowledge Discovery}, vol.~34, no.~3, pp. 742--775, 2020.

\bibitem{dempster2021minirocket}
A.~Dempster, D.~F. Schmidt, and G.~I. Webb, ``Minirocket: A very fast (almost) deterministic transform for time series classification,'' in \emph{Proceedings of the 27th ACM SIGKDD conference on knowledge discovery \& data mining}, 2021, pp. 248--257.

\bibitem{zhou2016learning}
B.~Zhou, A.~Khosla, A.~Lapedriza, A.~Oliva, and A.~Torralba, ``Learning deep features for discriminative localization,'' in \emph{Proceedings of the IEEE conference on computer vision and pattern recognition}, 2016, pp. 2921--2929.

\bibitem{lines2012shapelet}
J.~Lines, L.~M. Davis, J.~Hills, and A.~Bagnall, ``A shapelet transform for time series classification,'' in \emph{Proceedings of the 18th ACM SIGKDD international conference on Knowledge discovery and data mining}, 2012, pp. 289--297.

\end{thebibliography}


% \end{IEEEbiography}


%\newpage



%\vfill



% % \vspace{11pt}

% % \bf{If you will not include a photo:}\vspace{-33pt}
% \newpage

\begin{IEEEbiography}[{\includegraphics[width=1in,height=1.25in,clip,keepaspectratio]{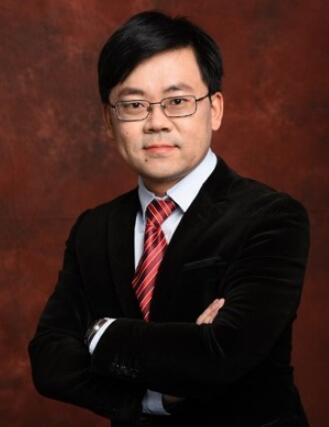}}]{Qianli Ma} (Member, IEEE) received the Ph.D. degree in computer science from the South China University of Technology, Guangzhou, China, in 2008. He is a Professor with the School of Computer Science and Engineering, South China University of Technology. From 2016 to 2017, he was a Visiting Scholar with the University of California, San Diego. His current research interests include machine learning algorithms, data-mining methodologies, and their applications. He is an associate editor of IEEE/ACM Transactions on Audio, Speech, and Language Processing. He has been recognized among the World's Top 2\% Scientists for 2023 and 2024.
\end{IEEEbiography}

\begin{IEEEbiography}[{\includegraphics[width=1in,height=1.25in,clip,keepaspectratio]{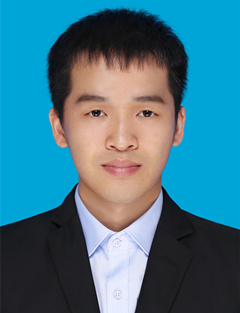}}]{Zhen Liu}
received the bachelor’s degree in software engineering from the South-Central Minzu University, Wuhan, China, in 2018. He is currently pursuing the  Ph.D. degree with the School of Computer Science and Engineering, South China University of Technology, Guangzhou, China.
His current research interests include machine
learning, deep learning, and time-series analysis.
\end{IEEEbiography}

\begin{IEEEbiography}[{\includegraphics[width=1in,height=1.25in,clip,keepaspectratio]{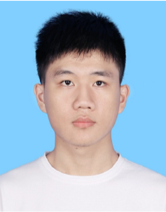}}]{Zhenjing Zheng}
received the bachelor's and master's degree in computer science from the South China University of Technology, Guangzhou, China, in 2019 and 2022, respectively.
His current research interests include machine learning, deep learning, and time-series analysis.
\end{IEEEbiography}

\newpage

\begin{IEEEbiography}[{\includegraphics[width=1in,height=1.25in,clip,keepaspectratio]{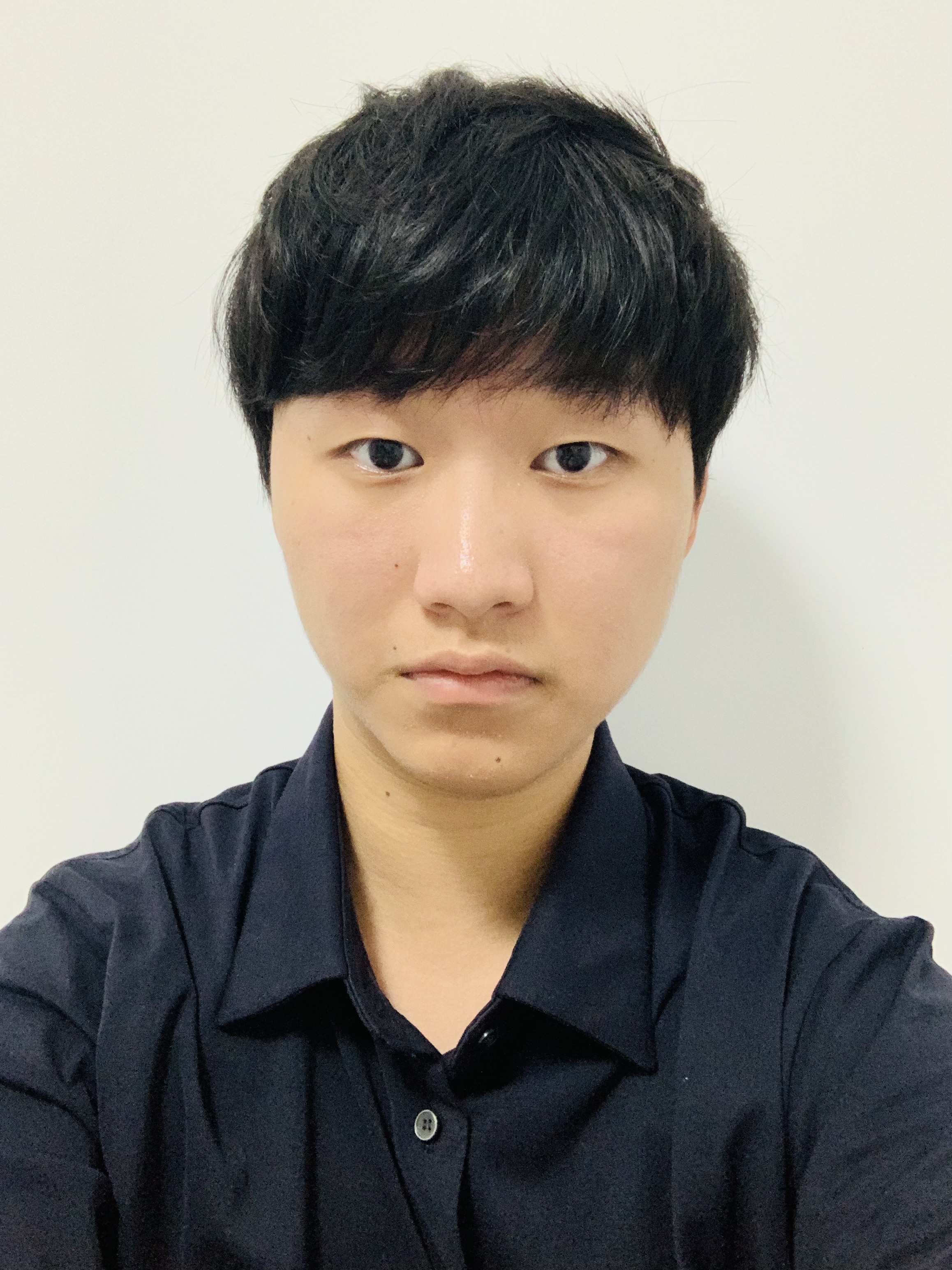}}] {Ziyang Huang} received the bachelor’s degree from the School of Computer Science and Engineering, South China University of Technology, Guangzhou, China{}{, in 2023}.
His current research interests include machine
learning, deep learning, and time-series analysis.
\end{IEEEbiography}

\begin{IEEEbiography}[{\includegraphics[width=1in,height=1.25in,clip,keepaspectratio]{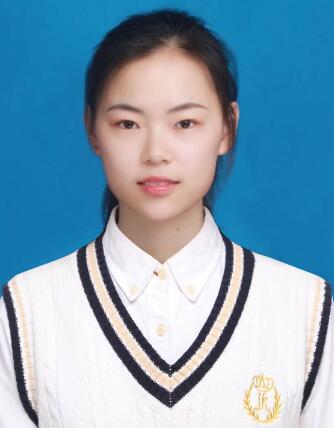}}]{Siying Zhu}
received the bachelor’s degree in computer science from the Southwest University, Chongqing, China, in 2020. 
She {}{received} the master’s degree from the School of Computer Science and Engineering, South China University of Technology, Guangzhou, China{}{, in 2023.}
Her current research interests include machine learning, deep learning, and time-series forecasting.
\end{IEEEbiography}

\begin{IEEEbiography}[{\includegraphics[width=1in,height=1.25in,clip,keepaspectratio]{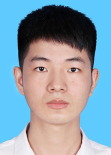}}]{Zhongzhong Yu}
received the bachelor's and master's degree in computer science from the South China University of Technology, Guangzhou, China, in 2019 and 2022, respectively.
His current research interests include machine learning, deep learning, and time-series anomaly detection.
\end{IEEEbiography}

\begin{IEEEbiography}[{\includegraphics[width=1in,height=1.25in,clip,keepaspectratio]{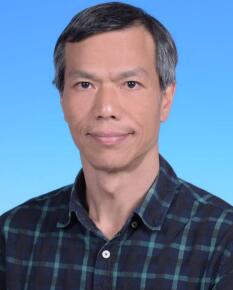}}]{James T. Kwok} (Fellow, IEEE) received the PhD degree in computer science from The Hong Kong University of Science and Technology, Hong Kong, in 1996. He is currently a Professor at the Department of Computer Science and Engineering, Hong Kong University of Science and Technology, Hong Kong. His research interests include machine learning, deep learning, and artificial intelligence. He is serving / served as an associate editor for the IEEE Transactions on Neural Networks and Learning Systems, Neural Networks, Neurocomputing, He is also serving / served as senior area chairs / area chairs of major conferences including NeurIPS, ICML, ICLR, IJCAI, and AAAI. He is on the IJCAI Board of Trustees. He is recognized as the Most Influential Scholar Award Honorable Mention for ``outstanding and vibrant contributions to the field of AAAI/IJCAI between 2009 and 2019". Prof Kwok is the IJCAI-2025 Program Chair.
\end{IEEEbiography}

\newpage

% \appendix
\appendices

% \renewcommand\thesection{\arabic{section A}}
% pass test

% \clearpage
% \centerline{\large{\textbf{Appendix: From Transfer to Transformer: A Survey on Time-Series Pre-Trained Models}}}

\section{{}{Related Survey Analysis~\label{appendix_A}}}

{}{Among the six surveys focus on time series pre-training, the first web online time of~\cite{eldele2024label} predates ours by three months, while~\cite{zhang2024self,meng2023unsupervised,jin2023large,jin2024position,liang2024foundation} appeared later.  Table~\ref{tab:appendix-survey-analysis} details the differences between these surveys:~\cite{eldele2024label,zhang2024self,meng2023unsupervised} focus on obtaining feature representations of time series through label-efficient, self-supervised, and unsupervised scenarios in the pre-training phase, while~\cite{jin2023large,jin2024position,liang2024foundation} explore transferring knowledge from pre-trained natural language models to the time series domain via fine-tuning. Our survey comprehensively analyzes pre-training methods for six tasks: classification, forecasting, clustering, anomaly detection, imputation, and extrinsic regression from supervised, unsupervised, and self-supervised learning perspectives. It also examines the effects of various transfer strategies during fine-tuning and measures the performance of different pre-training methods through extensive experiments.}

{}{Table~\ref{tab:appendix-survey-experiments} presents comprehensive statistics on the experimental scope, encompassing 27 distinct models (with a total of 35 models, adjusted by removing duplicates across different downstream tasks), 434 datasets (comprised of 166, 9, and 259 datasets), and 679 transfer learning experiments (calculated as 15 source datasets multiplied by 45 target datasets, plus an additional 4 independent time series scenarios for transfer learning) focused on classification tasks.
By reproducing experiments under consistent conditions across many datasets, we offer a more objective evaluation of the performance of various TS-PTMs. These experiments provide guidance for readers in selecting suitable deep learning models or pre-training techniques for the design of pre-training models.}

{}{In summarizing the existing literature, Eldele et al.~\cite{eldele2024label} systematically reviewed and analyzed work on learning time series representations in both in-domain and cross-domain settings under label-efficient scenarios with few labeled data and only unlabeled data. Although~\cite{eldele2024label} focuses on pre-training and fine-tuning techniques for time series, it mainly emphasizes RNN- and CNN-based pre-training in label-efficient scenarios. In contrast, our survey systematically summarizes pre-training methods based on five deep neural network architectures (RNN, CNN, TCN, Transformer, and GNN) from a transfer learning perspective.
Zhang et al.~\cite{zhang2024self} and Meng et al.~\cite{meng2023unsupervised} discuss obtaining time series feature representations through self-supervised or unsupervised learning in the pre-training stage but neglect other pre-training methods and the fine-tuning process within the transfer learning framework. Jin et al.~\cite{jin2023large} and Liang et al.~\cite{liang2024foundation} reviewed techniques for pre-training Large Language Models (LLMs) for fine-tuning on time series downstream tasks. However, the rationale behind the successful application of LLMs for downstream task fine-tuning in time series remains to be fully explored. Jin et al.~\cite{jin2024position} focused on transferring knowledge gained from pre-training LLMs to time series downstream tasks but overlooked leveraging domain-specific knowledge during the pre-training process to obtain appropriate time series feature representations.
In contrast to~\cite{jin2023large,jin2024position,liang2024foundation}, our survey focuses on the existing time series domain. We validate the importance of maintaining consistency between source and target datasets in time series transfer learning through extensive experiments. This demonstrates that the design of TS-PTMs, utilizing time series domain datasets for pre-training and fine-tuning paradigms, holds significant developmental potential.}

% \usepackage{booktabs}
% \usepackage{multirow}
\begin{table*}[]
\centering
\caption{{}{Comparison of our work with other related surveys. "\checkmark" indicates that the corresponding survey covers a specific point. A "1" means the survey fully covers the model with detailed discussion (Fully Covered), while a "0.5" indicates partial coverage (Partially Covered) with some articles using the model. Surveys marked with an asterisk (*) in the experiments collected existing experimental results from the literature but did not conduct uniform code replication experiments.}}
\label{tab:appendix-survey-analysis}
  \resizebox{\textwidth}{!}{
\begin{tabular}{@{}c|c|cc|ccc|ccccc|c@{}}
\toprule
\multirow{2}{*}{Survey} & \multirow{2}{*}{\begin{tabular}[c]{@{}c@{}}First Online \\ Time\end{tabular}} & \multicolumn{2}{c|}{Focus Topic} & \multicolumn{3}{c|}{Pre-training   Techniques} & \multicolumn{5}{c|}{Model Architecture} & \multirow{2}{*}{Experiments} \\ \cmidrule(lr){3-12}
 &  & Pre-training & Fine-tuning & Supervised & Unsupervised & Self-supervised & RNN & CNN & TCN & Transformer & GNN &  \\ \cmidrule(r){1-2} \cmidrule(l){1-13} 
Eldele et al.~\cite{eldele2024label} & Feb 2023 & \checkmark & \checkmark & \checkmark & \checkmark & \checkmark & 1 & 1 & 0.5 & 0.5 &  &  \\
Zhang et al.~\cite{zhang2024self} & Jun 2023 & \checkmark &  &  &  & \checkmark & 1 & 1 &  & 0.5 &  & * \\
Meng et al.~\cite{meng2023unsupervised} & Aug 2023 & \checkmark &  &  & \checkmark & \checkmark & 1 & 1 &  & 0.5 &  & \checkmark \\
Jin et al.~\cite{jin2023large} & Oct 2023 & \checkmark & \checkmark &  &  &  & 0.5 &  &  & 1 & 1 &  \\
Jin et al.~\cite{jin2024position} & Feb 2024 &  & \checkmark &  &  &  &  &  &  & 1 &  & \checkmark \\
Liang et al.~\cite{liang2024foundation} & Mar 2024 & \checkmark & \checkmark & \checkmark & \multicolumn{1}{l}{} & \checkmark &  & 1 & 1 & 1 &  &  \\ \hline
\textbf{Our Survey} & May 2023 & \checkmark & \checkmark & \checkmark & \checkmark & \checkmark & 1 & 1 & 1 & 1 & 1 & \checkmark \\ \bottomrule
\end{tabular}}
\end{table*}

% \usepackage{booktabs}
% \usepackage{multirow}
\begin{table*}[]
\centering
\caption{{}{Experimental comparison of our work with other related surveys. "Datasets" refers to the number of datasets used in the experiment. "Methods" indicates the number of comparison methods employed for the corresponding task. "Transfer Sets" represents the number of groups undergoing source domain-target domain migration in the transfer learning experiment.}}
\label{tab:appendix-survey-experiments}
\begin{tabular}{@{}c|cc|cc|cc|ccc@{}}
\toprule
\multirow{2}{*}{Survey} & \multicolumn{2}{c|}{Classification} & \multicolumn{2}{c|}{Forecasting} & \multicolumn{2}{c|}{Anomaly Detection} & \multicolumn{3}{c}{Transfer Learning} \\ \cmidrule(l){2-10} 
 & Datasets & Methods & Datasets & Methods & Datasets & Methods & Datasets & Methods & Transfer Sets \\ \cmidrule(r){1-10}
Meng et al.~\cite{meng2023unsupervised} & 15 & 17 & 0 & 0 & 0 & 0 & 0 & 0 & 0 \\
Jin et al.~\cite{jin2024position} & 1 & 1 & 0 & 0 & 0 & 0 & 0 & 0 & 0 \\ \hline
\textbf{Our Survey} & 166 & 13 & 9 & 12 & 259 & 10 & 53 & 4 & 679 \\ \bottomrule
\end{tabular}
\end{table*}

\section{Experimental Details and Resources~\label{appendix_B}}

\subsection{Source Datasets, Code and TS-PTMs~\label{appendix_B1}}
In time series domain, the statistics of publicly available benchmark datasets are shown in Table~\ref{tab:dataset}.
Also,  the statistics of existing TS-PTMs containing open-source code and the summary of TS-PTMs are shown in Table~\ref{tab:code} and Table~\ref{tab:summary}.
% , the summary of TS-PTMs is shown in .

\begin{table*}[htbp]
  \centering
  \caption{Benchmark Datasets in Time-Series Domain}
  \resizebox{\textwidth}{!}{
    \begin{tabular}{|c|c|c|}
    \hline
    Resource &  Description & URL \\
    \hline
    UCR   & 128 univariate UCR time series classification datasets & https://www.cs.ucr.edu/~eamonn/time\_series\_data\_2018/ \\
    UEA   & 30 multivariate UEA time series classification datasets & http://www.timeseriesclassification.com/ \\
    IHEPC &  Individual Household Electric Power Consumption (IHEPC) dataset & https://archive.ics.uci.edu/ml/datasets/individual+household+electric+power+consumption \\
    XJTU-SU & Run-to-failure data of 15 rolling element bearings & http://biaowang.tech/xjtu-sy-bearing-datasets \\
     MFPT & Bearing Fault Dataset & https://www.mfpt.org/fault-data-sets/ \\
    ECG Waveform & Long-term Electrocardiogram (ECG) recordings & https://archive.physionet.org/cgi-bin/atm/ATM \\
    HAR   & Human Activity Recognition Using Smartphones Data Set & https://archive.ics.uci.edu/ml/datasets/human+activity+recognition+using+smartphones \\
    Sleep-EDF & Sleep Stage Classification & https://github.com/emadeldeen24/TS-TCC \\
    Epilepsy & Epilepsy Seizure Prediction & https://github.com/emadeldeen24/TS-TCC \\
    FD    & Fault Diagnosis & https://github.com/emadeldeen24/TS-TCC \\
    EMG &   Electromyograms (EMG) measures muscle responses as electrical activity & https://github.com/mims-harvard/TFC-pretraining \\
    ETT   & A crucial indicator in the electric power long-term deployment & https://github.com/zhouhaoyi/ETDataset \\
    Electricity & The electricity consumption (Kwh) of 321 clients & https://archive.ics.uci.edu/ml/datasets/ElectricityLoadDiagrams20112014 \\
    
      {}{Traffic}   & {}{A collection of hourly data from the California Department of Transportation for forecasting} & {}{http://pems.dot.ca.gov} \\

     {}{Weather}   & {}{Recorded every 10 minutes, the dataset includes 21 meteorological indicators for forecasting} & {}{https://www.bgc-jena.mpg.de/wetter/} \\

       {}{Exchange}   & {}{Records the daily exchange rates of eight different countries from 1990 to 2016} & {}{https://github.com/laiguokun/multivariate-time-series-data} \\

      {}{ILI}   & {}{The weekly recorded data of influenza-like illness (ILI) patients for forecasting} & {}{https://gis.cdc.gov/grasp/fluview/fluportaldashboard.html} \\

{}{TSER}   & {}{19 time series extrinsic regression datasets} & {}{http://tseregression.org/} \\

     Yahoo & 367 hourly sampled time series with tagged anomaly points & https://webscope.sandbox.yahoo.com/ \\

     KPI   & Multiple minutely sampled real KPI curves from many internet companies & https://github.com/NetManAIOps/donut/tree/master/sample\_data \\

        {}{UCR-AT}   & {}{250 univariate UCR time series anomaly detection datasets} & {}{https://wu.renjie.im/research/anomaly-benchmarks-are-flawed/\#ucr-time-series-anomaly-archiv} \\

           {}{MSL}   & {}{Collected by NASA, the dataset reveals the condition of sensors and actuator data} & {}{https://github.com/khundman/telemanom} \\

          {}{SMAP}   & {}{Collected by NASA, the dataset presents soil samples and telemetry information} & {}{https://github.com/eBay/RANSynCoders} \\

      {}{PSM}   & {}{A public dataset
from eBay Server Machines with 25 dimensions} & {}{https://github.com/khundman/telemanom} \\

  {}{SMD}   & {}{A five-week-long dataset collected from an internet company compute cluster} & {}{https://github.com/NetManAIOps/OmniAnomaly} \\

   {}{SWaT}   & {}{A 51-dimension sensor-based dataset collected from
critical infrastructure systems} & {}{https://itrust.sutd.edu.sg/itrust-labs\_datasets/dataset\_info/\#swat} \\

 {}{NIPS-TS-SWAN}   & {}{A multivariate time series dataset extracted from solar photospheric
vector magnetograms} & {}{https://github.com/datamllab/tods/tree/benchmark/benchmark} \\

 {}{NIPS-TS-GECCO}   & {}{A drinking water quality
dataset for the ‘internet of things’} & {}{https://github.com/datamllab/tods/tree/benchmark/benchmark} \\

 % {}{Capnobase}   & {}{Simultaneous ECG, PPG and capnography respiratory signals for extrinsic regression} & {}{http://peterhcharlton.github.io/RRest/datasets.html} \\

 %  {}{ BIDMC}   & {}{Simultaneous ECG, PPG and thoracic impedance signals for extrinsic regression} & {}{http://peterhcharlton.github.io/RRest/datasets.html} \\

    \hline
    \end{tabular}}
  \label{tab:dataset}
\end{table*}

\begin{table*}[htbp]
  \centering
  \caption{The Open-Source Implementations of TS-PTMs.}
  \resizebox{\textwidth}{!}{
    \begin{tabular}{|c|c|c|}
    \hline
    Resource &  Description & URL \\
    \hline
    TL-FCN~\cite{fawaz2018transfer} & Framework: Keras (2018 Big Data)  & https://github.com/hfawaz/bigdata18 \\
    CPC~\cite{oord2018representation} &  Framework: PyTorch (2018 arXiv) & https://github.com/jefflai108/Contrastive-Predictive-Coding-PyTorch \\
    T-Loss~\cite{franceschi2019unsupervised} &  Framework: PyTorch (2019 NeurIPS)  & https://github.com/White-Link/UnsupervisedScalableRepresentationLearningTimeSeries \\
    SelfTime~\cite{fan2020self} & Framework: PyTorch (2020 arXiv)  & https://github.com/haoyfan/SelfTime \\
    TNC~\cite{tonekaboni2020unsupervised} & Framework: PyTorch (2021 ICLR)  & https://github.com/sanatonek/TNC\_representation\_learning \\
    Voice2Series~\cite{yang2021voice2series} & Framework: Tensorflow (2021 ICML)  & https://github.com/huckiyang/Voice2Series-Reprogramming \\
    TS-TCC~\cite{eldele2021time} & Framework: PyTorch (2021 IJCAI)  & https://github.com/emadeldeen24/TS-TCC \\
    TST~\cite{zerveas2021transformer} & Framework: PyTorch (2021 KDD)  & https://github.com/gzerveas/mvts\_transformer \\
    TS2Vec~\cite{yue2022ts2vec} & Framework: PyTorch (2022 AAAI)  & https://github.com/yuezhihan/ts2vec \\
    CoST~\cite{woo2022cost} & Framework: PyTorch (2022 ICLR)  & https://github.com/salesforce/CoST \\
     ExpCLR~\cite{nonnenmacher2022utilizing} & Framework: PyTorch (2022 ICML)  & https://github.com/boschresearch/expclr \\
    TF-C~\cite{zhang2022self} & Framework: PyTorch (2022 NeurIPS)  & https://github.com/mims-harvard/TFC-pretraining \\
     SLVM~\cite{jiangsequential} & Framework: PyTorch (2023 ICLR)  & https://github.com/john-x-jiang/meta\_ssm \\
    CLUDA~\cite{ozyurt2022contrastive} & Framework: PyTorch (2023 ICLR)  & https://github.com/oezyurty/CLUDA \\

     {}{PatchTST~\cite{nietime}} &  {}{Framework: PyTorch (2023 ICLR)}  &  {}{https://github.com/yuqinie98/PatchTST} \\

      {}{DCdetector~\cite{yang2023dcdetector}} &  {}{Framework: PyTorch (2023 KDD)}  &  {}{https://github.com/DAMO-DI-ML/KDD2023-DCdetector} \\

       {}{SimMTM~\cite{dong2024simmtm}} &  {}{Framework: PyTorch (2023 NeurIPS)}  &  {}{https://github.com/thuml/SimMTM} \\

      {}{STPT~\cite{hu2023self}} &  {}{Framework: PyTorch (2023 ICDM)}  &  {}{https://github.com/mhu3/STPT} \\

      {}{TimesURL~\cite{liu2024timesurl}} &  {}{Framework: PyTorch (2024 AAAI)}  &  {}{https://github.com/Alrash/TimesURL} \\

      {}{TimeSiam~\cite{dong2024timesiam}} &  {}{Framework: PyTorch (2024 ICML)}  &  {}{https://github.com/thuml/TimeSiam} \\

      {}{UP2ME~\cite{zhangup2me}} &  {}{Framework: PyTorch (2024 ICML)}  &  {}{https://github.com/Thinklab-SJTU/UP2ME} \\

      {}{TSLANet:~\cite{eldeletslanet}} &  {}{Framework: PyTorch (2024 ICML)}  &  {}{https://github.com/emadeldeen24/TSLANet} \\

      {}{MOMENT~\cite{goswamimoment}} &  {}{Framework: PyTorch (2024 ICML)}  &  {}{https://github.com/moment-timeseries-foundation-model/moment} \\

      {}{Timer~\cite{liu2024timer}} &  {}{Framework: PyTorch (2024 ICML)}  &  {}{https://github.com/thuml/Large-Time-Series-Model} \\

      {}{MOIRAI~\cite{woo2024unified}} &  {}{Framework: PyTorch (2024 ICML)}  &  {}{https://github.com/SalesforceAIResearch/uni2ts} \\

      {}{TimesFM~\cite{dasdecoder}} &  {}{Framework: PyTorch (2024 ICML)}  &  {}{https://github.com/google-research/timesfm} \\
     
    \hline
    \end{tabular}}
  \label{tab:code}
\end{table*}

\begin{table*}[htbp]
  \centering
  \caption{The summary of Time-Series Pre-Trained Models~(TS-PTMs).}
 \scalebox{0.95}{
    \begin{tabular}{c|c|cc|c|cc|c}
    \hline
    \multirow{2}[0]{*}{Year} & \multirow{2}[0]{*}{Method} & \multicolumn{2}{c|}{Superviesed PTMs} & \multicolumn{1}{c|}{Unsupervised PTMs} & \multicolumn{2}{c|}{Self-supervised PTMs} & \multirow{2}[0]{*}{Architecture} \\
    \cline{3-7}
    & & CLS & FORE & REC & CONSIS & PSEUDO \\
    \hline
    2016  & Audio Word2Vec~\cite{chung2016audio} & & & \checkmark & & & RNN \\
    2016  & DCFT~\cite{morales2016deep}  & \checkmark &  & & & & CNN\&RNN \\
    2016  & TCL~\cite{hyvarinen2016unsupervised}   & & &  & \checkmark & & MLP \\
    2016  & SHL-DNN~\cite{hu2016transfer} & & & \checkmark & & & MLP \\
    2017  & TimeNet~\cite{malhotra2017timenet} & & & \checkmark & & & RNN \\
    2017 & VRADA~\cite{purushotham2017variational} & \checkmark & & & & & RNN \\
    2018  & CPC~\cite{oord2018representation} & & \checkmark & & \checkmark & & CNN\&RNN \\
    2018  & TL-FCN~\cite{fawaz2018transfer} & \checkmark & & & & & CNN \\
    2018  & Encoder~\cite{serra2018towards} & \checkmark & & & & & CNN \\
    2018  & TL-CTSAD~\cite{xiong2018application} & & \checkmark & & & & RNN \\
    2018  & Speech2Vec~\cite{chung2018speech2vec} & & & & & \checkmark & RNN \\
    2018  & TL-LSTM~\cite{laptev2018reconstruction} & & \checkmark & \checkmark & & & RNN \\
    2018 & HDCNN~\cite{khan2018scaling} & \checkmark & & & & & CNN \\
    2018 & STL~\cite{wang2018stratified} & \checkmark & & & & & CNN \\
    2019  & vq-wav2vec~\cite{baevski2019vq} & & & & \checkmark & & Transformer Enc \\
    2019  & FRTL~\cite{mutegeki2019feature}  & \checkmark & & & & & CNN\&RNN \\
    % 2019  &  {\color{red}U-Net~\cite{wen2019time}} & & & & & & \checkmark & & & & & CNN \\
    2019  & ConvTimeNet~\cite{kashiparekh2019convtimenet} & \checkmark & & & & & CNN \\
    2019  & wav2vec~\cite{schneider2019wav2vec} & & \checkmark & & \checkmark & & CNN \\
    % 2019  & DTCR~\cite{ma2019learning}  & & & \checkmark & & & & & & & & RNN \\
    2019  & T-Loss~\cite{franceschi2019unsupervised}  & & & & \checkmark & & TCN \\
    2020  & SMS~\cite{meiseles2020source}  & \checkmark & & & & & CNN \\
    2020  & SelfTime~\cite{fan2020self} & & & & \checkmark & & CNN \\
    2020  & InceptionTime~\cite{ismail2020inceptiontime} & \checkmark & & & & & CNN \\
    2020  & Mockingjay~\cite{liu2020mockingjay} & & & \checkmark & & & Transformer Enc \\
    2020  & ML-TSC~\cite{narwariya2020meta} & \checkmark & & & & & CNN \\
    2020  & ML-TSF~~\cite{iwata2020few} & & \checkmark & & & & RNN \\
    2020  & DTL-TS~\cite{li2020deep} & \checkmark & & & & & CNN \\
    2020 & LSTM-DANN ~\cite{da2020remaining} & \checkmark & & & & & RNN \\
    2020 & CoDATS~\cite{wilson2020multi} & \checkmark & & & & & CNN \\
     {}{2021} & {}{TL-DCRNN}~\cite{mallick2021transfer} & & \checkmark & & & & {}{CNN\&RNN\&GNN} \\
    2021 & SASA~\cite{cai2021time} & \checkmark & & & & & RNN \\
    2021 & SLARDA~\cite{ragab2021self} & \checkmark & & & & & Transformer\&CNN \\
    2021 & GCA~\cite{li2021transferable} & \checkmark & & & & & RNN \\
    2021 & AdvSKM~\cite{liu2021adversarial} & \checkmark & & & & & CNN \\
    2021 & CALDA~\cite{wilson2021calda} & \checkmark & & & & & CNN \\
    2021 & CMTN~\cite{li2021causal} & \checkmark & & & & & RNN \\
    2021  & TSAD-TL~\cite{baireddy2021spacecraft} & & \checkmark & & & & RNN \\
    % 2021  & TE-TCN & & \checkmark & & & & & & & & & TCN \\
    2021  & TNC~\cite{tonekaboni2020unsupervised} & & & & \checkmark & & RNN/TCN \\
    2021  & Voice2Series~\cite{yang2021voice2series} & \checkmark & & & & & Trans\&CNN\&RNN \\
    % 2021  & TLS-AD & \checkmark & & \checkmark & & & & & & CNN \\ zhang2021sleeppriorcl
     2021  & SleepPriorCL~\cite{zhang2021sleeppriorcl} & &  & & \checkmark & \checkmark & CNN \\
    2021  & TS-TCC~\cite{eldele2021time} & & \checkmark & & \checkmark & & CNN/Transformer \\
    2021  & TS-Transformer~\cite{zerveas2021transformer} & & & \checkmark & & & Transformer Enc \\
    2021 & SSL-TS~\cite{shi2021self} & & & \checkmark & \checkmark & & Transformer Enc \\
    2021  & InfoTS~\cite{luo2021information} & & & & \checkmark & & TCN \\
    2021  & TabBERT~\cite{padhi2021tabular} & & & \checkmark & & & Transformer Enc \\
    2021  & TabAConvBERT~\cite{shankaranarayana2021attention} & & & \checkmark & & & Transformer Enc \\
    2021  & TERA~\cite{liu2021tera} & & & \checkmark & & & Transformer Enc \\
    % 2021  & CRLI~\cite{ma2021learning} & & & & \checkmark & & & & & & & RNN \\
    2021  & AdaRNN~\cite{du2021adarnn} & & \checkmark & & & & RNN \\
    2022  & TSSN~\cite{hou2022masked} & & & \checkmark & & & Transformer Enc \\
    2022  & ST-GSP~\cite{zhao2022st} & & & \checkmark & & & Transformer Enc \\
    2022  &  CoST~\cite{woo2022cost} & & & & \checkmark & & TCN \\ 
    2022  & TS2Vec~\cite{yue2022ts2vec} & & & & \checkmark & & TCN \\
     2022  & BTSF~\cite{yang2022unsupervised}  & & & & \checkmark & & TCN \\
     2022  & STEP~\cite{shao2022pre}  & &  & \checkmark &  & & {}{Transformer\&GNN}  \\ 
 2022  & TARNet~\cite{chowdhury2022tarnet}  & & & \checkmark &  & & Transformer \\   
 2022  &
 ST-GFSL~\cite{lu2022spatio}  & & \checkmark &  &  & & RNN \\   
  2022 & ExpCLR~\cite{nonnenmacher2022utilizing}  & & & & \checkmark & & TCN \\
     2022  & TimeHetNet~\cite{brinkmeyer2022few}  & & \checkmark & &  & & CNN\&RNN \\
      2022  & CRT~\cite{zhang2022cross}  & \checkmark & & & & & Transformer Enc \\
       2022  & SPGN~\cite{yang2022spectral}  & \checkmark & & &  & & CNN \\
      2022  & TF-C~\cite{zhang2022self}  & & &  & \checkmark & & CNN \\
       2023  & SLVM~\cite{jiangsequential}  &  & \checkmark & & & & RNN \\
      2023  & CLUDA~\cite{ozyurt2022contrastive}  & & & & \checkmark & & TCN \\
        {}{2023}  &  {}{PatchTST~\cite{nietime}}  & & & \checkmark &  & &  {}{Transformer Enc} \\
         {}{2023}  &  {}{DCdetector~\cite{yang2023dcdetector}}  & & &  & \checkmark & &  {}{Transformer Enc} \\
         
 {}{2023}  &  {}{SimMTM~\cite{dong2024simmtm}}  & & & \checkmark &  & &  {}{Transformer\&CNN} \\

{}{2023}  &  {}{STPT~\cite{hu2023self}}  & & &  & \checkmark & &  {}{Transformer Enc} \\

{}{2023}  &  {}{TriD-MAE~\cite{zhang2023trid}}  & & & \checkmark &  & &  {}{TCN} \\
 
          {}{2024}  &  {}{TimesURL~\cite{liu2024timesurl}}  & & &  & \checkmark & &  {}{TCN} \\

     {}{2024}  &  {}{TimeSiam~\cite{dong2024timesiam}}  & & & \checkmark &  & &  {}{Transformer\&TCN} \\
               
  {}{2024}  &  {}{UP2ME~\cite{zhangup2me}}  & & & \checkmark &  & &  {}{Transformer} \\

   {}{2024}  &  {}{TSLANet~\cite{eldeletslanet}}  & & & \checkmark &  &   &  {}{CNN} \\

   {}{2024}  &  {}{MOMENT~\cite{goswamimoment}}  & & & \checkmark &  & &  {}{Transformer Enc} \\

   {}{2024}  &  {}{Timer~\cite{liu2024timer}}  & & \checkmark &  &  & &  {}{Transformer} \\

    {}{2024}  &  {}{MOIRAI~\cite{woo2024unified}}  & &  \checkmark &  & & &  {}{Transformer} \\

    {}{2024}  &  {}{TimesFM~\cite{dasdecoder}}  & &  \checkmark &  & & &  {}{Transformer} \\
            
    \hline
    \end{tabular}
  }
  \label{tab:summary}
\end{table*}

% STPT, zhang2023trid
% zhangup2me  Transformer  liu2024timer  woo2024unified dong2024timesiam hu2023self zhang2023trid dasdecoder 

\subsection{Experimental Datasets~\label{appendix_B2}}

In the time series domain, 128 univariate UCR time series datasets~\cite{dau2019ucr} and 30 multivariate UEA time series datasets~\cite{bagnall2018uea} are commonly utilized in time-series classification studies.
However, each dataset in the UCR and UEA archives does not contain the validation set employed for hyperparameter selection.
For a fair analysis, we first combine the training and test sets for each dataset in UCR and UEA archives. Then, we adopt the five-fold cross-validation strategy to divide the training, validation, and test sets according to a ratio of 60\%, 20\%, and 20\%, respectively. 
Finally, these repartitioned datasets are used to evaluate the time-series classification task performance.
Unlike the time series forecasting task, time series classification does not require predicting past and future values of individual series.
Further, each dataset in UCR and UEA archives contains several independent time series.
Therefore, we can reasonably use a five-fold cross-validation strategy for time series classification.
For the time-series forecasting task, like ~\cite{yue2022ts2vec,wutimesnet}, we utilize {}{ETTh1, ETTh2, ETTm1, ETTm2, Electricity, Traffic, Weather, Exchange, and national illness (ILI) datasets} for the experimental analysis. In addition, following~\cite{yue2022ts2vec, yang2023dcdetector}, we employ the Yahoo, KPI, {}{UCR anomaly detection archive, seven multivariate datasets (SMD, MSL, SMAP, PSM, SWAT, NIPS\_TS\_Swan, and NIPS\_TS\_Water)} for the time-series anomaly detection task. 

\subsection{Baselines~\label{appendix_B3}}

Transfer learning has been a significant success in the study of PTMs. Fawaz et al.~\cite{fawaz2018transfer} utilized 85 UCR datasets for time series transfer learning. Firstly, a randomly selected dataset is used as the source dataset to pre-train the Fully Convolutional Networks~(FCN)~\cite{wang2017time} model via the supervised classification task. Then a randomly selected target UCR dataset is fine-tuned on the pre-trained FCN model. Meanwhile, Malhotra et al.~\cite{malhotra2017timenet} employed a symmetric RNN-based encoder-decoder architecture as the backbone. The encoder is first pre-trained using the unsupervised reconstruction task on various UCR datasets as source datasets. Then, the target dataset is fine-tuned on the pre-trained RNN-based encoder. Therefore, in terms of employing transfer learning for time series pre-training, we perform a comparative analysis utilizing a supervised classification transfer strategy and an unsupervised reconstruction transfer strategy.

Recently, TS-PTMs have achieved good performance in TSM tasks. 
For the time-series classification task, we select T-Loss~\cite{franceschi2019unsupervised}, SelfTime~\cite{fan2020self}, TS-TCC~\cite{eldele2021time}, TST~\cite{zerveas2021transformer}, TS2Vec~\cite{yue2022ts2vec}, {}{and GPT4TS~\cite{zhou2023one}} to analyze the performance of TS-PTMs and compare them with the supervised FCN~\cite{wang2017time}, {}{PatchTST~\cite{nietime}, and TimesNet~\cite{wutimesnet}} model. Meanwhile, we select TS2Vec~\cite{yue2022ts2vec} and CoST~\cite{woo2022cost} as benchmark TS-PTMs methods for the time-series forecasting task. Also, we compare them with supervised benchmark methods in time-series forecasting, including LogTrans~\cite{li2019enhancing}, TCN~\cite{bai2018empirical}, Informer~\cite{zhou2021informer}, and Autoformer~\cite{xu2021autoformer}, where LogTrans, Informer, and Autoformer are end-to-end models based on the Transformer. For the time-series anomaly detection task, we select TS2Vec~\cite{yue2022ts2vec} as the benchmark TS-PTM method. We compare it with the benchmark methods SPOT~\cite{siffer2017anomaly}, DSPOT~\cite{siffer2017anomaly}, LSTM-VAE~\cite{park2018multimodal}, DONUT~\cite{xu2018unsupervised}, Spectral Residual (SR)~\cite{ren2019time}, and Anomaly Transformer (AT)~\cite{xu2021anomaly} in time-series anomaly detection, where AT is a model based on the Transformer.

\subsubsection{Time-Series Classification}

In the time-series classification task, we first compare the performance of FCN, TCN, and Transformer for direct classification on 128 UCR time series datasets.
Then, we analyze the classification performance of T-Loss, SelfTime, TS-TCC, TST, and TS2Vec after pre-training on 128 UCR and 30 UEA time series datasets. In addition, for the datasets containing missing values in the UCR and UEA archives, we use the mean-imputation method to fill the divided training, validation, and test sets, respectively.
In this work, the classification accuracy of the test set is employed to evaluate classification performance.
Experiments are performed on eight 1080Ti GPUs and two 3090 GPUs. For the running time of the corresponding method, we use the test results on the 3090 GPU.

The specific methods used in this study are described as follows:

FCN: Based on literature~\cite{wang2017time}, we build the model using a three-layer one-dimensional \textbf{F}ully \textbf{C}onvolutional \textbf{N}etwork~(\textbf{FCN}) and a one-layer global average pooling layer.
We set the FCN structure and parameters based on the open source code from~\url{https://github.com/cauchyturing/UCR\_Time\_Series\_Classification\_Deep\_Learning\_Baseline}.

TCN: Based on the open source code of literature ~\cite{franceschi2019unsupervised}, we build the \textbf{T}emporal \textbf{C}onvolutional \textbf{N}etwork (\textbf{TCN}) using Pytorch.
The original TCN contains ten layers of residual blocks. To compare and analyze the three-layer structure of FCN, we employ a three-layer TCN for comparison experiments.
We use the open source code from~\url{https://github.com/White-Link/UnsupervisedScalableRepresentationLearningTimeSeries} for TCN structure and parameter setting.

Transformer: Zerveas et al.~\cite{zerveas2021transformer} proposed a Transformer-based model for unsupervised time series representation learning.
We use the \textbf{Transformer} model in open source code from~\url{https://github.com/gzerveas/mvts_transformer} for direct classification experiments on 128 UCR time series datasets.

T-Loss: Franceschi et al.~\cite{franceschi2019unsupervised} proposed an unsupervised time series representation learning method using TCN and a novel \textbf{T}riplet \textbf{L}oss (\textbf{T-Loss}).
We use the open source code from~\url{https://github.com/White-Link/UnsupervisedScalableRepresentationLearningTimeSeries} for experimental analysis.

SelfTime: Fan et al.~\cite{fan2020self} proposed a self-supervised time series representation learning framework called \textbf{SelfTime}, exploring the inter-sample and intra-temporal relation of time series for learning representations on unlabeled data.
We use the open source code from~\url{https://github.com/haoyfan/SelfTime} for experimental analysis.

TS-TCC: Eldele et al.~\cite{eldele2021time} proposed an unsupervised \textbf{T}ime \textbf{S}eries representation learning framework via \textbf{T}emporal and \textbf{C}ontextual \textbf{C}ontrasting (\textbf{TS-TCC}), thus learning feature representations of unlabeled time series.
We use the open source code from~\url{https://github.com/emadeldeen24/TS-TCC} for experimental analysis.

TST: Zerveas et al.~\cite{zerveas2021transformer} first proposed a multivariate time series unsupervised representation learning framework called \textbf{T}ime \textbf{S}eries \textbf{T}ransformer (\textbf{TST}).
We use the open source code from~\url{https://github.com/gzerveas/mvts_transformer} for experimental analysis.

TS2Vec: Yue et al.~\cite{yue2022ts2vec} proposed a general framework for time series representation learning in an arbitrary semantic level called \textbf{TS2Vec}.
For the time series classification task, the authors employed TCN as the backbone and the low-dimensional feature representation obtained from TCN to input into an SVM classifier with RBF kernel for classification.
We use the open source code from~\url{https://github.com/yuezhihan/ts2vec} for experimental analysis.

{}{TimesNet: Wu et al.~\cite{wutimesnet} propose TimesNet with TimesBlock to identify multiple periods and capture temporal 2D-variations from transformed 2D tensors using a parameter-efficient inception block. For the classification task, we employ TimesNet as a backbone for fully supervised training to analyze the potential in time series pre-training. We use the open source code from~\url{https://github.com/thuml/TimesNet} for experimental analysis.}

{}{PatchTST: Nie et al.~\cite{nietime} segment time series into patches and assume channels are independent for multivariate time series forecasting and self-supervised representation learning via a reconstruction-based pre-training strategy. Given that GPT4TS~\cite{zhou2023one} has employed similar patch and channel-independent strategies for fine-tuning, to reduce training time in the classification experiments, we utilize only the patches strategy and channel-independent strategy by combining a transformer for fully supervised training. We use the open source code from~\url{https://github.com/yuqinie98/PatchTST} for experimental analysis.}

{}{GPT4TS: Zhou et al.~\cite{zhou2023one} present a unified framework for time-series modeling using pre-trained Large Language Models (LLMs). Specifically, GPT4TS employs the same patch and channel-independent strategies in PatchTST~\cite{nietime} for LLMs fine-tuning for time series downstream tasks. We use the open source code from~\url{https://github.com/DAMO-DI-ML/NeurIPS2023-One-Fits-All} for experimental analysis.}

% GPT4TS~\cite{zhou2023one}
% PatchTST~\cite{nietime}, and TimesNet~\cite{wutimesnet}

\subsubsection{Time-Series Forecasting}

TS2Vec: Yue et al.~\cite{yue2022ts2vec} employed TCN as the backbone and the data from the last $T$ observations to predict the observations for the next $H$ timestamps for the time series forecasting task.
We use the open source code from~\url{https://github.com/yuezhihan/ts2vec} for experimental analysis.

CoST: Woo et al.~\cite{woo2022cost} proposed a representation learning framework for long time series forecasting, called \textbf{CoST}, which utilizes contrastive learning to learn time series disentangled seasonal-trend representations.
We use the open source code from~\url{https://github.com/salesforce/CoST} for experimental analysis.

LongTrans: Li et al.~\cite{li2019enhancing} proposed a Transformer-based model for \textbf{Long} time series forecasting with convolutional self-attention and LogSparse \textbf{Trans}former~(\textbf{LongTrans}) to capture the time series' local context and long-term dependencies information.
We use the open source code from~\url{https://github.com/AIStream-Peelout/flow-forecast} for experimental analysis.

TCN: Bai et al.~\cite{bai2018empirical}'s extensive experiments on the sequence modeling problem indicated that a simple convolutional architecture consisting of \textbf{T}emporal \textbf{C}onvolutional \textbf{N}etworks~(\textbf{TCN}) outperformed canonical recurrent networks such as LSTMs for time series forecasting on different datasets.
We use the open source code from~\url{https://github.com/locuslab/TCN} for experimental analysis.

Informer: Zhou et al.~\cite{zhou2021informer} designed an efficient transformer-based model called \textbf{Informer} for long time series forecasting.
We use the open source code from~\url{https://github.com/zhouhaoyi/Informer2020} for experimental analysis.

Autoformer: Wu et al.~\cite{xu2021autoformer} designed \textbf{Autoformer} as a novel decomposition architecture with an Auto-Correlation mechanism for long time series forecasting.
We use the open source code from~\url{https://github.com/thuml/autoformer} for experimental analysis.

{}{TimesNet: Wu et al.~\cite{wutimesnet} propose TimesNet with TimesBlock to identify multiple periods and capture temporal 2D-variations from transformed 2D tensors using a parameter-efficient inception block. For the forecasting task, we employ TimesNet as a backbone for fully supervised training to analyze the potential in time series pre-training. We use the open source code from~\url{https://github.com/thuml/TimesNet} for experimental analysis.}

{}{PatchTST: Nie et al.~\cite{nietime} propose an efficient design of Transformer-based models for multivariate time series forecasting and self-supervised representation learning. Given that GPT4TS~\cite{zhou2023one} has employed similar patch and channel-independent strategies for fine-tuning, to reduce training time in the forecasting experiments, we utilize only the patches strategy and channel-independent strategy by combining a transformer for fully supervised training. We use the open source code from~\url{https://github.com/yuqinie98/PatchTST} for experimental analysis.}

{}{DLinear: Zeng et al.~\cite{zeng2023transformers} introduce a set of embarrassingly simple one-layer linear models named DLinear for long-term time series forecasting. Given that DLinear runs faster and achieves good performance in the forecasting task, we select it as a supervised baseline for experimental analysis. We use the open source code from~\url{https://github.com/vivva/DLinear} for experimental analysis.}

{}{GPT4TS: Zhou et al.~\cite{zhou2023one} introduce a unified framework for time series modeling utilizing pre-trained Large Language Models (LLMs). Specifically, GPT4TS employs the patch and channel-independent strategies from PatchTST~\cite{nietime} for fine-tuning LLMs on time series downstream tasks. We use the open source code from~\url{https://github.com/DAMO-DI-ML/NeurIPS2023-One-Fits-All} for experimental analysis.}

{}{iTransformer: Liu et al.~\cite{liuitransformer} propose to investigate why Transformer-based models do not seem to be as efficient as Linear-based models~\cite{zeng2023transformers} for multivariate time series forecasting. Consequently, we select it as a supervised baseline for experimental analysis. We use the open source code from~\url{https://github.com/thuml/iTransformer} for experimental analysis.}

{}{TEMPO: Cao et al.~\cite{caotempo} propose TEMPO, which leverages a pre-trained language model for time-series forecasting tasks. The two main components of the proposed approach: the decomposition of time series into trend, seasonality, and residuals, as well as the prompt learning—effectively enhance the forecasting performance. We use the open source code from~\url{https://github.com/DC-research/TEMPO} for experimental analysis.}

\subsubsection{Time-Series Anomaly Detection}

TS2Vec: For the time-series anomaly detection task, Yue et al.~\cite{yue2022ts2vec} followed a streaming evaluation protocol~\cite{ren2019time} to perform point anomaly detection of time series.
We use the open source code from~\url{https://github.com/yuezhihan/ts2vec} for experimental analysis.

SPOT and DSPOT: Siffer et al.~\cite{siffer2017anomaly} proposed a time series anomaly detection method based on extreme value theory. 
The authors divided the proposed approach into two algorithms: \textbf{SPOT} for streaming data having any stationary distribution, and \textbf{DSPOT} for streaming data that can be subject to concept drift.
We use the open source code from~\url{https://github.com/Amossys-team/SPOT} for experimental analysis.

LSTM-VAE: Park et al.~\cite{park2018multimodal} proposed a \textbf{L}ong \textbf{S}hort-\textbf{T}erm \textbf{M}emory-based \textbf{V}ariational \textbf{a}uto\textbf{E}ncoder (\textbf{LSTM-VAE}) model for outlier detection of multimodal sensory signals.
We use the open source code from~\url{https://github.com/SchindlerLiang/VAE-for-Anomaly-Detection} for experimental analysis.

DONUT: Xu et al.~\cite{xu2018unsupervised} proposed an unsupervised anomaly detection algorithm based on a variational autoencoder called \textbf{DONUT}.
We use the open source code from~\url{https://github.com/NetManAIOps/donut} for experimental analysis.

SR: Ren et al.~\cite{ren2019time} proposed an algorithm based on \textbf{S}pectral \textbf{R}esidual (\textbf{SR}) and CNN for anomaly detection of time series.
Since the authors do not provide the open source code, we use the experimental results on Yahoo and KPI datasets in the original paper for comparative analysis.

AT: Xu et al.~\cite{xu2021anomaly} proposed \textbf{A}nomaly \textbf{T}ransformer~(\textbf{AT}) with a new anomaly-attention mechanism for unsupervised detection of anomaly points in time series.
We use the open source code from~\url{https://github.com/spencerbraun/anomaly_transformer_pytorch} for experimental analysis.

{}{TimesNet: Wu et al.~\cite{wutimesnet} propose TimesNet, which incorporates TimesBlock to identify multiple periods and capture temporal 2D variations from transformed 2D tensors using a parameter-efficient inception block. For the anomaly detection task, we utilize TimesNet as a backbone to directly train and analyze its potential in time series pre-training. We use the open source code from~\url{https://github.com/thuml/TimesNet} for experimental analysis.}

{}{GPT4TS: Zhou et al.~\cite{zhou2023one} introduce a unified framework for time series modeling using pre-trained Large Language Models (LLMs). Specifically, GPT4TS applies the patch and channel-independent strategies from PatchTST~\cite{nietime} to fine-tune LLMs for time series anomaly detection. We use the open source code from~\url{https://github.com/DAMO-DI-ML/NeurIPS2023-One-Fits-All} for experimental analysis.}

{}{DCdetector: Yang et al.~\cite{yang2023dcdetector} propose DCdetector, a multi-scale dual attention self-supervised contrastive representation learning model for time series anomaly detection. Specifically, DCdetector employs the patch strategy similar to PatchTST~\cite{nietime} to process raw time series data. We use the open source code from~\url{https://github.com/DAMO-DI-ML/KDD2023-DCdetector} for experimental analysis.}

For the above baselines in three major time-series mining tasks~(classification, forecasting, and anomaly detection), we use uniform random seeds for the model's training.
In addition, the dataset partitioning used by all baselines is kept consistent.
Further, the hyperparameter settings of all baselines are set according to the parameters provided by the original authors, and the details can be found in our open-source code~\url{https://github.com/qianlima-lab/time-series-ptms}.

\subsection{Implementation Details~\label{appendix_B4}}

{}{PTMs are usually trained in two stages. Firstly, supervised, unsupervised or self-supervised techniques are used to pre-train the base model on the source dataset. Then, the base model is fine-tuned using the training set of the target dataset. Finally, the base model is evaluated on the target test set to obtain the test results.
However, there is no available benchmark large-scale well-labelled time series dataset.
In other words, selecting a suitable source dataset to pre-train the encoder to obtain a positive transfer performance on the target dataset is difficult.
To address the above issues, existing studies (e.g., TS2Vec~\cite{yue2022ts2vec}, TST~\cite{zerveas2021transformer}, TS-TCC~\cite{eldele2021time}, and CoST~\cite{woo2022cost}) utilize self-supervised or unsupervised learning strategies to pre-train the base model using the target dataset and then fine-tune it on the target dataset.
We conducted extensive experiments on two fronts to evaluate the effectiveness of existing TS-PTMs. On the one hand, we selected the UCR archive and four scenarios time series datasets for transfer learning using the selected source and target datasets, thus analyzing the performance of different pre-training techniques on the downstream classification task. On the other hand, following the strategy of studies in TS2Vec~\cite{yue2022ts2vec} and CoST~\cite{woo2022cost}, we compare and analyze the strategy~(using the target dataset to pretrain the base model) on time-series downstream classification, forecasting, and anomaly detection tasks.
Through extensive experiments, we aim to provide guidance for the study of pre-training paradigms, techniques for TS-PTMs on different downstream tasks.}

For the time-series classification task, we normalize each series of the UCR and UEA datasets via z-score~\cite{ismail2019deep}, and build FCN and TCN models using Pytorch according to the settings of~\cite{ismail2019deep,franceschi2019unsupervised}.
Adam is adopted as the optimizer with a learning rate of 0.001 and a maximum batch size of 128. The maximum epoch of the source UCR times series dataset for transfer learning is 2000, while the maximum epoch of the target UCR times series dataset is 1000. Also, a uniform early stopping rule is used for all baselines in training.
The maximum epoch of the source dataset of four independent time series scenarios~(refer to Table~\ref{tab:my-table-transfer-scenario}) is 40, while the maximum epoch of the corresponding target dataset is 100.
For the classification accuracy of the target UCR time series, we use the average test accuracy on five-fold test sets by running one seed.
For the classification accuracy of the target datasets on four independent time series scenarios, we use the average test accuracy on the test set by running ten different seeds.
For the time-series forecasting and anomaly detection tasks, we follow the way of ~\cite{yue2022ts2vec} to preprocess the datasets and set the hyperparameters. In addition, the comparative benchmark methods for time-series classification, forecasting, and anomaly detection tasks are reproduced and analyzed using the open-source codes provided by the original authors.
To ensure the reproducibility of the experimental results, we use a uniform random seed for all baselines. Finally, all experiments are done on eight 1080Ti GPUs and two 3090 GPUs with Ubuntu 18.04 system. Also, the training times in the main experimental results are all run on 3090 GPUs.

\section{Full Results~\label{appendix_C}}
% In this section, we give detailed classification results based on time series PTMs on 128 UCR and 30 UEA datasets.

\begin{figure*}
	\centering 
	\includegraphics[width=1.1\textwidth]{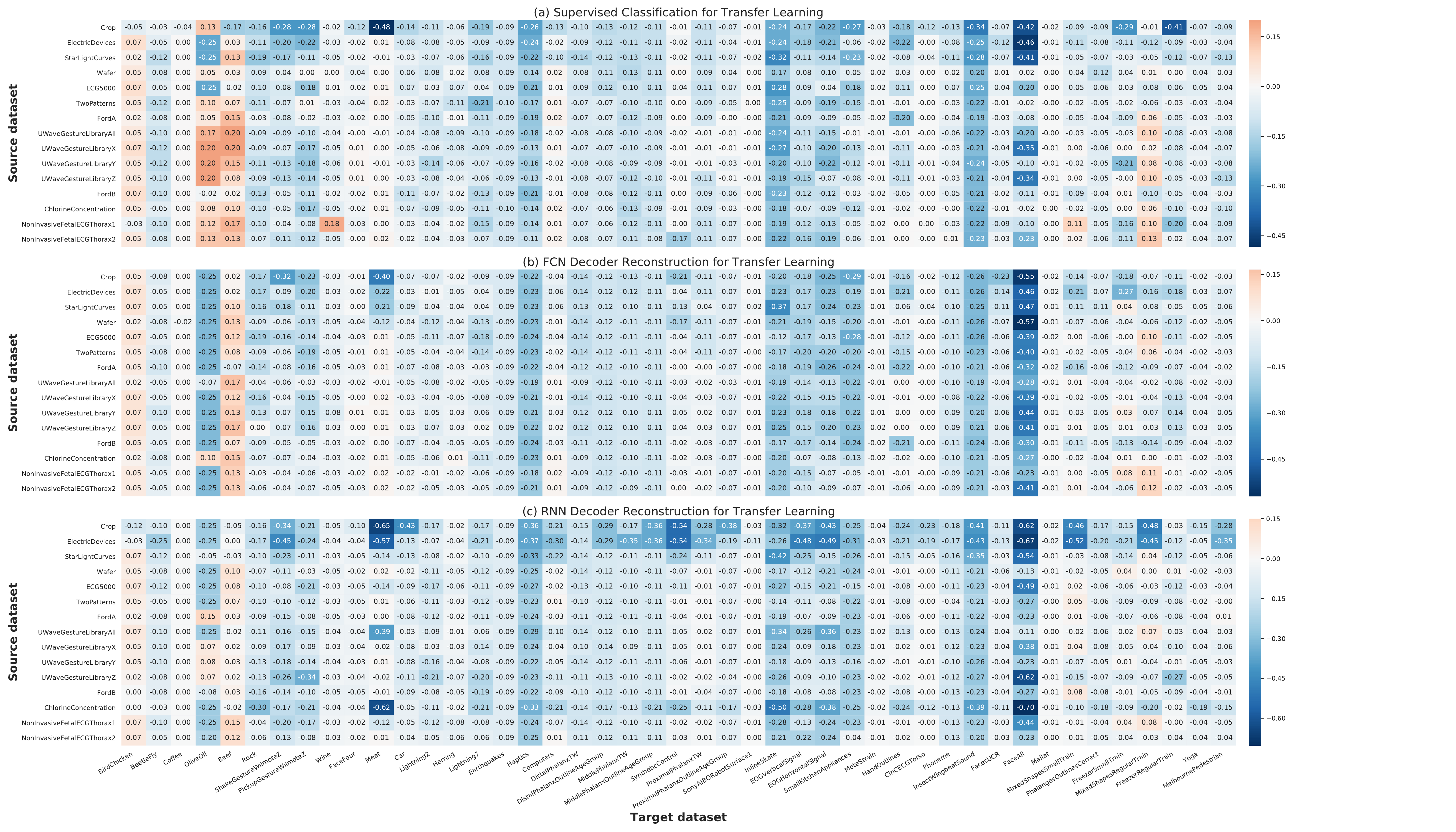}
	\caption{Classification comparison results based on transfer learning. The values indicate the difference in classification accuracy on the FCN encoder with and without transfer learning. A value greater than zero indicates positive transfer, while a value less than zero indicates negative transfer. Red background is for positive values and blue for negative, with larger magnitude values more saturated.} 
	\label{fig:transfer_cls_result}
\end{figure*}

% \usepackage{booktabs}
% \begin{table}[]
% \caption{Details of the source datasets used for transfer learning.}
% \label{tab:my-table-source}
% \center
% \scalebox{0.75}{
% \begin{tabular}{@{}cccccc@{}}
% \toprule
% ID & Dataset Name & Domain & Total Sample Size & Length & Class \\ \midrule
% 1 & Crop & Image & 24000 & 46 & 24 \\
% 2 & ElectricDevices & Device & 16637 & 96 & 7 \\
% 3 & StarLightCurves & Sensor & 9236 & 1024 & 3 \\
% 4 & Wafer & Sensor & 7164 & 152 & 2 \\
% 5 & ECG5000 & ECG & 5000 & 140 & 5 \\
% 6 & TwoPatterns & Simulated & 5000 & 128 & 4 \\
% 7 & FordA & Sensor & 4921 & 500 & 2 \\
% 8 & UWaveGestureLibraryAll & Motion & 4478 & 945 & 8 \\
% 9 & UWaveGestureLibraryX & Motion & 4478 & 315 & 8 \\
% 10 & UWaveGestureLibraryY & Motion & 4478 & 315 & 8 \\
% 11 & UWaveGestureLibraryZ & Motion & 4478 & 315 & 8 \\
% 12 & FordB & Sensor & 4446 & 500 & 2 \\
% 13 & ChlorineConcentration & Sensor & 4307 & 166 & 3 \\
% 14 & NonInvasiveFetalECGThorax1 & ECG & 3765 & 750 & 42 \\
% 15 & NonInvasiveFetalECGThorax2 & ECG & 3765 & 750 & 42 \\ \bottomrule
% \end{tabular}}
% \end{table}

% \usepackage{booktabs}
\begin{table}[]
\caption{Details of the source UCR time series datasets used for transfer learning.}
\label{tab:my-table-source}
\center
\scalebox{0.9}{
\begin{tabular}{@{}ccccc@{}}
\toprule
ID & Dataset Name & Total Sample Size & Length & Class \\ \midrule
1 & Crop & 24000 & 46 & 24 \\
2 & ElectricDevices & 16637 & 96 & 7 \\
3 & StarLightCurves & 9236 & 1024 & 3 \\
4 & Wafer & 7164 & 152 & 2 \\
5 & ECG5000 & 5000 & 140 & 5 \\
6 & TwoPatterns & 5000 & 128 & 4 \\
7 & FordA & 4921 & 500 & 2 \\
8 & UWaveGestureLibraryAll & 4478 & 945 & 8 \\
9 & UWaveGestureLibraryX & 4478 & 315 & 8 \\
10 & UWaveGestureLibraryY & 4478 & 315 & 8 \\
11 & UWaveGestureLibraryZ & 4478 & 315 & 8 \\
12 & FordB & 4446 & 500 & 2 \\
13 & ChlorineConcentration & 4307 & 166 & 3 \\
14 & NonInvasiveFetalECGThorax1 & 3765 & 750 & 42 \\
15 & NonInvasiveFetalECGThorax2 & 3765 & 750 & 42 \\ \bottomrule
\end{tabular}}
\end{table}

\begin{table}[]
\caption{Details of the target UCR time series datasets used for transfer learning.}
\label{tab:my-table-target}
\center
\scalebox{0.9}{
\begin{tabular}{@{}ccccc@{}}
\toprule
\multicolumn{5}{c}{\textbf{15 Minmum Sample Size Target Datasets}} \\ \midrule
ID & Dataset Name & Total Sample Size & Length & Class \\ \hline
1 & BirdChicken & 40 & 512 & 2 \\
2 & BeetleFly & 40 & 512 & 2 \\
3 & Coffee & 56 & 286 & 2 \\
4 & OliveOil & 60 & 570 & 4 \\
5 & Beef & 60 & 470 & 5 \\
6 & Rock & 70 & 2844 & 4 \\
7 & ShakeGestureWiimoteZ & 100 & 385 & 10 \\
8 & PickupGestureWiimoteZ & 100 & 361 & 10 \\
9 & Wine & 111 & 234 & 2 \\
10 & FaceFour & 112 & 350 & 4 \\
11 & Meat & 120 & 448 & 3 \\
12 & Car & 120 & 577 & 4 \\
13 & Lightning2 & 121 & 637 & 2 \\
14 & Herring & 128 & 512 & 2 \\
15 & Lightning7 & 143 & 319 & 7 \\\toprule
\multicolumn{5}{c}{\textbf{15 Medium Sample Size Target Datasets}} \\
\midrule
ID & Dataset Name & Total Sample Size & Length & Class \\ \hline
1 & Earthquakes & 461 & 512 & 2 \\
2 & Haptics & 463 & 1092 & 5 \\
3 & Computers & 500 & 720 & 2 \\
4 & DistalPhalanxTW & 539 & 80 & 6 \\
5 & DistalPhalanxOutlineAgeGroup & 539 & 80 & 3 \\
6 & MiddlePhalanxTW & 553 & 80 & 6 \\
7 & MiddlePhalanxOutlineAgeGroup & 554 & 80 & 3 \\
8 & SyntheticControl & 600 & 60 & 6 \\
9 & ProximalPhalanxTW & 605 & 80 & 6 \\
10 & ProximalPhalanxOutlineAgeGroup & 605 & 80 & 3 \\
11 & SonyAIBORobotSurface1 & 621 & 70 & 2 \\
12 & InlineSkate & 650 & 1882 & 7 \\
13 & EOGHorizontalSignal & 724 & 1250 & 12 \\
14 & EOGVerticalSignal & 724 & 1250 & 12 \\
15 & SmallKitchenAppliances & 750 & 720 & 3 \\ \toprule
\multicolumn{5}{c}{\textbf{15 Maxmum Sample Size Target Datasets}} \\
\midrule
ID & Dataset Name & Total Sample Size & Length & Class \\ \hline
1 & MoteStrain & 1272 & 84 & 2 \\
2 & HandOutlines & 1370 & 2709 & 2 \\
3 & CinCECGTorso & 1420 & 1639 & 4 \\
4 & Phoneme & 2110 & 1024 & 39 \\
5 & InsectWingbeatSound & 2200 & 256 & 11 \\
6 & FacesUCR & 2250 & 131 & 14 \\
7 & FaceAll & 2250 & 131 & 14 \\
8 & Mallat & 2400 & 1024 & 8 \\
9 & MixedShapesSmallTrain & 2525 & 1024 & 5 \\
10 & PhalangesOutlinesCorrect & 2658 & 80 & 2 \\
11 & FreezerSmallTrain & 2878 & 301 & 2 \\
12 & MixedShapesRegularTrain & 2925 & 1024 & 5 \\
13 & FreezerRegularTrain & 3000 & 301 & 2 \\
14 & Yoga & 3300 & 426 & 2 \\
15 & MelbournePedestrian & 3633 & 24 & 10 \\ \bottomrule
\end{tabular}}
\end{table}

% \usepackage{booktabs}
% \usepackage{multirow}
\begin{table*}[]
\caption{The detailed information of four independent time series scenarios datasets. For column \textit{Samples}, \textit{Train} is the training set, \textit{Val} is the validation set, and \textit{Test} is the test set.}
\begin{tabular}{@{}ccccccc@{}}
\toprule
Scenario & Transfer Learning & Dataset & \# Samples & \# Channels & \# Classes & \# Length \\ \midrule
\multirow{2}{*}{Neurological Stage Detection} & Pre-training (Source) & SleepEEG & Train (371055) & 1 & 5 & 200 \\ 
 & Fine-turning (Target) & Epilepsy & Train (60), Val (20), Test (11420) & 1 & 2 & 178 \\ \hline
\multirow{2}{*}{Mechanical Device Diagnosis} & Pre-training (Source) & FD-A & Train (8184) & 1 & 3 & 5120 \\ 
 & Fine-turning (Target) & FD-B & Train (60), Val (21), Test (13559) & 1 & 3 & 5120 \\ \hline
\multirow{2}{*}{Activity Recognition} & Pre-training (Source) & HAR & Train (10299) & 9 & 6 & 128 \\
 & Fine-turning (Target) & Gesture & Train (320), Val (120), Test (120) & 3 & 8 & 315 \\ \hline
\multirow{2}{*}{Physical Status Monitoring} & Pre-training (Source) & ECG & Train (43673) & 1 & 4 & 1500 \\ 
 & Fine-turning (Target) & EMG & Train (122), Val (41), Test (41) & 1 & 3 & 1500 \\ \cmidrule{1-7} 
\end{tabular}

\label{tab:my-table-transfer-scenario}
\end{table*}

\subsection{Comparison of Transfer Learning PTMs based on Supervised Classification and Unsupervised Reconstruction~\label{appendix_C1}}

% Still, the average standard deviations of the test classification accuracy on the 128 UCR time series datasets have been given in the main text.

{}{Advanced time series classification methods~(i.e., TS-CHIEF~\cite{shifaz2020ts}, miniRocket~\cite{dempster2021minirocket},  and OS-CNN~\cite{tang2021omni}) methods either integrate multiple techniques or require finding the optimal hyperparameters in training and are not suitable as the backbone of PTMs.
In recent years, related scholars have employed FCN~\cite{fawaz2018transfer}, TCN~\cite{yue2022ts2vec}, and Transformer~\cite{zerveas2021transformer} as the backbone for studying TS-PTMs. 
The test supervised classification accuracy of FCN, TCN, and Transformer on 128 UCR time series datasets is given in Table~\ref{tab:my-table-pre}.
To facilitate the layout and reading of the test classification results, the standard deviation of the classification accuracy for each dataset is not given in Table~\ref{tab:my-table-pre}.
Considering the training time and classification performance on 128 UCR time series datasets of the models~(please refer to Table~\ref{tab:my-table-pre}), we choose FCN for transfer learning.
% Like~\cite{fawaz2018transfer}, we utilize an FCN encoder combined with a linear classifier for transfer learning.
From the 128 UCR datasets,
the 15 datasets with the largest numbers of samples 
are employed as source datasets. The FCN encoder is pre-trained using the supervised classification task or unsupervised reconstruction task combined with a decoder~(symmetric FCN decoder or asymmetric RNN decoder).
From the remaining 113 UCR datasets,
we select a total of 45 time series datasets 
as 
target datasets for downstream classification task fine-tuning. 
15 of them have the smallest numbers of samples,
15 of them have the largest numbers of samples, and
15 of them have the medium numbers of samples.
% Please refer to
% Tables 4 and 5 for
% details on the source and target datasets.
For the datasets used in transfer learning PTMs, please refer to Tables~\ref{tab:my-table-source},~\ref{tab:my-table-target}, and~\ref{tab:my-table-transfer-scenario}.
For the source dataset, we employ all samples to pre-train the FCN encoder. 
Using the five-fold cross-validation strategy, 
each target dataset contains five different training sets, and
so we performed five fine-tunings for analysis.
For each transfer strategy, 
$15\times 15=
225$ sets of transfer results~(15 source datasets, 15 target datasets) are obtained for each 
target dataset
sample size~{}{(minimum, medium, and maximum)}.
The detailed transfer learning results are shown in Fig.~\ref{fig:transfer_cls_result}.}

\subsection{Comparison of TS-PTMs on Classification~\label{appendix_C2}}

The pre-training test classification accuracy using transformer-based and contrastive learning on 128 UCR and 30 UEA time series datasets are shown in Tables~\ref{tab:my-table2} and~\ref{tab:my-table3}.
Also, for the convenience of layout and reading
of the test classification results, the standard deviations of
the classification performance evaluated for each dataset are
not given in Tables~\ref{tab:my-table2} and~\ref{tab:my-table3}.

\subsubsection{{}{Visualization on the classification task}}

% In this section, we choose Multi-Dimensional Scaling (MDS)~\cite{kruskal1978multidimensional}, t-Distributed Stochastic Neighbor Embedding (t-SNE)~\cite{van2008visualizing}, 

{}{In this section, we use the Class Activation Map (CAM)~\cite{zhou2016learning}
and heatmap~\cite{yue2022ts2vec} for visualization of the time series classification models. 
CAM is a way to visualize CNNs and analyze which regions of the input data the CNN-based model focuses on.
The heatmap is used in~\cite{yue2022ts2vec} to analyze the heat distribution of the 16-dimensional variables with the largest variance among the feature variables, thus assisting to analyze the trend of the time series.
Hence, the ability of  the model to capture the change in time series distributions can be measured based on the heat distribution. 
We employ the validation sets from the five-fold cross-validation strategy to select models for all comparison settings. 
Among the five models for analysis,
we choose the one with the most significant visualization difference 
to highlight the visualization.}

% \begin{figure}[t]
%     \centering
%     \subfigure[Wine]{
%         \begin{minipage}[t]{0.23\textwidth}
%         \centering
%         \includegraphics[width=\textwidth]{figures/mds_seed_42_Wine.pdf}
%         \end{minipage}
%     }
%     \subfigure[FreezerSmallTrain]{
%         \begin{minipage}[t]{0.23\textwidth}
%         \centering
%         \includegraphics[width=\textwidth]{figures/tsne_seed_42_FreezerSmallTrain.pdf}
%         \end{minipage}
%     }
    
%     \caption{Embedding visualization analysis based on FCN combined with a linear classifier and nonlinear classifier.}
%     \label{fig:visualization_mds_tsne}
% \end{figure}

% We select the Wine and FreezerSmallTrain datasets from the UCR archive, and perform a dimensionality reduction visualization analysis to compare the linear and nonlinear classifiers in FCN.
% Among them, Fig.~\ref{fig:visualization_mds_tsne} (a) employs MDS for dimensionality reduction visualization of the Wine dataset, while Fig.~\ref{fig:visualization_mds_tsne} (b) utlizes t-SNE for dimensionality reduction visualization of the FreezerSmallTrain dataset.
% As shown in Fig.~\ref{fig:visualization_mds_tsne}, the use of a nonlinear classifier enables a significant difference in the embedding class clusters compared with the FCN combined with a linear classifier.
% In addition, we choose Class Activation Map~\cite{zhou2016learning} for visualization  on the Gunpoint dataset to analyze the transfer learning strategy, and please refer to Fig.~\ref{fig:visualization_cam} in the Appendix~\ref{appendix_B3} for detailed analysis.

{}{The Gunpoint dataset in the UCR archive contains two types of series, representing a male or female performing two actions (using a replica gun or finger to point to the target)~\cite{lines2012shapelet}. 
Fawaz et al.~\cite{ismail2019deep} employed the Gunpoint dataset for CAM visualization due to its 100\% classification accuracy on FCN, low noise, and containing only two types of data.
Therefore, we perform a CAM visualization on the Gunpoint dataset, as shown in Fig.~\ref{fig:visualization_cam}.}

{}{We select the sample with the smallest variance in each class of series in the Gunpoint dataset for the CAM visualization.
As shown in Fig.~\ref{fig:visualization_cam}, the above two series are
discriminative between the two fragments at [45, 60] and [90, 105].
The direct classification using FCN can learn the discriminative subseries information 
between the two class of series
well.
However, when the TCN is utilized for direct classification, only one class of
series is marked in red (on the fragment [90, 105]), while the other class is
marked in blue, resulting in low classification performance.
Meanwhile, we select UWaveGestureLibraryX as the source dataset for transfer learning of the Gunpoint dataset.
Regarding the supervised classification-based transfer strategy, the FCN model can
simultaneously make the two classes of series marked in red for
the [90, 105] fragment.
For the unsupervised transfer strategy, the FCN model can
make one class of series marked in dark red on the [45, 60] and [90,
105] fragments, and another class of series marked in yellow on the [45, 60] and [90, 105] fragments.}
% width=3.4in,height=0.28
% width=0.5
\begin{figure}
	\centering 
	\includegraphics[width=3.6in,height=0.4\textwidth]{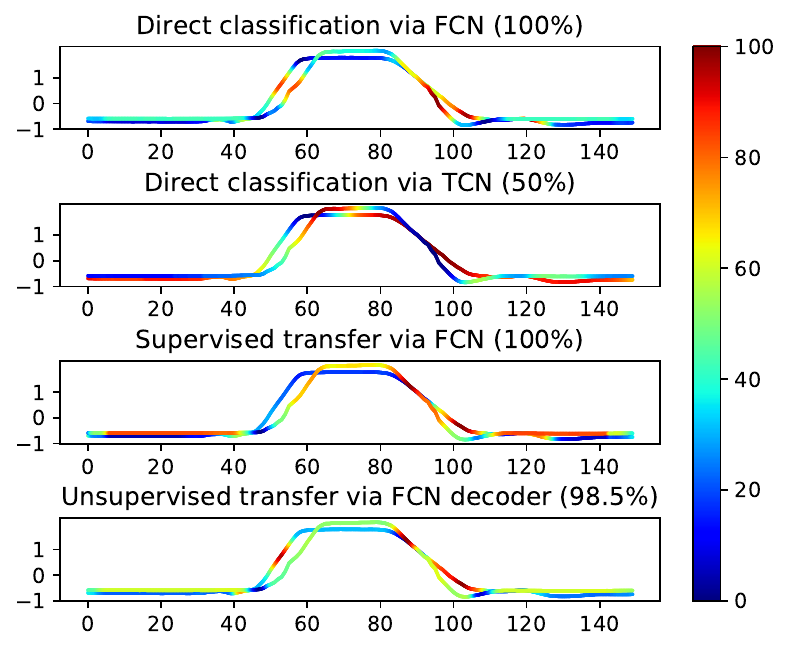}
	\caption{Visualization of the Gunpoint dataset using CAM. 
 For the discriminative features learned by the CNN-based model, {red} represents high contribution, {blue} indicates almost no contribution, and {yellow} is medium contribution~(each subplot title gives the accuracy).} 
	\label{fig:visualization_cam}
\end{figure}

{}{The MixedShapesSmallTrain and Wine datasets from the UCR archive are selected as the target datasets to analyze the learning of positive and negative transfer on time-series classification.
Based on the classification results of transfer learning in Fig. 1 in the Appendix, we select the source datasets with the best accuracy for positive transfer and the worst accuracy for negative transfer.
From Figs.~\ref{fig:visualization_heatmap} and~\ref{fig:visualization_heatmap1}, the heat distribution of the variable with the largest 16-dimensional variance obtained by direct classification and positive transfer is more similar to the trend of the original time series, while the negative transfer is difficult to present a heat distribution that matches the trend of the original time series.
The heatmap visualization indicates that negative transfer may make it difficult for the model to capture the dynamic change information of the original time series, leading to degraded classification performance.}

\begin{figure}
	\centering 
	\includegraphics[width=3.6in,height=0.4\textwidth]{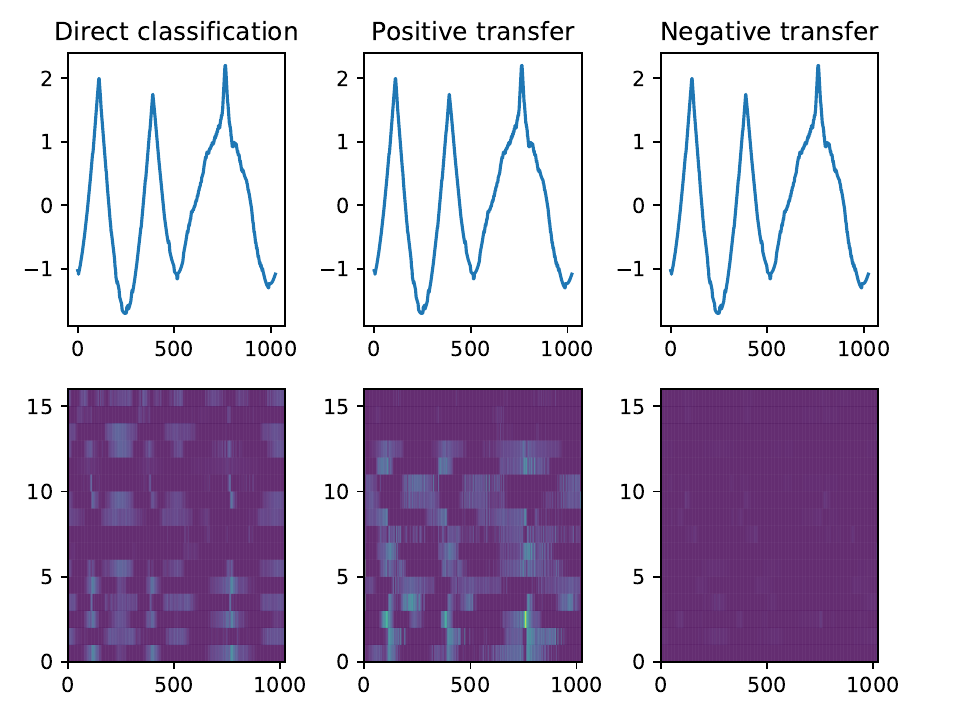}
	\caption{Comparison of positive and negative transfer based on supervised classification using the MixedShapesSmallTrain dataset.} 
	\label{fig:visualization_heatmap}
\end{figure}

\begin{figure}
	\centering 
	\includegraphics[width=3.6in,height=0.4\textwidth]{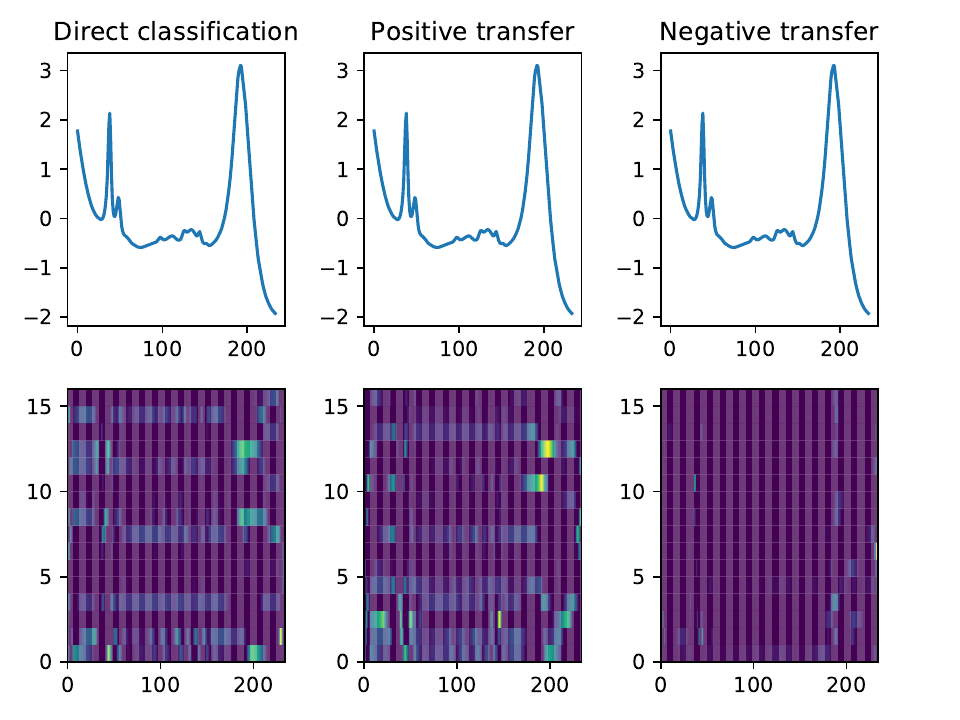}
	\caption{Comparison of positive and negative transfer based on supervised classification using the Wine dataset.} 
	\label{fig:visualization_heatmap1}
\end{figure}

\begin{table}[]
\centering
\caption{Test classification accuracy using FCN, TCN, and Transformer for directly supervised classification training on 128 UCR datasets.}
% \center
\scalebox{0.56}{
\begin{tabular}{@{}clcccc@{}}
\toprule
\textbf{ID} & {Dataset} & {FCN} & {TCN (three layers)} & {TCN (ten layers)} & {Transformer} \\ \midrule
1 & ACSF1 & {\ul \textbf{0.6250}} & 0.3950 & 0.5400 & 0.3450 \\
2 & Adiac & {\ul \textbf{0.6492}} & 0.2549 & 0.2894 & 0.6415 \\
3 & AllGestureWiimoteX & {\ul \textbf{0.7110}} & 0.2370 & 0.6630 & 0.3720 \\
4 & AllGestureWiimoteY & 0.6740 & 0.4980 & {\ul \textbf{0.6960}} & 0.4950 \\
5 & AllGestureWiimoteZ & {\ul \textbf{0.7200}} & 0.7050 & 0.6300 & 0.3380 \\
6 & ArrowHead & {\ul \textbf{0.8958}} & 0.3838 & 0.3838 & 0.8676 \\
7 & Beef & 0.6000 & 0.3167 & 0.4167 & {\ul \textbf{0.6500}} \\
8 & BeetleFly & 0.8250 & 0.6500 & 0.7000 & {\ul \textbf{0.8500}} \\
9 & BirdChicken & 0.9250 & 0.6250 & 0.8500 & {\ul \textbf{0.9944}} \\
10 & BME & 0.7833 & 0.9944 & {\ul \textbf{1.0000}} & 0.7167 \\
11 & Car & 0.9417 & 0.8750 & 0.9417 & {\ul \textbf{1.0000}} \\
12 & CBF & {\ul \textbf{1.0000}} & {\ul \textbf{1.0000}} & {\ul \textbf{1.0000}} & 0.7250 \\
13 & Chinatown & 0.9753 & 0.9808 & {\ul \textbf{0.9836}} & {\ul \textbf{0.9836}} \\
14 & ChlorineConcentration & {\ul \textbf{0.9984}} & 0.5356 & 0.5407 & 0.9970 \\
15 & CinCECGTorso & {\ul \textbf{0.9986}} & 0.9915 & 0.9500 & 0.9951 \\
16 & Coffee & {\ul \textbf{1.0000}} & {\ul \textbf{1.0000}} & {\ul \textbf{1.0000}} & {\ul \textbf{1.0000}} \\
17 & Computers & {\ul \textbf{0.8800}} & 0.8740 & 0.5000 & 0.6680 \\
18 & CricketX & {\ul \textbf{0.9064}} & 0.7615 & 0.6974 & 0.5513 \\
19 & CricketY & {\ul \textbf{0.9000}} & 0.6551 & 0.7013 & 0.5526 \\
20 & CricketZ & {\ul \textbf{0.8833}} & 0.8026 & 0.7679 & 0.5641 \\
21 & Crop & 0.1312 & 0.0417 & 0.0417 & {\ul \textbf{0.7239}} \\
22 & DiatomSizeReduction & {\ul \textbf{0.9969}} & 0.8913 & 0.6569 & {\ul \textbf{0.9969}} \\
23 & DistalPhalanxOutlineAgeGroup & {\ul \textbf{0.9091}} & 0.5974 & 0.8032 & 0.8107 \\
24 & DistalPhalanxOutlineCorrect & 0.8917 & {\ul \textbf{0.9020}} & 0.8962 & 0.8185 \\
25 & DistalPhalanxTW & {\ul \textbf{0.8423}} & 0.4731 & 0.4731 & 0.7792 \\
26 & DodgerLoopDay & 0.7486 & {\ul \textbf{0.7990}} & 0.2833 & 0.4869 \\
27 & DodgerLoopGame & 0.8167 & 0.5190 & 0.5190 & {\ul \textbf{0.8798}} \\
28 & DodgerLoopWeekend & 0.9812 & 0.7089 & 0.7089 & {\ul \textbf{0.9871}} \\
29 & Earthquakes & {\ul \textbf{0.8482}} & 0.7983 & 0.7983 & 0.5995 \\
30 & ECG200 & 0.9350 & {\ul \textbf{0.9400}} & 0.6650 & 0.7374 \\
31 & ECG5000 & {\ul \textbf{0.9258}} & 0.5838 & 0.5380 & 0.8750 \\
32 & ECGFiveDays & {\ul \textbf{1.0000}} & {\ul \textbf{1.0000}} & {\ul \textbf{1.0000}} & 0.9494 \\
33 & ElectricDevices & {\ul \textbf{0.6731}} & 0.1845 & 0.2633 & 0.5428 \\
34 & EOGHorizontalSignal & 0.7376 & 0.5347 & 0.7336 & {\ul \textbf{1.0000}} \\
35 & EOGVerticalSignal & 0.6519 & 0.7003 & 0.7514 & {\ul \textbf{0.8053}} \\
36 & EthanolLevel & 0.8307 & 0.3717 & {\ul \textbf{0.9154}} & 0.2720 \\
37 & FaceAll & 0.9462 & {\ul \textbf{0.9911}} & 0.9711 & 0.9556 \\
38 & FaceFour & 0.9557 & {\ul \textbf{0.9913}} & 0.9648 & 0.9296 \\
39 & FacesUCR & {\ul \textbf{0.9969}} & 0.9893 & 0.9876 & 0.9618 \\
40 & FiftyWords & 0.5072 & 0.2155 & 0.4243 & {\ul \textbf{0.6740}} \\
41 & Fish & {\ul \textbf{0.9686}} & 0.4086 & 0.9400 & 0.8086 \\
42 & FordA & {\ul \textbf{0.9734}} & 0.9651 & 0.9705 & 0.5237 \\
43 & FordB & 0.9312 & 0.9517 & {\ul \textbf{0.9559}} & 0.5461 \\
44 & FreezerRegularTrain & 0.9790 & 0.5000 & 0.5000 & {\ul \textbf{0.9987}} \\
45 & FreezerSmallTrain & 0.8359 & 0.5000 & 0.5000 & {\ul \textbf{0.9997}} \\
46 & Fungi & 0.9118 & 0.3987 & 0.0980 & {\ul \textbf{1.0000}} \\
47 & GestureMidAirD1 & {\ul \textbf{0.6954}} & 0.2781 & 0.4621 & 0.5087 \\
48 & GestureMidAirD2 & {\ul \textbf{0.5860}} & 0.2930 & 0.4650 & 0.4972 \\
49 & GestureMidAirD3 & {\ul \textbf{0.3819}} & 0.2486 & 0.3407 & 0.2694 \\
50 & GesturePebbleZ1 & {\ul \textbf{0.9541}} & 0.5330 & 0.8852 & 0.6875 \\
51 & GesturePebbleZ2 & {\ul \textbf{0.9343}} & 0.3584 & 0.3185 & 0.6615 \\
52 & GunPoint & {\ul \textbf{1.0000}} & 0.9950 & 0.5900 & 0.9900 \\
53 & GunPointAgeSpan & {\ul \textbf{0.9956}} & {\ul \textbf{0.9956}} & {\ul \textbf{0.9956}} & 0.9822 \\
54 & GunPointMaleVersusFemale & 0.9978 & 0.9956 & 0.9978 & {\ul \textbf{1.0000}} \\
55 & GunPointOldVersusYoung & {\ul \textbf{0.9934}} & 0.9823 & 0.9557 & 0.9756 \\
56 & Ham & 0.6632 & 0.8828 & {\ul \textbf{0.9442}} & 0.8750 \\
57 & HandOutlines & {\ul \textbf{0.8956}} & 0.6752 & 0.6387 & 0.8757 \\
58 & Haptics & 0.6158 & 0.5623 & {\ul \textbf{0.7261}} & 0.4083 \\
59 & Herring & {\ul \textbf{0.6406}} & 0.5938 & 0.6092 & 0.6243 \\
60 & HouseTwenty & {\ul \textbf{0.9875}} & 0.5597 & 0.5597 & 0.7292 \\
61 & InlineSkate & 0.7046 & 0.4923 & {\ul \textbf{0.8169}} & 0.3369 \\
62 & InsectEPGRegularTrain & {\ul \textbf{0.9935}} & 0.8232 & 0.9457 & 0.7651 \\
63 & InsectEPGSmallTrain & {\ul \textbf{0.8761}} & 0.8270 & 0.6836 & 0.7971 \\
64 & InsectWingbeatSound & 0.7309 & {\ul \textbf{0.7527}} & 0.7364 & 0.6691 \\
65 & ItalyPowerDemand & {\ul \textbf{0.9891}} & 0.9781 & 0.9754 & 0.9735 \\
66 & LargeKitchenAppliances & {\ul \textbf{0.9507}} & 0.7680 & 0.8707 & 0.5187 \\
67 & Lightning2 & 0.7940 & 0.6113 & {\ul \textbf{0.8843}} & 0.7593 \\
68 & Lightning7 & {\ul \textbf{0.8406}} & 0.2655 & 0.8123 & 0.6352 \\
69 & Mallat & {\ul \textbf{0.9975}} & 0.9858 & 0.8213 & 0.9800 \\
70 & Meat & 0.9667 & 0.4417 & 0.3917 & {\ul \textbf{0.9917}} \\
71 & MedicalImages & {\ul \textbf{0.8397}} & 0.5206 & 0.5206 & 0.7388 \\
72 & MelbournePedestrian & 0.8145 & 0.8748 & {\ul \textbf{0.9219}} & 0.8792 \\
73 & MiddlePhalanxOutlineAgeGroup & {\ul \textbf{0.8177}} & 0.7635 & 0.7274 & 0.7383 \\
74 & MiddlePhalanxOutlineCorrect & 0.8834 & {\ul \textbf{0.9372}} & 0.9092 & 0.8451 \\
75 & MiddlePhalanxTW & {\ul \textbf{0.7074}} & 0.5967 & 0.5967 & 0.6184 \\
76 & MixedShapesRegularTrain & 0.7463 & 0.2010 & 0.1908 & {\ul \textbf{0.8732}} \\
77 & MixedShapesSmallTrain & 0.7339 & 0.1893 & 0.1893 & {\ul \textbf{0.8745}} \\
78 & MoteStrain & {\ul \textbf{0.9819}} & 0.9811 & 0.9772 & 0.9584 \\
79 & NonInvasiveFetalECGThorax1 & {\ul \textbf{0.8542}} & 0.3057 & 0.4167 & 0.8510 \\
80 & NonInvasiveFetalECGThorax2 & 0.8093 & 0.5216 & 0.4733 & {\ul \textbf{0.8951}} \\
81 & OliveOil & {\ul \textbf{0.7000}} & 0.5000 & 0.4667 & 0.4955 \\
82 & OSULeaf & {\ul \textbf{0.9887}} & 0.6668 & 0.9167 & 0.8667 \\
83 & PhalangesOutlinesCorrect & 0.8691 & 0.9079 & {\ul \textbf{0.9090}} & 0.1469 \\
84 & Phoneme & 0.3019 & 0.1128 & 0.1128 & {\ul \textbf{0.5600}} \\
85 & PickupGestureWiimoteZ & {\ul \textbf{0.7800}} & 0.6300 & 0.6700 & 0.0352 \\
86 & PigAirwayPressure & 0.1091 & 0.0706 & 0.0351 & {\ul \textbf{0.1345}} \\
87 & PigArtPressure & {\ul \textbf{0.5164}} & 0.1251 & 0.0446 & 0.0607 \\
88 & PigCVP & 0.4822 & 0.1764 & 0.0446 & {\ul \textbf{0.5708}} \\
89 & PLAID & 0.4348 & 0.4004 & 0.4321 & {\ul \textbf{0.8295}} \\
90 & Plane & {\ul \textbf{1.0000}} & {\ul \textbf{1.0000}} & 0.6524 & 0.9857 \\
91 & PowerCons & 0.9139 & 0.9222 & {\ul \textbf{0.9694}} & 0.9528 \\
92 & ProximalPhalanxOutlineAgeGroup & {\ul \textbf{0.8727}} & 0.7818 & 0.7884 & 0.8496 \\
93 & ProximalPhalanxOutlineCorrect & 0.9214 & 0.9181 & {\ul \textbf{0.9237}} & 0.8833 \\
94 & ProximalPhalanxTW & 0.8083 & 0.7074 & 0.4165 & {\ul \textbf{0.8182}} \\
95 & RefrigerationDevices & 0.5747 & 0.7307 & {\ul \textbf{0.7640}} & 0.4013 \\
96 & Rock & 0.6429 & {\ul \textbf{0.7000}} & 0.3714 & 0.6571 \\
97 & ScreenType & 0.6093 & {\ul \textbf{0.7573}} & 0.7240 & 0.4373 \\
98 & SemgHandGenderCh2 & {\ul \textbf{0.9378}} & 0.6111 & 0.6000 & 0.8422 \\
99 & SemgHandMovementCh2 & {\ul \textbf{0.7156}} & 0.6444 & 0.5711 & 0.4122 \\
100 & SemgHandSubjectCh2 & 0.6922 & 0.3889 & {\ul \textbf{0.7744}} & 0.7567 \\
101 & ShakeGestureWiimoteZ & {\ul \textbf{0.9200}} & 0.3700 & 0.6600 & 0.5300 \\
102 & ShapeletSim & {\ul \textbf{0.9700}} & 0.5000 & 0.5000 & 0.4550 \\
103 & ShapesAll & {\ul \textbf{0.7267}} & 0.2750 & 0.5250 & 0.6675 \\
104 & SmallKitchenAppliances & {\ul \textbf{0.7987}} & 0.3480 & 0.3333 & 0.5320 \\
105 & SmoothSubspace & 0.9467 & 0.9433 & 0.9600 & {\ul \textbf{0.9800}} \\
106 & SonyAIBORobotSurface1 & {\ul \textbf{0.9984}} & 0.9952 & 0.9888 & 0.9952 \\
107 & SonyAIBORobotSurface2 & {\ul \textbf{0.9990}} & 0.9980 & 0.9908 & 0.9643 \\
108 & StarLightCurves & {\ul \textbf{0.9803}} & 0.8555 & 0.6883 & 0.9153 \\
109 & Strawberry & {\ul \textbf{0.9756}} & 0.9705 & 0.9654 & 0.9613 \\
110 & SwedishLeaf & {\ul \textbf{0.9929}} & 0.9040 & 0.7884 & 0.8880 \\
111 & Symbols & {\ul \textbf{0.9961}} & 0.8206 & 0.9843 & 0.9529 \\
112 & SyntheticControl & {\ul \textbf{0.9700}} & 0.5000 & 0.8283 & {\ul \textbf{0.9700}} \\
113 & ToeSegmentation1 & {\ul \textbf{0.9664}} & 0.9518 & 0.5628 & 0.6864 \\
114 & ToeSegmentation2 & {\ul \textbf{0.9282}} & 0.7471 & 0.7471 & 0.8316 \\
115 & Trace & {\ul \textbf{1.0000}} & {\ul \textbf{1.0000}} & {\ul \textbf{1.0000}} & {\ul \textbf{1.0000}} \\
116 & TwoLeadECG & {\ul \textbf{1.0000}} & 0.9991 & 0.9991 & {\ul \textbf{1.0000}} \\
117 & TwoPatterns & 0.9454 & 0.6958 & 0.5568 & {\ul \textbf{0.9862}} \\
118 & UMD & 0.8778 & 0.3333 & 0.9944 & {\ul \textbf{1.0000}} \\
119 & UWaveGestureLibraryAll & 0.9245 & 0.9359 & {\ul \textbf{0.9884}} & 0.8957 \\
120 & UWaveGestureLibraryX & 0.8665 & 0.8761 & {\ul \textbf{0.9134}} & 0.7347 \\
121 & UWaveGestureLibraryY & 0.7906 & 0.8091 & {\ul \textbf{0.8627}} & 0.6876 \\
122 & UWaveGestureLibraryZ & 0.8484 & 0.8569 & {\ul \textbf{0.8629}} & 0.6648 \\
123 & Wafer & {\ul \textbf{1.0000}} & 0.8936 & 0.8936 & 0.9987 \\
124 & Wine & 0.5411 & 0.5316 & 0.5229 & {\ul \textbf{0.7012}} \\
125 & WordSynonyms & 0.5967 & 0.2210 & 0.2210 & {\ul \textbf{0.6243}} \\
126 & Worms & {\ul \textbf{0.7906}} & 0.4225 & 0.4225 & 0.4028 \\
127 & WormsTwoClass & 0.8025 & {\ul \textbf{0.8528}} & 0.5968 & 0.5504 \\
128 & Yoga & 0.9718 & 0.9645 & {\ul \textbf{0.9788}} & 0.8906 \\ \hline
\multicolumn{2}{c}{\textbf{Avg Acc}} & {\ul \textbf{0.8296}} & 0.6610 & 0.6835 & 0.7415 \\
\multicolumn{2}{c}{\textbf{Avg Rank}} & {\ul \textbf{1.68}} & 2.74 & 2.64 & 2.56 \\
\multicolumn{2}{c}{\textbf{P-value}} & - & 1.38E-16 & 1.44E-13 & 5.41E-07 \\
\multicolumn{2}{c}{\textbf{Training Time (hours)}} & 4.87 & {\ul \textbf{3.06}} & 10.99 & 263.48 \\ \bottomrule
\end{tabular}}
\label{tab:my-table-pre}
\end{table}

% \usepackage{booktabs}
\begin{table*}[]
\centering
\caption{{}{Test classification accuracy using different time series PTMs on 128 UCR datasets. The best results are in bold.}}
\label{tab:my-table2}
\scalebox{0.56}{
\begin{tabular}{@{}clccccccccc@{}}
\toprule
ID & Dataset & Supervised~(FCN) & T-Loss & Selftime & TS-TCC & TST & TS2Vec & {}{TimesNet} & {}{PatchTST} & {}{GPT4TS} \\ \midrule
1 & ACSF1 & 0.6250 & 0.8050 & 0.7250 & 0.5700 & 0.7050 & \textbf{0.8800} & 0.6000 & 0.7600 & 0.6200 \\
2 & Adiac & 0.6492 & 0.7811 & 0.7322 & 0.4003 & 0.7363 & 0.8066 & 0.8082 & \textbf{0.8358} & 0.8069 \\
3 & AllGestureWiimoteX & 0.7110 & 0.7170 & 0.6461 & 0.7201 & 0.5280 & \textbf{0.7820} & 0.5120 & 0.5730 & 0.7220 \\
4 & AllGestureWiimoteY & 0.6740 & 0.8110 & 0.7170 & 0.7615 & 0.3850 & \textbf{0.8400} & 0.5820 & 0.6730 & 0.7800 \\
5 & AllGestureWiimoteZ & 0.7200 & 0.7470 & 0.6485 & 0.6567 & 0.4780 & \textbf{0.7730} & 0.6620 & 0.5890 & 0.7300 \\
6 & ArrowHead & \textbf{0.8958} & 0.8770 & 0.8155 & 0.8200 & 0.8722 & 0.8908 & 0.7821 & 0.7078 & 0.8348 \\
8 & Beef & 0.8250 & 0.8750 & 0.5167 & 0.6333 & 0.6667 & 0.6833 & 0.8000 & 0.4167 & \textbf{0.9333} \\
9 & BeetleFly & 0.9250 & \textbf{0.9750} & 0.8000 & 0.6250 & 0.6750 & 0.9250 & 0.7000 & 0.8500 & 0.7250 \\
10 & BirdChicken & 0.7833 & \textbf{0.9833} & 0.9000 & 0.6750 & 0.8500 & 0.9750 & 0.8250 & 0.8500 & 0.7500 \\
7 & BME & 0.6000 & 0.5333 & 0.9722 & 0.9889 & 0.9722 & \textbf{0.9944} & 0.9778 & 0.9500 & 0.9611 \\
12 & Car & \textbf{1.0000} & \textbf{1.0000} & 0.6583 & 0.7167 & 0.7583 & 0.8833 & 0.8417 & 0.9083 & 0.9000 \\
11 & CBF & 0.9417 & 0.8250 & 0.9936 & 0.9967 & 0.9946 & \textbf{1.0000} & \textbf{1.0000} & \textbf{1.0000} & 0.9978 \\
13 & Chinatown & 0.9753 & 0.9807 & 0.9669 & 0.9507 & 0.9699 & 0.9726 & \textbf{0.9863} & 0.9836 & \textbf{0.9863} \\
14 & ChlorineConcentration & 0.9984 & 0.9988 & 0.6116 & 0.8424 & 0.9974 & \textbf{0.9998} & 0.9993 & 0.9995 & \textbf{0.9998} \\
15 & CinCECGTorso & 0.9986 & 0.9944 & 0.9877 & 0.9995 & 0.9993 & 0.9972 & 0.9930 & \textbf{1.0000} & 0.9930 \\
16 & Coffee & \textbf{1.0000} & 0.9818 & 0.9273 & 0.9455 & 0.8758 & \textbf{1.0000} & \textbf{1.0000} & \textbf{1.0000} & \textbf{1.0000} \\
17 & Computers & \textbf{0.8800} & 0.7140 & 0.7920 & 0.6080 & 0.6880 & 0.6940 & 0.8120 & 0.8380 & 0.8080 \\
18 & CricketX & \textbf{0.9064} & 0.7795 & 0.7316 & 0.7327 & 0.7026 & 0.8321 & 0.7974 & 0.7821 & 0.7910 \\
19 & CricketY & \textbf{0.9000} & 0.7487 & 0.6904 & 0.7161 & 0.6962 & 0.8256 & 0.7808 & 0.7821 & 0.8026 \\
20 & CricketZ & \textbf{0.8833} & 0.7833 & 0.7411 & 0.7384 & 0.6936 & 0.8487 & 0.7962 & 0.7923 & 0.8026 \\
21 & Crop & 0.1312 & 0.7063 & 0.6634 & 0.7801 & 0.7709 & 0.7448 & 0.8503 & 0.1193 & \textbf{0.8764} \\
22 & DiatomSizeReduction & 0.9969 & 0.9969 & 0.9723 & 0.9783 & \textbf{1.0000} & 0.9969 & 0.9938 & \textbf{1.0000} & 0.9907 \\
23 & DistalPhalanxOutlineAgeGroup & \textbf{0.9091} & 0.8274 & 0.7959 & 0.8200 & 0.8404 & 0.8125 & 0.8590 & 0.8779 & 0.8890 \\
24 & DistalPhalanxOutlineCorrect & 0.8917 & 0.8471 & 0.7895 & 0.7646 & 0.8254 & 0.8185 & 0.8860 & \textbf{0.9009} & 0.8929 \\
25 & DistalPhalanxTW & 0.8423 & 0.7421 & 0.7514 & 0.7755 & 0.7551 & 0.7792 & \textbf{0.8499} & 0.8485 & 0.8263 \\
26 & DodgerLoopDay & \textbf{0.7486} & 0.4819 & 0.4742 & 0.5885 & 0.5377 & 0.6391 & 0.7107 & 0.6694 & 0.7403 \\
27 & DodgerLoopGame & 0.8167 & 0.8921 & 0.8282 & 0.8857 & 0.8988 & \textbf{0.9563} & 0.8484 & 0.8872 & 0.9001 \\
28 & DodgerLoopWeekend & \textbf{0.9812} & 0.9677 & 0.9681 & 0.9742 & 0.9679 & \textbf{0.9812} & 0.9688 & 0.9492 & 0.9165 \\
34 & Earthquakes & 0.7376 & 0.7417 & 0.7983 & 0.7505 & 0.7961 & 0.7983 & \textbf{0.9136} & 0.9115 & 0.9052 \\
29 & ECG200 & 0.8482 & 0.7983 & 0.7900 & 0.7950 & 0.8600 & 0.8800 & \textbf{0.9100} & 0.8763 & 0.8951 \\
30 & ECG5000 & 0.9350 & 0.8500 & 0.9317 & 0.9525 & 0.9502 & 0.9528 & 0.9708 & 0.9364 & \textbf{0.9746} \\
31 & ECGFiveDays & 0.9258 & 0.9478 & 0.9873 & 0.9972 & 0.9977 & \textbf{1.0000} & \textbf{1.0000} & \textbf{1.0000} & 0.9978 \\
35 & ElectricDevices & 0.6519 & 0.6547 & 0.7173 & 0.8639 & 0.8758 & \textbf{0.8848} & 0.8610 & 0.4883 & 0.8610 \\
32 & EOGHorizontalSignal & \textbf{1.0000} & \textbf{1.0000} & 0.7579 & 0.6295 & 0.6781 & 0.7279 & 0.6092 & 0.8010 & 0.7581 \\
33 & EOGVerticalSignal & 0.6731 & \textbf{0.8736} & 0.6774 & 0.6190 & 0.4806 & 0.6519 & 0.6369 & 0.8013 & 0.7804 \\
36 & EthanolLevel & 0.8307 & 0.5627 & 0.4864 & 0.2985 & 0.7919 & 0.5897 & 0.7690 & 0.7248 & \textbf{0.8337} \\
37 & FaceAll & 0.9462 & 0.9720 & 0.9552 & \textbf{0.9914} & 0.9720 & 0.9871 & 0.9733 & 0.8057 & 0.9877 \\
38 & FaceFour & 0.9557 & 0.9281 & 0.8830 & 0.9379 & 0.9375 & 0.9640 & 0.9557 & 0.9068 & \textbf{0.9826} \\
39 & FacesUCR & \textbf{0.9969} & 0.9729 & 0.9544 & 0.9899 & 0.9702 & 0.9902 & 0.9800 & 0.9824 & 0.9796 \\
40 & FiftyWords & 0.5072 & 0.8088 & 0.6705 & 0.7628 & 0.7381 & 0.8022 & 0.8276 & \textbf{0.8417} & 0.7709 \\
41 & Fish & \textbf{0.9686} & 0.8743 & 0.8514 & 0.8114 & 0.8628 & 0.9343 & 0.8000 & 0.9343 & 0.9143 \\
42 & FordA & \textbf{0.9734} & 0.9114 & 0.8787 & 0.9342 & 0.8990 & 0.9303 & 0.9348 & 0.9500 & 0.9403 \\
43 & FordB & 0.9312 & 0.9035 & 0.8771 & 0.9087 & 0.8650 & 0.9132 & 0.9046 & \textbf{0.9377} & 0.9287 \\
44 & FreezerRegularTrain & 0.9790 & 0.9980 & 0.9974 & 0.9969 & 0.9987 & 0.9977 & \textbf{0.9990} & 0.9393 & 0.9987 \\
45 & FreezerSmallTrain & 0.8359 & 0.9965 & 0.9975 & 0.9966 & 0.9979 & 0.9969 & 0.9969 & 0.8492 & \textbf{0.9983} \\
46 & Fungi & 0.9118 & \textbf{1.0000} & 0.9316 & 0.9557 & 0.9902 & \textbf{1.0000} & 0.8337 & 0.7276 & 0.9049 \\
47 & GestureMidAirD1 & 0.6954 & 0.5919 & \textbf{0.7067} & 0.6120 & 0.6181 & 0.6479 & 0.6099 & 0.6103 & 0.6449 \\
48 & GestureMidAirD2 & \textbf{0.5860} & 0.4973 & 0.5504 & 0.4853 & 0.5475 & 0.5270 & 0.5449 & 0.5692 & 0.5410 \\
49 & GestureMidAirD3 & 0.3819 & 0.3108 & 0.3664 & 0.3137 & 0.3462 & 0.3315 & 0.4559 & 0.4962 & \textbf{0.5282} \\
50 & GesturePebbleZ1 & \textbf{0.9541} & 0.6420 & 0.9043 & 0.8977 & 0.7761 & 0.9507 & \textbf{0.9541} & 0.8316 & 0.9158 \\
51 & GesturePebbleZ2 & 0.9343 & 0.6614 & 0.8817 & 0.8914 & 0.8159 & \textbf{0.9473} & 0.8850 & 0.7800 & 0.9382 \\
52 & GunPoint & \textbf{1.0000} & 0.9850 & 0.9550 & 0.9500 & 0.9700 & 0.9950 & 0.9850 & 0.9800 & 0.9800 \\
53 & GunPointAgeSpan & \textbf{0.9956} & 0.9756 & 0.9800 & 0.9578 & 0.9778 & 0.9889 & 0.9823 & 0.9889 & 0.9890 \\
54 & GunPointMaleVersusFemale & 0.9978 & 0.9911 & 0.9845 & 0.9823 & 0.9867 & 0.9956 & 0.9956 & \textbf{1.0000} & 0.9956 \\
55 & GunPointOldVersusYoung & 0.9934 & 0.9468 & 0.9490 & 0.9512 & 0.9667 & \textbf{0.9956} & 0.9800 & 0.9912 & 0.9691 \\
56 & Ham & 0.6632 & 0.8080 & 0.7193 & 0.8275 & 0.8450 & 0.8741 & 0.9023 & 0.5420 & \textbf{0.9349} \\
57 & HandOutlines & 0.8956 & 0.9073 & 0.8425 & 0.8657 & 0.7170 & 0.9168 & 0.9029 & \textbf{0.9350} & 0.8956 \\
58 & Haptics & 0.6158 & 0.4879 & 0.4730 & 0.4990 & 0.4794 & 0.5594 & 0.6422 & 0.6672 & \textbf{0.6842} \\
59 & Herring & 0.6406 & 0.5929 & 0.6086 & 0.5452 & 0.6400 & 0.6806 & 0.6560 & 0.7604 & \textbf{0.8003} \\
60 & HouseTwenty & \textbf{0.9875} & 0.8935 & 0.9746 & 0.8244 & 0.7673 & 0.9496 & 0.8615 & 0.8617 & 0.8683 \\
61 & InlineSkate & 0.7046 & 0.6354 & 0.1922 & 0.4172 & 0.4754 & 0.6415 & 0.5369 & \textbf{0.8985} & 0.6499 \\
62 & InsectEPGRegularTrain & \textbf{0.9935} & 0.9839 & 0.9616 & 0.8554 & 0.8327 & 0.9807 & 0.8267 & 0.9617 & 0.8307 \\
63 & InsectEPGSmallTrain & 0.8761 & \textbf{0.9850} & 0.9774 & 0.8308 & 0.8235 & 0.9699 & 0.8011 & 0.9511 & 0.8198 \\
64 & InsectWingbeatSound & 0.7309 & 0.6673 & 0.6092 & 0.7119 & 0.6982 & 0.7077 & 0.7355 & \textbf{0.8609} & 0.8518 \\
65 & ItalyPowerDemand & \textbf{0.9891} & 0.9671 & 0.9674 & 0.9660 & 0.9726 & 0.9745 & 0.9863 & 0.9827 & 0.9818 \\
66 & LargeKitchenAppliances & \textbf{0.9507} & 0.8907 & 0.9403 & 0.7401 & 0.5693 & 0.9147 & 0.7520 & 0.8253 & 0.6747 \\
67 & Lightning2 & 0.7940 & \textbf{0.9090} & 0.7860 & 0.7187 & 0.6787 & 0.9010 & 0.8937 & 0.8219 & 0.8946 \\
68 & Lightning7 & 0.8406 & \textbf{0.8596} & 0.7419 & 0.7419 & 0.7059 & 0.8389 & 0.8264 & 0.7025 & 0.8378 \\
69 & Mallat & 0.9975 & 0.9875 & 0.9901 & 0.9857 & 0.9954 & 0.9962 & 0.9954 & \textbf{0.9996} & 0.9983 \\
70 & Meat & 0.9667 & \textbf{1.0000} & 0.8917 & 0.5417 & 0.9833 & \textbf{1.0000} & \textbf{1.0000} & 0.9750 & \textbf{1.0000} \\
71 & MedicalImages & 0.8397 & 0.8291 & 0.7530 & 0.8175 & 0.8010 & 0.8431 & 0.8677 & \textbf{0.8823} & 0.8772 \\
72 & MelbournePedestrian & 0.8145 & 0.8489 & 0.8485 & 0.9141 & 0.5888 & 0.8986 & \textbf{0.9567} & 0.6573 & 0.9321 \\
73 & MiddlePhalanxOutlineAgeGroup & \textbf{0.8177} & 0.7544 & 0.7527 & 0.7563 & 0.7600 & 0.7347 & 0.8140 & 0.8092 & 0.7692 \\
74 & MiddlePhalanxOutlineCorrect & 0.8834 & 0.8294 & 0.7936 & 0.7838 & 0.8507 & 0.8507 & \textbf{0.9339} & 0.9170 & 0.8924 \\
75 & MiddlePhalanxTW & 0.7074 & 0.6202 & 0.6184 & 0.6256 & 0.6456 & 0.6455 & 0.7309 & \textbf{0.7611} & 0.7027 \\
76 & MixedShapesRegularTrain & 0.7463 & 0.9586 & 0.9580 & 0.9310 & 0.9296 & 0.9450 & 0.9053 & 0.7318 & \textbf{0.9645} \\
77 & MixedShapesSmallTrain & 0.7339 & 0.9541 & 0.9530 & 0.9255 & 0.9275 & 0.9410 & 0.9038 & 0.6924 & \textbf{0.9545} \\
78 & MoteStrain & \textbf{0.9819} & 0.9670 & 0.9505 & 0.9435 & 0.9772 & 0.9717 & 0.9623 & 0.9757 & 0.9694 \\
79 & NonInvasiveFetalECGThorax1 & 0.8542 & 0.9384 & 0.8799 & 0.9253 & 0.9129 & 0.9392 & 0.9349 & 0.9504 & \textbf{0.9639} \\
80 & NonInvasiveFetalECGThorax2 & 0.8093 & 0.9461 & 0.9012 & 0.9400 & 0.9286 & 0.9490 & 0.9575 & 0.9451 & \textbf{0.9657} \\
82 & OliveOil & \textbf{0.9887} & 0.8304 & 0.5333 & 0.4000 & 0.5334 & 0.8500 & 0.8500 & 0.6967 & 0.9000 \\
81 & OSULeaf & 0.7000 & 0.8833 & 0.8370 & 0.6493 & 0.6111 & \textbf{0.8937} & 0.8127 & 0.8799 & 0.8085 \\
84 & PhalangesOutlinesCorrect & 0.3019 & 0.3664 & 0.7466 & 0.7803 & 0.8423 & 0.8439 & \textbf{0.9316} & 0.8619 & 0.8548 \\
85 & Phoneme & \textbf{0.7800} & 0.7200 & 0.4144 & 0.3587 & 0.1929 & 0.4204 & 0.1981 & 0.2781 & 0.5569 \\
86 & PickupGestureWiimoteZ & 0.1091 & 0.1700 & 0.8000 & 0.7800 & 0.6200 & \textbf{0.8400} & 0.7200 & 0.6300 & 0.7000 \\
87 & PigAirwayPressure & 0.5164 & \textbf{0.6446} & 0.2182 & 0.0575 & 0.0353 & 0.4038 & 0.2446 & 0.3500 & 0.4280 \\
88 & PigArtPressure & 0.4822 & 0.6731 & \textbf{0.9456} & 0.0929 & 0.1312 & 0.9424 & 0.3826 & 0.5308 & 0.4893 \\
89 & PigCVP & 0.4348 & 0.5121 & 0.7083 & 0.0449 & 0.0321 & \textbf{0.8753} & 0.4181 & 0.5269 & 0.6079 \\
83 & PLAID & \textbf{0.8691} & 0.8424 & 0.3912 & 0.4832 & 0.7421 & 0.5503 & 0.5951 & 0.4107 & 0.5867 \\
90 & Plane & \textbf{1.0000} & \textbf{1.0000} & 0.9762 & 0.9714 & 0.9381 & 0.9905 & 0.9810 & 0.9952 & 0.9810 \\
91 & PowerCons & 0.9139 & 0.9194 & 0.8889 & 0.9444 & 0.9722 & \textbf{0.9861} & 0.9722 & 0.9333 & 0.9833 \\
92 & ProximalPhalanxOutlineAgeGroup & 0.8727 & 0.8496 & 0.8446 & 0.8231 & 0.8496 & 0.8545 & \textbf{0.8860} & 0.8727 & 0.8793 \\
93 & ProximalPhalanxOutlineCorrect & 0.9214 & 0.8889 & 0.8383 & 0.8303 & 0.8799 & 0.8833 & \textbf{0.9406} & 0.9170 & 0.8945 \\
94 & ProximalPhalanxTW & 0.8083 & 0.8215 & 0.8000 & 0.7653 & 0.8347 & 0.8281 & 0.8512 & \textbf{0.8798} & 0.8645 \\
95 & RefrigerationDevices & 0.5747 & \textbf{0.7773} & 0.7193 & 0.5170 & 0.5146 & 0.7627 & 0.5693 & 0.6400 & 0.7267 \\
96 & Rock & 0.6429 & 0.7143 & 0.6857 & 0.7857 & 0.7714 & 0.8286 & 0.8143 & 0.6940 & \textbf{0.9143} \\
97 & ScreenType & 0.6093 & 0.5480 & \textbf{0.6852} & 0.4805 & 0.4827 & 0.5520 & 0.6347 & 0.5093 & 0.6480 \\
98 & SemgHandGenderCh2 & 0.9378 & 0.7456 & 0.8745 & 0.9516 & 0.9111 & \textbf{0.9544} & 0.9367 & 0.9378 & 0.9389 \\
99 & SemgHandMovementCh2 & 0.7156 & 0.4056 & 0.5769 & 0.6980 & 0.4667 & 0.7767 & 0.7800 & 0.6311 & \textbf{0.7989} \\
100 & SemgHandSubjectCh2 & 0.6922 & 0.6356 & 0.8286 & 0.9310 & 0.8467 & 0.9344 & \textbf{0.9411} & 0.8278 & 0.9089 \\
101 & ShakeGestureWiimoteZ & 0.9200 & 0.9100 & 0.8900 & 0.8500 & 0.6800 & \textbf{0.9300} & 0.5900 & 0.6900 & 0.7200 \\
102 & ShapeletSim & 0.9700 & 0.9200 & 0.9450 & 0.6050 & 0.5600 & \textbf{0.9850} & 0.5500 & 0.9150 & 0.6300 \\
103 & ShapesAll & 0.7267 & 0.8925 & 0.8835 & 0.7777 & 0.7292 & \textbf{0.9133} & 0.8325 & 0.8833 & 0.8500 \\
104 & SmallKitchenAppliances & 0.7987 & 0.7227 & \textbf{0.8256} & 0.7095 & 0.5747 & 0.7347 & 0.7840 & 0.6387 & 0.7213 \\
105 & SmoothSubspace & 0.9467 & 0.9233 & 0.9367 & 0.9433 & \textbf{0.9800} & 0.9633 & 0.9767 & 0.9633 & 0.9133 \\
106 & SonyAIBORobotSurface1 & \textbf{0.9984} & 0.9920 & 0.9726 & 0.9775 & 0.9823 & 0.9952 & 0.9952 & 0.9888 & 0.9888 \\
107 & SonyAIBORobotSurface2 & \textbf{0.9990} & 0.9878 & 0.9376 & 0.9849 & 0.9847 & 0.9949 & 0.9939 & 0.9939 & 0.9949 \\
108 & StarLightCurves & 0.9803 & 0.9783 & 0.9793 & 0.9738 & 0.9734 & 0.9801 & 0.9616 & \textbf{0.9912} & 0.9818 \\
109 & Strawberry & 0.9756 & 0.9614 & 0.9426 & 0.9414 & 0.9735 & 0.9685 & \textbf{0.9766} & 0.9624 & 0.9736 \\
110 & SwedishLeaf & \textbf{0.9929} & 0.9404 & 0.9158 & 0.9423 & 0.9253 & 0.9502 & 0.9449 & 0.9778 & 0.9671 \\
111 & Symbols & \textbf{0.9961} & 0.9824 & 0.9798 & 0.9733 & 0.9765 & 0.9863 & 0.9618 & 0.9922 & 0.9667 \\
112 & SyntheticControl & 0.9700 & 0.9883 & 0.9683 & 0.9950 & 0.9767 & \textbf{0.9983} & 0.9767 & 0.9700 & 0.9717 \\
113 & ToeSegmentation1 & \textbf{0.9664} & 0.9551 & 0.9143 & 0.9181 & 0.6903 & 0.9588 & 0.8514 & 0.9219 & 0.8846 \\
114 & ToeSegmentation2 & 0.9282 & 0.9219 & 0.9283 & 0.8250 & 0.8435 & \textbf{0.9517} & 0.8316 & 0.8510 & 0.8143 \\
115 & Trace & \textbf{1.0000} & \textbf{1.0000} & 0.9950 & 0.9150 & 0.9700 & \textbf{1.0000} & 0.9400 & \textbf{1.0000} & 0.8900 \\
116 & TwoLeadECG & \textbf{1.0000} & 0.9983 & 0.9932 & 0.9965 & 0.9914 & 0.9991 & 0.9983 & 0.9991 & 0.9948 \\
117 & TwoPatterns & 0.9454 & \textbf{1.0000} & 0.9513 & 0.9990 & 0.9994 & \textbf{1.0000} & 0.9980 & 0.9990 & 0.9962 \\
118 & UMD & 0.8778 & 0.9944 & 0.9389 & 0.9611 & 0.9778 & 0.9944 & 0.9833 & \textbf{1.0000} & 0.9778 \\
119 & UWaveGestureLibraryAll & 0.9245 & 0.9518 & 0.8595 & 0.9750 & 0.9602 & 0.9652 & 0.9504 & \textbf{0.9913} & 0.9708 \\
120 & UWaveGestureLibraryX & 0.8665 & 0.8457 & 0.6680 & 0.8385 & 0.8091 & 0.8513 & 0.8924 & \textbf{0.9214} & 0.9002 \\
121 & UWaveGestureLibraryY & 0.7906 & 0.7887 & 0.5342 & 0.7573 & 0.7300 & 0.7874 & 0.8455 & \textbf{0.8955} & 0.8399 \\
122 & UWaveGestureLibraryZ & 0.8484 & 0.8021 & 0.6076 & 0.7762 & 0.4790 & 0.8024 & 0.8669 & \textbf{0.9006} & 0.8535 \\
123 & Wafer & \textbf{1.0000} & 0.9989 & 0.9587 & 0.9982 & 0.9992 & 0.9990 & 0.9986 & 0.9992 & 0.9992 \\
124 & Wine & 0.5411 & 0.9553 & 0.6743 & 0.5221 & 0.5854 & \textbf{0.9648} & 0.8289 & 0.5135 & 0.8561 \\
125 & WordSynonyms & 0.5967 & 0.8022 & 0.5718 & 0.7444 & 0.7083 & 0.7989 & 0.8442 & \textbf{0.8690} & 0.7823 \\
126 & Worms & 0.7906 & 0.6821 & 0.6860 & 0.5423 & 0.4148 & 0.7210 & 0.7446 & \textbf{0.8072} & 0.7097 \\
127 & WormsTwoClass & 0.8025 & 0.7634 & 0.7557 & 0.6898 & 0.6240 & 0.7716 & 0.7833 & \textbf{0.9037} & 0.8221 \\
128 & Yoga & 0.9718 & 0.9661 & 0.7699 & 0.9207 & 0.9497 & 0.9709 & 0.9539 & \textbf{0.9739} & 0.9476 \\ \hline
\multicolumn{2}{c}{\textbf{Avg. Acc}} & 0.8296 & 0.8325 & 0.8017 & 0.7807 & 0.7755 & \textbf{0.8691} & 0.8367 & 0.8265 & 0.8593 \\
\multicolumn{2}{c}{\textbf{Avg. Rank}} & 4.15 & 5.07 & 6.59 & 6.59 & 6.39 & \textbf{3.28} & 4.4 & 4.16 & 3.73 \\ \bottomrule
\end{tabular}}
\end{table*}

\begin{table*}[]
\centering
\caption{{}{Test classification accuracy using time series PTMs on 30 UEA datasets. The best results are in \textbf{bold}.}}
\label{tab:my-table3}
\begin{tabular}{@{}clcccccccc@{}}
\toprule
ID & Dataset & Supervised~(FCN) & T-Loss & TS-TCC & TST & TS2Vec & {}{TimesNet} & {}{PatchTST} & {}{GPT4TS} \\ \midrule
1 & ArticularyWordRecognition & \textbf{0.9983} & 0.8974 & 0.9757 & 0.9826 & 0.9913 & 0.9739 & 0.9513 & 0.9739 \\
2 & AtrialFibrillation & 0.3667 & 0.4000 & 0.3667 & 0.5667 & 0.1667 & 0.5667 & 0.5333 & \textbf{0.7667} \\
3 & BasicMotions & \textbf{1.0000} & 0.6375 & 0.9625 & 0.9000 & 0.9500 & 0.8500 & 0.8000 & 0.8750 \\
4 & CharacterTrajectories & 0.9738 & 0.9598 & 0.9882 & 0.9815 & \textbf{0.9941} & 0.9843 & 0.8982 & 0.9916 \\
5 & Cricket & 0.9500 & 0.6778 & 0.9278 & 0.9611 & \textbf{0.9889} & 0.9111 & 0.9778 & 0.9111 \\
6 & DuckDuckGeese & \textbf{0.7200} & 0.2500 & 0.2800 & 0.5100 & 0.4000 & 0.5500 & 0.5000 & 0.6200 \\
7 & EigenWorms & 0.8839 & 0.5056 & 0.7476 & 0.5791 & \textbf{0.8996} & 0.7265 & 0.8153 & 0.7574 \\
8 & Epilepsy & 0.9855 & 0.5600 & 0.9236 & 0.7018 & 0.9818 & 0.5564 & \textbf{0.9891} & 0.8109 \\
9 & ERing & \textbf{0.9867} & 0.9200 & 0.9667 & 0.9533 & 0.9600 & 0.9500 & 0.9700 & 0.9800 \\
10 & EthanolConcentration & 0.3321 & 0.2500 & 0.2633 & 0.2596 & 0.2157 & \textbf{0.5807} & 0.4568 & 0.5235 \\
11 & FaceDetection & 0.7987 & 0.5255 & 0.6973 & 0.6935 & 0.5228 & \textbf{0.9028} & 0.8506 & 0.8628 \\
12 & FingerMovements & 0.6638 & 0.5145 & 0.5047 & 0.5265 & 0.5239 & \textbf{0.7073} & 0.6783 & 0.7024 \\
13 & HandMovementDirection & 0.6550 & 0.2992 & 0.4144 & 0.4660 & 0.3079 & 0.5862 & 0.6245 & \textbf{0.7744} \\
14 & Handwriting & \textbf{0.9520} & 0.3480 & 0.7266 & 0.3080 & 0.8740 & 0.8280 & 0.6470 & 0.8580 \\
15 & Heartbeat & \textbf{0.8363} & 0.7213 & 0.7066 & 0.7164 & 0.7115 & 0.6995 & 0.7335 & 0.7167 \\
16 & InsectWingbeat & 0.1503 & 0.2676 & 0.6004 & 0.6253 & 0.2805 & 0.1771 & 0.4594 & \textbf{0.7628} \\
17 & JapaneseVowels & \textbf{0.9937} & 0.7406 & 0.9594 & 0.9609 & 0.9656 & 0.9609 & 0.9781 & 0.9844 \\
18 & Libras & \textbf{0.9250} & 0.7306 & 0.7444 & 0.7972 & 0.8500 & 0.7833 & 0.8389 & 0.7917 \\
19 & LSST & 0.6016 & 0.3155 & 0.5682 & 0.5127 & 0.6268 & 0.5978 & 0.4357 & \textbf{0.8258} \\
20 & MotorImagery & 0.7572 & 0.5025 & 0.5318 & 0.5157 & 0.5267 & 0.7520 & 0.7862 & \textbf{0.8076} \\
21 & NATOPS & \textbf{0.9278} & 0.5944 & 0.7806 & 0.8722 & 0.8417 & 0.8139 & 0.8250 & 0.8444 \\
22 & PEMS-SF & 0.9318 & 0.6864 & 0.8159 & 0.8068 & 0.8682 & 0.9341 & 0.8659 & \textbf{0.9773} \\
23 & PenDigits & 0.9979 & 0.9933 & 0.9932 & 0.9895 & 0.9953 & \textbf{0.9986} & 0.9974 & 0.9976 \\
24 & PhonemeSpectra & 0.1570 & 0.0964 & 0.2356 & 0.0562 & 0.2478 & 0.4309 & 0.1101 & \textbf{0.7010} \\
25 & RacketSports & \textbf{0.9408} & 0.6169 & 0.8317 & 0.8181 & 0.8709 & 0.9277 & 0.8683 & 0.8683 \\
26 & SelfRegulationSCP1 & 0.8486 & 0.8128 & 0.8288 & 0.7968 & 0.7736 & 0.8040 & \textbf{0.9163} & 0.9093 \\
27 & SelfRegulationSCP2 & 0.6395 & 0.5000 & 0.5105 & 0.4868 & 0.5132 & \textbf{0.7684} & 0.6737 & 0.6605 \\
28 & SpokenArabicDigits & 0.8131 & 0.9519 & 0.9960 & 0.9849 & \textbf{0.9975} & 0.9953 & 0.8454 & 0.9932 \\
29 & StandWalkJump & 0.4067 & 0.4533 & 0.4000 & 0.5133 & 0.5067 & 0.5333 & 0.5200 & \textbf{0.8600} \\
30 & UWaveGestureLibrary & 0.9614 & 0.8591 & 0.9273 & 0.9205 & 0.9500 & 0.8591 & \textbf{0.9659} & 0.9568 \\ \hline
\multicolumn{2}{c}{\textbf{Avg. Acc}} & 0.7718 & 0.5863 & 0.7059 & 0.6921 & 0.7101 & 0.7570 & 0.7504 & \textbf{0.8355} \\
\multicolumn{2}{c}{\textbf{Avg. Rank}} & 3.07 & 7.07 & 5.27 & 5.37 & 4.27 & 4.13 & 3.9 & \textbf{2.7} \\ \bottomrule
\end{tabular}
\end{table*}

\subsection{{}{Details of Results on Time-Series Forecasting~\label{appendix_C3}}}

{}{The detailed test results of prediction lengths on time series forecasting tasks are presented in Table~\ref{tab:appendix-forecasting}. For the ETTh1, ETTh2, ETTm1, and Electricity datasets, we selected the prediction lengths and train-val-test set configurations based on the experimental settings of TSVec~\cite{yue2022ts2vec} and CoST~\cite{woo2022cost}. For the ETTm2, Traffic, Weather, Exchange, and National Illness (ILI) datasets, we followed the experimental settings of TimesNet~\cite{wutimesnet} to determine the appropriate prediction lengths and train-val-test set configurations for analysis.
In addition,  for the LogTrans, TCN, Informer, Autformer, PatchTST, DLinear, GPT4TS, TEMPO, and iTransformer models, we adhere to the unified settings for the maximum number of epochs as used in TimesNet for our experimental analysis. For other parameters, we follow the default settings of the respective models to ensure consistency in our experimental analysis.
}

% \usepackage{booktabs}
% \usepackage{multirow}
% \usepackage[normalem]{ulem}
\begin{table*}[]
\caption{{}{Comparison of test results for time-series forecasting. "-" indicates that the results could not be obtained due to memory errors or excessive training time. The best results are in \underline{underline} and \textbf{bold}.}}
\label{tab:appendix-forecasting}
\resizebox{\linewidth}{!}{ 
\begin{tabular}{@{}c|c|cccccccccccccccccccccccc@{}}
\toprule
\multicolumn{2}{c}{Models} & \multicolumn{2}{c}{LogTrans} & \multicolumn{2}{c}{TCN} & \multicolumn{2}{c}{Informer} & \multicolumn{2}{c}{Autoformer} & \multicolumn{2}{c}{TS2Vec} & \multicolumn{2}{c}{CoST} & \multicolumn{2}{c}{TimesNet} & \multicolumn{2}{c}{PatchTST} & \multicolumn{2}{c}{DLinear} & \multicolumn{2}{c}{GPT4TS} & \multicolumn{2}{c}{TEMPO} & \multicolumn{2}{c}{iTransformer} \\ \midrule
\multicolumn{2}{c}{Metric} & MSE & MAE & MSE & MAE & MSE & MAE & MSE & MAE & MSE & MAE & MSE & MAE & MSE & MAE & MSE & MAE & MSE & MAE & MSE & MAE & MSE & MAE & MSE & MAE \\ \hline
\multirow{6}{*}{ETTh1} & 24 & 0.6149 & 0.5823 & 0.6476 & 0.5765 & 0.5972 & 0.5680 & 0.3729 & 0.4176 & 0.5952 & 0.5313 & 0.3861 & 0.4289 & 0.3485 & 0.3873 & 0.3890 & 0.4128 & 0.3885 & 0.4105 & 0.3102 & 0.3626 & 0.4262 & 0.4264 & {\ul \textbf{0.3065}} & {\ul \textbf{0.3589}} \\
 & 48 & 0.8360 & 0.7075 & 0.7417 & 0.6292 & 0.7344 & 0.6460 & 0.4296 & 0.4419 & 0.6316 & 0.5566 & 0.4358 & 0.4634 & 0.3991 & 0.4169 & 0.4362 & 0.4379 & 0.4111 & 0.4232 & 0.3529 & 0.3904 & 0.4677 & 0.4518 & {\ul \textbf{0.3451}} & {\ul \textbf{0.3818}} \\
 & 168 & 1.0119 & 0.8084 & 0.9388 & 0.7324 & 0.9414 & 0.7480 & 0.4796 & 0.4687 & 0.7669 & 0.6405 & 0.6415 & 0.5817 & 0.4846 & 0.4670 & 0.5304 & 0.4879 & 0.4762 & 0.4613 & 0.4600 & 0.4661 & 0.5128 & 0.4857 & {\ul \textbf{0.4307}} & {\ul \textbf{0.4304}} \\
 & 336 & 1.0821 & 0.8458 & 1.1151 & 0.8305 & 1.1217 & 0.8375 & 0.4939 & 0.4815 & 0.9419 & 0.7334 & 0.8099 & 0.6776 & 0.5583 & 0.5175 & 0.5928 & 0.5193 & 0.5243 & 0.4903 & 0.5167 & 0.4960 & 0.6184 & 0.5496 & {\ul \textbf{0.4889}} & {\ul \textbf{0.4602}} \\
 & 720 & 1.0381 & 0.8167 & 0.9641 & 0.7802 & 1.1921 & 0.8591 & {\ul \textbf{0.5112}} & 0.5100 & 1.0948 & 0.8098 & 0.9678 & 0.7700 & 0.5886 & 0.5287 & 0.6123 & 0.5442 & 0.5483 & 0.5310 & 0.6878 & 0.5858 & 0.6454 & 0.5724 & 0.5184 & {\ul \textbf{0.4982}} \\
 & Avg. & 0.9166 & 0.7521 & 0.8815 & 0.7098 & 0.9174 & 0.7317 & 0.4574 & 0.4639 & 0.8061 & 0.6543 & 0.6482 & 0.5843 & 0.4758 & 0.4635 & 0.5121 & 0.4804 & 0.4697 & 0.4633 & 0.4655 & 0.4602 & 0.5341 & 0.4972 & {\ul \textbf{0.4179}} & {\ul \textbf{0.4259}} \\ \hline
\multirow{6}{*}{ETTh2} & 24 & 0.7676 & 0.6854 & 0.9376 & 0.7760 & 1.2772 & 0.9142 & 0.2933 & 0.3689 & 0.4478 & 0.5032 & 0.4463 & 0.5032 & 0.2129 & 0.2929 & 0.2176 & 0.3043 & 0.2332 & 0.3289 & 0.2009 & 0.2917 & 0.2507 & 0.3258 & {\ul \textbf{0.1831}} & {\ul \textbf{0.2719}} \\
 & 48 & 1.2485 & 0.9004 & 1.2930 & 0.9183 & 1.4506 & 0.9663 & 0.3349 & 0.3922 & 0.6460 & 0.6184 & 0.7022 & 0.6397 & 0.2824 & 0.3429 & 0.2722 & 0.3388 & 0.2902 & 0.3677 & 0.2738 & 0.3443 & 0.3078 & 0.3659 & {\ul \textbf{0.2413}} & {\ul \textbf{0.3120}} \\
 & 168 & 5.4984 & 1.8806 & 4.0537 & 1.6939 & 5.6799 & 1.9599 & 0.4627 & 0.4638 & 1.7771 & 1.0569 & 1.5428 & 0.9822 & 0.4461 & 0.4361 & 0.4092 & 0.4172 & 0.4807 & 0.4829 & 0.4436 & 0.4476 & 0.4891 & 0.4739 & {\ul \textbf{0.3725}} & {\ul \textbf{0.3940}} \\
 & 336 & 4.4323 & 1.6926 & 5.3142 & 2.0093 & 4.6730 & 1.8000 & 0.4753 & 0.4794 & 2.1157 & 1.1759 & 1.7560 & 1.0478 & 0.4875 & 0.4687 & 0.4670 & 0.4610 & 0.6314 & 0.5638 & 0.5011 & 0.4876 & 0.5549 & 0.5214 & {\ul \textbf{0.4368}} & {\ul \textbf{0.4433}} \\
 & 720 & 3.1966 & 1.4905 & 9.1991 & 2.2230 & 3.6218 & 1.5993 & 0.4970 & 0.5015 & 2.5823 & 1.3521 & 1.9716 & 1.0938 & 0.5193 & 0.4893 & 0.4721 & 0.4734 & 0.8289 & 0.6648 & 0.5389 & 0.5146 & 0.6687 & 0.5851 & {\ul \textbf{0.4479}} & {\ul \textbf{0.4585}} \\
 & Avg. & 3.0287 & 1.3299 & 4.1595 & 1.5241 & 3.3405 & 1.4479 & 0.4126 & 0.4412 & 1.5138 & 0.9413 & 1.2838 & 0.8533 & 0.3896 & 0.4060 & 0.3676 & 0.3989 & 0.4929 & 0.4816 & 0.3917 & 0.4172 & 0.4542 & 0.4544 & {\ul \textbf{0.3363}} & {\ul \textbf{0.3759}} \\ \hline
\multirow{6}{*}{ETTm1} & 24 & 0.1576 & 0.2780 & 0.2003 & 0.3345 & 0.2123 & 0.3418 & 0.1473 & 0.2610 & 0.1970 & 0.3179 & {\ul \textbf{0.1331}} & {\ul \textbf{0.2589}} & 0.2416 & 0.3115 & 0.2522 & 0.3186 & 0.2777 & 0.3328 & 0.2004 & 0.2769 & 0.2324 & 0.3011 & 0.2251 & 0.2967 \\
 & 48 & 0.2533 & 0.3658 & 0.2432 & 0.3641 & 0.3220 & 0.4262 & {\ul \textbf{0.1897}} & {\ul \textbf{0.2898}} & 0.2682 & 0.3784 & 0.1959 & 0.3192 & 0.3194 & 0.3637 & 0.3202 & 0.3593 & 0.3302 & 0.3643 & 0.3006 & 0.3344 & 0.3026 & 0.3472 & 0.3019 & 0.3474 \\
 & 96 & 0.4699 & 0.5022 & 0.3856 & 0.4599 & 0.4427 & 0.5227 & 0.3424 & 0.3546 & 0.3735 & 0.4496 & {\ul \textbf{0.3043}} & {\ul \textbf{0.4034}} & 0.3558 & 0.3864 & 0.3553 & 0.3793 & 0.3543 & 0.3785 & 0.3008 & 0.3519 & 0.3683 & 0.3889 & 0.3393 & 0.3740 \\
 & 288 & 1.2230 & 0.8335 & 0.8695 & 0.7408 & 1.1282 & 0.8074 & {\ul \textbf{0.3264}} & {\ul \textbf{0.3663}} & 0.7566 & 0.6672 & 0.7554 & 0.6609 & 0.4411 & 0.4304 & 0.4207 & 0.4162 & 0.4117 & 0.4131 & 0.3712 & 0.3989 & 0.4491 & 0.4441 & 0.4147 & 0.4163 \\
 & 672 & 3.0955 & 1.3050 & 1.9628 & 1.1394 & 3.0900 & 1.3206 & {\ul \textbf{0.4497}} & {\ul \textbf{0.4357}} & 1.8217 & 1.0452 & 1.5897 & 0.9840 & 0.6567 & 0.5324 & 0.4878 & 0.4527 & 0.4725 & 0.4496 & 0.4570 & 0.4483 & 0.4964 & 0.4723 & 0.4880 & 0.4577 \\
 & Avg. & 1.0399 & 0.6569 & 0.7323 & 0.6077 & 1.0390 & 0.6837 & {\ul \textbf{0.2911}} & {\ul \textbf{0.3415}} & 0.6834 & 0.5717 & 0.5957 & 0.5253 & 0.4029 & 0.4049 & 0.3672 & 0.3852 & 0.3693 & 0.3876 & 0.3260 & 0.3621 & 0.3698 & 0.3907 & 0.3538 & 0.3784 \\ \hline
\multirow{5}{*}{ETTm2} & 96 & 0.4411 & 0.5149 & 0.3053 & 0.4097 & 0.3951 & 0.4839 & 0.2220 & 0.3040 & 0.3502 & 0.4381 & 0.2958 & 0.3955 & 0.1877 & 0.2678 & 0.1876 & 0.2728 & 0.2015 & 0.3025 & 0.1861 & 0.2781 & 0.2031 & 0.2830 & {\ul \textbf{0.1830}} & {\ul \textbf{0.2652}} \\
 & 192 & 0.7027 & 0.6550 & 0.5954 & 0.5979 & 0.7435 & 0.6675 & 0.2763 & 0.3342 & 0.5684 & 0.5732 & 0.5266 & 0.5391 & 0.2748 & 0.3220 & 0.2544 & 0.3140 & 0.2878 & 0.3650 & 0.2624 & 0.3271 & 0.2653 & 0.3206 & {\ul \textbf{0.2507}} & {\ul \textbf{0.3101}} \\
 & 336 & 1.0915 & 0.8209 & 1.1618 & 0.8573 & 1.2908 & 0.8660 & 0.3308 & 0.3676 & 0.9589 & 0.7532 & 0.8752 & 0.7091 & 0.3922 & 0.3830 & 0.3173 & 0.3527 & 0.3833 & 0.4285 & {\ul \textbf{0.3164}} & 0.3650 & 0.3214 & 0.3646 & 0.3179 & {\ul \textbf{0.3524}} \\
 & 720 & 2.7116 & 1.4048 & 1.8699 & 1.0495 & 3.3230 & 1.3981 & 0.4256 & 0.4203 & 2.5705 & 1.2483 & 1.9638 & 1.0948 & 0.4477 & 0.4244 & 0.4179 & {\ul \textbf{0.4075}} & 0.5472 & 0.5202 & 0.4246 & 0.4331 & 0.4232 & 0.4286 & {\ul \textbf{0.4161}} & 0.4090 \\
 & Avg. & 1.2367 & 0.8489 & 0.9831 & 0.7286 & 1.4381 & 0.8539 & 0.3137 & 0.3565 & 1.1120 & 0.7532 & 0.9154 & 0.6846 & 0.3256 & 0.3493 & 0.2943 & 0.3367 & 0.3550 & 0.4041 & 0.2974 & 0.3508 & 0.3032 & 0.3492 & {\ul \textbf{0.2919}} & {\ul \textbf{0.3342}} \\ \hline
\multirow{6}{*}{Electricity} & 24 & 0.3667 & 0.4054 & 0.4616 & 0.4673 & 0.5750 & 0.5159 & 0.1888 & 0.3031 & 0.3038 & 0.3836 & 0.1438 & 0.2471 & 0.1612 & 0.2630 & 0.1681 & 0.2502 & 0.2443 & 0.3408 & {\ul \textbf{0.1019}} & {\ul \textbf{0.1970}} & - & - & 0.1434 & 0.2336 \\
 & 48 & 0.3842 & 0.4182 & 0.4581 & 0.4694 & 0.5808 & 0.5281 & 0.2043 & 0.3148 & 0.3311 & 0.4014 & 0.1602 & 0.2611 & 0.1692 & 0.2704 & 0.2017 & 0.2765 & 0.2687 & 0.3565 & {\ul \textbf{0.1166}} & {\ul \textbf{0.2108}} & - & - & 0.1774 & 0.2603 \\
 & 168 & 0.3911 & 0.4206 & 0.5079 & 0.4938 & 0.5486 & 0.5220 & 0.2446 & 0.3401 & 0.3581 & 0.4195 & 0.1776 & 0.2739 & 0.1936 & 0.2907 & 0.2019 & 0.2839 & 0.2628 & 0.3576 & {\ul \textbf{0.1459}} & {\ul \textbf{0.2376}} & - & - & 0.1930 & 0.2754 \\
 & 336 & 0.4065 & 0.4300 & 0.5209 & 0.5024 & 0.6021 & 0.5571 & 0.2551 & 0.3464 & 0.3865 & 0.4367 & 0.2134 & 0.3018 & 0.2081 & 0.3034 & 0.2260 & 0.3063 & 0.2823 & 0.3746 & {\ul \textbf{0.1721}} & {\ul \textbf{0.2649}} & - & - & 0.2203 & 0.3014 \\
 & 720 & 0.4118 & 0.4308 & 0.5505 & 0.5116 & 0.7690 & 0.6517 & 0.3075 & 0.3808 & 0.4313 & 0.4618 & 0.2629 & 0.3384 & 0.2257 & 0.3183 & 0.2704 & 0.3397 & 0.3170 & 0.4009 & {\ul \textbf{0.2118}} & {\ul \textbf{0.2942}} & - & - & 0.2682 & 0.3385 \\
 & Avg. & 0.3921 & 0.4210 & 0.4998 & 0.4889 & 0.6151 & 0.5550 & 0.2401 & 0.3370 & 0.3622 & 0.4206 & 0.1916 & 0.2845 & 0.1915 & 0.2891 & 0.2136 & 0.2913 & 0.2750 & 0.3661 & {\ul \textbf{0.1497}} & {\ul \textbf{0.2409}} & - & - & 0.2005 & 0.2818 \\ \hline
\multirow{5}{*}{Traffic} & 96 & 0.9029 & 0.5171 & 0.8102 & 0.4755 & 0.6845 & 0.3824 & 0.6348 & 0.3876 & - & - & - & - & 0.6087 & 0.3164 & 0.6389 & 0.3960 & 0.8068 & 0.4901 & {\ul \textbf{0.3725}} & {\ul \textbf{0.2571}} & 0.4036 & 0.2966 & 0.4185 & 0.2845 \\
 & 192 & 0.9389 & 0.5377 & 0.7557 & 0.4582 & 0.6899 & 0.3879 & 0.6329 & 0.3935 & - & - & - & - & 0.6264 & 0.3314 & 0.6053 & 0.3817 & 0.7679 & 0.4747 & {\ul \textbf{0.3848}} & {\ul \textbf{0.2612}} & 0.4125 & 0.3027 & 0.4394 & 0.2921 \\
 & 336 & 0.9559 & 0.5441 & 0.6804 & 0.4326 & 0.7360 & 0.4147 & 0.6349 & 0.3930 & - & - & - & - & 0.6548 & 0.3462 & 0.6173 & 0.3874 & 0.7768 & 0.4773 & {\ul \textbf{0.3924}} & {\ul \textbf{0.2647}} & 0.4231 & 0.3056 & 0.4578 & 0.3016 \\
 & 720 & 1.0025 & 0.5596 & 0.5696 & 0.3777 & 0.8059 & 0.4507 & 0.6566 & 0.4028 & - & - & - & - & 0.6770 & 0.3506 & 0.6617 & 0.4097 & 0.8157 & 0.4917 & {\ul \textbf{0.4316}} & {\ul \textbf{0.2879}} & 0.4383 & 0.3091 & 0.4917 & 0.3226 \\
 & Avg. & 0.9500 & 0.5396 & 0.7040 & 0.4360 & 0.7291 & 0.4089 & 0.6398 & 0.3942 & - & - & - & - & 0.6417 & 0.3362 & 0.6308 & 0.3937 & 0.7918 & 0.4835 & {\ul \textbf{0.3953}} & {\ul \textbf{0.2677}} & 0.4194 & 0.3035 & 0.4518 & 0.3002 \\ \hline
\multirow{5}{*}{Weather} & 96 & 0.2476 & 0.3360 & 0.1899 & 0.2645 & 0.5056 & 0.5075 & 0.2936 & 0.3491 & 0.1858 & 0.2668 & 0.1590 & 0.2389 & 0.1817 & 0.2291 & 0.2013 & 0.2406 & 0.1980 & 0.2602 & {\ul \textbf{0.1534}} & {\ul \textbf{0.2079}} & 0.1575 & 0.2139 & 0.1792 & 0.2191 \\
 & 192 & 0.2733 & 0.3527 & 0.2171 & 0.2872 & 0.4882 & 0.4902 & 0.3222 & 0.3707 & 0.2252 & 0.3035 & 0.2038 & 0.2804 & 0.2423 & 0.2793 & 0.2466 & 0.2770 & 0.2378 & 0.2973 & {\ul \textbf{0.1996}} & {\ul \textbf{0.2532}} & 0.2029 & 0.2542 & 0.2278 & 0.2600 \\
 & 336 & 0.3160 & 0.3805 & 0.2738 & 0.3339 & 0.6144 & 0.5359 & 0.3607 & 0.3927 & 0.2805 & 0.3466 & {\ul \textbf{0.2557}} & 0.3219 & 0.2895 & 0.3099 & 0.2984 & 0.3129 & 0.2826 & 0.3331 & 0.2613 & 0.2980 & 0.2715 & {\ul \textbf{0.2976}} & 0.2835 & 0.3002 \\
 & 720 & 0.3926 & 0.4230 & 0.3365 & 0.3819 & 1.5827 & 0.9240 & 0.4268 & 0.4286 & 0.3565 & 0.3984 & 0.3203 & 0.3691 & 0.3975 & 0.3784 & 0.3704 & 0.3586 & 0.3463 & 0.3831 & {\ul \textbf{0.3269}} & {\ul \textbf{0.3428}} & 0.3493 & 0.3549 & 0.3596 & 0.3505 \\
 & Avg. & 0.3074 & 0.3730 & 0.2543 & 0.3169 & 0.7977 & 0.6144 & 0.3508 & 0.3853 & 0.2620 & 0.3288 & {\ul \textbf{0.2347}} & 0.3026 & 0.2777 & 0.2992 & 0.2791 & 0.2973 & 0.2662 & 0.3184 & 0.2353 & {\ul \textbf{0.2755}} & 0.2453 & 0.2802 & 0.2625 & 0.2825 \\ \hline
\multirow{5}{*}{Exchange} & 96 & 1.3842 & 0.9759 & 0.6826 & 0.5662 & 0.8626 & 0.7415 & 0.1551 & 0.2850 & 0.2186 & 0.3501 & 0.3092 & 0.4261 & 0.1289 & 0.2537 & 0.1049 & 0.2301 & 0.1168 & 0.2594 & 0.1067 & 0.2346 & - & - & {\ul \textbf{0.0899}} & {\ul \textbf{0.2113}} \\
 & 192 & 1.4420 & 1.0010 & 0.5730 & 0.5741 & 1.3249 & 0.9216 & 0.2944 & 0.3954 & 0.4687 & 0.5080 & 0.4843 & 0.5387 & 0.2479 & 0.3510 & 0.1983 & 0.3205 & 0.2049 & 0.3458 & 0.2008 & 0.3242 & - & - & {\ul \textbf{0.1848}} & {\ul \textbf{0.3088}} \\
 & 336 & 1.5682 & 1.0433 & 0.7006 & 0.6566 & 1.2885 & 0.9438 & 0.4476 & 0.4967 & 0.7984 & 0.6789 & 0.8584 & 0.7139 & 0.4307 & 0.4832 & 0.3517 & 0.4316 & {\ul \textbf{0.3261}} & 0.4442 & 0.4064 & 0.4729 & - & - & 0.3388 & {\ul \textbf{0.4222}} \\
 & 720 & 1.9746 & 1.1628 & 2.0907 & 1.1803 & 1.5762 & 1.0094 & 1.1311 & 0.8274 & 1.2703 & 0.8875 & 1.3870 & 0.9669 & 1.3950 & 0.8633 & 0.8876 & 0.7148 & {\ul \textbf{0.5293}} & {\ul \textbf{0.5717}} & 1.0583 & 0.7784 & - & - & 0.8761 & 0.7120 \\
 & Avg. & 1.5922 & 1.0458 & 1.0117 & 0.7443 & 1.2631 & 0.9041 & 0.5071 & 0.5011 & 0.6890 & 0.6061 & 0.7597 & 0.6614 & 0.5506 & 0.4878 & 0.3856 & 0.4242 & {\ul \textbf{0.2943}} & {\ul \textbf{0.4053}} & 0.4430 & 0.4525 & - & - & 0.3724 & 0.4136 \\ \hline
\multirow{5}{*}{ILI} & 24 & 7.0450 & 1.8932 & 7.0507 & 1.8967 & 5.2708 & 1.5313 & 3.4776 & 1.3505 & 3.2963 & 1.1271 & {\ul \textbf{1.7753}} & {\ul \textbf{0.8399}} & 2.4637 & 1.0337 & 3.6716 & 1.3056 & 3.9658 & 1.4726 & 2.9939 & 1.2228 & 4.1965 & 1.5388 & 2.3113 & 1.0427 \\
 & 36 & 7.0539 & 1.8949 & 7.1350 & 1.9084 & 5.1110 & 1.5148 & 3.3136 & 1.3154 & 3.2484 & 1.1215 & {\ul \textbf{2.0151}} & {\ul \textbf{0.8936}} & 2.2281 & 0.9843 & 2.7695 & 1.1426 & 3.8974 & 1.4451 & 3.4103 & 1.3562 & 4.2705 & 1.5554 & 2.1656 & 1.0194 \\
 & 48 & 7.2207 & 1.9205 & 7.2418 & 1.9256 & 5.2915 & 1.5469 & 3.3103 & 1.2902 & 3.2821 & 1.1190 & 2.2033 & {\ul \textbf{0.9327}} & 2.1243 & {\ul \textbf{0.9546}} & 2.2878 & 0.9802 & 3.9395 & 1.4410 & 2.7575 & 1.2639 & 4.4945 & 1.5472 & {\ul \textbf{2.0375}} & 0.9793 \\
 & 60 & 7.4870 & 1.9667 & 7.4951 & 1.9698 & 5.6127 & 1.6050 & 3.3582 & 1.2956 & 3.1346 & 1.0879 & 2.3573 & {\ul \textbf{0.9627}} & 2.1243 & 0.9672 & 2.1775 & 0.9778 & 4.1150 & 1.4639 & 2.8209 & 1.2752 & 4.5043 & 1.5694 & {\ul \textbf{2.0194}} & 0.9991 \\
 & Avg. & 7.2016 & 1.9188 & 7.2306 & 1.9251 & 5.3215 & 1.5495 & 3.3649 & 1.3129 & 3.2403 & 1.1139 & {\ul \textbf{2.0877}} & {\ul \textbf{0.9073}} & 2.2351 & 0.9849 & 2.7266 & 1.1015 & 3.9794 & 1.4557 & 2.9957 & 1.2795 & 4.3664 & 1.5527 & 2.1335 & 1.0102 \\ \cmidrule(l){1-26} 
\end{tabular}}
\end{table*}

\subsection{{}{Details of Results on Time-Series Anomaly Detection~\label{appendix_C4}}}

{}{We select SPOT, DSPOT, LSTM-VAE, DOUNT, AI, TS2Vec, TimesNet, GPT4TS, and DCdetector for time series anomaly detection analysis. Due to memory or training errors, TimesNet, GPT4TS, and DCdetector could not obtain results on the Yahoo and KPI datasets; therefore, we provide the test results for the remaining six methods on these datasets.
The detailed test results for time series anomaly detection on the UCR anomaly detection archive are presented in the following tables: Table~\ref{tab:appendix_ad_f1} (F1-score), Table~\ref{tab:appendix_ad_precision} (Precision), Table~\ref{tab:appendix_ad_recall} (Recall), Table~\ref{tab:appendix_ad_affp} (Aff-P), Table~\ref{tab:appendix_ad_affr} (Aff-R), 
Table~\ref{tab:appendix_ad_rar} (R\_A\_R), 
Table~\ref{tab:appendix_ad_rap} (R\_A\_P), 
Table~\ref{tab:appendix_ad_vroc} (V\_ROC), 
Table~\ref{tab:appendix_ad_vpr} (V\_PR), 
Table~\ref{tab:appendix_ad_f1pa10} (F1-PA-10), 
Table~\ref{tab:appendix_ad_f1pa50} (F1-PA-50), and
Table~\ref{tab:appendix_ad_f1pa90} (F1-PA-90). The \textit{UCR ID} column in these tables refers to the dataset IDs in the UCR anomaly detection archive. In addition, for the four datasets with IDs [79, 108, 187, 203] in the UCR anomaly detection archive, we uniformly set their experimental results to zero because their true labels contain no anomalous data points. This absence leads to errors in calculating nine assessment metrics, aside from F1, P, and R metrics, by the relevant methods. To reasonably analyze the performance of different methods on the UCR anomaly detection archive, we use the number of the most recent performance datasets (Win) for statistical analysis to avoid errors introduced by the large discrepancies in assessment indicator values across different datasets. Also, detailed test results for time series anomaly detection on seven multivariate datasets are presented in Table~\ref{tab:appendix_ad_multi}.}

% \usepackage{booktabs}
\begin{table*}[]
\centering
\caption{{}{The test F1-score of time series anomaly detection in UCR 250 sub-datasets. The best results are in \textbf{bold}.}}
\label{tab:appendix_ad_f1}
\scalebox{0.6}{
\begin{tabular}{@{}c|ccccccccc|c|ccccccccc@{}}
\toprule
UCR ID & SPOT & DSPOT & LSTM-VAE & DONUT & AT & TS2Vec & TimesNet & GPT4TS & DCdetector & UCR ID & SPOT & DSPOT & LSTM-VAE & DONUT & AT & TS2Vec & TimesNet & GPT4TS & DCdetector \\ \midrule
1 & 0.0000 & 0.0000 & 0.7456 & 0.7529 & 0.6978 & 0.8569 & 0.7588 & 0.7652 & \textbf{0.8593} & 126 & 0.0000 & 0.0000 & 0.0000 & 0.4717 & 0.4776 & 0.6024 & 0.0000 & 0.4989 & \textbf{0.6849} \\
2 & 0.0000 & 0.0000 & 0.5725 & 0.5769 & 0.4893 & 0.0000 & 0.0001 & 0.0287 & \textbf{0.7472} & 127 & 0.0000 & 0.0000 & \textbf{0.5535} & 0.0000 & 0.0000 & 0.0000 & 0.0000 & 0.0000 & 0.0000 \\
3 & 0.0000 & 0.0000 & 0.5720 & 0.5720 & 0.4979 & 0.7444 & 0.6083 & 0.5613 & \textbf{0.7500} & 128 & 0.0000 & 0.0000 & 0.8537 & 0.8623 & 0.7007 & 0.0000 & 0.6403 & 0.7744 & \textbf{0.9301} \\
4 & 0.0000 & 0.0000 & 0.0000 & 0.8247 & 0.6839 & \textbf{0.9732} & 0.7758 & 0.7050 & 0.9091 & 129 & 0.0000 & 0.0000 & 0.0000 & 0.6164 & 0.3481 & 0.0000 & 0.1485 & 0.3658 & \textbf{0.7500} \\
5 & 0.0000 & 0.0000 & 0.0000 & 0.0000 & 0.0000 & 0.0000 & 0.0000 & \textbf{0.0009} & 0.0000 & 130 & 0.0000 & 0.0000 & 0.0000 & 0.7638 & 0.4980 & 0.0000 & 0.3415 & 0.6202 & \textbf{0.8645} \\
6 & 0.0000 & 0.0000 & \textbf{0.5106} & 0.4898 & 0.0000 & 0.0000 & 0.0000 & 0.0320 & 0.0000 & 131 & 0.0000 & 0.0000 & 0.0000 & 0.7376 & 0.5053 & 0.0000 & 0.0961 & 0.5880 & \textbf{0.8851} \\
7 & 0.0000 & 0.0000 & 0.0000 & \textbf{0.4486} & 0.0950 & 0.0000 & 0.0883 & 0.0279 & 0.0000 & 132 & 0.0000 & 0.0000 & 0.0000 & 0.0000 & 0.2782 & 0.0000 & 0.0000 & 0.0508 & \textbf{0.8333} \\
\bottomrule
\end{tabular}}
\end{table*}

% \usepackage{booktabs}
\begin{table*}[]
\caption{{}{The test precision of time series anomaly detection in UCR 250 sub-datasets. The best results are in \textbf{bold}.}}
\label{tab:appendix_ad_precision}
\scalebox{0.6}{
\begin{tabular}{@{}c|ccccccccc|c|ccccccccc@{}}
\toprule
UCR ID & SPOT & DSPOT & LSTM-VAE & DONUT & AT & TS2Vec & TimesNet & GPT4TS & DCdetector & UCR ID & SPOT & DSPOT & LSTM-VAE & DONUT & AT & TS2Vec & TimesNet & GPT4TS & DCdetector \\ \midrule
1 & 0.0000 & 0.0000 & 0.5944 & 0.6037 & 0.5813 & 0.7497 & 0.6896 & 0.6949 & \textbf{0.7533} & 126 & 0.0000 & 0.0000 & 0.0000 & 0.3086 & 0.3748 & 0.4310 & 0.0000 & 0.3915 & \textbf{0.5208} \\
2 & 0.0000 & 0.0000 & 0.4011 & 0.4054 & 0.3613 & 0.0000 & 0.0001 & 0.0322 & \textbf{0.5964} & 127 & 0.0000 & 0.0000 & \textbf{0.3826} & 0.0000 & 0.0000 & 0.0000 & 0.0000 & 0.0000 & 0.0000 \\
3 & 0.0000 & 0.0000 & 0.4005 & 0.4005 & 0.3680 & 0.5929 & 0.4990 & 0.4719 & \textbf{0.6000} & 128 & 0.0000 & 0.0000 & 0.7448 & 0.7580 & 0.8004 & 0.0000 & 0.7743 & 0.8231 & \textbf{0.8694} \\
4 & 0.0000 & 0.0000 & 0.0000 & 0.7018 & 0.7543 & \textbf{0.9479} & 0.7952 & 0.7564 & 0.8333 & 129 & 0.0000 & 0.0000 & 0.0000 & 0.4455 & 0.3283 & 0.0000 & 0.1559 & 0.3420 & \textbf{0.6000} \\
5 & 0.0000 & 0.0000 & 0.0000 & 0.0000 & 0.0000 & 0.0000 & 0.0000 & \textbf{0.0005} & 0.0000 & 130 & 0.0000 & 0.0000 & 0.0000 & 0.6178 & 0.5789 & 0.0000 & 0.4531 & 0.6501 & \textbf{0.7613} \\
6 & 0.0000 & 0.0000 & \textbf{0.3429} & 0.3243 & 0.0000 & 0.0000 & 0.0000 & 0.0342 & 0.0000 & 131 & 0.0000 & 0.0000 & 0.0000 & 0.5843 & 0.5787 & 0.0000 & 0.1689 & 0.6287 & \textbf{0.7939} \\
7 & 0.0000 & 0.0000 & 0.0000 & \textbf{0.2892} & 0.0995 & 0.0000 & 0.1026 & 0.0298 & 0.0000 & 132 & 0.0000 & 0.0000 & 0.0000 & 0.0000 & 0.2794 & 0.0000 & 0.0000 & 0.0593 & \textbf{0.7143} \\
\bottomrule
\end{tabular}}
\end{table*}

% \usepackage{booktabs}
\begin{table*}[]
\caption{{}{The test recall of time series anomaly detection in UCR 250 sub-datasets. The best results are in \textbf{bold}.}}
\label{tab:appendix_ad_recall}
\scalebox{0.6}{
\begin{tabular}{@{}c|ccccccccc|c|ccccccccc@{}}
\toprule
UCR ID & SPOT & DSPOT & LSTM-VAE & DONUT & AT & TS2Vec & TimesNet & GPT4TS & DCdetector & UCR ID & SPOT & DSPOT & LSTM-VAE & DONUT & AT & TS2Vec & TimesNet & GPT4TS & DCdetector \\ \midrule
1 & 0.0000 & 0.0000 & \textbf{1.0000} & \textbf{1.0000} & 0.8727 & \textbf{1.0000} & 0.8435 & 0.8513 & \textbf{1.0000} & 126 & 0.0000 & 0.0000 & 0.0000 & \textbf{1.0000} & 0.6580 & \textbf{1.0000} & 0.0000 & 0.6875 & \textbf{1.0000} \\
2 & 0.0000 & 0.0000 & \textbf{1.0000} & \textbf{1.0000} & 0.7577 & 0.0000 & 0.0001 & 0.0258 & \textbf{1.0000} & 127 & 0.0000 & 0.0000 & \textbf{1.0000} & 0.0000 & 0.0000 & 0.0000 & 0.0000 & 0.0000 & 0.0000 \\
3 & 0.0000 & 0.0000 & \textbf{1.0000} & \textbf{1.0000} & 0.7695 & \textbf{1.0000} & 0.7790 & 0.6923 & \textbf{1.0000} & 128 & 0.0000 & 0.0000 & \textbf{1.0000} & \textbf{1.0000} & 0.6230 & 0.0000 & 0.5458 & 0.7311 & \textbf{1.0000} \\
4 & 0.0000 & 0.0000 & 0.0000 & \textbf{1.0000} & 0.6255 & \textbf{1.0000} & 0.7574 & 0.6601 & \textbf{1.0000} & 129 & 0.0000 & 0.0000 & 0.0000 & \textbf{1.0000} & 0.3704 & 0.0000 & 0.1418 & 0.3931 & \textbf{1.0000} \\
5 & 0.0000 & 0.0000 & 0.0000 & 0.0000 & 0.0000 & 0.0000 & 0.0000 & \textbf{0.0100} & 0.0000 & 130 & 0.0000 & 0.0000 & 0.0000 & \textbf{1.0000} & 0.4369 & 0.0000 & 0.2741 & 0.5929 & \textbf{1.0000} \\
6 & 0.0000 & 0.0000 & \textbf{1.0000} & \textbf{1.0000} & 0.0000 & 0.0000 & 0.0000 & 0.0300 & 0.0000 & 131 & 0.0000 & 0.0000 & 0.0000 & \textbf{1.0000} & 0.4485 & 0.0000 & 0.0671 & 0.5522 & \textbf{1.0000} \\
7 & 0.0000 & 0.0000 & 0.0000 & \textbf{1.0000} & 0.0908 & 0.0000 & 0.0775 & 0.0263 & 0.0000 & 132 & 0.0000 & 0.0000 & 0.0000 & 0.0000 & 0.2770 & 0.0000 & 0.0000 & 0.0443 & \textbf{1.0000} \\
\bottomrule
\end{tabular}}
\end{table*}

% \usepackage{booktabs}
\begin{table*}[]
\centering
\caption{{}{The test precision of affiliation of time series anomaly detection in UCR 250 sub-datasets. The best results are in \textbf{bold}.}}
\label{tab:appendix_ad_affp}
\scalebox{0.6}{
\begin{tabular}{@{}c|ccccccccc|c|ccccccccc@{}}
\toprule
UCR ID & SPOT & DSPOT & LSTM-VAE & DONUT & AT & TS2Vec & TimesNet & GPT4TS & DCdetector & UCR ID & SPOT & DSPOT & LSTM-VAE & DONUT & AT & TS2Vec & TimesNet & GPT4TS & DCdetector \\ \midrule
1 & 0.4090 & 0.4340 & \textbf{0.7925} & \textbf{0.7973} & 0.5759 & \textbf{0.8733} & 0.6697 & \textbf{0.8979} & \textbf{0.4961} & 126 & 0.0000 & 0.0000 & 0.4992 & \textbf{0.5000} & 0.9586 & \textbf{0.4992} & 0.8155 & \textbf{0.9857} & \textbf{0.5040} \\
2 & 0.4730 & 0.5176 & \textbf{0.6823} & \textbf{0.7018} & 0.6484 & 0.4983 & 0.1161 & 0.3632 & \textbf{0.5071} & 127 & 0.2293 & 0.2600 & \textbf{0.5169} & 0.5348 & 0.3807 & 0.0000 & 0.1867 & 0.3611 & \textbf{0.5395} \\
3 & 0.4291 & 0.4200 & \textbf{0.6981} & \textbf{0.7136} & 0.5393 & \textbf{0.7986} & \textbf{0.8196} & 0.4588 & \textbf{0.5051} & 128 & 0.2293 & 0.2600 & \textbf{0.4437} & \textbf{0.5113} & 0.8652 & 0.0000 & 0.8869 & \textbf{0.9056} & \textbf{0.5862} \\
4 & 0.4679 & 0.6378 & 0.4833 & \textbf{0.8511} & 0.4663 & \textbf{0.9712} & 0.9668 & 0.8493 & \textbf{0.4469} & 129 & 0.2293 & 0.2600 & 0.4410 & \textbf{0.5015} & \textbf{0.7551} & 0.0000 & 0.7144 & 0.6575 & \textbf{0.5017} \\
5 & 0.0000 & 0.0000 & 0.4387 & 0.4426 & 0.4021 & 0.6587 & \textbf{0.8404} & \textbf{0.6251} & 0.4833 & 130 & 0.2646 & 0.2275 & 0.4891 & \textbf{0.4888} & \textbf{0.9753} & 0.0000 & 0.5775 & 0.9131 & \textbf{0.5506} \\
6 & 0.0000 & 0.0000 & \textbf{0.7010} & \textbf{0.6349} & 0.4571 & 0.5504 & 0.2876 & 0.6208 & 0.4422 & 131 & 0.5054 & 0.5703 & 0.5058 & \textbf{0.4966} & 0.6409 & 0.0000 & 0.3799 & \textbf{0.7442} & \textbf{0.6075} \\
7 & 0.0000 & 0.0000 & 0.5382 & \textbf{0.6364} & 0.6201 & \textbf{0.8482} & 0.7904 & 0.6689 & 0.5536 & 132 & 0.0000 & 0.0000 & 0.5076 & 0.5183 & \textbf{0.6503} & 0.0000 & 0.6090 & 0.6015 & \textbf{0.5117} \\
\bottomrule
\end{tabular}}
\end{table*}

% \usepackage{booktabs}
\begin{table*}[]
\centering
\caption{{}{The test recall of affiliation of time series anomaly detection in UCR 250 sub-datasets. The best results are in \textbf{bold}.}}
\label{tab:appendix_ad_affr}
\scalebox{0.6}{
\begin{tabular}{@{}c|ccccccccc|c|ccccccccc@{}}
\toprule
UCR ID & SPOT & DSPOT & LSTM-VAE & DONUT & AT & TS2Vec & TimesNet & GPT4TS & DCdetector & UCR ID & SPOT & DSPOT & LSTM-VAE & DONUT & AT & TS2Vec & TimesNet & GPT4TS & DCdetector \\ \midrule
1 & 0.8970 & 0.8970 & \textbf{1.0000} & \textbf{1.0000} & 0.4844 & \textbf{1.0000} & 0.0149 & \textbf{0.0369} & \textbf{0.9939} & 126 & 0.0000 & 0.0000 & 0.9955 & \textbf{0.9982} & 0.4275 & \textbf{0.9969} & 0.0344 & \textbf{0.3782} & \textbf{0.9979} \\
2 & 0.9741 & 0.9741 & \textbf{1.0000} & \textbf{1.0000} & 0.3930 & 0.9944 & 0.0642 & 0.1064 & \textbf{0.9986} & 127 & 0.3393 & 0.3397 & \textbf{0.9963} & 0.9614 & 0.1373 & 0.0000 & 0.0680 & 0.2334 & \textbf{0.9838} \\
3 & 0.9547 & 0.9547 & \textbf{1.0000} & \textbf{1.0000} & 0.4170 & \textbf{1.0000} & \textbf{0.2048} & 0.1430 & \textbf{0.9984} & 128 & 0.3431 & 0.3437 & \textbf{0.9894} & \textbf{0.9871} & 0.1569 & 0.0000 & 0.0436 & \textbf{0.2462} & \textbf{0.9903} \\
4 & 0.6381 & 0.6391 & 0.9642 & \textbf{1.0000} & 0.2022 & \textbf{1.0000} & 0.4152 & 0.3633 & \textbf{0.9876} & 129 & 0.3393 & 0.3397 & 0.9759 & \textbf{0.9939} & \textbf{0.4392} & 0.0000 & 0.3926 & 0.5553 & \textbf{0.9966} \\
5 & 0.0000 & 0.0000 & \textbf{0.9940} & 0.9888 & 0.1028 & 0.7813 & \textbf{0.4673} & \textbf{0.0849} & 0.9622 & 130 & 0.4515 & 0.2973 & 0.9862 & \textbf{0.9958} & \textbf{0.2235} & 0.0000 & 0.1778 & 0.3196 & \textbf{0.9954} \\
6 & 0.0000 & 0.0000 & \textbf{1.0000} & \textbf{1.0000} & 0.3136 & 0.7887 & 0.0766 & 0.2455 & 0.9925 & 131 & 0.7254 & 0.7263 & 0.9722 & \textbf{0.9889} & 0.4038 & 0.0000 & 0.0604 & \textbf{0.4965} & \textbf{0.9948} \\
7 & 0.0000 & 0.0000 & 0.9895 & \textbf{1.0000} & 0.3308 & \textbf{0.8676} & 0.2035 & 0.1199 & 0.9897 & 132 & 0.0000 & 0.0000 & 0.9888 & 0.9823 & \textbf{0.2302} & 0.0000 & 0.2587 & 0.2113 & \textbf{0.9924} \\
\bottomrule
\end{tabular}}
\end{table*}

% \usepackage{booktabs}
\begin{table*}[]
\centering
\caption{{}{The test Range-AUC-ROC of time series anomaly detection in UCR 250 sub-datasets. The best results are in \textbf{bold}.}}
\label{tab:appendix_ad_rar}
\scalebox{0.6}{
\begin{tabular}{@{}c|ccccccccc|c|ccccccccc@{}}
\toprule
UCR ID & SPOT & DSPOT & LSTM-VAE & DONUT & AT & TS2Vec & TimesNet & GPT4TS & DCdetector & UCR ID & SPOT & DSPOT & LSTM-VAE & DONUT & AT & TS2Vec & TimesNet & GPT4TS & DCdetector \\ \midrule
1 & 0.5064 & 0.5054 & \textbf{0.7237} & \textbf{0.7096} & 0.9023 & \textbf{0.6215} & 0.8896 & \textbf{0.8933} & \textbf{0.9671} & 126 & 0.0000 & 0.0000 & 0.6970 & \textbf{0.7334} & 0.7229 & \textbf{0.6734} & 0.4991 & \textbf{0.7326} & \textbf{0.8520} \\
2 & 0.5100 & 0.5090 & \textbf{0.7267} & \textbf{0.7291} & 0.8248 & 0.6142 & 0.4988 & 0.5096 & \textbf{0.9352} & 127 & 0.5061 & 0.5030 & \textbf{0.7394} & 0.6811 & 0.5032 & 0.0000 & 0.5026 & 0.5053 & \textbf{0.5105} \\
3 & 0.5100 & 0.5090 & \textbf{0.7360} & \textbf{0.7328} & 0.8306 & \textbf{0.6225} & \textbf{0.8352} & 0.7978 & \textbf{0.9353} & 128 & 0.4938 & 0.4907 & \textbf{0.7357} & \textbf{0.7017} & 0.7549 & 0.0000 & 0.7232 & \textbf{0.7991} & \textbf{0.9162} \\
4 & 0.4923 & 0.4907 & 0.7546 & \textbf{0.7332} & 0.7536 & \textbf{0.5445} & 0.8064 & 0.7671 & \textbf{0.9101} & 129 & 0.5060 & 0.5029 & 0.7142 & \textbf{0.6868} & \textbf{0.5933} & 0.0000 & 0.5383 & 0.6022 & \textbf{0.7611} \\
5 & 0.0000 & 0.0000 & \textbf{0.6867} & \textbf{0.7148} & 0.5048 & 0.5174 & \textbf{0.5101} & \textbf{0.5039} & 0.5042 & 130 & 0.5034 & 0.4979 & 0.7404 & \textbf{0.7153} & \textbf{0.6564} & 0.0000 & 0.5991 & 0.7119 & \textbf{0.8657} \\
6 & 0.0000 & 0.0000 & \textbf{0.7339} & \textbf{0.7948} & 0.5084 & 0.5396 & 0.5059 & 0.5110 & 0.5126 & 131 & 0.5017 & 0.4986 & 0.7291 & \textbf{0.7335} & 0.6570 & 0.0000 & 0.5246 & \textbf{0.6940} & \textbf{0.8535} \\
7 & 0.0000 & 0.0000 & 0.7506 & \textbf{0.8350} & 0.5224 & \textbf{0.5080} & 0.5186 & 0.5086 & 0.5179 & 132 & 0.0000 & 0.0000 & \textbf{0.7592} & 0.6915 & \textbf{0.5585} & 0.0000 & 0.5069 & 0.5146 & \textbf{0.7087} \\
\bottomrule
\end{tabular}}
\end{table*}

% \usepackage{booktabs}
\begin{table*}[]
\centering
\caption{{}{The test Range-AUC-PR of time series anomaly detection in UCR 250 sub-datasets. The best results are in \textbf{bold}.}}
\label{tab:appendix_ad_rap}
\scalebox{0.6}{
\begin{tabular}{@{}c|ccccccccc|c|ccccccccc@{}}
\toprule
UCR ID & SPOT & DSPOT & LSTM-VAE & DONUT & AT & TS2Vec & TimesNet & GPT4TS & DCdetector & UCR ID & SPOT & DSPOT & LSTM-VAE & DONUT & AT & TS2Vec & TimesNet & GPT4TS & DCdetector \\ \midrule
1 & 0.0278 & 0.0259 & \textbf{0.8135} & \textbf{0.8001} & 0.6994 & \textbf{0.7010} & 0.7387 & \textbf{0.7450} & \textbf{0.8485} & 126 & 0.0000 & 0.0000 & 0.4779 & \textbf{0.7780} & 0.4152 & \textbf{0.5666} & 0.0052 & \textbf{0.4328} & \textbf{0.6184} \\
2 & 0.0277 & 0.0258 & \textbf{0.8151} & \textbf{0.8173} & 0.5120 & 0.2433 & 0.0068 & 0.0329 & \textbf{0.7367} & 127 & 0.0194 & 0.0141 & \textbf{0.8026} & 0.2723 & 0.0183 & 0.0000 & 0.0159 & 0.0257 & \textbf{0.0453} \\
3 & 0.0276 & 0.0258 & \textbf{0.8236} & \textbf{0.8207} & 0.5215 & \textbf{0.7002} & \textbf{0.5890} & 0.5385 & \textbf{0.7385} & 128 & 0.0194 & 0.0141 & \textbf{0.6531} & \textbf{0.5518} & 0.6670 & 0.0000 & 0.6232 & \textbf{0.7212} & \textbf{0.8582} \\
4 & 0.0085 & 0.0057 & 0.3549 & \textbf{0.8241} & 0.6422 & \textbf{0.5817} & 0.7129 & 0.6560 & \textbf{0.8320} & 129 & 0.0194 & 0.0141 & 0.3065 & \textbf{0.7143} & \textbf{0.2692} & 0.0000 & 0.1314 & 0.2905 & \textbf{0.5724} \\
5 & 0.0000 & 0.0000 & \textbf{0.7399} & \textbf{0.8024} & 0.0170 & 0.0343 & \textbf{0.0395} & \textbf{0.0142} & 0.0135 & 130 & 0.0230 & 0.0123 & 0.5119 & \textbf{0.7987} & \textbf{0.4559} & 0.0000 & 0.3388 & 0.5457 & \textbf{0.7514} \\
6 & 0.0000 & 0.0000 & \textbf{0.8217} & \textbf{0.8731} & 0.0369 & 0.0796 & 0.0255 & 0.0478 & 0.0695 & 131 & 0.0194 & 0.0140 & 0.3263 & \textbf{0.6173} & 0.4608 & 0.0000 & 0.1243 & \textbf{0.5239} & \textbf{0.7560} \\
7 & 0.0000 & 0.0000 & 0.7758 & \textbf{0.9031} & 0.0942 & \textbf{0.0218} & 0.0869 & 0.0375 & 0.0818 & 132 & 0.0000 & 0.0000 & \textbf{0.7774} & 0.6899 & \textbf{0.2128} & 0.0000 & 0.0320 & 0.0653 & \textbf{0.5743} \\
\bottomrule
\end{tabular}}
\end{table*}

% \usepackage{booktabs}
\begin{table*}[]
\centering
\caption{{}{The test V\_ROC of time series anomaly detection in UCR 250 sub-datasets. The best results are in \textbf{bold}.}}
\label{tab:appendix_ad_vroc}
\scalebox{0.6}{
\begin{tabular}{@{}c|ccccccccc|c|ccccccccc@{}}
\toprule
UCR ID & SPOT & DSPOT & LSTM-VAE & DONUT & AT & TS2Vec & TimesNet & GPT4TS & DCdetector & UCR ID & SPOT & DSPOT & LSTM-VAE & DONUT & AT & TS2Vec & TimesNet & GPT4TS & DCdetector \\ \midrule
1 & 0.5062 & 0.5054 & \textbf{0.7106} & \textbf{0.7015} & 0.8519 & \textbf{0.6337} & 0.6990 & \textbf{0.7163} & \textbf{0.9673} & 126 & 0.0000 & 0.0000 & 0.7001 & \textbf{0.7241} & 0.6815 & \textbf{0.6949} & 0.4987 & \textbf{0.6709} & \textbf{0.8637} \\
2 & 0.5099 & 0.5090 & \textbf{0.7120} & \textbf{0.7173} & 0.7759 & 0.6258 & 0.4983 & 0.5045 & \textbf{0.9382} & 127 & 0.5061 & 0.5030 & \textbf{0.7238} & 0.6695 & 0.5018 & 0.0000 & 0.5011 & 0.5030 & \textbf{0.5068} \\
3 & 0.5099 & 0.5090 & \textbf{0.7196} & \textbf{0.7171} & 0.7835 & \textbf{0.6350} & \textbf{0.7088} & 0.6754 & \textbf{0.9379} & 128 & 0.4938 & 0.4907 & \textbf{0.7198} & \textbf{0.6897} & 0.6561 & 0.0000 & 0.6278 & \textbf{0.7011} & \textbf{0.9203} \\
4 & 0.4923 & 0.4907 & 0.7306 & \textbf{0.7185} & 0.6616 & \textbf{0.5459} & 0.7331 & 0.7192 & \textbf{0.9150} & 129 & 0.5061 & 0.5029 & 0.7021 & \textbf{0.6804} & \textbf{0.5789} & 0.0000 & 0.5324 & 0.5932 & \textbf{0.7850} \\
5 & 0.0000 & 0.0000 & \textbf{0.6794} & \textbf{0.7043} & 0.5028 & 0.5175 & \textbf{0.5095} & \textbf{0.5023} & 0.5077 & 130 & 0.5034 & 0.4979 & 0.7240 & \textbf{0.6990} & \textbf{0.6123} & 0.0000 & 0.5618 & 0.6480 & \textbf{0.8756} \\
6 & 0.0000 & 0.0000 & \textbf{0.7154} & \textbf{0.7601} & 0.5059 & 0.5384 & 0.5038 & 0.5075 & 0.5137 & 131 & 0.5018 & 0.4986 & 0.7176 & \textbf{0.7240} & 0.6327 & 0.0000 & 0.5133 & \textbf{0.6813} & \textbf{0.8652} \\
7 & 0.0000 & 0.0000 & 0.7258 & \textbf{0.7877} & 0.5166 & \textbf{0.5071} & 0.5124 & 0.5053 & 0.5161 & 132 & 0.0000 & 0.0000 & \textbf{0.7320} & 0.6793 & \textbf{0.5393} & 0.0000 & 0.5045 & 0.5095 & \textbf{0.7413} \\
\bottomrule
\end{tabular}}
\end{table*}

% \usepackage{booktabs}
\begin{table*}[]
\centering
\caption{{}{The test V\_PR of time series anomaly detection in UCR 250 sub-datasets. The best results are in \textbf{bold}.}}
\label{tab:appendix_ad_vpr}
\scalebox{0.6}{
\begin{tabular}{@{}c|ccccccccc|c|ccccccccc@{}}
\toprule
UCR ID & SPOT & DSPOT & LSTM-VAE & DONUT & AT & TS2Vec & TimesNet & GPT4TS & DCdetector & UCR ID & SPOT & DSPOT & LSTM-VAE & DONUT & AT & TS2Vec & TimesNet & GPT4TS & DCdetector \\ \midrule
1 & 0.0273 & 0.0256 & \textbf{0.7813} & \textbf{0.7726} & 0.6497 & \textbf{0.7001} & 0.5508 & \textbf{0.5705} & \textbf{0.8483} & 126 & 0.0000 & 0.0000 & 0.4773 & \textbf{0.6601} & 0.3739 & \textbf{0.5440} & 0.0047 & \textbf{0.3714} & \textbf{0.6299} \\
2 & 0.0271 & 0.0255 & \textbf{0.7815} & \textbf{0.7852} & 0.4635 & 0.2439 & 0.0061 & 0.0277 & \textbf{0.7398} & 127 & 0.0193 & 0.0140 & \textbf{0.7012} & 0.2440 & 0.0153 & 0.0000 & 0.0129 & 0.0211 & \textbf{0.0339} \\
3 & 0.0271 & 0.0255 & \textbf{0.7879} & \textbf{0.7855} & 0.4745 & \textbf{0.6993} & \textbf{0.4635} & 0.4169 & \textbf{0.7411} & 128 & 0.0193 & 0.0140 & \textbf{0.5770} & \textbf{0.5093} & 0.5714 & 0.0000 & 0.5308 & \textbf{0.6264} & \textbf{0.8620} \\
4 & 0.0084 & 0.0056 & 0.3121 & \textbf{0.7906} & 0.5520 & \textbf{0.5818} & 0.6415 & 0.6090 & \textbf{0.8369} & 129 & 0.0193 & 0.0140 & 0.3330 & \textbf{0.6049} & \textbf{0.2540} & 0.0000 & 0.1242 & 0.2800 & \textbf{0.5949} \\
5 & 0.0000 & 0.0000 & \textbf{0.6472} & \textbf{0.6370} & 0.0131 & 0.0341 & \textbf{0.0364} & \textbf{0.0113} & 0.0324 & 130 & 0.0230 & 0.0123 & 0.4936 & \textbf{0.6705} & \textbf{0.4123} & 0.0000 & 0.3017 & 0.4828 & \textbf{0.7613} \\
6 & 0.0000 & 0.0000 & \textbf{0.7866} & \textbf{0.8230} & 0.0306 & 0.0760 & 0.0199 & 0.0421 & 0.0698 & 131 & 0.0193 & 0.0140 & 0.3378 & \textbf{0.5814} & 0.4359 & 0.0000 & 0.1127 & \textbf{0.5106} & \textbf{0.7677} \\
7 & 0.0000 & 0.0000 & 0.6491 & \textbf{0.8454} & 0.0861 & \textbf{0.0199} & 0.0793 & 0.0329 & 0.0733 & 132 & 0.0000 & 0.0000 & \textbf{0.6366} & 0.5472 & \textbf{0.1917} & 0.0000 & 0.0275 & 0.0573 & \textbf{0.6069} \\
\bottomrule
\end{tabular}}
\end{table*}

% \usepackage{booktabs}
\begin{table*}[]
\centering
\caption{{}{The test F1-PA-10 of time series anomaly detection in UCR 250 sub-datasets. The best results are in \textbf{bold}.}}
\label{tab:appendix_ad_f1pa10}
\scalebox{0.6}{
\begin{tabular}{@{}c|cccccc|c|cccccc@{}}
\toprule
UCR ID & SPOT & DSPOT & LSTM-VAE & DONUT & TS2Vec & DCdetector & UCR ID & SPOT & DSPOT & LSTM-VAE & DONUT & TS2Vec & DCdetector \\ \midrule
1 & 0.0920 & 0.1968 & 0.1940 & 0.1961 & \textbf{0.3301} & 0.1650 & 126 & 0.0501 & 0.0285 & 0.0630 & \textbf{0.0789} & 0.0614 & 0.0464 \\
2 & 0.0328 & 0.0627 & 0.1040 & 0.0936 & 0.0303 & \textbf{0.1592} & 127 & 0.0262 & 0.0191 & \textbf{0.1259} & 0.0950 & 0.0360 & 0.0000 \\
3 & \textbf{0.3396} & 0.1791 & 0.1233 & 0.1153 & 0.1774 & 0.0816 & 128 & 0.3641 & 0.3405 & 0.3447 & \textbf{0.4189} & 0.3845 & 0.0361 \\
4 & 0.3135 & \textbf{0.4158} & 0.3228 & 0.3350 & 0.3604 & 0.0552 & 129 & 0.1557 & 0.0881 & 0.0864 & 0.1095 & \textbf{0.4167} & 0.2041 \\
5 & \textbf{0.0000} & \textbf{0.0000} & \textbf{0.0000} & \textbf{0.0000} & \textbf{0.0000} & \textbf{0.0000} & 130 & \textbf{0.4591} & 0.2391 & 0.2220 & 0.2810 & 0.3220 & 0.3089 \\
6 & 0.0921 & 0.1023 & \textbf{0.2892} & 0.1752 & 0.0958 & 0.1595 & 131 & 0.7939 & 0.2755 & 0.2327 & 0.2147 & \textbf{0.9630} & 0.2515 \\
7 & 0.0726 & \textbf{0.1408} & 0.0734 & 0.1360 & 0.0520 & 0.0000 & 132 & 0.0534 & 0.0405 & 0.0929 & 0.0942 & 0.0284 & \textbf{0.2765} \\
\bottomrule
\end{tabular}}
\end{table*}

% \usepackage{booktabs}
\begin{table*}[]
\centering
\caption{{}{The test F1-PA-50 of time series anomaly detection in UCR 250 sub-datasets. The best results are in \textbf{bold}.}}
\label{tab:appendix_ad_f1pa50}
\scalebox{0.6}{
\begin{tabular}{@{}c|cccccc|c|cccccc@{}}
\toprule
UCR ID & SPOT & DSPOT & LSTM-VAE & DONUT & TS2Vec & DCdetector & UCR ID & SPOT & DSPOT & LSTM-VAE & DONUT & TS2Vec & DCdetector \\ \midrule
1 & 0.0307 & 0.0616 & 0.0529 & 0.0563 & \textbf{0.0676} & 0.0285 & 126 & 0.0139 & 0.0149 & 0.0184 & \textbf{0.0168} & \textbf{0.0193} & 0.0183 \\
2 & 0.0256 & \textbf{0.0349} & 0.0250 & 0.0255 & 0.0244 & \textbf{0.0252} & 127 & 0.0253 & 0.0156 & \textbf{0.0273} & 0.0195 & \textbf{0.0340} & 0.0000 \\
3 & \textbf{0.0524} & 0.0333 & 0.0239 & 0.0256 & 0.0290 & 0.0191 & 128 & 0.1124 & \textbf{0.1677} & 0.1049 & \textbf{0.1033} & 0.1059 & 0.0361 \\
4 & \textbf{0.1410} & \textbf{0.1373} & 0.0962 & 0.0945 & 0.1086 & 0.0552 & 129 & \textbf{0.0925} & 0.0572 & 0.0280 & 0.0312 & \textbf{0.0833} & 0.0460 \\
5 & \textbf{0.0000} & \textbf{0.0000} & \textbf{0.0000} & \textbf{0.0000} & \textbf{0.0000} & \textbf{0.0000} & 130 & \textbf{0.1056} & \textbf{0.1297} & 0.0637 & 0.0597 & 0.0579 & 0.0481 \\
6 & \textbf{0.0812} & 0.0696 & \textbf{0.0516} & 0.0755 & 0.0406 & 0.0483 & 131 & 0.1351 & 0.0514 & 0.0574 & 0.0576 & \textbf{0.2378} & 0.0583 \\
7 & 0.0345 & \textbf{0.0457} & 0.0199 & 0.0323 & 0.0324 & 0.0000 & 132 & 0.0419 & 0.0387 & 0.0210 & 0.0250 & 0.0185 & \textbf{0.0976} \\
\bottomrule
\end{tabular}}
\end{table*}

% \usepackage{booktabs}
\begin{table*}[]
\centering
\caption{{}{The test F1-PA-90 of time series anomaly detection in UCR 250 sub-datasets. The best results are in \textbf{bold}.}}
\label{tab:appendix_ad_f1pa90}
\scalebox{0.6}{
\begin{tabular}{@{}c|cccccc|c|cccccc@{}}
\toprule
UCR ID & SPOT & DSPOT & LSTM-VAE & DONUT & TS2Vec & DCdetector & UCR ID & SPOT & DSPOT & LSTM-VAE & DONUT & TS2Vec & DCdetector \\ \midrule
1 & 0.0277 & 0.0374 & 0.0297 & 0.0316 & \textbf{0.0676} & 0.0285 & 126 & 0.0127 & 0.0130 & 0.0100 & \textbf{0.0157} & \textbf{0.0102} & \textbf{0.0183} \\
2 & 0.0210 & \textbf{0.0247} & 0.0147 & 0.0143 & 0.0186 & \textbf{0.0252} & 127 & 0.0239 & 0.0129 & \textbf{0.0267} & 0.0178 & \textbf{0.0232} & 0.0000 \\
3 & \textbf{0.0431} & 0.0333 & 0.0149 & 0.0145 & 0.0290 & 0.0191 & 128 & 0.0703 & \textbf{0.1282} & 0.0643 & \textbf{0.0644} & 0.0679 & 0.0361 \\
4 & \textbf{0.1410} & \textbf{0.1373} & 0.0522 & 0.0511 & 0.1086 & 0.0552 & 129 & \textbf{0.0504} & 0.0318 & 0.0154 & 0.0244 & \textbf{0.0833} & 0.0460 \\
5 & \textbf{0.0000} & \textbf{0.0000} & \textbf{0.0000} & \textbf{0.0000} & \textbf{0.0000} & \textbf{0.0000} & 130 & \textbf{0.0657} & \textbf{0.0716} & 0.0341 & 0.0345 & 0.0579 & 0.0481 \\
6 & \textbf{0.0556} & 0.0385 & \textbf{0.0516} & \textbf{0.0755} & 0.0259 & 0.0483 & 131 & 0.1351 & 0.0463 & 0.0326 & 0.0358 & \textbf{0.2378} & 0.0583 \\
7 & 0.0212 & \textbf{0.0320} & 0.0130 & \textbf{0.0323} & 0.0198 & 0.0000 & 132 & 0.0315 & 0.0278 & 0.0154 & 0.0250 & 0.0165 & \textbf{0.0976} \\
\bottomrule
\end{tabular}}
\end{table*}

% \usepackage{booktabs}
% \usepackage[normalem]{ulem}
\begin{table*}[]
\centering
\caption{{}{The test results of time series anomaly detection on seven real-world multivariate datasets. The best results are in bold. The best model in the corresponding dataset is in \underline{underline} and \textbf{bold}. "-" indicates that the metric could not be obtained due to the excessive time required to compute it from the model's output. In addition, LSTM-VAE, DOUNT, and TS2Vec are unable to produce test results on the PSM and SWAT datasets due to memory errors. Furthermore, SPOT and DSPOT can not obtain test results on multivariate datasets because the original code provided by the authors does not include settings for multivariate scenarios.} }
\label{tab:appendix_ad_multi}
\begin{tabular}{@{}ccccccccccccc@{}}
\toprule
\multicolumn{13}{c}{{}{The SMD Dataset}} \\ \midrule
Models & F1 & P & R & F1-PA-10 & F1-PA-50 & F1-PA-90 & Aff-P & Aff-R & R\_A\_R & R\_A\_P & V\_ROC & V\_PR \\ \hline
LSTM-VAE & 0.5002 & 0.9964 & 0.3339 & 0.4244 & 0.1040 & 0.0911 & \textbf{0.8785} & 0.2177 & 0.7460 & 0.3208 & 0.7484 & 0.3188 \\
DONUT & 0.4965 & \textbf{0.9964} & 0.3306 & 0.4247 & 0.1043 & 0.0915 & 0.8763 & 0.2143 & 0.7405 & 0.3136 & 0.7440 & 0.3130 \\
AT & 0.8259 & 0.8694 & 0.7864 & - & - & - & - & - & - & - & - & - \\
TS2Vec & \textbf{0.8571} & 0.9447 & 0.7843 & \textbf{0.4477} & \textbf{0.1149} & \textbf{0.1128} & 0.7972 & 0.6678 & 0.5579 & 0.2971 & 0.5546 & 0.2740 \\
TimesNet & 0.8031 & 0.8577 & 0.7551 & - & - & - & - & - & - & - & - & - \\
GPT4TS & 0.8421 & 0.8731 & 0.8133 & - & - & - & - & - & - & - & - & - \\ 
{\ul \textbf{DCdetector}} & 0.8453 & 0.7971 & \textbf{0.8997} & 0.3086 & 0.1073 & 0.0735 & 0.5096 & \textbf{0.9480} & \textbf{0.7989} & \textbf{0.7186} & \textbf{0.7547} & \textbf{0.6767} \\ \midrule
\multicolumn{13}{c}{{}{The MSL Dataset}} \\ \midrule
Models & F1 & P & R & F1-PA-10 & F1-PA-50 & F1-PA-90 & Aff-P & Aff-R & R\_A\_R & R\_A\_P & V\_ROC & V\_PR \\ \hline
LSTM-VAE & 0.3936 & 0.9643 & 0.2472 & 0.5800 & \textbf{0.2383} & \textbf{0.1915} & 0.5296 & 0.4924 & 0.5060 & 0.1361 & 0.5054 & 0.1279 \\
DONUT & 0.3936 & \textbf{0.9648} & 0.2472 & 0.5907 & 0.2373 & 0.1913 & 0.5310 & 0.4924 & 0.5056 & 0.1354 & 0.5054 & 0.1278 \\
AT & 0.8683 & 0.9083 & 0.8318 & - & - & - & 0.6569 & 0.5020 & 0.8436 & 0.8246 & 0.8010 & 0.7860 \\
TS2Vec & 0.8748 & 0.9473 & 0.8126 & 0.5852 & 0.2192 & 0.1907 & 0.5402 & 0.8830 & 0.5860 & 0.3798 & 0.5808 & 0.3504 \\
TimesNet & 0.8014 & 0.8834 & 0.7334 & - & - & - & 0.6764 & 0.3371 & 0.7691 & 0.7590 & 0.7004 & 0.6952 \\
GPT4TS & 0.8114 & 0.8865 & 0.7480 & - & - & - & \textbf{0.6820} & 0.3752 & 0.7786 & 0.7628 & 0.7154 & 0.7050 \\
{\ul \textbf{DCdetector}} & \textbf{0.9563} & 0.9256 & \textbf{0.9892} & \textbf{0.6719} & 0.1608 & 0.1596 & 0.5177 & \textbf{0.9738} & \textbf{0.9224} & \textbf{0.9012} & \textbf{0.9150} & \textbf{0.8949} \\ \midrule
\multicolumn{13}{c}{{}{The SMAP Dataset}} \\ \midrule
Models & F1 & P & R & F1-PA-10 & F1-PA-50 & F1-PA-90 & Aff-P & Aff-R & R\_A\_R & R\_A\_P & V\_ROC & V\_PR \\ \hline
LSTM-VAE & 0.6863 & 0.9896 & 0.5253 & \textbf{0.7123} & 0.3311 & 0.2433 & 0.4700 & 0.5768 & 0.4659 & 0.0679 & 0.4648 & 0.0657 \\
DONUT & 0.6864 & \textbf{0.9899} & 0.5253 & 0.7059 & \textbf{0.3347} & \textbf{0.2439} & 0.4804 & 0.5745 & 0.4660 & 0.0677 & 0.4650 & 0.0655 \\
AT & 0.9435 & 0.9451 & 0.9420 & - & - & - & 0.6920 & 0.4431 & 0.9424 & 0.9266 & 0.8713 & 0.8648 \\
TS2Vec & 0.7013 & 0.9467 & 0.5569 & 0.5450 & 0.3114 & 0.2433 & 0.4024 & 0.6587 & 0.5280 & 0.2378 & 0.5172 & 0.2087 \\
TimesNet & 0.6705 & 0.9209 & 0.5271 & - & - & - & 0.7317 & 0.1699 & 0.6549 & 0.6710 & 0.6016 & 0.6234 \\
GPT4TS & 0.6750 & 0.9117 & 0.5359 & - & - & - & \textbf{0.7844} & 0.2190 & 0.6697 & 0.6747 & 0.6200 & 0.6311 \\
{\ul \textbf{DCdetector}} & \textbf{0.9606} & 0.9367 & \textbf{0.9857} & 0.6729 & 0.1701 & 0.1701 & 0.5109 & \textbf{0.9886} & \textbf{0.9645} & \textbf{0.9424} & \textbf{0.9517} & \textbf{0.9313} \\ \midrule
\multicolumn{13}{c}{{}{The PSM Dataset}} \\ \midrule
Models & F1 & P & R & F1-PA-10 & F1-PA-50 & F1-PA-90 & Aff-P & Aff-R & R\_A\_R & R\_A\_P & V\_ROC & V\_PR \\ \hline
AT & 0.9632 & 0.9687 & 0.9577 & - & - & - & 0.6776 & 0.4548 & 0.9338 & 0.9434 & 0.8651 & 0.8942 \\
TimesNet & 0.9183 & 0.9850 & 0.8600 & - & - & - & \textbf{0.8416} & 0.5546 & 0.8162 & 0.8704 & 0.7761 & 0.8406 \\
GPT4TS & 0.9706 & \textbf{0.9860} & 0.9557 & - & - & - & 0.8361 & 0.5706 & \textbf{0.9383} & \textbf{0.9540} & 0.8575 & 0.8963 \\
{\ul \textbf{DCdetector}} & \textbf{0.9744} & 0.9699 & \textbf{0.9789} & \textbf{0.8144} & \textbf{0.2443} & \textbf{0.2442} & 0.5244 & \textbf{0.8173} & 0.9292 & 0.9385 & \textbf{0.8739} & \textbf{0.8974} \\ \midrule
\multicolumn{13}{c}{{}{The SWAT Dataset}} \\ \midrule
Models & F1 & P & R & F1-PA-10 & F1-PA-50 & F1-PA-90 & Aff-P & Aff-R & R\_A\_R & R\_A\_P & V\_ROC & V\_PR \\ \hline
AT & 0.9466 & 0.9219 & 0.9726 & - & - & - & \textbf{0.6967} & 0.5510 & 0.9674 & 0.9378 & 0.9398 & 0.9135 \\
TimesNet & 0.9265 & 0.9201 & 0.9330 & - & - & - & 0.6646 & 0.4193 & 0.8999 & 0.8834 & 0.8101 & 0.8037 \\
GPT4TS & 0.9284 & 0.9218 & 0.9352 & - & - & - & 0.6579 & 0.4320 & 0.9108 & 0.8902 & 0.8278 & 0.8170 \\
{\ul \textbf{DCdetector}} & \textbf{0.9588} & \textbf{0.9325} & \textbf{0.9867} & \textbf{0.6906} & \textbf{0.3289} & \textbf{0.2364} & 0.5352 & \textbf{0.9622} & \textbf{0.9758} & \textbf{0.9497} & \textbf{0.9677} & \textbf{0.9426} \\ \midrule
\multicolumn{13}{c}{{}{The NIPS\_TS\_Swan Dataset}} \\ \midrule 
Models & F1 & P & R & F1-PA-10 & F1-PA-50 & F1-PA-90 & Aff-P & Aff-R & R\_A\_R & R\_A\_P & V\_ROC & V\_PR \\ \hline
LSTM-VAE & 0.7387 & \textbf{0.9997} & 0.5857 & \textbf{0.7833} & 0.6286 & 0.4992 & 0.8722 & 0.0020 & 0.8282 & 0.5284 & 0.8266 & 0.5239 \\
DONUT & 0.7387 & \textbf{0.9997} & 0.5857 & 0.7820 & 0.6308 & \textbf{0.4995} & 0.8722 & 0.0020 & 0.8298 & 0.5315 & \textbf{0.8281} & 0.5269 \\
AT & 0.7346 & 0.9701 & 0.5911 & - & - & - & - & - & - & - & - & - \\
{\ul \textbf{TS2Vec}} & \textbf{0.7545} & 0.9938 & \textbf{0.6081} & 0.7650 & \textbf{0.6441} & 0.4991 & \textbf{0.9159} & \textbf{0.0604} & 0.7654 & 0.5930 & 0.7485 & 0.5623 \\
TimesNet & 0.7438 & 0.9930 & 0.5945 & - & - & - & - & - & - & - & - & - \\
GPT4TS & 0.7389 & 0.9956 & 0.5874 & - & - & - & - & - & - & - & - & - \\
DCdetector & 0.4009 & 0.8824 & 0.2594 & 0.3879 & 0.3015 & 0.2990 & 0.4751 & 0.0386 & \textbf{0.8416} & \textbf{0.9124} & 0.8259 & \textbf{0.8981} \\ \midrule
\multicolumn{13}{c}{{}{The NIPS\_TS\_Water Dataset}} \\ \midrule
Models & F1 & P & R & F1-PA-10 & F1-PA-50 & F1-PA-90 & Aff-P & Aff-R & R\_A\_R & R\_A\_P & V\_ROC & V\_PR \\ \hline
LSTM-VAE & 0.5155 & \textbf{0.9625} & 0.3521 & 0.4053 & 0.1321 & 0.1136 & 0.6465 & 0.2356 & 0.5575 & 0.1587 & 0.5461 & 0.1403 \\
DONUT & 0.5150 & 0.9590 & 0.3521 & 0.3779 & 0.1062 & 0.1062 & 0.6458 & 0.2948 & 0.5489 & 0.1505 & 0.5394 & 0.1339 \\
AT & 0.1821 & 0.1504 & 0.2307 & - & - & - & 0.5199 & 0.4796 & 0.5556 & 0.1603 & 0.5476 & 0.1492 \\
TS2Vec & 0.7736 & 0.9206 & 0.6671 & \textbf{0.9035} & \textbf{0.6459} & \textbf{0.5086} & 0.7530 & 0.6161 & 0.5428 & 0.3035 & 0.5433 & 0.2817 \\
TimesNet & 0.7048 & 0.6079 & 0.8385 & - & - & - & \textbf{0.8654} & 0.7332 & 0.7261 & 0.6211 & 0.7276 & 0.6080 \\
{\ul \textbf{GPT4TS}} & \textbf{0.7844} & 0.6994 & \textbf{0.8928} & - & - & - & 0.8372 & 0.8647 & \textbf{0.7264} & \textbf{0.6513} & \textbf{0.7505} & \textbf{0.6657} \\
DCdetector & 0.3694 & 0.3277 & 0.4233 & 0.1086 & 0.0166 & 0.0166 & 0.5124 & \textbf{0.8886} & 0.5954 & 0.2852 & 0.5873 & 0.2774 \\ \bottomrule
\end{tabular}
\end{table*}


\ifCLASSOPTIONcaptionsoff
  \newpage
\fi

\end{document}